\documentclass{article}

\usepackage{soul,xcolor}

\setstcolor{red}
\usepackage{mhchem}

\usepackage{xr-hyper}

\usepackage{arxiv}
\usepackage[utf8]{inputenc} 
\usepackage[T1]{fontenc}    
\usepackage{hyperref}       
\usepackage{url}            
\usepackage{booktabs}
\usepackage{longtable}

\usepackage{amsmath}    
\DeclareMathOperator{\sech}{sech}

\usepackage{amsfonts}       
\usepackage{nicefrac}       
\usepackage{microtype}      
\usepackage{lipsum}

\DeclareOldFontCommand{\bf}{\normalfont\bfseries}{\mathbf}

\usepackage[]{subcaption}
\usepackage{graphicx}
\usepackage[outdir=./]{epstopdf}
\DeclareGraphicsExtensions{.pdf,eps}
\newcommand{\beginsupplement}{%
    \setcounter{table}{0}
    \renewcommand{\thetable}{S\arabic{table}}%
    \setcounter{figure}{0}
    \renewcommand{\thefigure}{S\arabic{figure}}%
}

\usepackage{placeins}
\makeatletter
\AtBeginDocument{%
  \expandafter\renewcommand\expandafter\subsection\expandafter
    {\expandafter\@fb@secFB\subsection}%
  \newcommand\@fb@secFB{\FloatBarrier
    \gdef\@fb@afterHHook{\@fb@topbarrier \gdef\@fb@afterHHook{}}}%
  \g@addto@macro\@afterheading{\@fb@afterHHook}%
  \gdef\@fb@afterHHook{}%
}
\makeatother

\usepackage{titlesec}
\titleformat{\section}[hang]{\bfseries}{\thesection.\ }{5pt}{}
\titleformat{\subsection}[hang]{\itshape}{\thesubsection.\ }{5pt}{}

\usepackage[style=numeric,citestyle=numeric-comp,sorting=none,doi=true,url=false,giveninits=true,hyperref]{biblatex}
\title{SVD Perspectives for Augmenting \\DeepONet Flexibility and Interpretability
}

\author{
  Simone Venturi \\
  Extreme Scale Data Science \& Analytics\\
  Sandia National Laboratories \\
  Livermore, CA, 94550\\
  \texttt{sventur@sandia.gov} \\
   \And
  Tiernan Casey \\
  Extreme Scale Data Science \& Analytics\\
  Sandia National Laboratories \\
  Livermore, CA, 94550\\
  \texttt{tcasey@sandia.gov} \\
}

\begin{document}
\maketitle

\begin{abstract}

Deep operator networks (DeepONets) are powerful and flexible architectures that are attracting attention in multiple fields due to their utility for fast and accurate emulation of complex dynamics. As their remarkable generalization capabilities are primarily enabled by their projection-based attribute, in this paper, we investigate connections with low-rank techniques derived from the singular value decomposition (SVD). We demonstrate that some of the concepts behind proper orthogonal decomposition (POD)-neural networks can improve DeepONet’s design and training phases. These ideas lead us to a methodology extension that we name SVD-DeepONet. Moreover, through multiple SVD analyses of scenario- and time-aggregated snapshot matrices, we find that DeepONet inherits from its projection-based attribute strong inefficiencies in representing dynamics characterized by symmetries. Inspired by the work on shifted-POD, we develop flexDeepONet, an architecture enhancement that relies on a pre-transformation network for generating a moving reference frame and isolating the rigid components of the dynamics. In this way, the physics can be represented on a latent space free from rotations, translations, and stretches, and an accurate projection can be performed to a low-dimensional basis. In addition to improving DeepONet’s flexibility and interpretability, the proposed perspectives increase its generalization capabilities and computational efficiencies. For instance, we show flexDeepONet can accurately surrogate the dynamics of 19 thermodynamic variables in a combustion chemistry application by relying on 95\% fewer trainable parameters than that of the ``vanilla'' architecture. As stressed in the paper, we argue that DeepONet and SVD-based methods can reciprocally benefit from each other. In particular, the flexibility of the former in leveraging multiple data sources and multifidelity knowledge in the form of both unstructured data and physics-informed constraints has the potential to greatly extend the applicability of methodologies such as POD and principal component analysis (PCA).

\end{abstract}


\vspace{1mm}
\section{Introduction}

The computational burden required to numerically solve partial differential equations (PDEs) for complex systems still precludes their application to numerous real-world problems that need fast predictions (e.g., predictive control) or a large number of replicas (e.g., optimization and uncertainty quantification). As significant progress has been made in exploiting machine learning (ML) techniques to reduce these computational costs in many fields, expensive fully physics-based simulations are being replaced with efficient ML-based surrogates~\cite{Surrogates_Brunton_2020,Surrogates_Qian_2020,ROMS_Kim_2022, ROMS_Lauzon_2022}. Given that they can integrate multifidelity observational data~\cite{Surrogates_Reichstein_2019} and enforce physical laws~\cite{PINNs_Raissi_2019,PINNs_Sirignano_2018}, these emulators are able to achieve high predictive accuracy for a broad spectrum of problems~\cite{PINNs_Karniadakis_2021}: from contexts involving small amounts of available data and extensively known physics, to cases of big data and fully missing physics. \\
In the last few years, multiple research communities worldwide have been attracted by a particular subclass of ML-based surrogates, called neural operators~\cite{NeuralOs_Li_1_2020,NeuralOs_Li_2_2020,NeuralOs_Kovachki_2021,UATs_Chen_1995,DeepONet_Lu_2019,DeepONet_Lu_2021,PCANN_Bhattacharya_2020,NeuralOs_Trask_2019,NeuralOs_Gin_2021,NeuralOs_Patel_2021,NeuralOs_You_2022,NeuralOs_Zhang_2022,NeuralOs_Kissas_2022}. These emulators can learn the physical system at the operator level, and they are then effective in predicting the system responses under different initial/boundary conditions, forcing terms, or PDE parameters. Among the variety of techniques that have been recently proposed, the Deep Operator Network (DeepONet) by Lu \textit{et al.}~\cite{DeepONet_Lu_2019,DeepONet_Lu_2021} has become one of the methods of choice primarily because of the low generalization errors, simplicity of implementation, fast learning with respect to the training data, and high flexibility. In fact, one of its main strengths is the capability of being trained by multifidelity data or by heterogeneous sources of experimental data and simulations~\cite{MFidDeepONet_De_2022,MFidDeepONet_Howard_2022, MFidDeepONet_Lu_2022}. Additionally, DeepONet’s building blocks are not constrained to a particular architecture, and they can be implemented in multiple ways (e.g., as feed-forward neural networks (FNNs) or convolutional neural networks (CNNs)). Since it was introduced in 2019, DeepONet has been successfully applied to operator surrogation in various fields, including fluid and gas dynamics~\cite{DeepONetApps_Lin_1_2021,DeepONetApps_Lin_2_2021,DeepONetApps_Cai_2021,DeepONetApps_DiLeoni_2021}, combustion~\cite{DeepONetApps_Gitushi_2022}, hypersonics~\cite{DeepONetApps_Mao_2021,DeepONetApps_Sharma_2021,DeepONetApps_Zanardi_2022}, energy conversion~\cite{DeepONetApps_Osorio_2022}, material science~\cite{DeepONetApps_Goswami_2022, ArchiDeepONet_Oommen_2022}, medicine~\cite{DeepONetApps_Yin_2022}, seismology~\cite{DeepONetApps_Liu_2021}, and finance~\cite{DeepONetApps_Leite_2021,DeepONetApps_Remlinger_2022}. Physics-informed extensions (PI-DeepONets) have also been proposed to enforce known physical constraints~\cite{PIDeepONet_Wang2021_1,PIDeepONet_Wang2021_2,PIDeepONet_Wang2021_3,DeepONetApps_Goswami_2022}. \\
DeepONet's approach to surrogate construction relies on uncovering a projection that approximates the operator with good accuracy. The bases of this projection are assumed to be non-linear functions of the operator's independent variables (e.g., time and spatial coordinates), while the coefficients are constructed as non-linear functions of the operator input (e.g., initial conditions). DeepONet autonomously discovers the appropriate projection from the data at the training phase by learning the parameters of two separate networks, called branch and trunk nets, which approximate the coefficients-to-input and basis-to-independent-variables mappings. \\
As the projection is the keystone of DeepONet, this paper aims to draw analogies with other techniques that detect suitable bases for representing data and/or describing dynamics. In particular, we focus on linking the DeepONet to methodologies derived from the singular value decomposition (SVD)~\cite{SVD_Beltrami_1873,SVD_Jordan_1875,SVD_Stewart_2001}, such as the proper orthogonal decomposition (POD)~\cite{POD_Sirovich_1987,POD_Berkooz_1993,POD_Rathinam_2004,POD_Hinze_2005}. While ongoing work is focused on formalizing and contextualizing these parallelisms to the standpoint of PDEs, similarities and contrasts are highlighted in this paper mainly through the investigation of test cases built upon ordinary differential equations (ODEs). It is worth noticing that PDEs reduce to coupled ODEs in the method of lines discretization used in most numerical treatments of differential equations in continuum mechanics. We choose a mass-spring-damper model as the first problem to address the DeepONet components' interpretability. 
A novel SVD-DeepONet is constructed and compared with the recently developed POD-DeepONet approach by Lu~\textit{et al.}~\cite{PODDeepONet_Lu_2021,DeepONetApps_Kontolati_2022}. The latter relies on performing POD on the training data, employing the resulting modes as the trunk net, and using the branch net to learn the decomposition coefficients while training the DeepONet. As will be discussed in detail, the SVD-DeepONet presented in this work relies on three main modifications of the POD-DeepONet approach, the main being that SVD-DeepONet’s training of trunks and branches are embarrassingly parallelizable, as they happen fully independently. In fact, in analogy with the offline stage of the POD-neural network (POD-NN) method by Hesthaven \textit{et al.}~\cite{NNPOD_Hesthaven_2018, NNPOD_Wang_2019}, SVD-DeepONet computes the $\ell_2$-optimal projection of the scenario-aggregated training data via SVD, and it trains fully-separated FNNs for learning the resulting bases and coefficients. These blocks are then coupled, and they are respectively used as DeepONet's trunks and branches in the prediction phase.
Despite the instructive attribute of the SVD-formulation and its practical benefits for some fully data-driven applications, the analysis also allows us to stress the numerous advantages of the original DeepONet~\cite{DeepONet_Lu_2019,DeepONet_Lu_2021} for real-world problems. \\
The second test case that we study is a toy problem in which the evolution of a dynamical state is described as a hyperbolic tangent shifted in time as a function of an initial condition. While lacking an explicit physical analog, the system calls attention to the limitation of DeepONet in describing dynamics characterized by translations. We conjecture that this shortcoming is traceable to the DeepONet being a linear projection-based method and, as such, being inefficient in processing symmetries that include translations, rotations, and stretchings~\cite{SVD_Brunton_2019}. However, rather than fully revolutionizing the architecture and renouncing the advantages of a linear subspace for the complexity of non-linear manifolds, we propose a simple but effective modification. Similar to what was independently developed by Hadorn~\cite{SymmPOD_Hadorn_2022}, we introduce an additional building block, which we call a pre-transformation network. This aims to discover a moving reference frame with respect to which fewer modes can efficiently summarize the multiple scenarios. We refer to this augmented structure as a flexible DeepONet (flexDeepONet). We also discuss similarities with the shifted-POD methods~\cite{SymmPOD_Reiss_2018,SymmPOD_Reiss_2_2018,SymmPOD_Papapicco_2022} developed for overcoming the issue arising from symmetries in transport-dominated flows.\\
This toy problem is propaedeutic to the third test case, a combustion chemistry model in an idealized homogeneous reactor with a zero-spatial variation assumption, such that gradient-based mass transport effects are ignored. As for the previous system, the dynamics are also characterized by symmetries that cause noticeable oscillations in the vanilla DeepONet architecture's predictions and prevent it from achieving high precision. As formerly adopted by Lu \textit{et al.}~\cite{PODDeepONet_Lu_2021}, here and in what follows, the adjective ``vanilla'' refers to the unmodified version of the DeepONet architecture as firstly introduced in~\cite{DeepONet_Lu_2019,DeepONet_Lu_2021}. This regression task for the chemical system time integration is also complicated by the large number of thermodynamic variables involved and by their different orders of magnitude.\\
While the first three test cases aim to highlight that various ideas previously developed for SVD-based methods can further improve DeepONet's flexibility and generalization capabilities, the fourth problem we present is to show that DeepONet's architecture and training paradigm can also benefit the generic projection-based methods. In fact, the last test case studies the dynamics of a rigid body that rotates, translates, and stretches with time. Due to the complex rigid motions involved, classical POD analysis would require a large number of POD modes (more than 150) to characterize the dynamics precisely. Despite that, flex-DeepONet is found to be an efficient and effective surrogate capable of accurate predictions even at spatial locations outside the training domain.\\
The paper is organized as follows. Sec.~\ref{sec:Methods} summarizes the DeepONet approach and the SVD-derived methodologies. In Sec.~\ref{sec:Results}, we apply the DeepONet to the four test cases: i) a mass-spring-damper system, ii) a shifting hyperbolic tangent function, iii) a chemical system in an isobaric reactor undergoing combustion processes, and iv) a two-dimensional rigid body that translates, rotates, and stretches in time. We analyze the drawbacks of the vanilla architecture and propose improvements in light of analogies with SVD-based methods. Finally, Sec.~\ref{sec:Conclusions} summarizes the findings and outlines future directions.


\section{Methods}\label{sec:Methods}

This section introduces the singular value decomposition (SVD) and the Deep Operator Network (DeepONet) approach to operator regression along with the techniques derived from it: POD-DeepONet by Lu \textit{et al.}~\cite{PODDeepONet_Lu_2021}, and SVD-DeepONet and flexDeepONet, both presented in this paper for the first time. By analyzing four test cases, Sec.~\ref{sec:Results} will link the SVD and DeepONet families of methodologies and motivate the proposed extensions in light of their connections. \\

\subsection{Singular Value Decomposition (SVD)}\label{subsec:SVD}

The singular value decomposition (SVD)~\cite{SVD_Beltrami_1873,SVD_Jordan_1875,SVD_Stewart_2001} is a technique that extracts low-rank patterns from high-dimensional data without any knowledge or constraints arising from the underlying physics. The main strengths of SVD compared to other decompositions are i) its numerical robustness and efficiency, ii) the fact that it provides a hierarchical representation of the data, and iii) its existence for any matrix, including those that are non-square. SVD is the underlying algorithm of many ubiquitous analysis methods in science and engineering. Most of them have been independently proposed for dimensionality reduction, and they mainly differ in the way they pre-process the data~\cite{SVD_Brunton_2019}. Some examples include principal component analysis (PCA)~\cite{PCA_Pearson_1901,PCA_Hotelling_1933,PCA_Jolliffe_2011,PCA_Jolliffe_2016}, proper orthogonal decomposition (POD)~\cite{POD_Sirovich_1987,POD_Berkooz_1993,POD_Rathinam_2004,POD_Hinze_2005}, the Karhunen-Loeve transform~\cite{KLE_Karhunen_1947,KLE_Loeve_1963}, singular value expansion (SVE)~\cite{SVE_Hansen_1988}, and dynamic mode decomposition (DMD)~\cite{DMD_Schmid_2008,DMD_Schmid_2010,DMD_Brunton_2019}. In the following, we will briefly outline the SVD approach, and details about the methodology can be found in~\cite{SVD_Lloyd_1999,SVD_Brunton_2019}.\\
For a given data matrix $\mathbf{X} \in \mathbb{C}^{n \times m}$, there exists a unique decomposition:
\begin{equation}
    \mathbf{X} = \mathbf{U} \mathbf{\Sigma} \mathbf{V}^{\ast},
\end{equation}
where $\mathbf{U} \in \mathbb{C}^{n \times n}$ and $\mathbf{V} \in \mathbb{C}^{m \times m}$ are unitary matrices. $\mathbf{\Sigma} \in \mathbb{C}^{n \times m}$ is a real non-negative diagonal matrix, and its non-zero values (i.e., $\sigma_i=\Sigma_{i,i}$) are ranked in descending order (i.e., $\sigma_1 \ge \sigma_i \ge \sigma_m$). The columns of $\mathbf{X}$ are typically referred to as snapshots, those of $\mathbf{U}$ and $\mathbf{V}$ are named left and right singular vectors respectively, and $\sigma_i$ are referred to as singular values. If we now truncate $\mathbf{\Sigma}$, $\mathbf{U}$, and $\mathbf{V}$ by retaining only the largest $r$ singular values and their associated vectors, it can be proved that the resulting rank-$r$ approximation:
\begin{equation}\label{eq:low_rank}
    \mathbf{X} \approx \mathbf{\tilde{U}} \mathbf{\tilde{\Sigma}} \mathbf{\tilde{V}}^{\ast}
\end{equation}
is optimal in the $\ell_2$ sense, where $\mathbf{\tilde{U}}$, $\mathbf{\tilde{\Sigma}}$, and $\mathbf{\tilde{V}}$ represent the truncated matrices. POD-based methods rely on the columns of $\mathbf{\tilde{U}}$ as optimal modes, and employs them as a low-rank, orthogonal basis to represent the dynamics~\cite{SVD_Brunton_2019}.\\
If we assume that $\mathbf{X}$ is centered (i.e., the mean values of its columns are equal to zero), we can construct its $m \times m$ column-wise covariance matrix as:
\begin{equation}\label{eq:diag}
    \mathbf{C} = \dfrac{\mathbf{X}^{\ast} \mathbf{X}}{n-1} = \mathbf{V} \dfrac{\mathbf{\Sigma}^2}{n-1} \mathbf{V}^{\ast}.
\end{equation}
Recalling that $\mathbf{V}$ is unitary, Eq.~\ref{eq:diag} corresponds to the diagonalization of $\mathbf{C}$ (i.e., $\mathbf{C} = \mathbf{V} \mathbf{\Lambda} \mathbf{V}^{\ast}$, with $\mathbf{\Lambda}= \mathbf{\Sigma}^2/(n-1)$). Following the PCA nomenclature, the columns of $\mathbf{V}$ are also called principal directions (or principal axes), as they are the eigenvectors corresponding to the column-wise covariance matrix. Moreover, because they are organized in descending order based on $\lambda_i=\sigma_i^2/(n-1)$), these columns capture decreasing contributions to the covariance with increasing $i$. The projection of $\mathbf{X}$ in the principal directions gives the principal components (or scores):
\begin{equation}\label{eq:pca_1}
    \mathbf{U} \mathbf{\Sigma} = \mathbf{X} \mathbf{V}.
\end{equation}
From this perspective, the approximation of $\mathbf{X}$ in Eq.~\ref{eq:low_rank} can be rewritten as:
\begin{equation}\label{eq:pca_2}
    \mathbf{X} \approx \mathbf{\Phi}_{x} \mathbf{A}_{x}^{\ast} = \sum_{i=1}^{N_{\phi}} \boldsymbol{\phi_{x}}_i \boldsymbol{\alpha}_{{x}_i}^{\ast}
\end{equation}
where $N_{\phi}=r$ is the number of retained singular values, $\mathbf{\Phi}_{x} = \tilde{\mathbf{U}} \tilde{\mathbf{\Sigma}}$ represents the reduced principal components and has dimensions $[n \times N_{\phi}]$, while $\mathbf{A}_{x}=\tilde{\mathbf{V}}$ corresponds to the reduced principal directions and has dimensions $[m \times N_{\phi}]$. \\
For some of the SVD-derived methods to work properly and so that the decomposition can focus only on the relevant variations, the snapshot matrix needs to be first centered by representing the components of each column as deviations about their mean values. Additional common practice is to scale the centered snapshot matrix for (de-)emphasizing correlations among different groups of columns, especially if these have different units and vary over different scales. For example, auto-scaling produces snapshots with unitary standard deviations in order for the SVD to analyze the data based on correlations instead of covariances. In contrast, variable stability (VAST) scaling reserves higher importance to columns that do not show strong variation. Descriptions of the centering and scaling techniques and their effects on data reduction can be found in \cite{Scaling_VanDenBerg_2006,Scaling_Parente_2013,Scaling_Armstrong_2021,Scaling_Zdybal_2020}.
To include centering and scaling into our analysis, we allow for a generic pre-processing step to transform the raw-data snapshot matrix, $\mathbf{X}_{Raw}$, before SVD is performed:
\begin{equation}\label{eq:pca_3}
    \boldsymbol{x}_i = \dfrac{\boldsymbol{x}_{{Raw}_i} - c_{x_i}}{d_{x_i}},
\end{equation}
where  $\boldsymbol{x}_{{Raw}_i}$ and $\boldsymbol{x}_i$ respectively represent the $i$-th columns of the raw-data and pre-processed snapshot matrix. Additionally, $c_{x_i}$ and $d_{x_i}$ respectively indicate the components of the $\boldsymbol{c}_x$ and $\boldsymbol{d}_x$ centering and scaling vectors, both with dimensions $[n \times 1]$.

\subsection{Deep Operator Network (DeepONet)}\label{subsec:DeepONets}

\begin{figure}[!tb]
    \centering
    \includegraphics[width=6.0in]{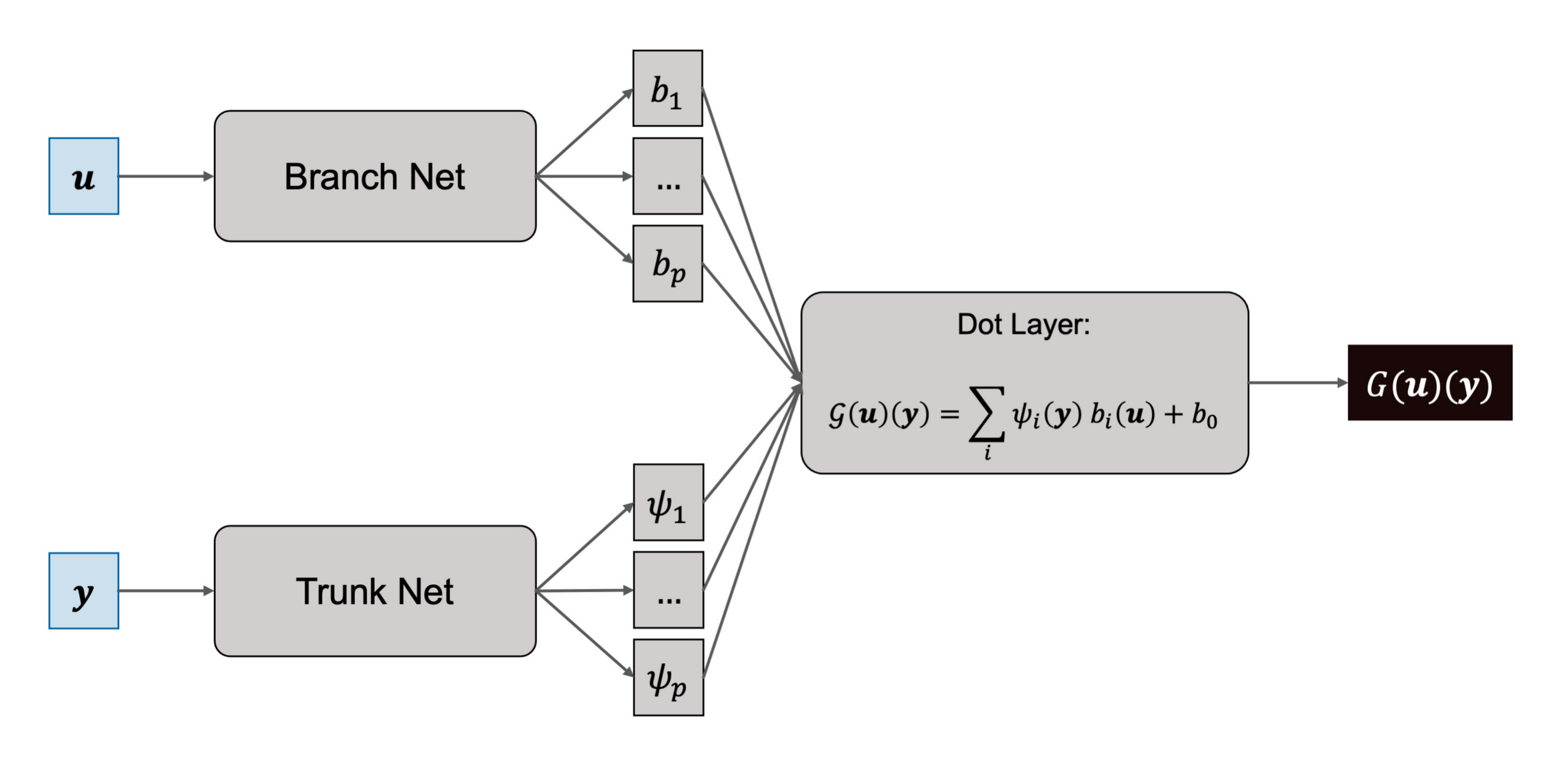}
    \caption{\textbf{Architecture of the (unstacked) vanilla DeepONet}~\cite{DeepONet_Lu_2019,DeepONet_Lu_2021}.}
    \label{fig:DeepONet0}
\end{figure}
As in all the neural operator approaches, a deep operator network (DeepONet)~\cite{DeepONet_Lu_2019,DeepONet_Lu_2021} aims to learn an operator, $\mathcal{G}$, that takes an input function $\boldsymbol{u}$ (e.g., initial conditions, boundary conditions, forcing terms) and gives an output function $\mathcal{G}(\boldsymbol{u})$. When this last quantity is evaluated at any point in the space of the independent variables, $\boldsymbol{y}$ (e.g., spatial coordinates and/or time instants), the result is a real number, $\mathcal{G}(\boldsymbol{u})(\boldsymbol{y})$. In the following, we will refer to the values of $\boldsymbol{y}$ as time instants or locations, and the dynamics corresponding to different $\boldsymbol{u}$ as scenarios. Therefore, a scenario is a time-evolving solution corresponding to any particular initial value of $\boldsymbol{u}$.\\
The DeepONet architecture has been heavily inspired by the universal approximation theorem for operators by Chen and Chen~\cite{UATs_Chen_1995}, which states that
a neural network (NN) with a single hidden layer can accurately approximate any non-linear continuous functional and operator. The theorem has been later extended to deep NNs~\cite{DeepONet_Lu_2021}. Of the four marginally different DeepONet structures introduced by the original paper~\cite{DeepONet_Lu_2021}, we consider ``vanilla'' the architecture corresponding to an unstacked configuration with bias, which is also the one with the lowest generalization errors. The DeepONet is composed of two subnetworks: i) the branch net, that generates the $p$-dimensional vector $\boldsymbol{b}$ by encoding the input functions, $\boldsymbol{u}$, at fixed sensor points, and ii) a trunk net, that produces the $p$-dimensional vector $\boldsymbol{\psi}$ by encoding the independent variables, $\boldsymbol{y}$. In order to predict the operator's value, the latent outputs $\boldsymbol{b}$ and $\boldsymbol{\psi}$ are then merged via a dot product layer, which can also include a trainable bias $b_0$~\cite{DeepONet_Lu_2021}. The resulting architecture is illustrated in Fig.~\ref{fig:DeepONet0}.
The DeepONet approach comes with theoretical guarantees of universal approximation by construction. Additionally, recent works theorized the upper bounds for the approximation error in terms of network size, operator type, and data regularity~\cite{UATs_Lanthaler_2021,UATs_Deng_2021} and showed that DeepONets can approximate the solution operators of elliptic PDEs with exponential accuracy~\cite{UATs_Mercati_2021}.\\
To construct and train DeepONets, as well as all the surrogate architectures introduced in the following, we utilize an in-house software package, ROMNet~\cite{Soft_ROMNet_2022}, a toolbox developed within Sandia National Laboratories built on the Tensorflow library~\cite{Soft_TensorFlow_2015}. In this work, all the building blocks of these architectures are implemented as feed-forward neural networks (FNNs), and all the training is performed by applying the Adam optimizer~\cite{Soft_Adam_2014} and using mean squared error (MSE) as the loss function.


\subsection{POD-DeepONet}\label{subsec:POD-DeepONet}

The POD-DeepONet was developed by Lu \textit{et al.}~\cite{PODDeepONet_Lu_2021,DeepONetApps_Kontolati_2022} as an alteration of DeepONet's training paradigm without requiring any modification to the vanilla architecture. The training approach consists of the following steps:
\begin{enumerate}
    \item SVD is performed on the training data snapshot-matrix to compute modes and coefficients.
    \item A FNN is constructed and trained for representing the modes as function of $\boldsymbol{y}$.
    \item A DeepONet architecture is formed by relying on the just-trained FNN block as a trunk network and by freezing its parameters' values.
    \item The DeepONet branch is finally trained with the data used for SVD.
\end{enumerate}


\subsection{SVD-DeepONet}\label{subsec:SVD-DeepONet}

\begin{figure}[!tb]
    \centering
    \includegraphics[width=6.0in]{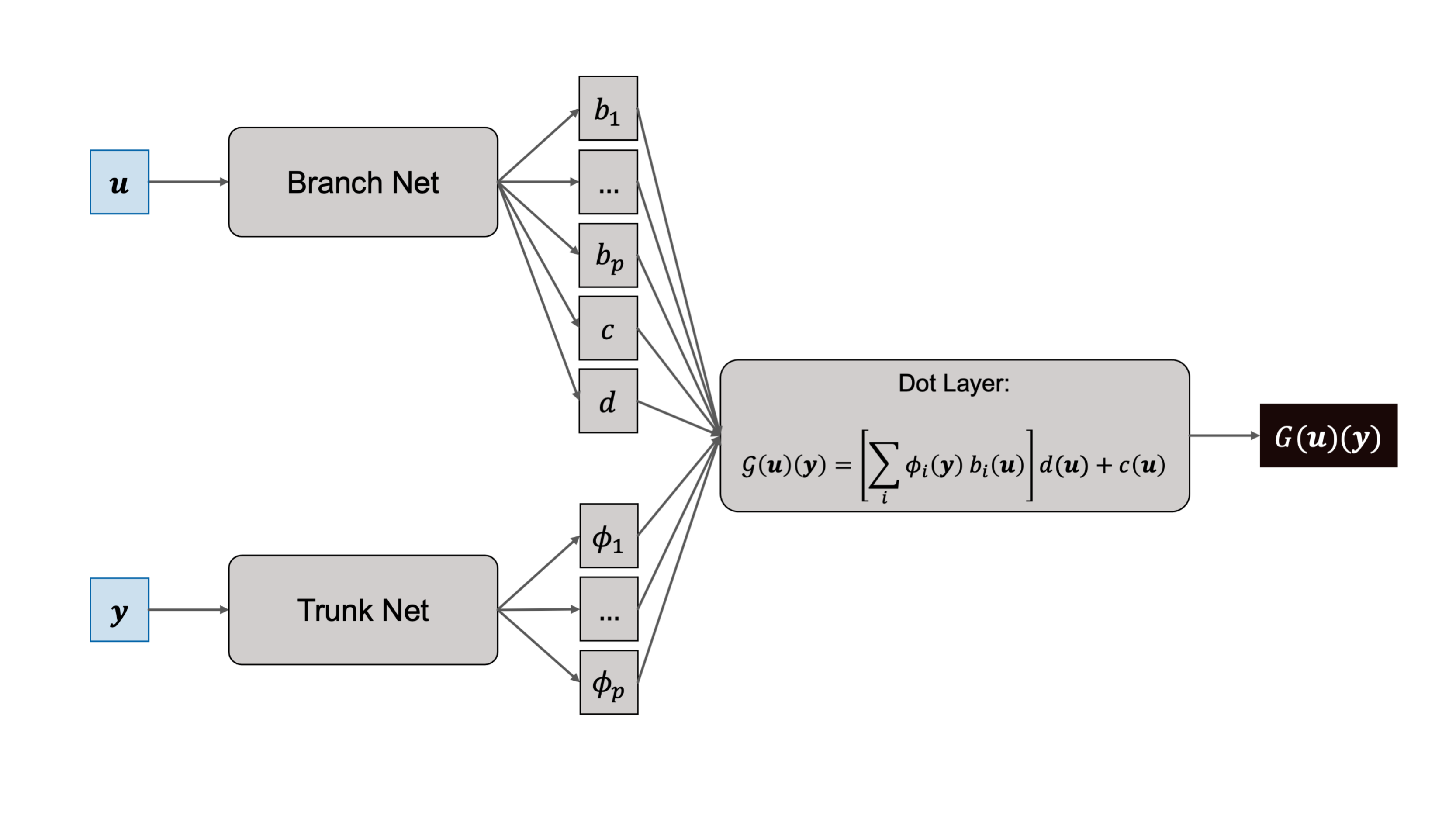}
    \caption{\textbf{Architecture of the newly developed SVD-DeepONet}.}
    \label{fig:SVDDeepONet0}
\end{figure}
The SVD-DeepONet consists of two modifications to the vanilla DeepONet, the motivations of which will be provided in the next section. The alterations can be summarized as follows:
\begin{itemize}

    \item \textit{Adjustment to the vanilla architecture:} as sketched in Fig.~\ref{fig:SVDDeepONet0}, the branch produces two additional outputs, which are used as scenario-specific shifting and stretching coefficients for the surrogate's predictions.
    
    \item \textit{Modification to the surrogate's training paradigm}. This consists of five steps:
    \begin{enumerate}
        \item Each of the columns of the training data snapshot-matrix is centered and scaled.
        \item Similarly to what is done in POD-DeepONet, SVD is performed on the resulting snapshot-matrix to compute modes and coefficients.
        \item Similarly to what is done in POD-DeepONet, a FNN is constructed and trained for representing the modes as function of $\boldsymbol{y}$.
        \item A FNN is constructed and trained for representing the coefficients, the centering coefficient, and the stretching coefficients, all as function of $\boldsymbol{u}$.
        \item A DeepONet architecture is formed by assembling as trunk and branch networks the two FNN blocks produced in the two previous steps. The resulting surrogate does not require further training and is ready for the prediction phase.
    \end{enumerate}

\end{itemize}
Test Case 1 in Sec.~\ref{subsec:TestCase1} will serve as a toy problem for a step-by-step exemplification of SVD-DeepONet's working procedure.


\subsection{Flexible DeepONet (flexDeepONet)}\label{subsec:flexDeepONet}

\begin{figure}
    \centering
    \includegraphics[width=3.2in]{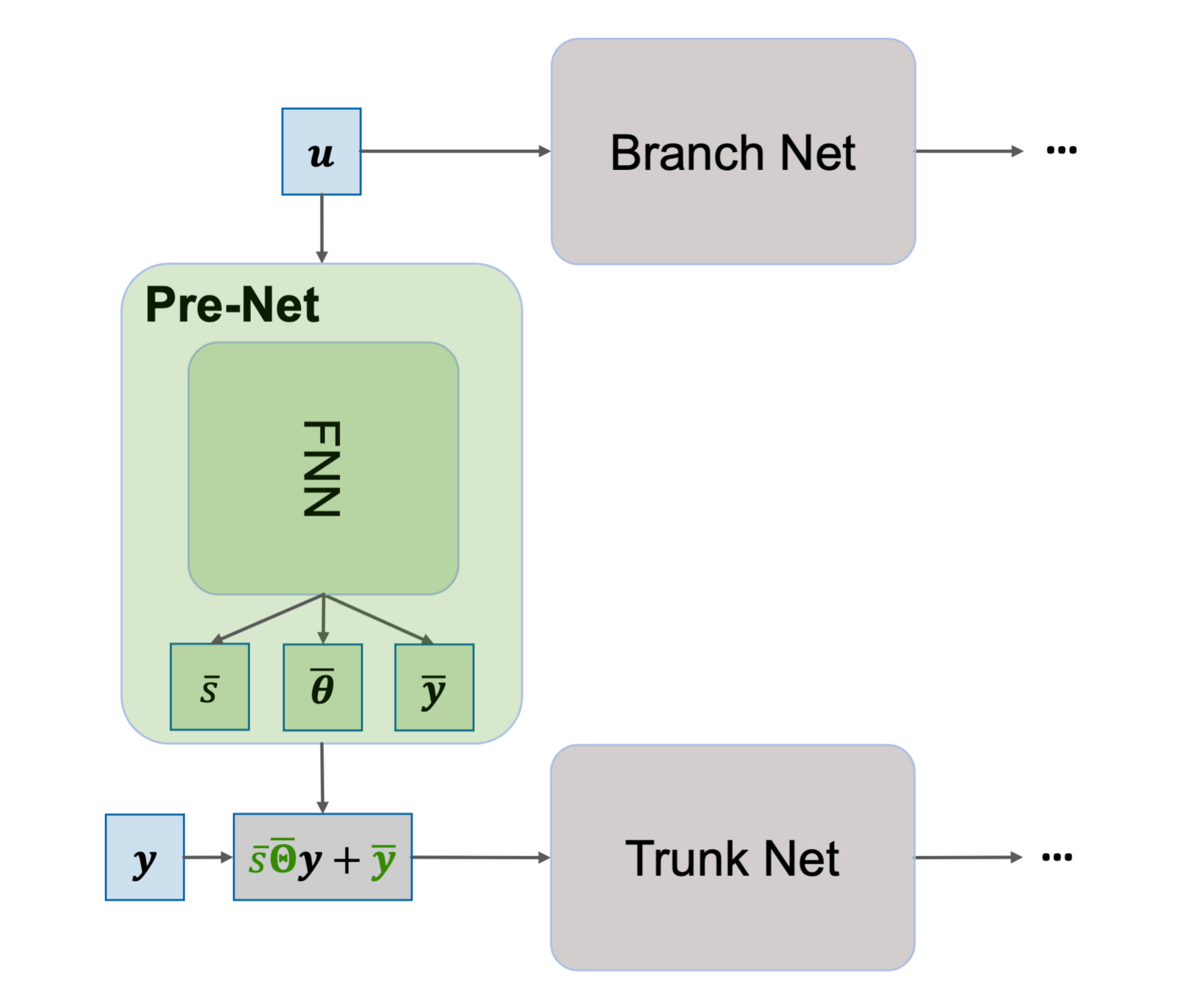}
    \caption{\textbf{FlexDeepONet's pre-transformation network (Pre-Net), as proposed in the present work}. The block implements a transformation that automatically discovers a moving frame of reference with respect to which the multiple scenarios are efficiently compressible in fewer modes. In this way, the Pre-Net extends DeepONet's flexibility with learning operators characterized by dynamics that shift, rotate, and/or scale. Note: in the transformation layer subsequent to the Pre-Net, a rotation matrix $\mathbf{\bar{\Theta}}$ composed of sines and cosines is constructed from $\boldsymbol{\bar{\theta}}$.}
    \label{fig:flexDeepONet}
\end{figure}
Flexible DeepONet (flexDeepONet) consists of two modifications to the vanilla architecture:
\begin{itemize}
    
    \item \textit{First adjustment to the vanilla architecture:} an additional block called pre-transformation network (Pre-Net) is added to the architecture. As sketched in Fig.~\ref{fig:flexDeepONet} for a generic PDE with independent variables $\boldsymbol{y}$, this new component takes the same inputs as the branch nets, $\boldsymbol{u}$, and it outputs a scaling factor, $\bar{s}$, a vector of angles, $\boldsymbol{\bar{\theta}}$, and a vector of shifting coefficients, $\boldsymbol{\bar{y}}$. In a subsequent transformation layer, a rotation matrix $\mathbf{\bar{\Theta}}$ composed of sines and cosines is constructed from $\boldsymbol{\bar{\theta}}$, and the independent variables (i.e., the input of vanilla DeepONet's trunk) are rotated, scaled, and shifted before they enter the trunk net. \\
    As it will be motivated in details later, the goal of this additional block is to eliminate the DeepONet's inefficiencies in representing dynamics characterized by rigid-body components. In fact, the Pre-Net allows the automatic discovery of a moving frame of reference with respect to which the multiple scenarios result in being as overlapping as possible, and they then can be efficiently compressed to a lower number of modes. \\
    While this paper was being written, a very similar approach, called Shift-DeepONet, has been independently proposed by Hadorn~\cite{SymmPOD_Hadorn_2022} to improve the performance of DeepONet applied to problems involving discontinuities. However, there are three main differences between the two techniques: i) flexDeepONet allows for the realignment of data affected by rotations, as it includes the matrix $\mathbf{\bar{\Theta}}$. In this sense, Shift-DeepONet's pre-processing block is a particular instance of a flexDeepONet's Pre-Net characterized by diagonal $\mathbf{\bar{\Theta}}$. ii) FlexDeepONet's Pre-Net has $N_y$ outputs, where $N_y$ represents the dimensionality of $\boldsymbol{y}$. In contrast, Shift-DeepONet's pre-processing block produces $N_y \times p$ outputs, which are all used as trunk's inputs. From this perspective, Shift-DeepONet has more expressivity than flexDeepONet, as each of the trunk's modes is constructed as a non-linear combination of the $N_y \times p$ pre-processing block's outputs. Nevertheless, such versatility might come at the expense of interpretability and generalization in low-data regimes. iii) Shift-DeepONet's branch produces $p$ outputs, while FlexDeepONet's branch has $p+1$ outputs, as motivated in the section.\\
    For reasons of completeness, it should be mentioned that the idea of linking $\boldsymbol{u}$ and $\boldsymbol{y}$ before DeepONet's inner product layer was already proposed by Wang \textit{et al.}~\cite{PIDeepONet_Wang2021_2}. However, the trunk-branch connections those authors devised were multiple (i.e., the outputs of each branch's hidden layers entered the trunk) and bidirectional (i.e., the outputs of each trunk's hidden layers also entered the branch). Their improvement, in fact, was consistent with different reasoning. As Wang \textit{et al.} mentioned, ``(in the vanilla architecture), the final information fusion may be inefficient if the DeepONet input signals fail to propagate through a deep branch network or trunk network at initialization, leading to an ineffective training process and poor model performance''~\cite{PIDeepONet_Wang2021_2}.
    
    \item \textit{Second adjustment to the vanilla architecture:} an additional output is produced by the branch and used as a scenario-specific shifting coefficient for the surrogate's predictions.
    
\end{itemize}
Test Case 2 in Sec.~\ref{subsec:TestCase2} will serve as a toy problem for a step-by-step exemplification of flexDeepONet's working procedure.



\section{Results}\label{sec:Results}

In this section, the DeepONet is applied to four test cases, which have been selected \textit{ad hoc} to highlight the approach's effectiveness, draw connections with SVD-derived methodologies, and propose improvements in light of these.
The first analysis that we carry out involves employing DeepONets to learn the operator characterizing a mass-spring-damper model. This physical system is chosen for the simplicity of the underlying ODEs and the presence of multiple state variables as a prelude to the multi-dimensional state spaces typical of physics applications. The second investigation focuses on a toy problem constructed by representing the time evolution of a single state variable as a hyperbolic tangent affected by a delay proportional to the initial condition. Despite the resulting ODE lacking a direct physical analog, this simple test case highlights and motivates some limitations of the original DeepONet architecture. Moreover, the problem is propaedeutical to the third test case of direct physical relevance, involving combustion chemistry in a zero-dimensional isobaric reactor, characterized by 19 thermodynamic variables. Finally, the fourth problem focuses on a rigid body that rotates, translates, and stretches as non-linear functions of time.

\subsection{Test Case 1: Mass-Spring-Damper System}\label{subsec:TestCase1}

\begin{figure}[!htb]
    \centering
    \includegraphics[width=6.0in]{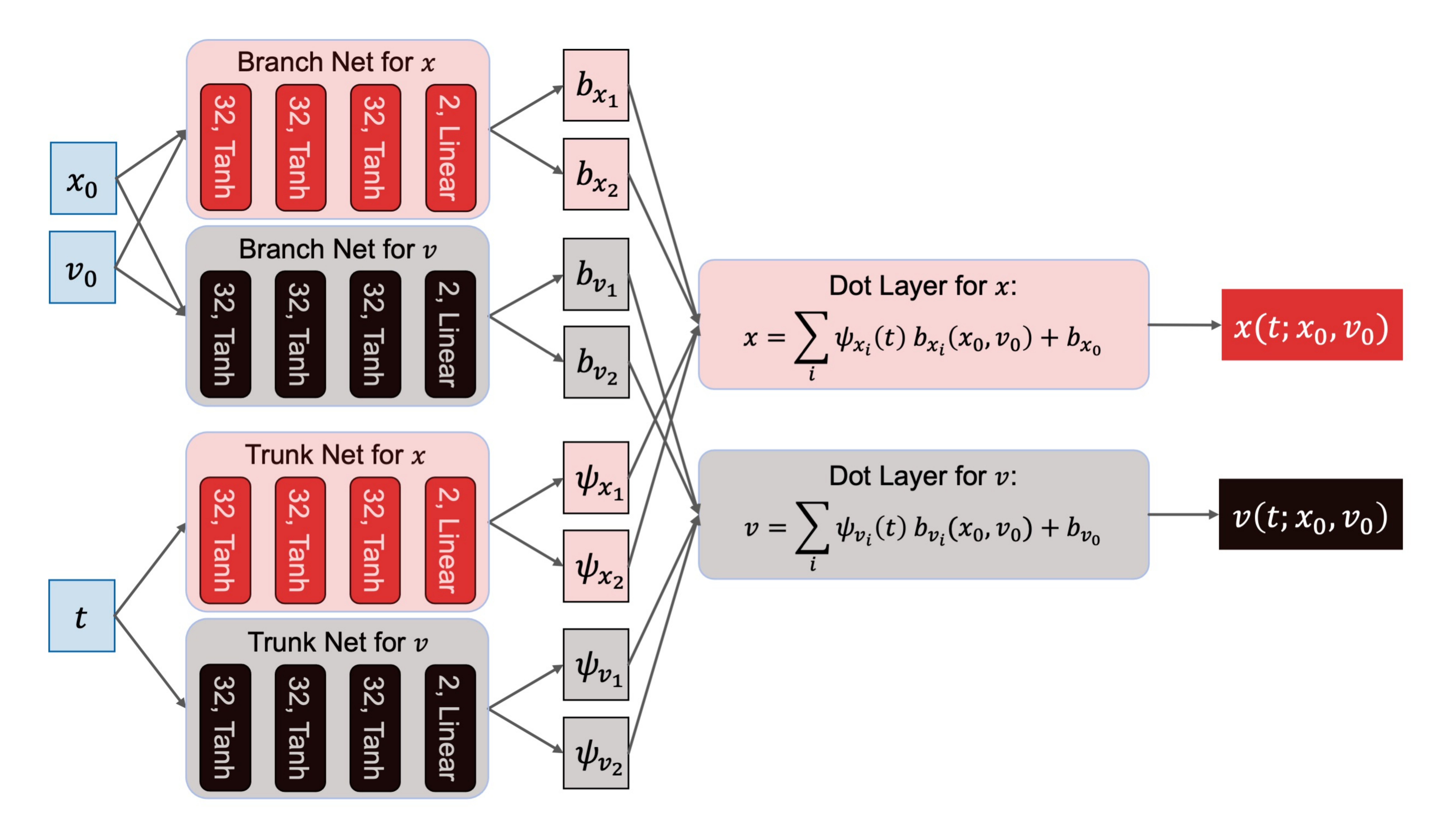}
    \caption{\textbf{Architecture of the vanilla DeepONet for the mass-spring-damper test case}. The same architecture is used at both training and prediction phases.}
    \label{fig:MSD_DeepONet_1}
\end{figure}

\begin{figure}[!htb]
    \centering
    \includegraphics[width=6.0in]{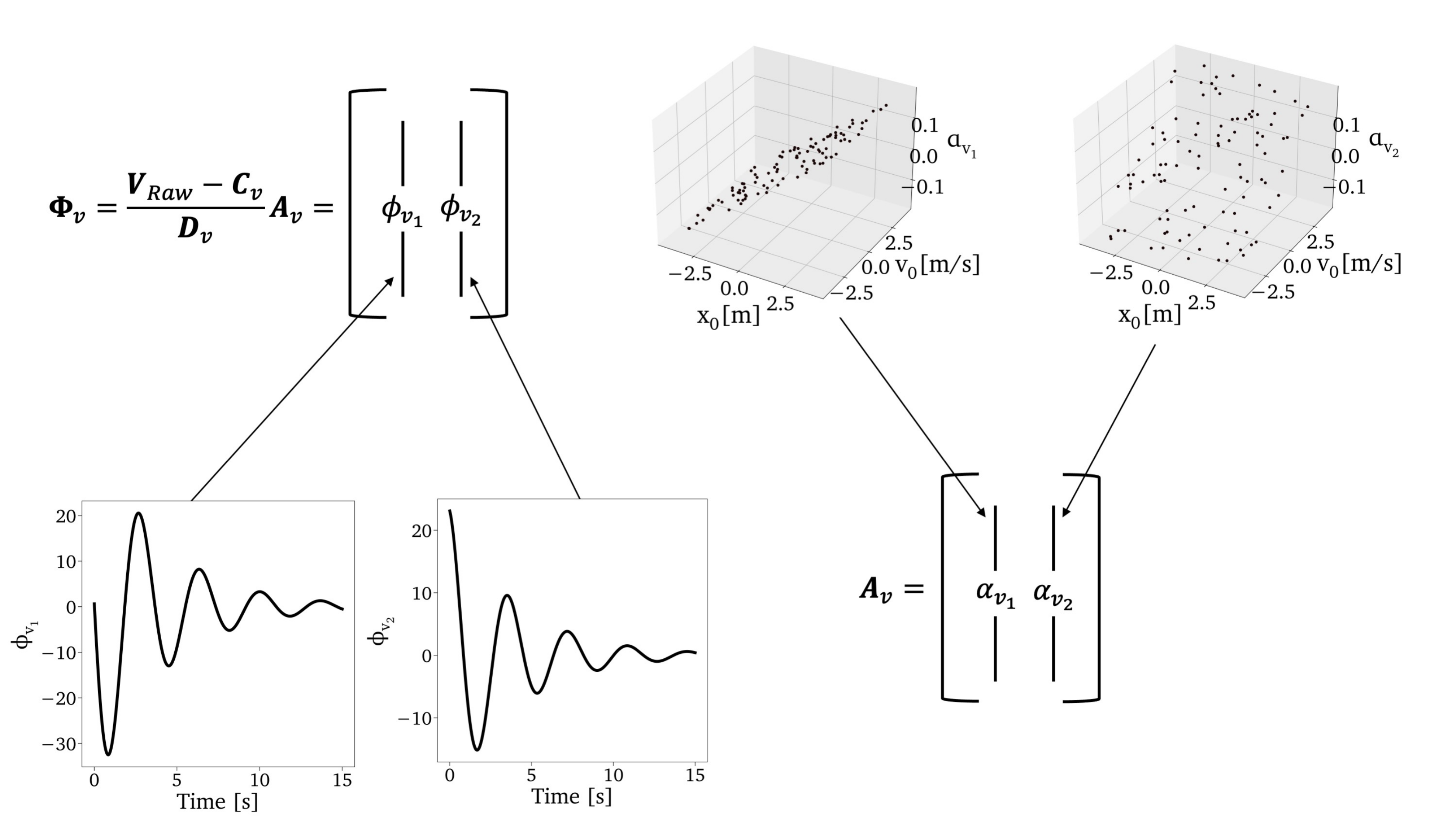}
    \caption{\textbf{SVD of the velocity data matrix from the mass-spring-damper test case.} Schematics of the columns resulting from the decomposition.}
    \label{fig:SVD_DeepONet_Inter_v}
\end{figure}

\begin{figure}[!htb]
    \begin{subfigure}{0.49\textwidth}
        \centering
        \caption{}
        \label{fig:MSD_DeepONet_Inter_x_1}
        \includegraphics[width=3.1in]{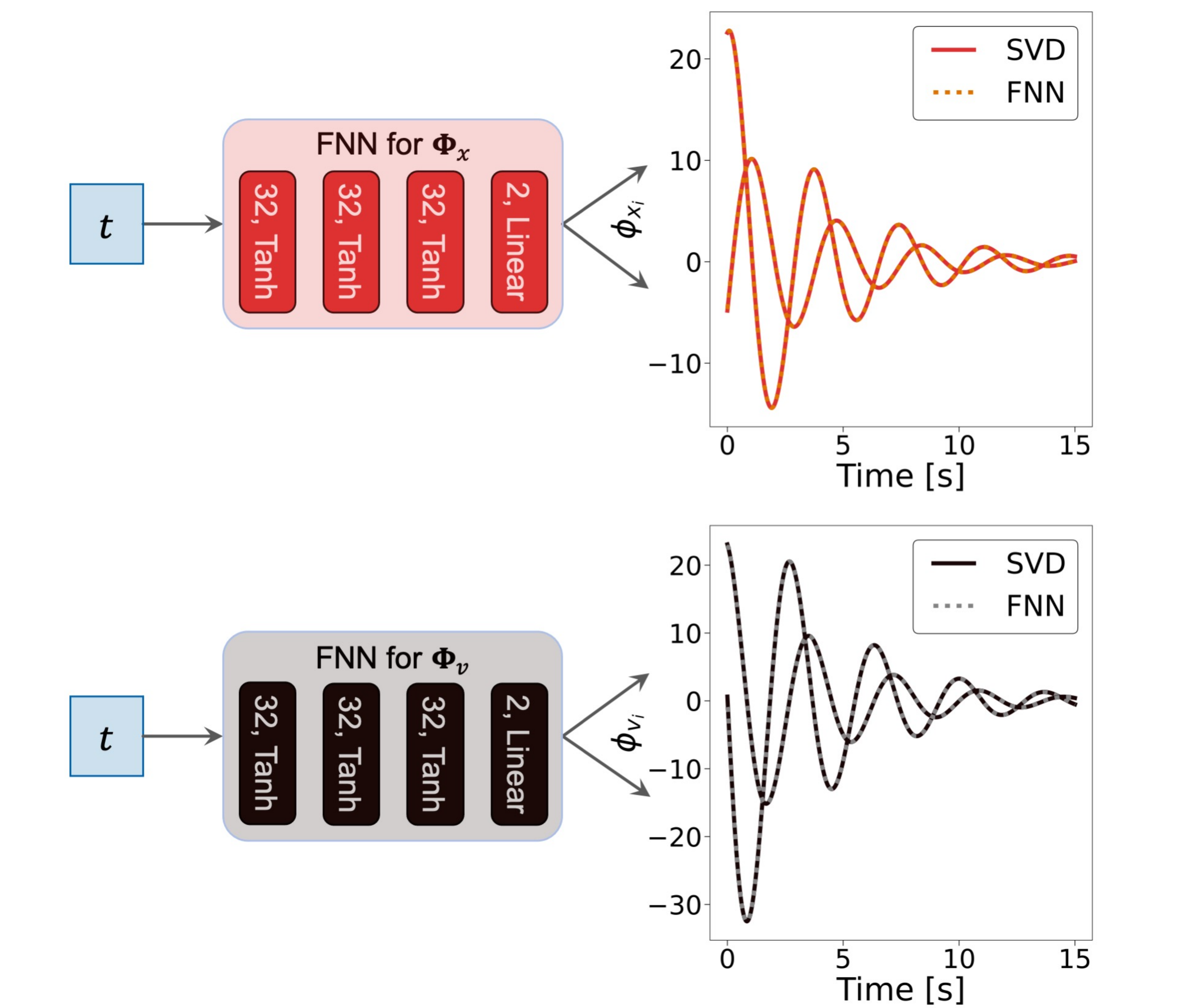}
    \end{subfigure}
    \begin{subfigure}{0.49\textwidth}
        \centering
        \caption{}
        \label{fig:MSD_DeepONet_Inter_x_2}
        \includegraphics[width=3.6in]{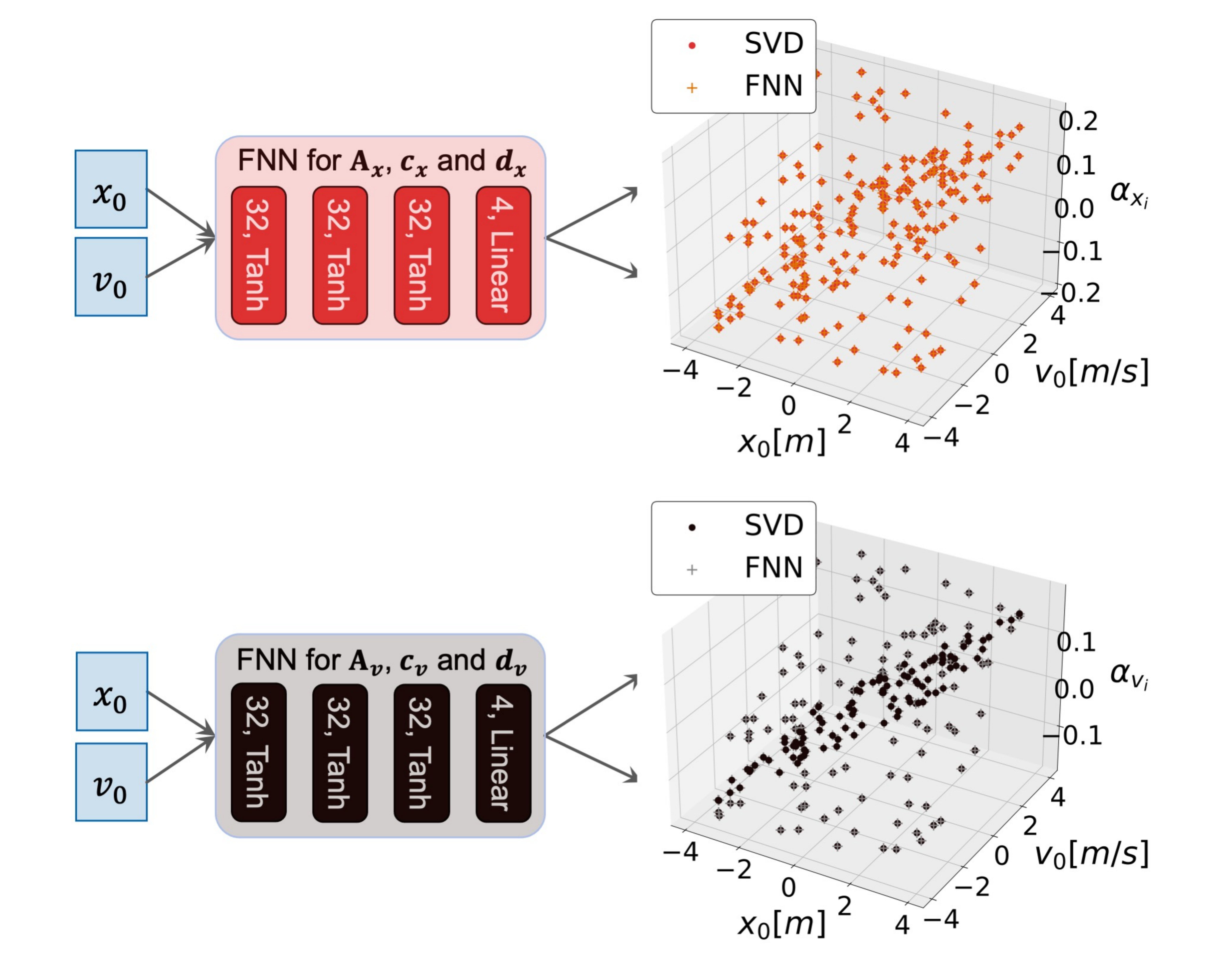}
    \end{subfigure}
    \caption{\textbf{Architectures of the building blocks for the SVD-DeepONet in the mass-spring-damper test case}. (\textbf{A}): Two feed-forward neural networks (FNN) are trained to reproduce the principal components of $\mathbf{X}$ and $\mathbf{V}$, respectively. (\textbf{B}): Two FNNs are independently trained to retrieve the two columns principal directions, the centering coefficient, and the scaling coefficient for each of the two state variables.}
    \label{fig:MSD_DeepONet_Inter_x}
\end{figure}

\begin{figure}[!htb]
    \centering
    \includegraphics[width=6.0in]{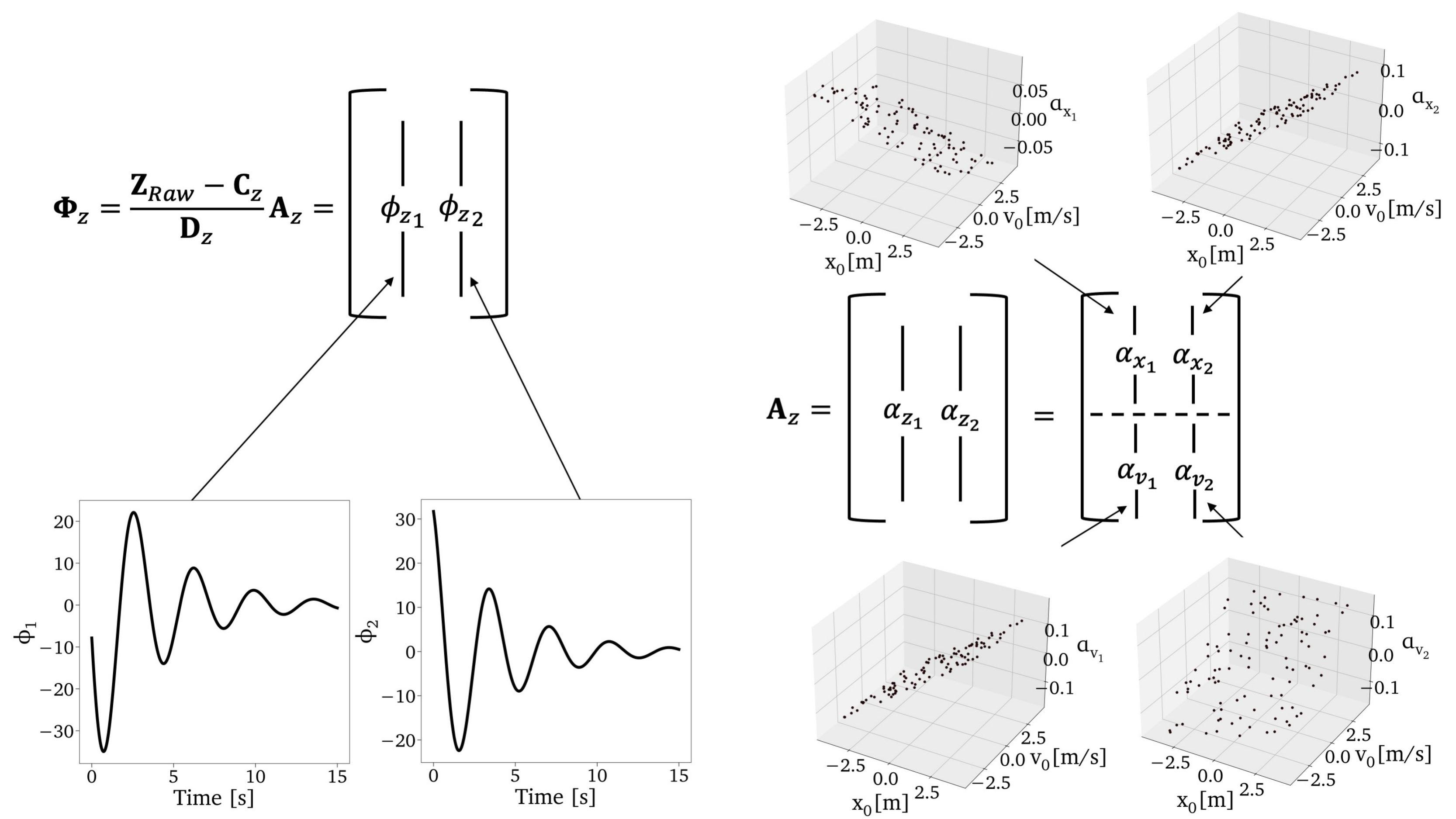}
    \caption{\textbf{SVD of the displacement-velocity combined data matrix,} $\mathbf{Z}$ \textbf{, from the mass-spring-damper test case}. Schematics of the columns resulting from the decomposition.}
    \label{fig:SVD_DeepONet_Inter_All}
\end{figure}

\begin{figure}[!htb]
    \begin{subfigure}{0.49\textwidth}
        \centering
        \caption{}
        \label{fig:MSD_DeepONet_Inter_All_1}
        \includegraphics[width=3.1in]{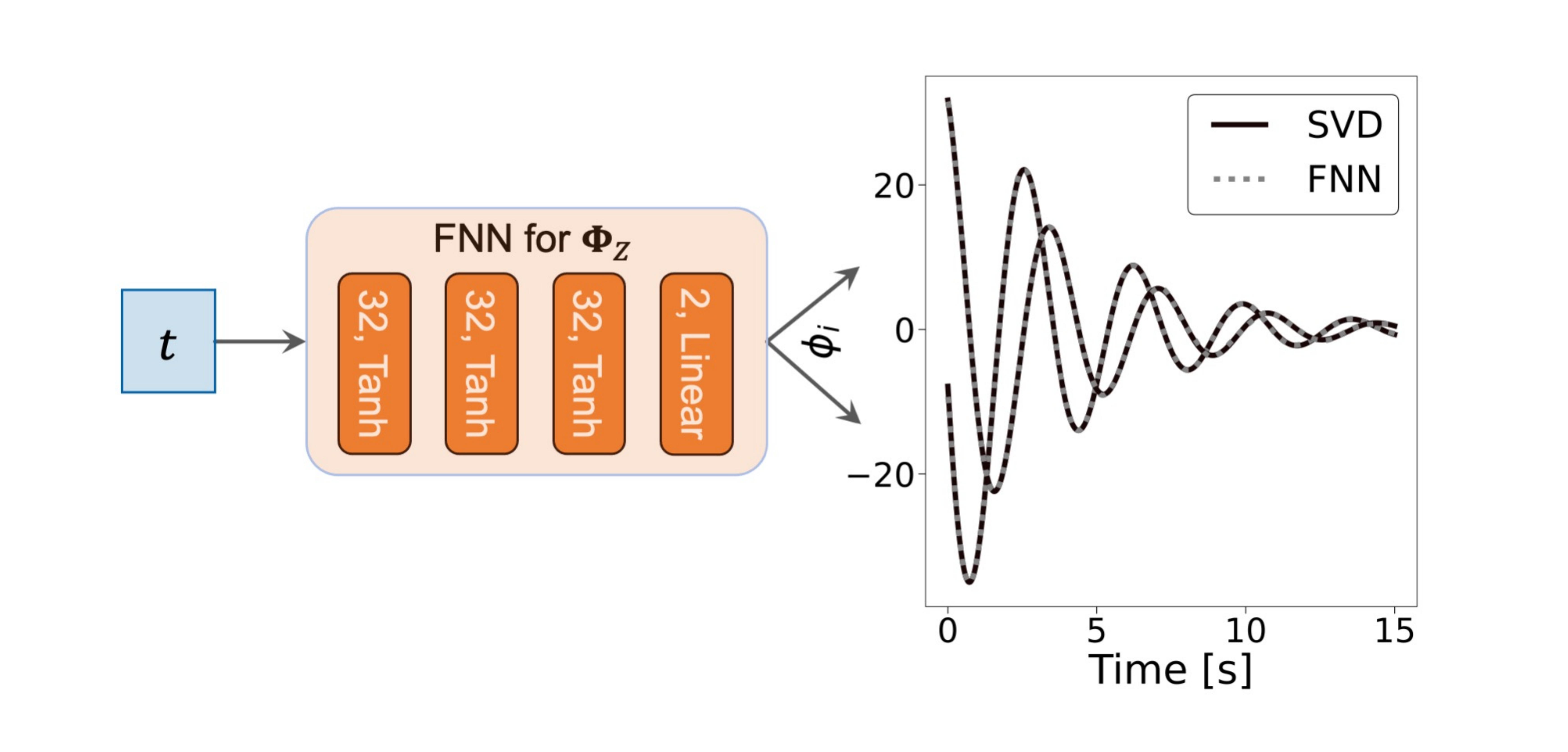}
    \end{subfigure}
    \begin{subfigure}{0.49\textwidth}
        \centering
        \caption{}
        \label{fig:MSD_DeepONet_Inter_All_2}
        \includegraphics[width=3.5in]{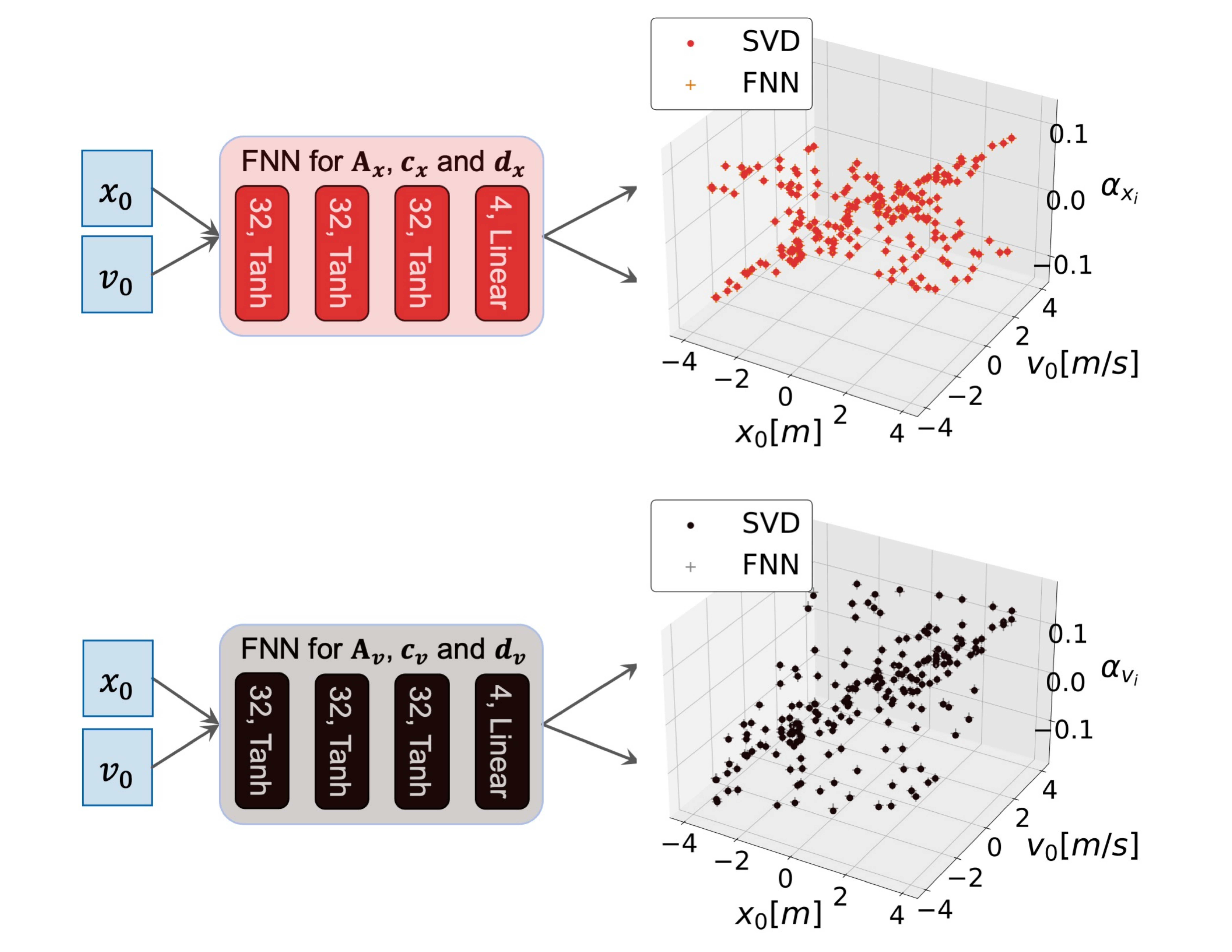}
    \end{subfigure}
    \caption{\textbf{Architectures of the building blocks for the shared-trunk SVD-DeepONet in the mass-spring-damper test case}. (\textbf{A}): One feed-forward neural networks (FNNs) is trained for reproducing the two principal components of $\mathbf{Z}$. (\textbf{B}): Two FNNs are independently trained for retrieving the two principal directions, the centering coefficient, and the scaling coefficient for each of the two state variables.}
    \label{fig:MSD_DeepONet_Inter_All}
\end{figure}

\begin{figure}[!htb]
    \centering
    \includegraphics[width=7.0in]{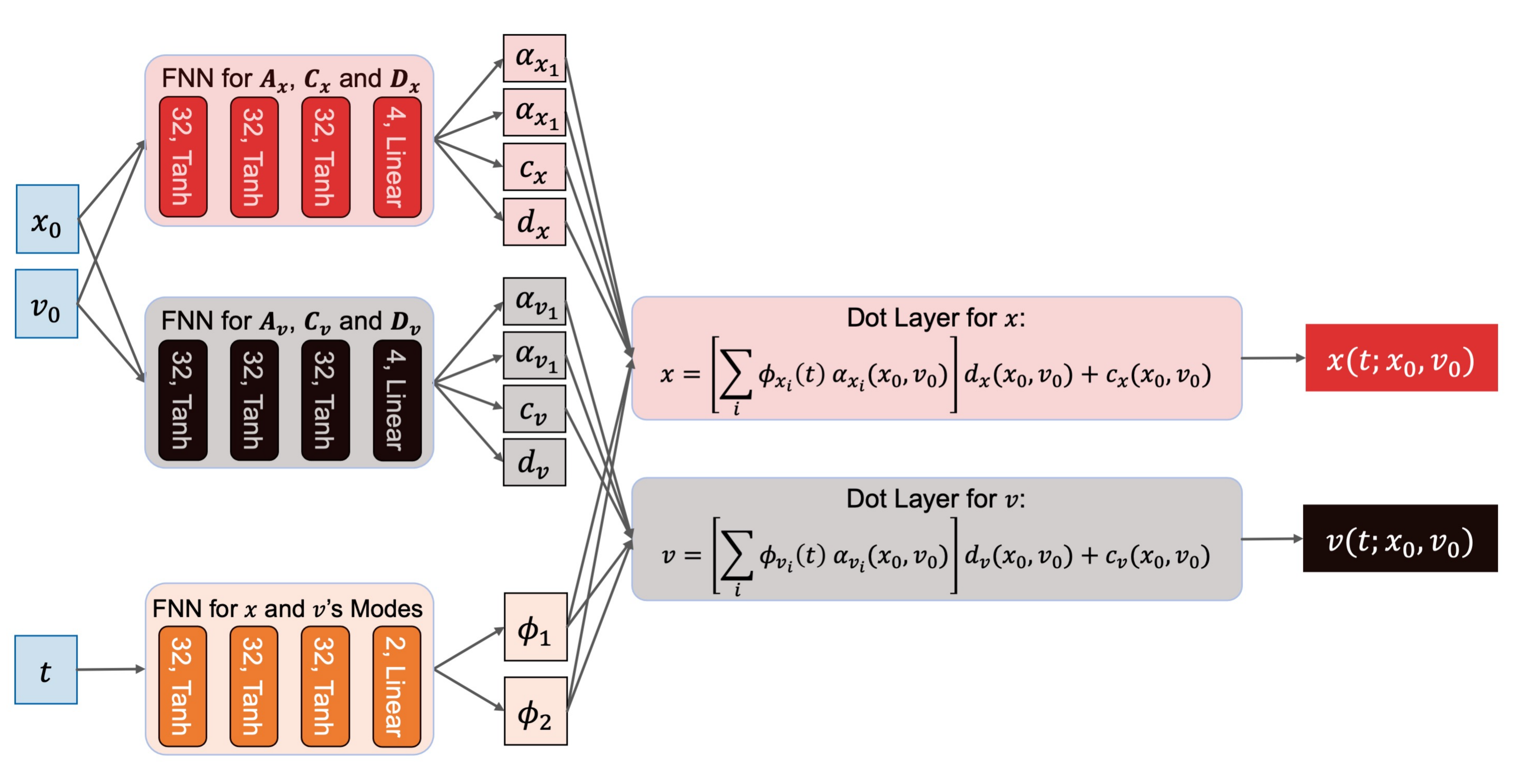}
    \caption{\textbf{Assembled architecture of the shared-trunk SVD-DeepONet for the mass-spring-damper test case}. After being independently trained as feed-forward neural networks (FNNs), the three blocks are assembled as DeepONet's trunk net and branch nets to predict displacements and velocities at unseen times and for unseen initial conditions.}
    \label{fig:MSD_DeepONet_3}
\end{figure}

\begin{figure}[!htb]
    \centering
    \includegraphics[width=3.2in]{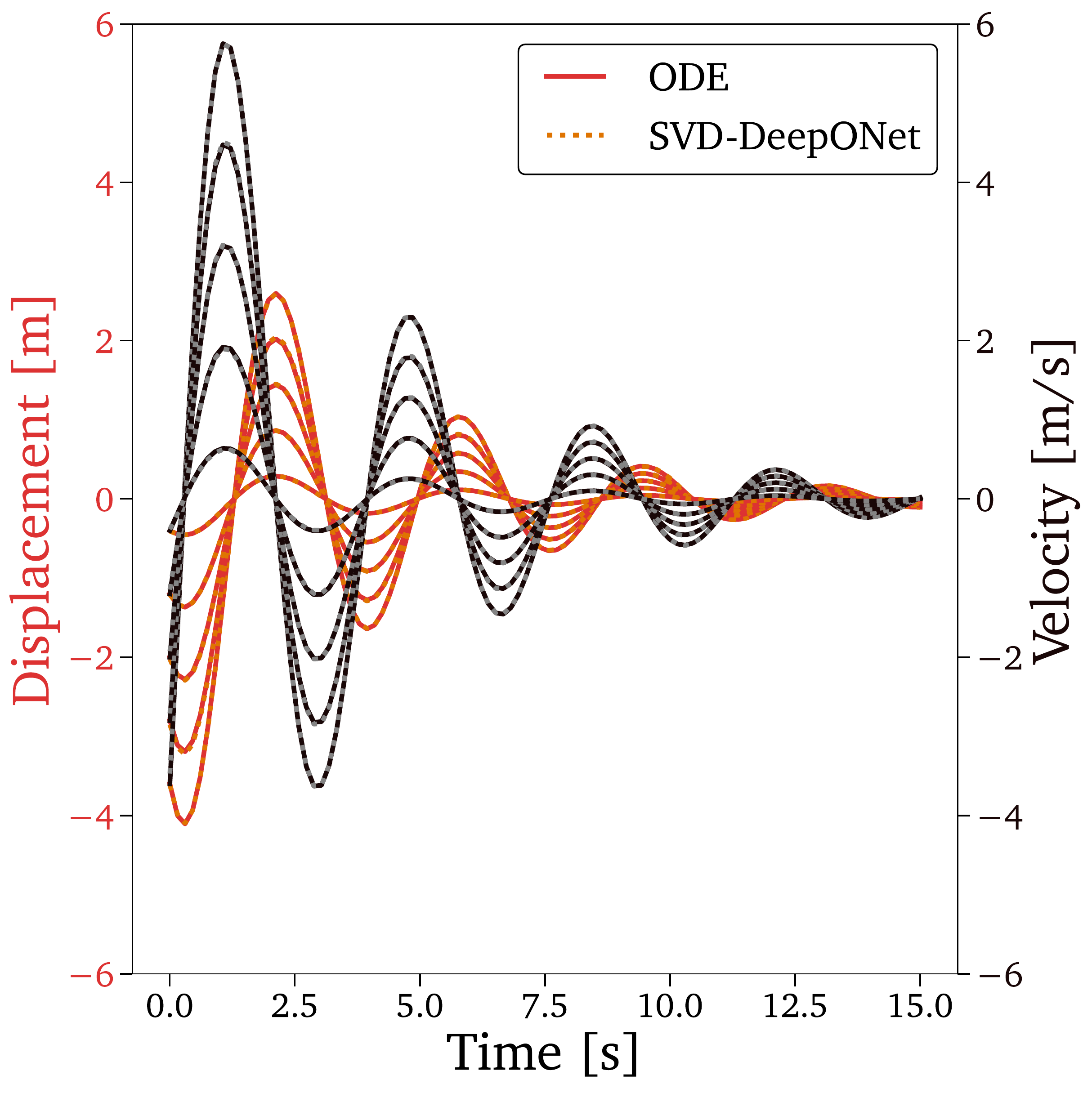}
    \caption{\textbf{SVD-DeepONet applied to the mass-spring-damper test case}. Displacements and velocities at five test scenarios from the ODE integration (solid lines) and as predicted by the shared-trunk SVD-DeepONet (overlapping dotted lines).}
    \label{fig:MSD_SVDDeepONet_Tests_All}
\end{figure}

We start by analyzing a simple mass-spring-damper model under the assumption of no external forcing term acting on the body. The system can be described by the following linear ODE:
\begin{equation}\label{eq1}
    \renewcommand{\arraystretch}{2.0}
    \begin{bmatrix}
        \dfrac{dx(t)}{dt} \\
        \dfrac{dv(t)}{dt}
    \end{bmatrix}
    =
    \left[
        \begin{array}{cc}
        0 & 1 \\
        -\dfrac{k}{m} & -\dfrac{c}{m}
        \end{array}
    \right]
    \begin{bmatrix}
        x \\
        v
    \end{bmatrix},
\end{equation}
with initial conditions:
\begin{equation}\label{eq2}
    \begin{bmatrix}
        x(0) \\
        v(0)
    \end{bmatrix}
    =
    \begin{bmatrix}
        x_0 \\
        v_0
    \end{bmatrix},
\end{equation}
where $x$ and $v$ represent the body's displacement and velocity respectively, while $k$, $c$, and $m$ represent the spring constant, the damping coefficients, and the body's mass. In the following, the values of these three parameters are set to 3 [N/m], 0.5 [N s/m], and 1 [kg], respectively. One hundred pairs of $x_0$ and $v_0$ initial conditions (i.e., $N_S$=100) are randomly sampled in the space $(x_0,v_0) \in (-4, 4)$ [m] $ \times (-4, 4)$ [m/s] based on a Latin hypercube strategy~\cite{LHS_McKay_1979}, and collected in a $[N_S \times 2]$ matrix, $[\boldsymbol{x}_0,\boldsymbol{v}_0]$. For each of these scenarios, the system is integrated in time, and the $x$ and $v$ variables are collected at five hundred time instances (i.e., $N_t$=500) equally spaced between 0 and 15 [s], composing the vector $\boldsymbol{t}_S$.
\begin{figure}
    \centering
    \begin{subfigure}{0.49\textwidth}
        \caption{}
        \includegraphics[width=3.1in]{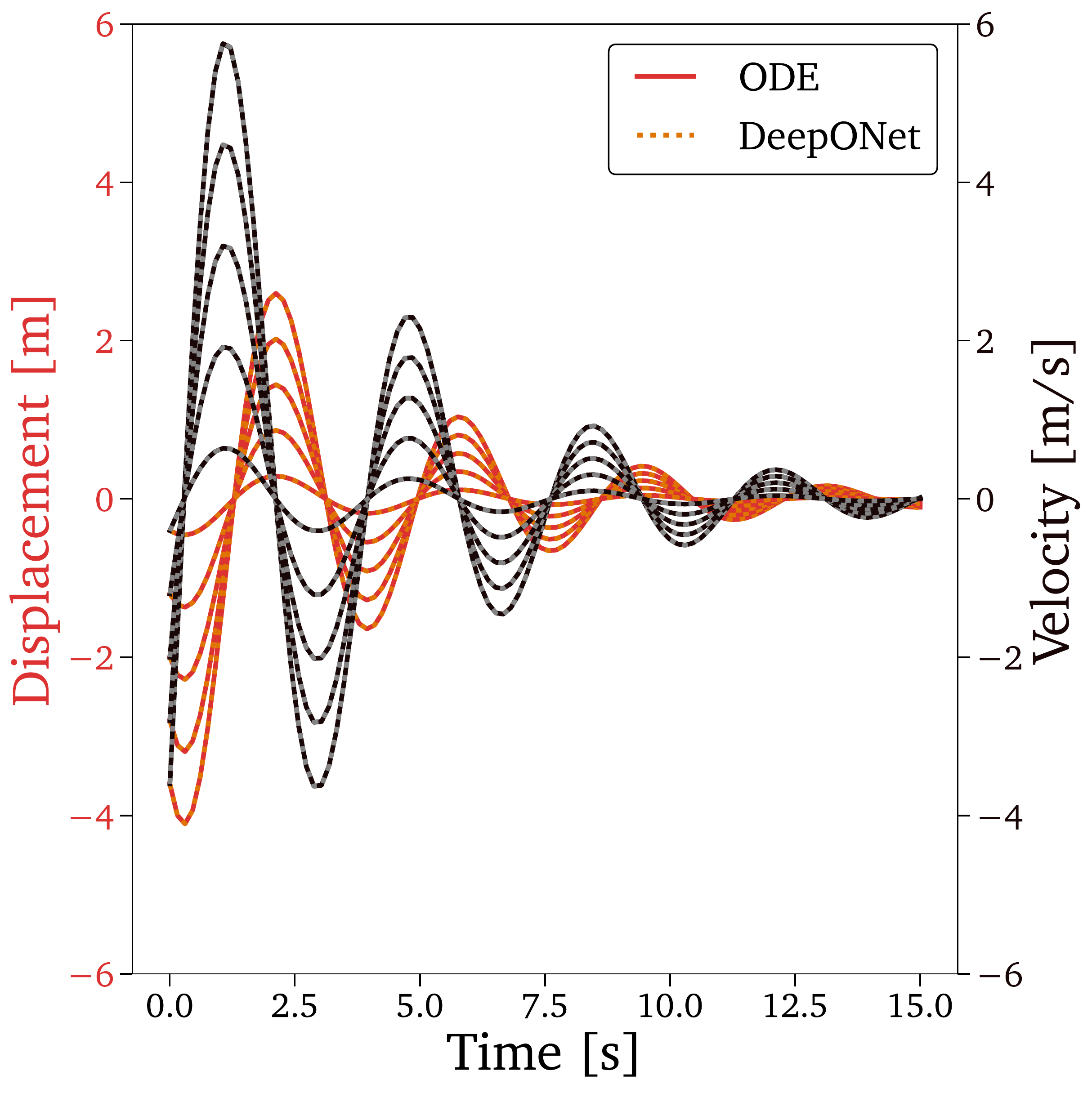}
        \label{fig:MSD_DeepONet_Tests}
    \end{subfigure}
    \hfil
    \begin{subfigure}{0.49\textwidth}
        \caption{}
        \includegraphics[width=3.1in]{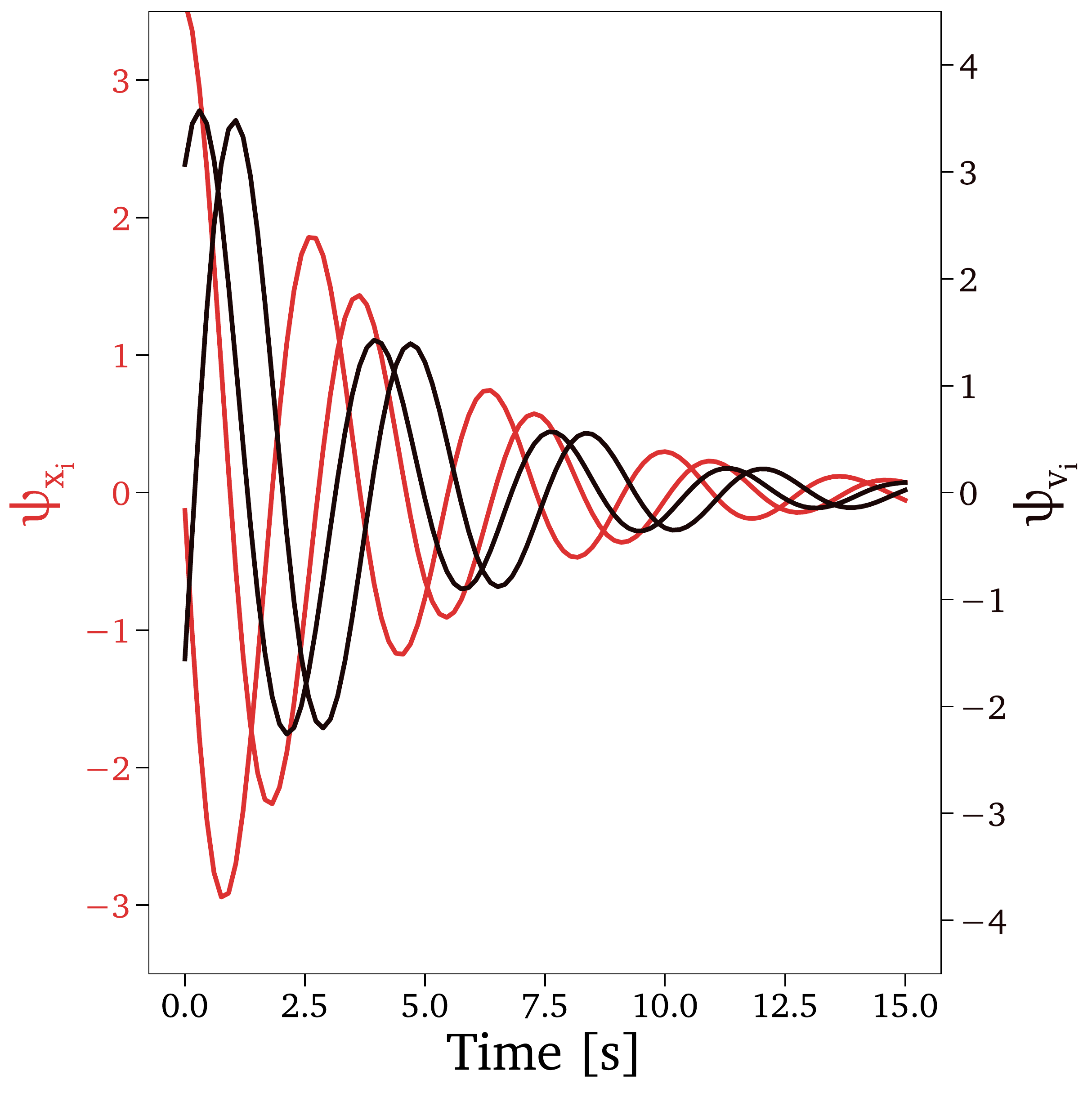}
        \label{fig:MSD_DeepONet_Modes}
    \end{subfigure}
    \caption{{Vanilla DeepONet applied to the mass-spring-damper test case}. ({A}): Displacements and velocities at five test scenarios as the results of the ODE integration (solid lines) and as predicted by the DeepONet (overlapping dotted lines). ({B}): Outputs of the trunk nets for displacements (red curves) and velocities (black curves). See Fig.~\ref{fig:MSD_DeepONet_1} for details about the vanilla DeepONet's structure.}
    \label{fig:MSD_DeepONet}
\end{figure}
The resulting 50,000 data points are used to train a vanilla DeepONet characterized by the structure in Fig.~\ref{fig:MSD_DeepONet_1}. Figure~\ref{fig:MSD_DeepONet_Tests} compares the displacements and velocities predicted by the surrogate at five test scenarios, unseen during training, with those obtained via direct time integration. The DeepONet performs well despite the output layers of its branch and trunk nets being composed of only two neurons (i.e., $p=2$). This means that the operator information content learned through the sampled scenarios is effectively summarized by the four outputs of the $x$ and $v$ trunk nets, which are shown in Fig.~\ref{fig:MSD_DeepONet_Modes}.
\begin{figure}
    \centering
    \includegraphics[width=6.0in]{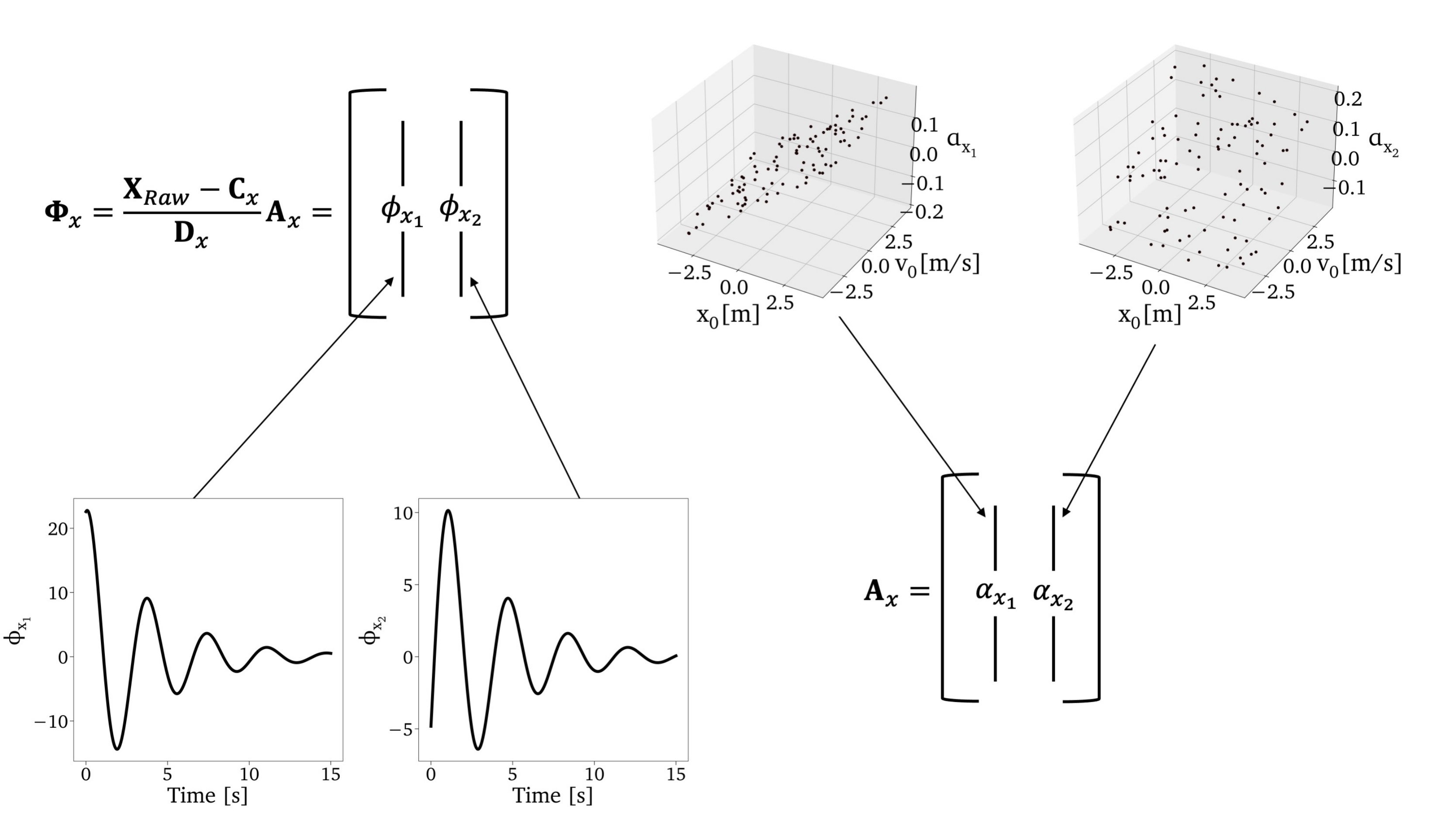}
    \caption{\textbf{SVD of the displacement data matrix}. Schematics of the columns resulting from the decomposition.}
    \label{fig:SVD_DeepONet_Inter_x}
\end{figure}
\begin{figure}
    \centering
    \includegraphics[width=6.0in]{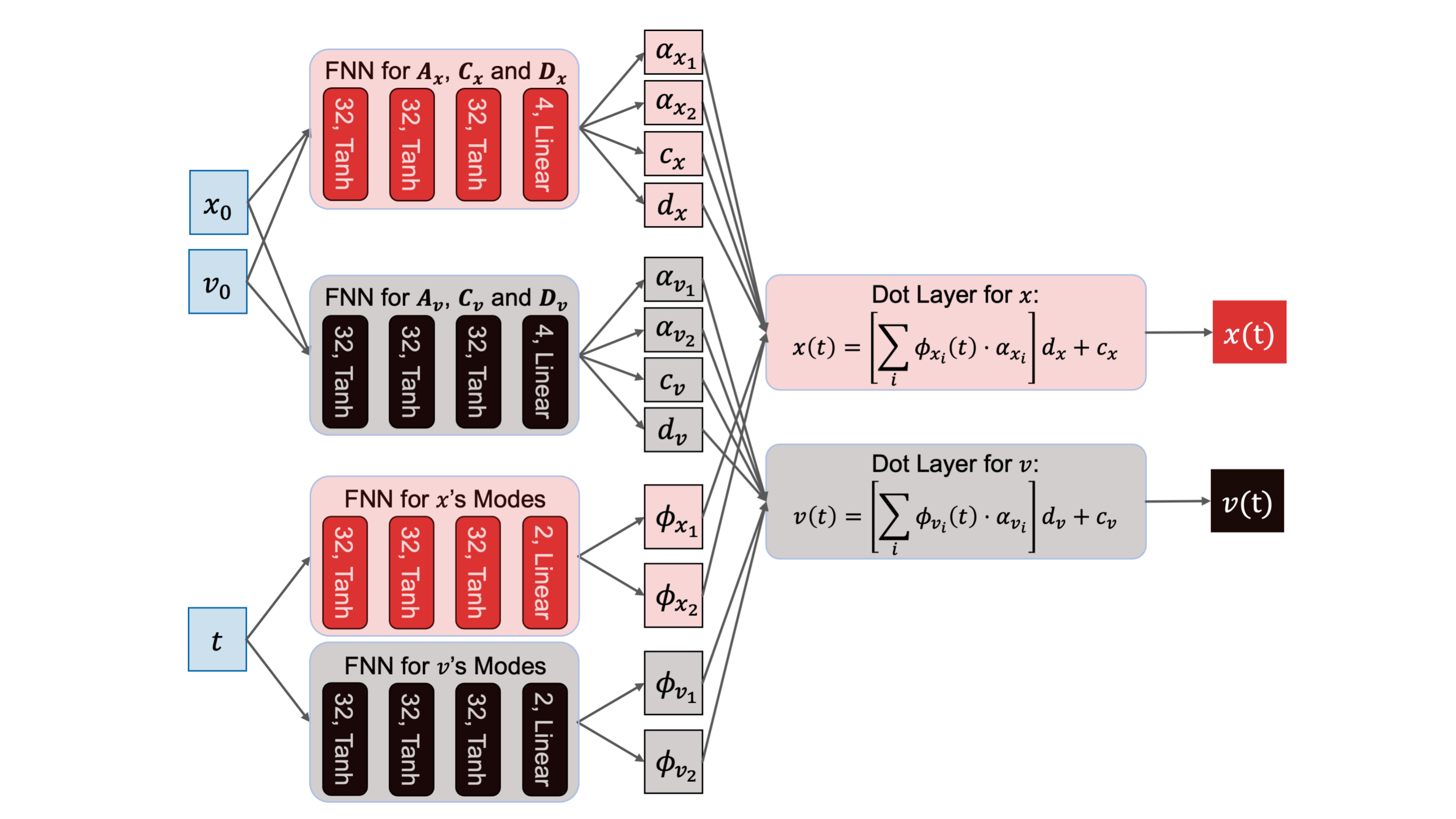}
    \caption{\textbf{Assembled architecture of the SVD-DeepONet for the mass-spring-damper test case}. After being independently trained as feed-forward neural networks (FNNs), the four blocks are assembled as DeepONet's trunk nets and branch nets to predict displacements and velocities at unseen times and for unseen initial conditions.}
    \label{fig:MSD_DeepONet_2}
\end{figure}
We now detail the application of the alternative approach to the construction of DeepONet-based surrogates developed under the perspective of the singular value expansion (SVE)~\cite{SVE_Hansen_1988,SymmSVE_Constantine_2014} and introduced in Sec.~\ref{subsec:SVD-DeepONet} as SVD-DeepONet. Beforehand, we concatenate displacements and velocities from different scenarios as multiple columns of two data matrices, $\mathbf{X}_{Raw}$ and $\mathbf{V}_{Raw}$, with dimensions $[N_t \times N_S]$. When performing SVD on matrices assembled in this way we refer to this as scenario-aggregated SVD, as each column is a snapshot of the dynamics arising from a specific initial condition, or scenario. We highlight that in the case of a generic time-and-space-dependent PDE, the column of the resulting scenario-aggregated snapshot matrices would contain the full spatio-temporal solution for a given scenario. This contrasts with the POD-based methods, for which each snapshot represents a spatially varying solution, and different columns correspond to different time instants. That being said, we center each of the matrices' columns and generate the scenario-dependent vectors $\boldsymbol{c}_x$ and $\boldsymbol{c}_v$. We then perform auto-scaling (i.e., scale each of the columns based on its standard deviation), obtaining the scenario-dependent vectors $\boldsymbol{d}_x$ and $\boldsymbol{d}_v$. At this point, we carry out the singular value decompositions and produce the corresponding matrices $\mathbf{\Phi}_x$, $\mathbf{A}_x$, $\mathbf{\Phi}_v$, and $\mathbf{A}_v$ based on Eq.~\ref{eq:pca_2}. Schematics are presented in Figs.~\ref{fig:SVD_DeepONet_Inter_x} and~\ref{fig:SVD_DeepONet_Inter_v}.
Consistent with the vanilla DeepONet's predictions for $x$ and $v$ being effectively summarized by two trunk outputs each, two singular values are sufficient for effectively decomposing both $\mathbf{X}$ and $\mathbf{V}$. In fact, the cumulative energy content of their first two singular values exceeds 99.9999\%, and the resulting encoding-decoding errors are below machine precision. For the application under analysis, we also note that such effectiveness of the reduction does not depend on the choice of centering and scaling. \\
Subsequently, two feed-forward neural networks (FNNs) with one input and two outputs each are constructed to regress $x$ and $v$'s principal components as functions of time. These blocks are independently trained in a fully data-driven fashion by relying on the data points $(\boldsymbol{t}_s; \mathbf{\Phi}_{x})$ and $(\boldsymbol{t}_s; \mathbf{\Phi}_{v})$, respectively. Here and in the following, we want to communicate through the notation $(\mathbf{I}; \mathbf{O})$ that the surrogate's training is performed by taking the $\mathbf{I}$ data matrix as input and comparing the resulting output with the $\mathbf{O}$ data matrix. Two additional FNNs with two inputs and four outputs each are generated to regress $x$ and $v$'s principal directions as functions of the initial conditions. These blocks are also independently trained in a fully data-driven fashion by relying on the data points $([\boldsymbol{x}_0,\boldsymbol{v}_0]; \mathbf{B}_x)$ and $([\boldsymbol{x}_0,\boldsymbol{v}_0]; \mathbf{B}_v)$, respectively. Here, the generic $\mathbf{B}_j$ matrix is constructed by concatenating $\mathbf{A}_j$, the vector $\boldsymbol{c}_j$, and the vector $\boldsymbol{d}_j$:
\begin{equation}\label{eq:pca_4}
    \mathbf{B}_j = \begin{bmatrix}
        \vert & \vert & \vert & \vert \\
        \boldsymbol{\alpha}_{1_j} & \boldsymbol{\alpha}_{2_j} & \boldsymbol{c}_{j} & \boldsymbol{d}_{j}  \\
        \vert & \vert & \vert & \vert
    \end{bmatrix},
\end{equation}
with $j \in \{x,v\}$. A visual outline of these steps is shown in Fig.~\ref{fig:MSD_DeepONet_Inter_x}. Finally, in the last phase, the four FNNs are assembled as trunk and branch nets of a DeepONet that does not require any additional training, as its prediction phase relies on the parameters already learned during the sub-nets' calibrations and simply involves the evaluation of the blocks and dot product layer. Figures~\ref{fig:MSD_DeepONet_2} and \ref{fig:MSD_SVDDeepONet_Tests} respectively sketch the architecture for this particular problem and show the resulting predictions for unseen scenarios, which are in excellent agreement with the time-integrated dynamics. Together with the training and prediction times, Table~\ref{table:MSD_TestErrors} provides quantitative comparisons between the errors from vanilla DeepONet, POD-DeepONet, and SVD-DeepONet, which are very similar to each other. 
\begin{figure}
    \centering
    \includegraphics[width=3.2in]{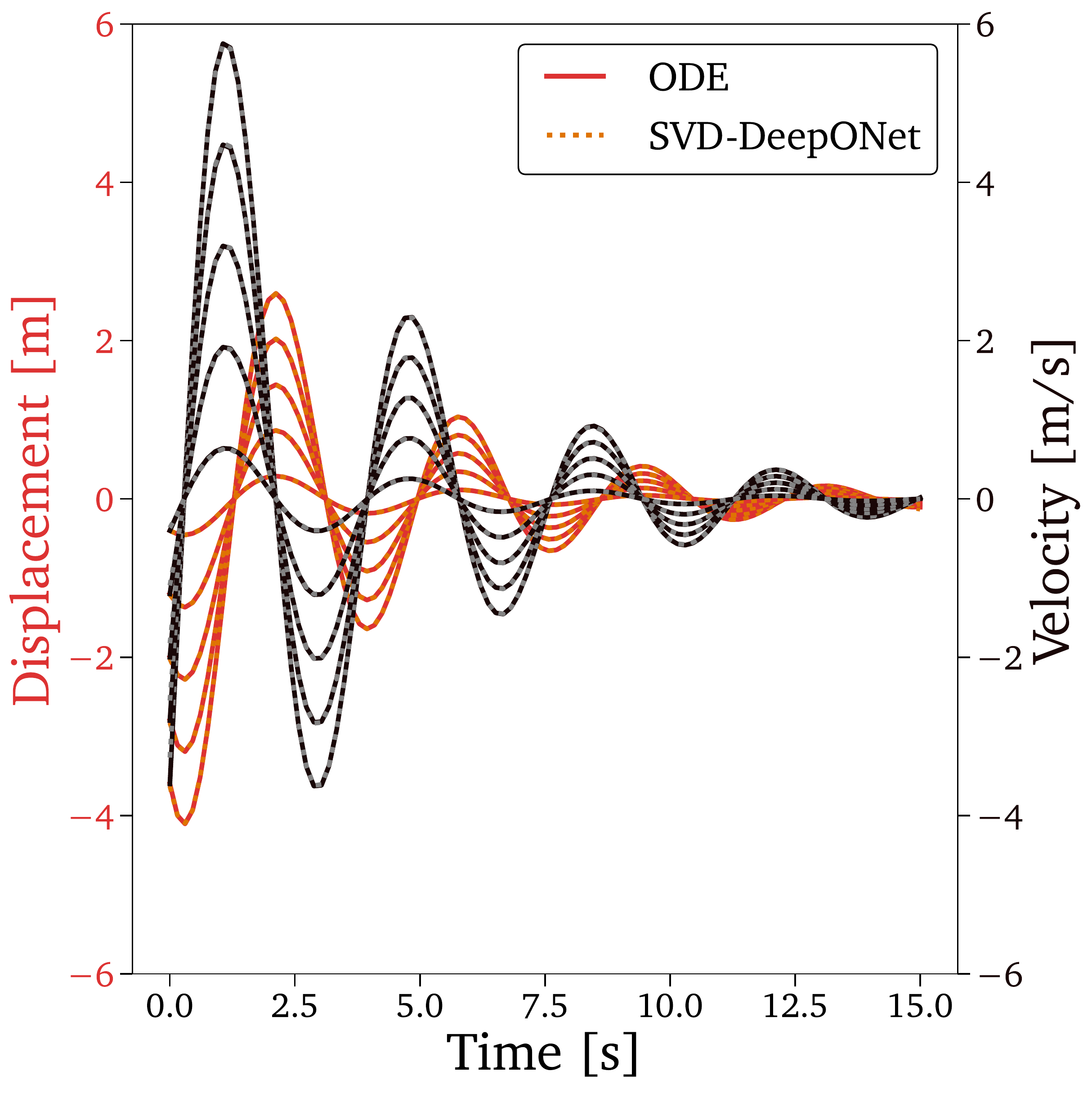}
    \caption{\textbf{SVD-DeepONet applied to the mass-spring-damper test case}. Displacements and velocities at five test scenarios as the results of the ODE integration (solid lines) and as predicted by the SVD-DeepONet (overlapping dotted lines) with structure as in Fig.~\ref{fig:MSD_DeepONet_2}.}
    \label{fig:MSD_SVDDeepONet_Tests}
\end{figure}
While the SVD method bounds the best performance that the SVD-DeepONet can achieve, the former does not require the initial data to be aligned on a grid, and it can predict the values of modes, coefficients, and reconstructed solutions at generic times and for generic initial conditions.\\
The present work draws similar connections between SVD-derived techniques and DeepONets as some recent works. In particular, Meuris \textit{et al.} interpreted the trunk outputs as unprocessed POD modes. After extracting and transforming them into hierarchical orthonormal bases, they used these outputs to expand the solutions of time-dependent PDEs~\cite{SVDDeepONet_Meuris_2021}. 
Instead, with respect to the POD-DeepONet approach introduced in Sec.~\ref{subsec:POD-DeepONet}, the SVD-DeepONet presented in this work consists of three main modifications: i) the trunks are FNNs that fit the principal components, rather than being POD modes, ii) the branches recover the principal directions, rather than the POD coefficients, and iii) the training of the branches does not involve the trunk nets. This last attribute is in analogy with the offline stage of the POD-neural network (POD-NN) method by Hesthaven \textit{et al.}~\cite{NNPOD_Hesthaven_2018, NNPOD_Wang_2019}. In other words, SVD-DeepONet trains trunks and branches as fully-independent FNNs, and it assembles these building blocks as the DeepONet architecture only at the prediction phase.
Compared to the POD-DeepONet, one advantage of the proposed approach is a simplification of the training stage. Particularly, the resulting embarrassingly parallelizable workload has vast implications in the case of stacked~\cite{DeepONet_Lu_2019, DeepONet_Lu_2021,PODDeepONet_Lu_2021} POD-DeepONets, which contain one branch net for each of the projection matrix's columns. As an additional example, if the DeepONet's building blocks involve multiple libraries or coding languages (e.g., trunk nets constructed as polynomial chaos expansions by relying on legacy code or libraries), the training of the assembled DeepONet requires the breakage of the full-architecture's graph within the training and increases the computational costs. In contrast, SVD-DeepONet would not have to face these issues, as each of the building blocks can separately rely on its own library during calibration. For the sake of completeness, it is worth mentioning that, as for POD-DeepONet, SVD-DeepONet shares similarities with the PCA-NN approach of Bhattacharya \textit{et al.}~\cite{PCANN_Bhattacharya_2020}. As stated by Kovachki~\textit{et al.}, both POD- and SVD-DeepONet ``finite-dimensionalize the output in the span of PCA modes, bringing them (the DeepONets) closer to the method introduced in Bhattacharya \textit{et al.}~\cite{PCANN_Bhattacharya_2020}, but with a different finite-dimensionalization of the input space.''~\cite{NeuralOs_Kovachki_2021}.\\
For operators that involve multiple states, as the mass-spring-damper system analyzed here, the SVD-DeepONet strategy applied to surrogate learning can be pursued differently. In fact, instead of computing separate principal components and related principal directions for each of the snapshot matrices (e.g., for $\mathbf{X}$ and $\mathbf{V}$ in the test case under analysis), one can concatenate all or some of them and perform the SVD on the resulting matrix, $\mathbf{Z}$ (e.g., $\mathbf{Z} = [\mathbf{X}, \mathbf{V}]$). The dimensions of $\mathbf{Z}$ are $[N_t \times N_o N_S]$, with $N_o$ being the number of states sharing the same trunk. The shared components, $\mathbf{\Phi}_z$, can then be fit with a single FNN, and the columns of the $\mathbf{A}_z$ matrix together with the related $\boldsymbol{c}_z$ and $\boldsymbol{d}_z$ vectors can be retrieved by $N_o$ FNNs. Figures~\ref{fig:SVD_DeepONet_Inter_All}-\ref{fig:MSD_SVDDeepONet_Tests_All} outline and report the application of this approach to the mass-spring-damper test case. It should be noted that the idea of sharing trunks between variables was one of the four strategies suggested by Lu \textit{et al.}~\cite{PODDeepONet_Lu_2021} for dealing with multiple outputs. However, in light of SVD-DeepONet, it must also be stressed that this technique has the potential of being significantly advantageous only if the dynamics of the state variables that share the trunk take place on comparable temporal/spatial scales and can be represented by similar principal components. Moreover, scenario-dependent scaling and centering become crucial in SVD-DeepONets if the state variables sharing the trunk have different units or span different ranges.\\
Finally, in view of what has been discussed above concerning its SVD-extension, we want to call attention to the benefits of the vanilla DeepONet as originally proposed in~\cite{DeepONet_Lu_2019, DeepONet_Lu_2021}:
\begin{itemize}
    \item In POD- and SVD-DeepONets, the eigenbasis of the empirical covariance operator needs to be computed \textit{a priori} via POD/SVD. On the contrary, as pointed out by Lanthaler \textit{et al.}: ``(vanilla) DeepOnets do not require any explicit knowledge of the covariance operator. In fact, our analysis shows that DeepOnets implicitly and concurrently learn a suitable basis in output space along with an approximation of the projected operator''~\cite{UATs_Lanthaler_2021}. This generality, however, comes at the expense of predictive accuracy, as proven by the fact that the POD-DeepONet significantly outperformed the vanilla approach in the 16 benchmarks of~\cite{PODDeepONet_Lu_2021}. Indeed, POD- and SVD-DeepONets produce orthonormal basis that are optimal in the $\ell_2$ sense.
    \item The vanilla DeepONet autonomously learns a scenario-dependent scaling, as this variable is implicitly absorbed into the branches' $p$ outputs as a function of the initial conditions (i.e., $b_i(x_0)=\tilde{b}_i(x_0)*d(x_0)$). On the contrary, because it uses a scalar bias~\cite{DeepONet_Lu_2021,DeepONetApps_Yin_2022} that corresponds to a shift of the entire snapshot matrix based on its overall mean value, the vanilla architecture cannot learn a scenario-dependent centering. Instead, by adding a $p+1$-th output to the branch net, $c(x_0,v_0)$, we enable the discovery of the proper scenario-dependent bias, which can further improve the generalization capabilities of the overall architecture. 
    \item The vanilla DeepONet can theoretically learn a good approximation of the operator by simply relying on the residuals of the governing equations without using data points~\cite{PIDeepONet_Wang2021_1,PIDeepONet_Wang2021_2,PIDeepONet_Wang2021_3}. However, a fully physics-based approach to training can be impractical in real-world problems, as it could require a large set of collocation points. Recent work, for example, recommends hybrid physics-data
  strategies~\cite{DeepONetApps_Goswami_2022,PODDeepONet_Lu_2021}. Additionally, a physics-informed attribute can be given to the SVD-DeepONet by relying on transfer learning. In fact, after having performed the SVD decomposition of the available data and having independently learned the FNNs representing modes and coefficients, these blocks can be assembled into a DeepONet. At this point, after having initialized the surrogate's parameters from the learned values, the DeepONet can be further refined via the physics-informed training of the entire architecture (i.e., by enforcing ODE/PDE residuals in the loss).
    \item The vanilla DeepONet can work with sparse datasets. Most importantly, while it still requires a discretization of the branch input functions, it does not require any particular discretization for the inputs to the trunks.
    \item Similarly to incremental implementations of the SVD method~\cite{SVD_Brand_2002,SVD_Choi_2021}, the SVD-DeepONet can be trained on-the-fly (i.e., while more data is being acquired from sensors or additional simulations).
\end{itemize}


\subsection{Test Case 2: A Shifting Hyperbolic Tangent Function}\label{subsec:TestCase2}

\begin{figure}[!htb]
    \begin{subfigure}{0.49\textwidth}
        \caption{}
        \label{fig:Tanh_DeepONet_2_Tests}
        \centering
        \includegraphics[width=3.2in]{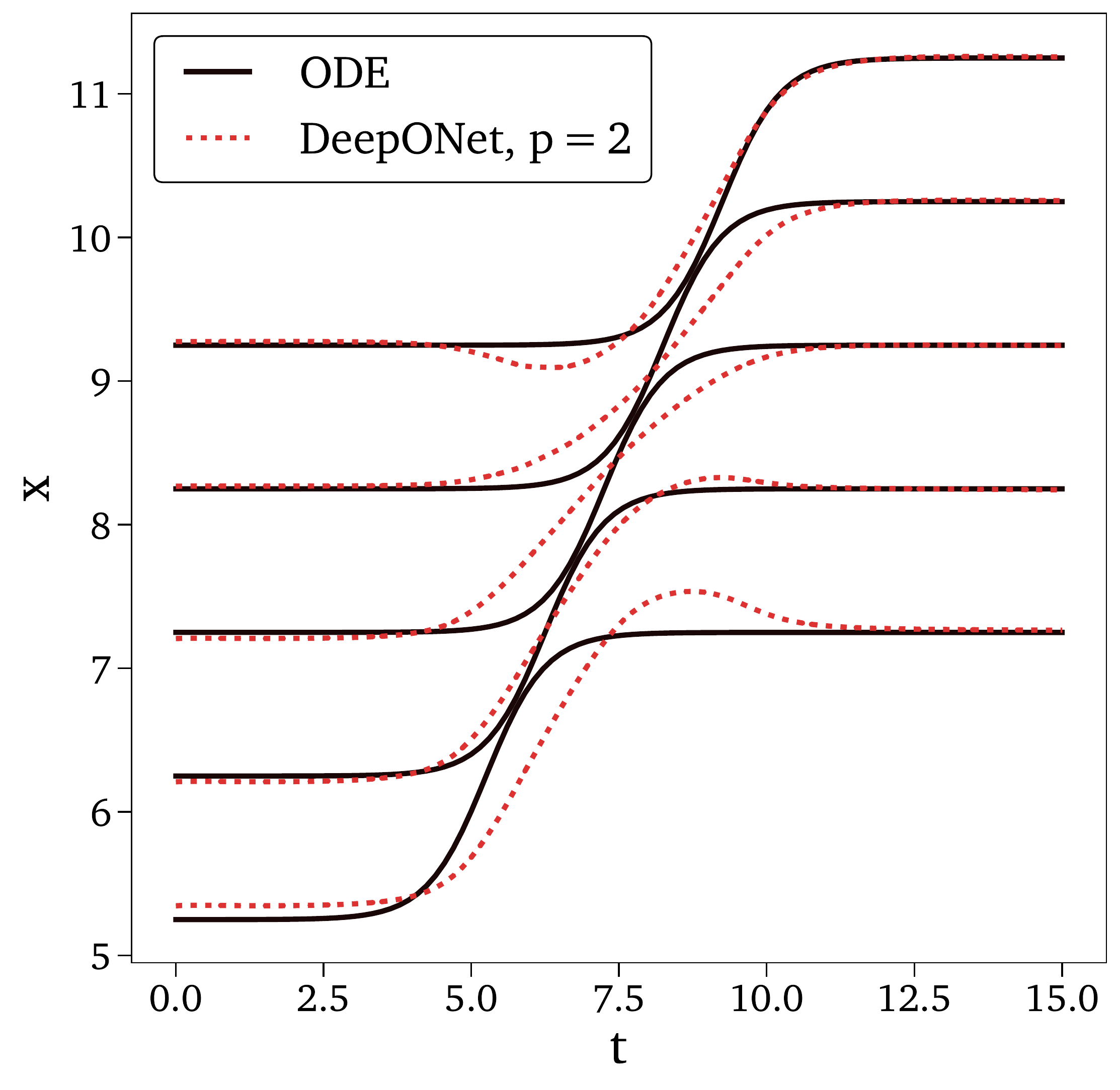}
    \end{subfigure}
    \begin{subfigure}{0.49\textwidth}
        \caption{}
        \label{fig:Tanh_DeepONet_8_Tests}
        \centering
        \includegraphics[width=3.2in]{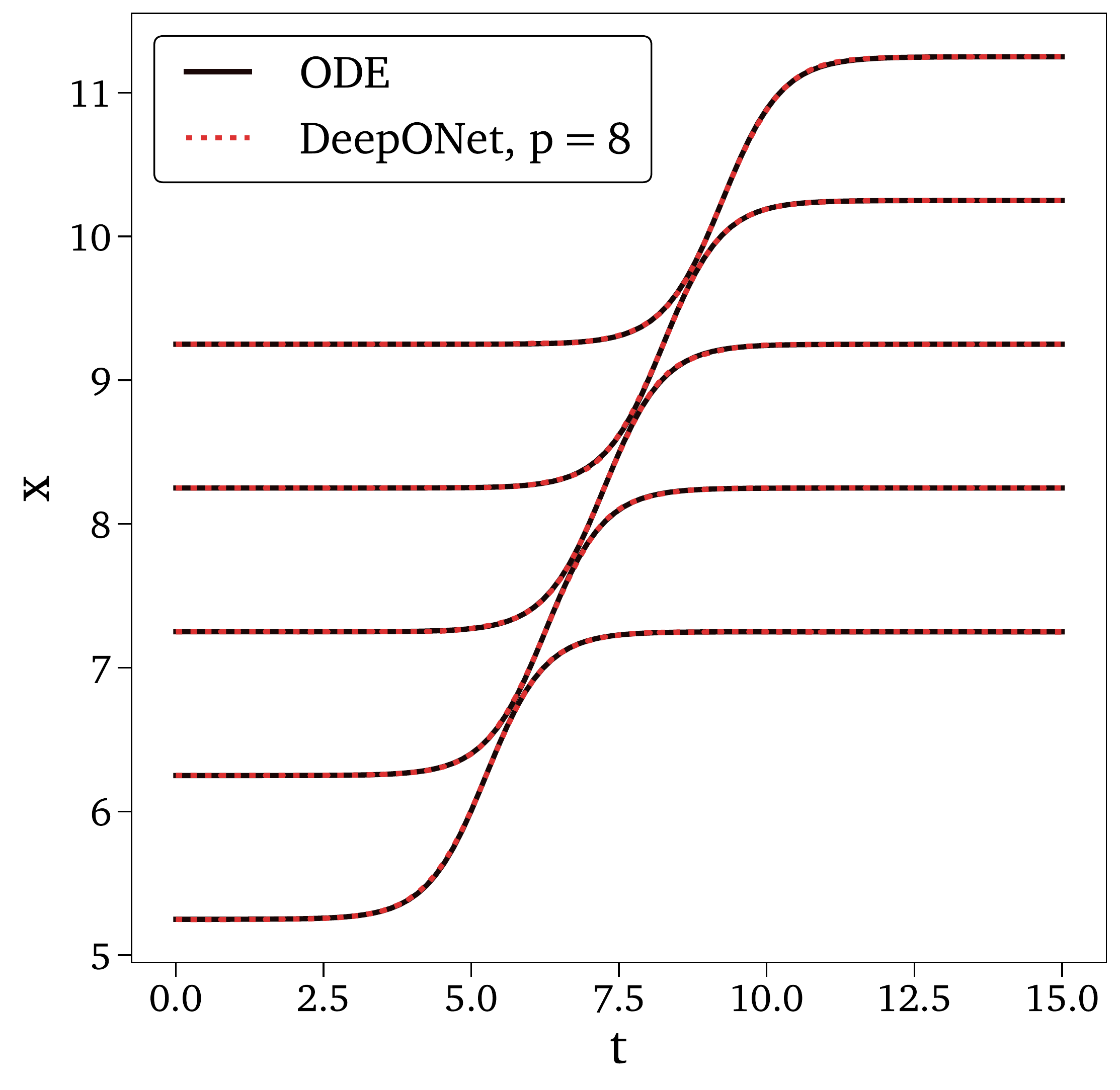}
    \end{subfigure}
    \caption{\textbf{Vanilla DeepONet applied to the shifting hyperbolic function test case}. $x$ at five test scenarios as the results of the ODE integration (solid black lines) and as predicted by vanilla DeepONets (red dotted lines) with trunk and branch's output layers composed of two (\textbf{A}) and eight (\textbf{B}) neurons. See Fig.~\ref{fig:Tanh_DeepONet_1} for details on the vanilla DeepONet's structure.}
    \label{fig:Tanh_DeepONets_Tests}
\end{figure}

\begin{figure}[!htb]
    \centering
    \includegraphics[width=6.0in]{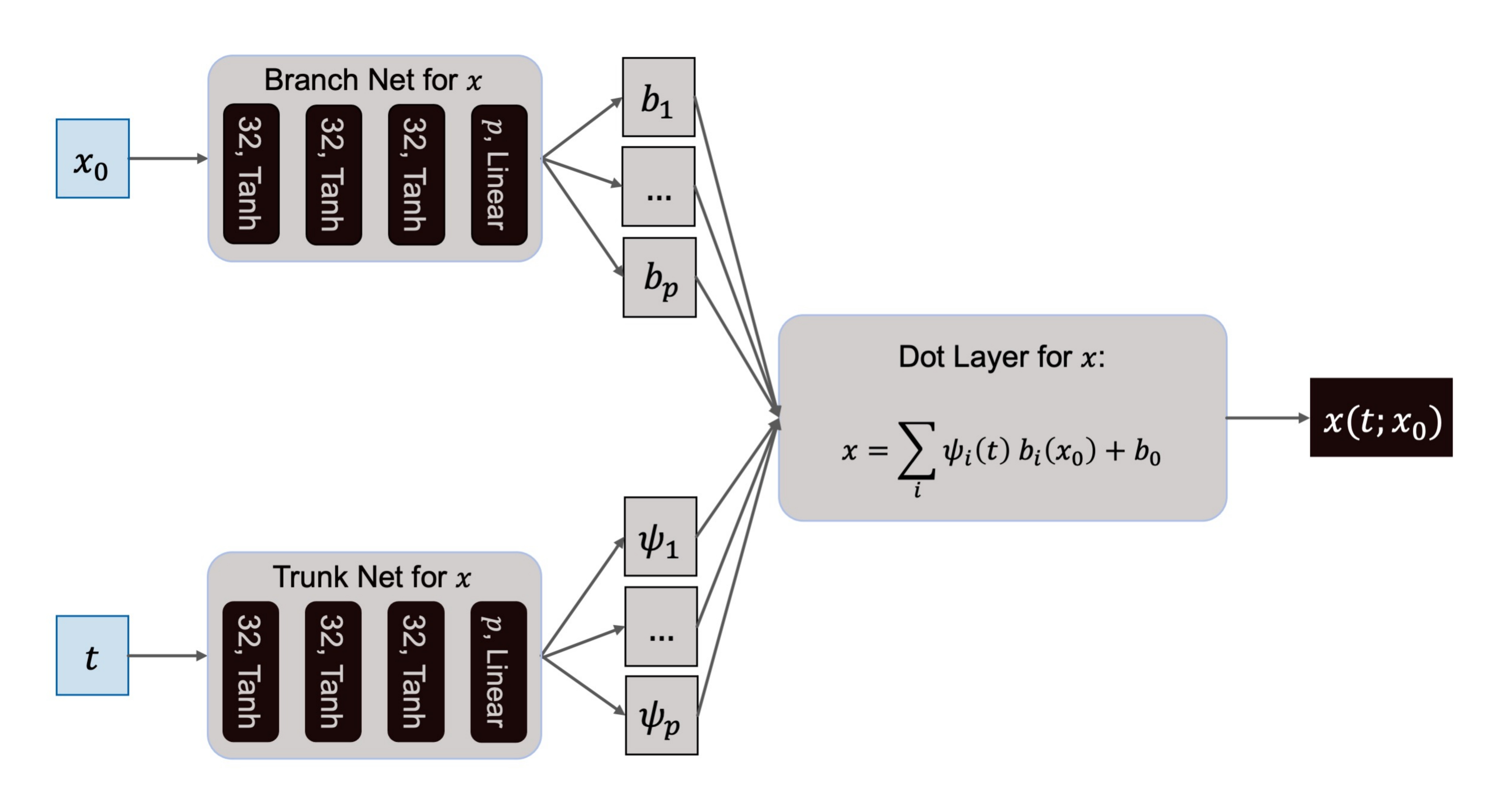}
    \caption{\textbf{Architecture of the vanilla DeepONet for the shifting hyperbolic function test case}. We tested multiple configurations with different numbers of trunk and branch's outputs, $p$.}
    \label{fig:Tanh_DeepONet_1}
\end{figure}

\begin{figure}[!htb]
    \centering
    \includegraphics[width=3.2in]{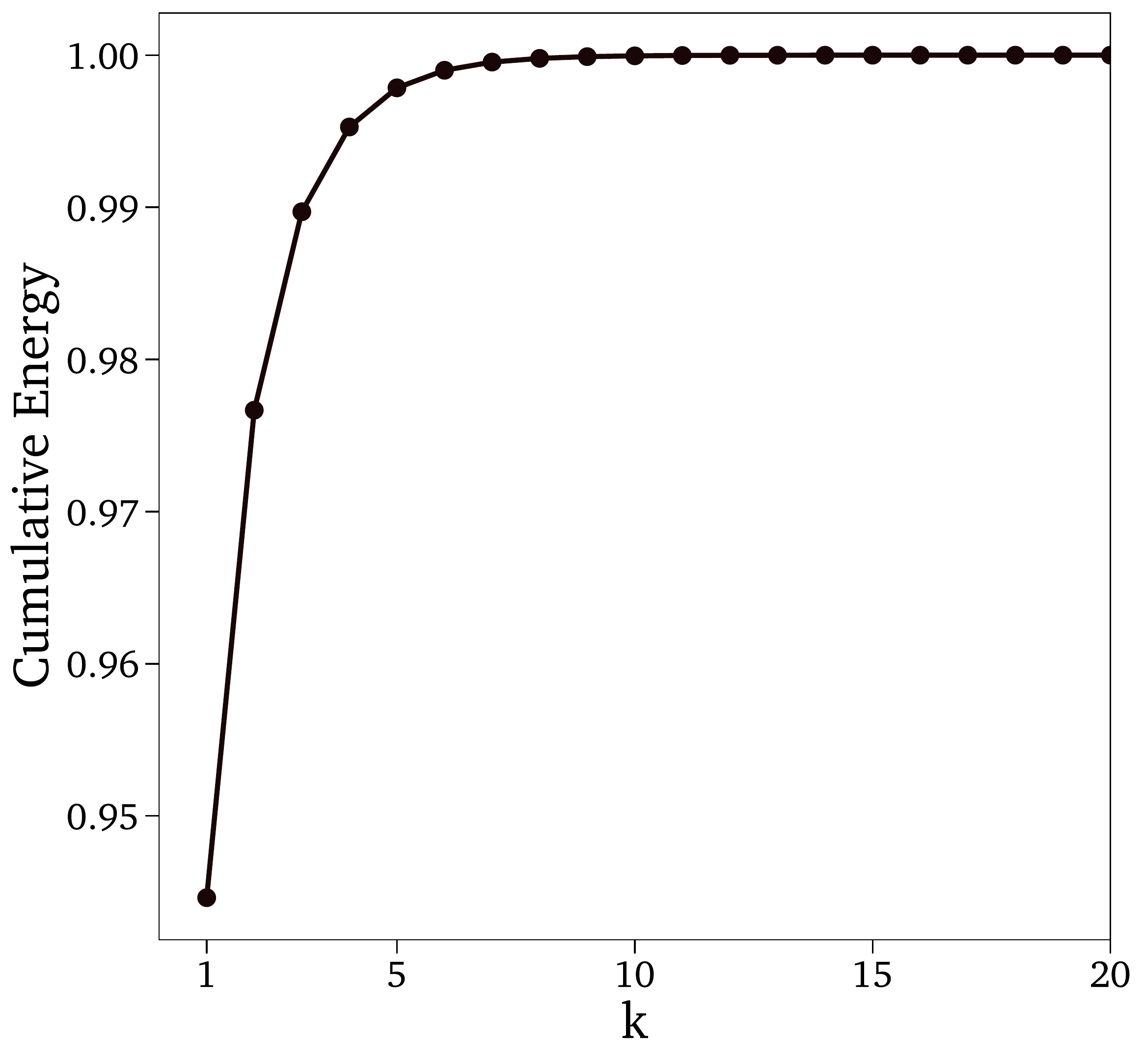}
    \caption{\textbf{Cumulative energies of the position data matrix from the shifting hyperbolic function test case}. Cumulative energies contained in $\mathbf{X}$'s first k singular values.}
    \label{fig:Tanh_CumEnergy}
\end{figure}

\begin{figure}[!htb]
    \begin{subfigure}{0.49\textwidth}
        \centering
        \caption{}
        \label{fig:Tanh_Mode1}
        \includegraphics[width=3.1in]{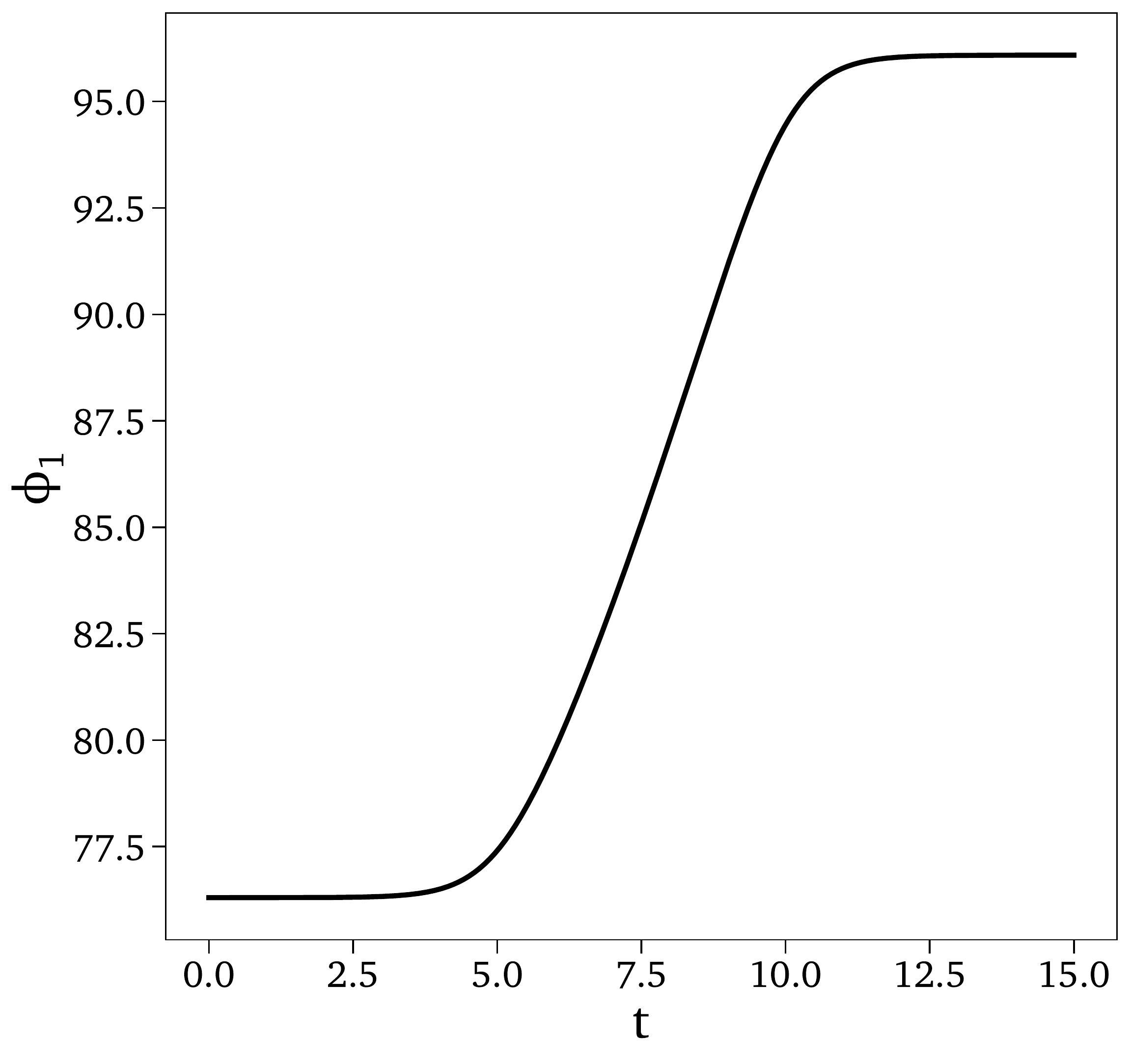}
    \end{subfigure}
    \begin{subfigure}{0.49\textwidth}
        \centering
        \caption{}
        \label{fig:Tanh_Mode8}
        \includegraphics[width=3.2in]{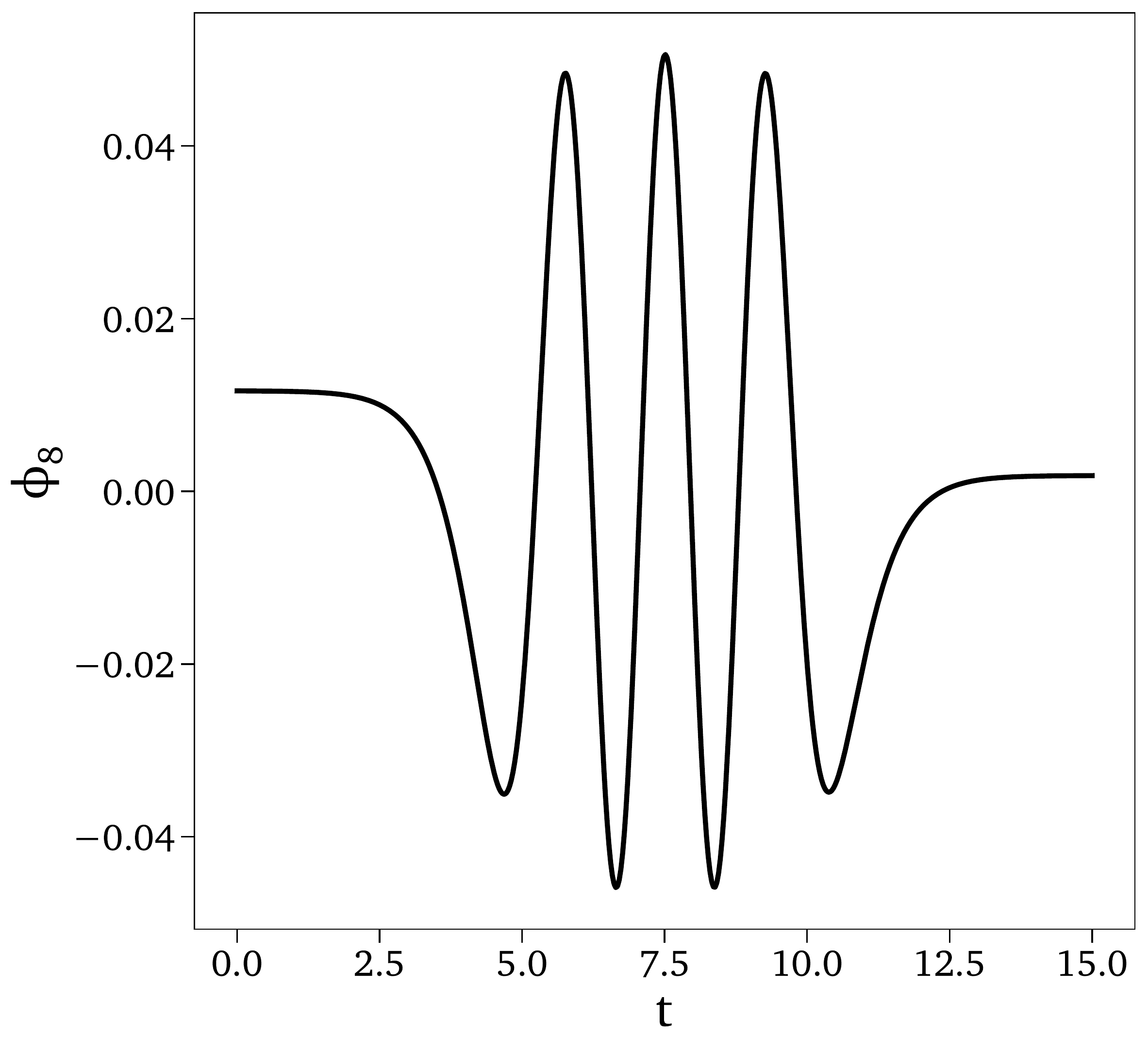}
    \end{subfigure}
    \caption{\textbf{Principal components of the position data matrix from the shifting hyperbolic function test case}. First (\textbf{A}) and eight (\textbf{B}) principal components of $\mathbf{X}$.}
    \label{fig:Tanh_Modes}
\end{figure}

\begin{figure}[!htb]
    \begin{subfigure}{0.49\textwidth}
        \centering
        \caption{}
        \label{fig:Tanh_Coeff1}
        \includegraphics[width=3.1in]{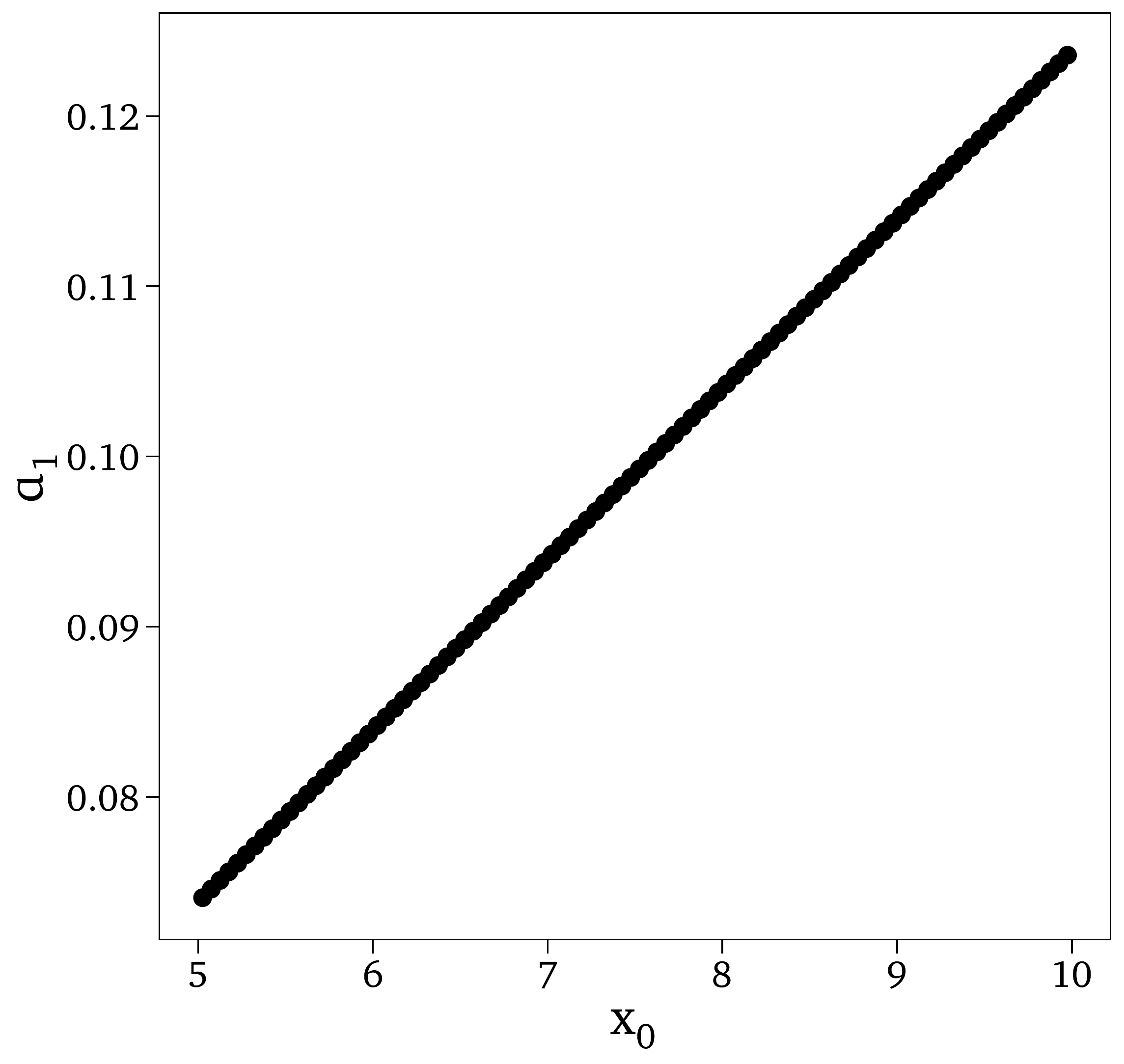}
    \end{subfigure}
    \begin{subfigure}{0.49\textwidth}
        \centering
        \caption{}
        \label{fig:Tanh_Coeff8}
        \includegraphics[width=3.2in]{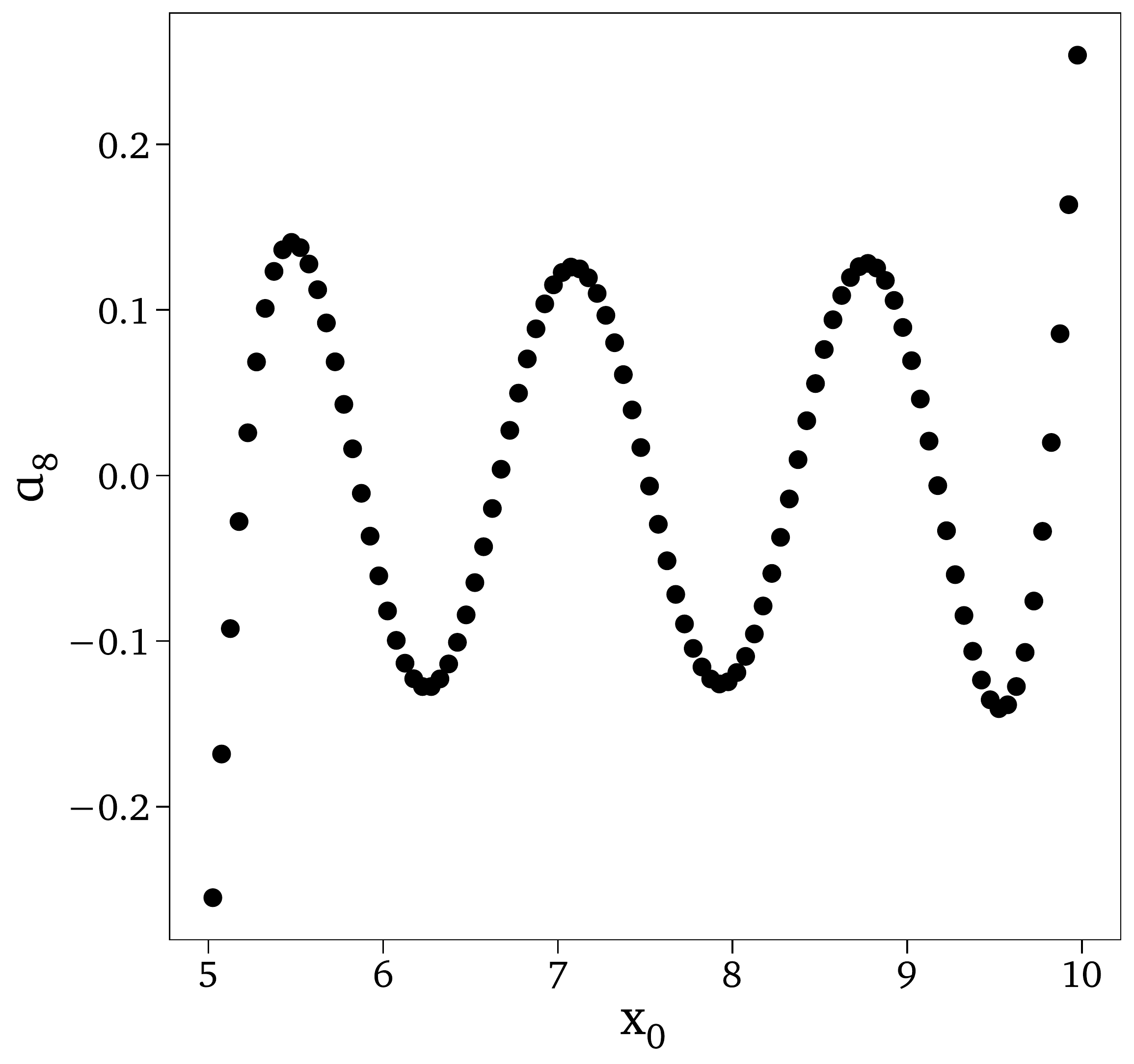}
    \end{subfigure}
    \caption{\textbf{Coefficients of the principal direction matrix for the shifting hyperbolic function test case}. First (\textbf{A}) and eight (\textbf{B}) columns of $\mathbf{A}_x$.}
    \label{fig:Tanh_Coeffs}
\end{figure}

The second test case that we analyze is a toy problem constructed from the following ODE:
\begin{equation}\label{Tanh_Eq_1}
    \dfrac{dx(t)}{dt} = a \sech^2{(b x_0 - t)}
\end{equation}
\begin{equation}\label{Tanh_IC}
    x(0) = x_0
\end{equation}
where $t\in(0,15)$, the coefficients $a$ and $b$ are both set to $1$, and the initial condition, $x_0$, is randomly selected from the interval $(5,10)$.
The ODE has the analytical solution,
\begin{equation}\label{Tanh_Eq_2}
    x(t) = a \tanh{(t - b x_0)} + a \tanh{(b x_0)} + x_0,
\end{equation}
which corresponds to a hyperbolic tangent shifted in time and space as a function of the initial conditions $x_0$. \\
To construct a surrogate of the operator, we simulate 100 scenarios characterized by different initial conditions, and for each, we uniformly sample 500 time instants. We use the resulting 50,000 data points to train vanilla DeepONets with multiple numbers of neurons for trunk and branch output layers, $p$. \\
For clarity, in this test case we employ a vanilla DeepONet architecture as the starting point in order to motivate distinct improvements required to represent the underlying dynamics.
In Fig.~\ref{fig:Tanh_DeepONets_Tests}, the resulting predictions at five test scenarios are compared to the time-integrated solutions. In relation to the simplicity of the ODE, a relatively large network structure is required in the DeepONet (see Fig.~\ref{fig:Tanh_DeepONet_1}), which also employs $\tanh{}$ as activation functions consistently with the analytical solution. Notwithstanding, all the surrogates with $p<8$ (i.e., many modes) show noticeable inaccuracies.\\
\begin{figure}[!htb]
    \centering
    \includegraphics[width=3.2in]{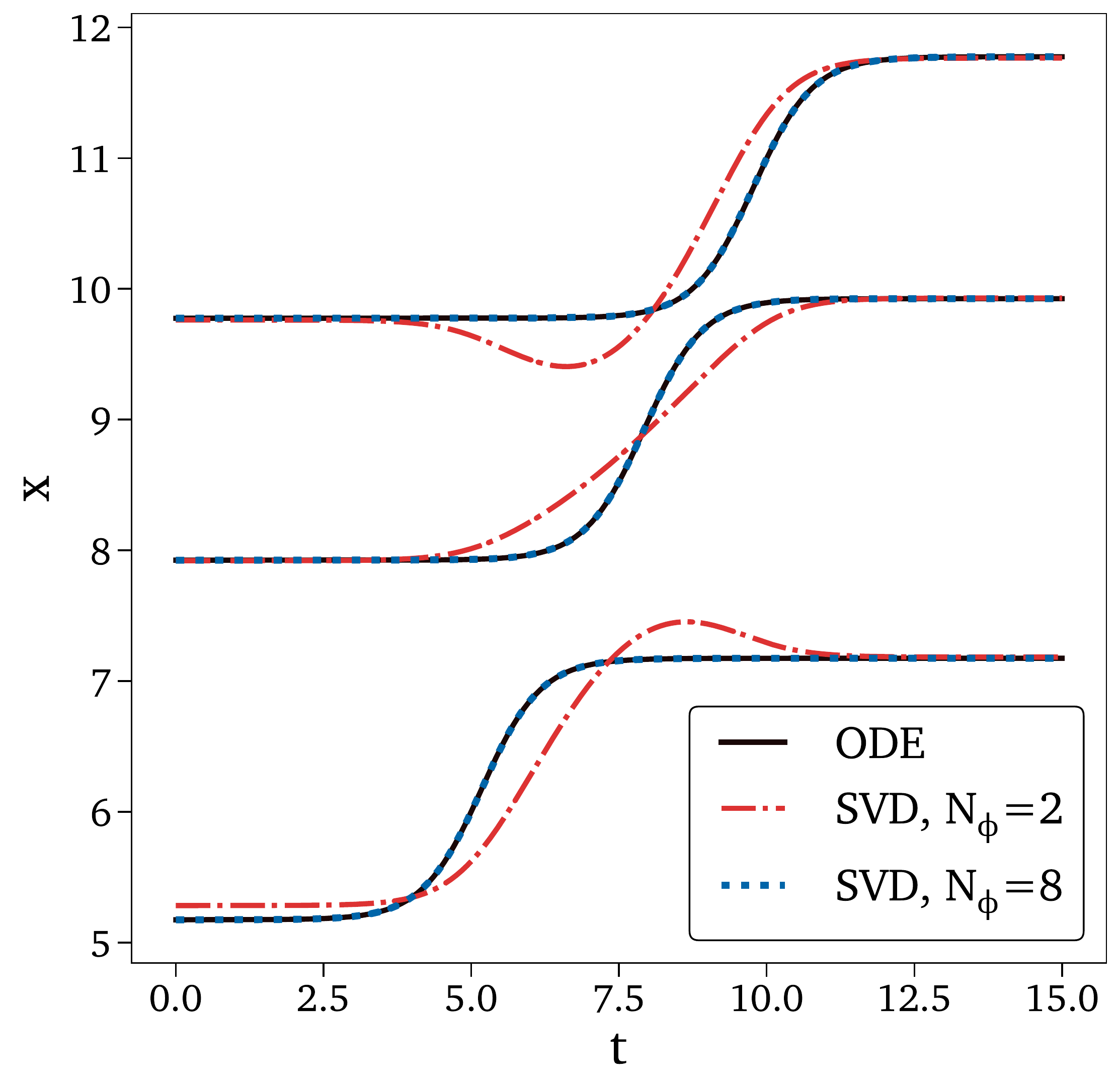}
    \caption{\textbf{Training scenarios reconstructed via SVD}. Three of the training scenarios for the shifting hyperbolic function from the ODE integration (black solid lines) and after being encoded-decoded based on SVD's first two (red dashed lines) and first eight (blue dotted lines) singular values.}
    \label{fig:Tanh_SVD_Reconstructed}
\end{figure}
In order to investigate this limitation on the expressive power of the vanilla DeepONet, we perform an analysis analogous to that carried out for the previous test case. We start by constructing a scenario-aggregated snapshot matrix, $\mathbf{X}_{Raw}$, that concatenates the 100 different scenarios as columns of dimensions $[500 \times 1]$, and, after having centered and auto-scaled, we decompose it via SVD. Two results are highlighted. Firstly, the energy contained in the singular values shows a slower decay (Fig.~\ref{fig:Tanh_CumEnergy}) than the one observed for the mass-spring-damper test case. This fact is also testified by the large encoding-decoding error generated by data matrix compressions that rely on less than eight modes, as reported in Fig.~\ref{fig:Tanh_SVD_Reconstructed} for three training scenarios. From comparing this last figure with Fig.~\ref{fig:Tanh_DeepONet_2_Tests}, we can draw analogies between the dynamics predicted by the under-parameterized DeepONet and that reconstructed from the over-compressed data matrix. An insufficient number of neurons in the output layers of the trunk and branch nets acts as a bottleneck similar to scarce cumulative energy content in the retained singular values. 
Secondly, the complexity of the basis increases with the number of preserved singular values, as principal components corresponding to smaller singular values are characterized by higher frequencies (Fig.~\ref{fig:Tanh_Modes}), and large oscillations also affect the related columns of the principal directions (Fig.~\ref{fig:Tanh_Coeffs})~\cite{SymmSVE_Constantine_2014}. In analogy, this observation also has repercussions on the DeepONet's performance. In order to fully exploit additional outputs, trunk and branch nets demand higher capacity as their regression tasks are complicated by oscillatory latent dynamics. This requirement may translate into deeper and/or wider trunks and branches, increasing the training efforts and the predictive computational costs.\\
The leading cause of slow singular value decay is the translational symmetry that characterizes this particular dynamical system. In fact, it is well known that SVD is highly contingent on the coordinate system adopted for representing the data. As a consequence, as stated by Brunton and Kutz~\cite{SVD_Brunton_2019}, ``the SVD rank explodes when objects in the columns translate, rotate, or scale, which severely limits its use for data that has not been heavily pre-processed''. Given the strong parallelisms between SVD and DeepONet discussed so far, it is not surprising to find the latter is also affected by similar limitations in processing shifts, rotations and scaling.
Here we adopt a simple but effective modification to DeepONet's original structure that, if viewed under the SVD paradigm, acts as an artificial intelligence (AI)-operated data alignment aimed at symmetry removal. As discussed in Sec.~\ref{subsec:flexDeepONet}, the proposed improvement relies on a pre-transformation network (Pre-Net) to automatically discover a moving frame of reference with respect to which the multiple scenarios become as overlapping as possible and then can be efficiently compressed to a lower number of modes. \\
Recently, Oommen \textit{et al.} proposed to learn complex dynamics via DeepONet in a latent and low-dimensional space through the employment of autoencoders~\cite{ ArchiDeepONet_Oommen_2022}. In their application, they reduced the dimensionality of images with resolution $[128\times128]$ into coded variables with dimensions of the order $\mathcal{O}(10-100)$. The dynamics represented in the latter quantities resulted in being less affected by high gradients and more easily representable via DeepONets. While they both benefit from transformations into latent spaces of independent variables that simplify the surrogation by disentangling the dynamics, the autoencoded-DeepONet and the flexDeepONet approaches have different primary focuses. Indeed, while an autoencoder is an extremely powerful addition in applications that require dimensionality reduction, the so performed transformations to the latent spaces lack interpretability and do not necessarily guarantee the separation of the rigid components of the motion. 
However, we believe that these aspects can be improved by introducing in the autoencoded-DeepONets paradigm some of the concepts from~\cite{SymmPOD_Mojgani_2020}. As stated in their work, Mojgani and Balajewicz trained “a diffeomorphic spatio-temporal grid, that registers the output sequence of the PDEs on a non-uniform parameter/time-varying grid, such that the Kolmogorov n-width of the mapped data on the learned grid is minimized.” In addition, in autoencoded-DeepONets the detection of the effective (reduced-dimensional) reference frame precedes the DeepONet, as the training of the autoencoder takes place before the training of the surrogate. This fact might represent a limitation in streaming machine learning applications. In contrast, flexDeepONet discovers the new reference frame while learning the dynamics. 
On the other hand, it should also be noted that the flexDeepONet approach is particularly effective in applications that involve rigid body motions, which characterize many physics and engineering problems. When these motions are negligible (e.g., modeling of material heterogeneity and defects~\cite{IFNOs_You_2022}), we expect the improvements resulting from the Pre-Net component to be minimal. We also acknowledge that, almost simultaneously with the work of Oommen \textit{et al.}, the idea of coupling DeepONets and autoencoders was suggested by Zhang \textit{et al.}~\cite{MODeepONet_Zhang_2022} for surrogating stochastic differential equations (SDE).\\
\begin{figure}[!b]
    \centering
    \includegraphics[width=6.0in]{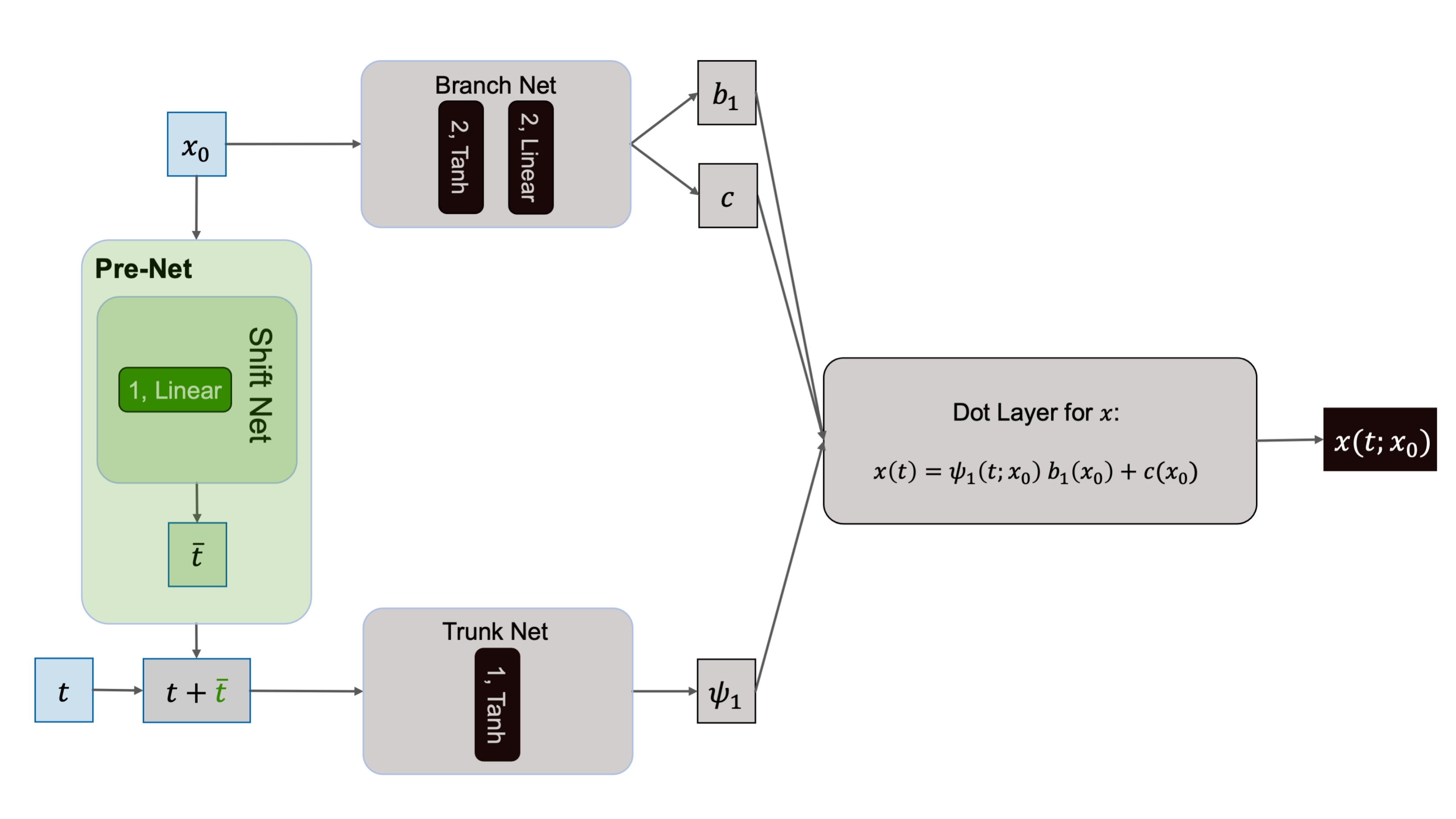}
    \caption{\textbf{FlexDeepONet architecture for the shifting hyperbolic function test case}. A Pre-Net composed of a shifting FNN is introduced to guarantee the alignment of the different scenarios. An additional neuron is added to the output layer of the branch net to permit scenario-specific centering.}
    \label{fig:Tanh_flexDeepONet}
\end{figure}
\begin{figure}[!htb]
    \begin{subfigure}{0.49\textwidth}
        \caption{}
        \label{fig:Tanh_flexDeepONet_Shift}
        \centering
        \includegraphics[width=3.2in]{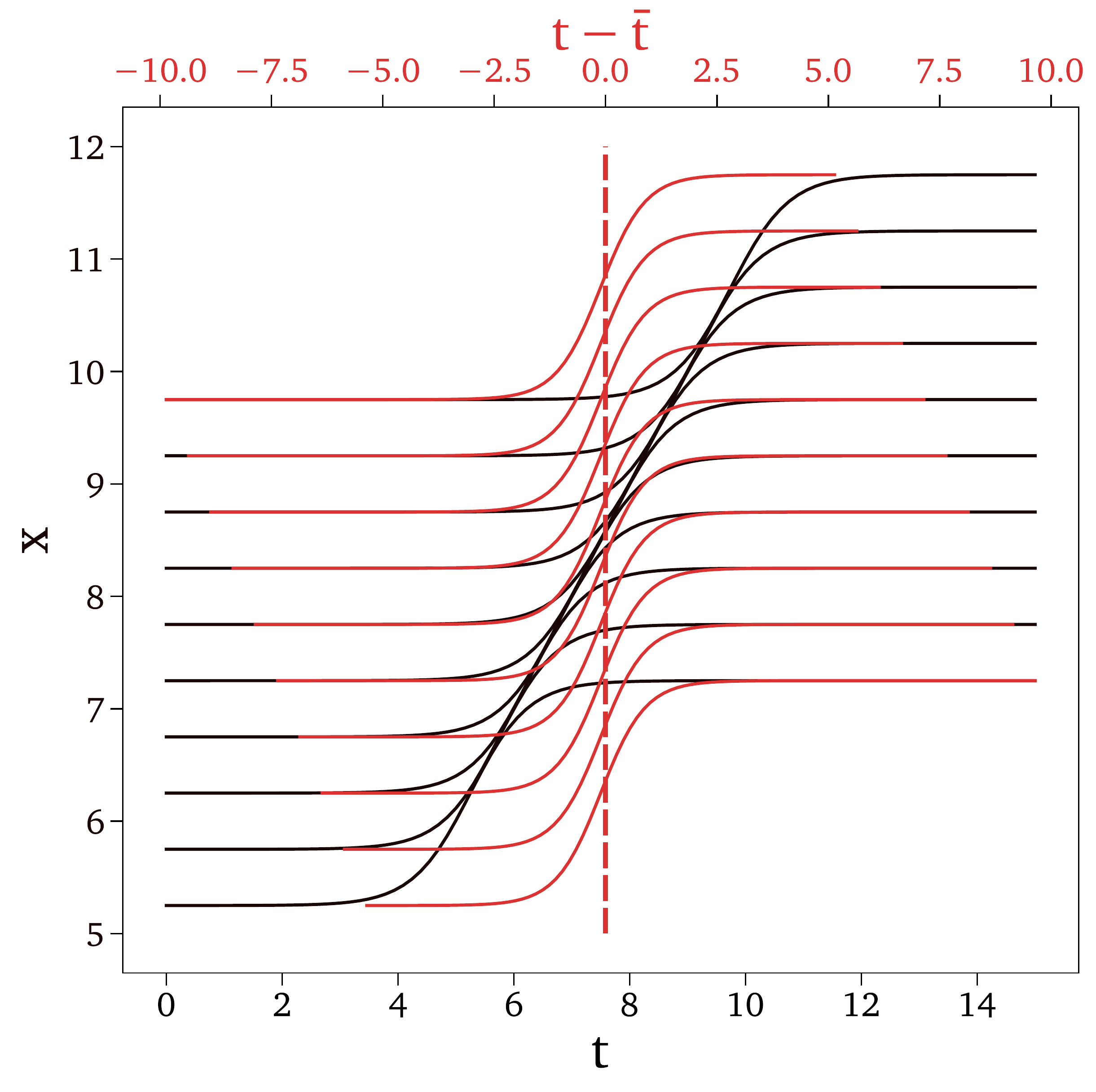}
    \end{subfigure}
    \begin{subfigure}{0.49\textwidth}
        \caption{}
        \label{fig:Tanh_flexDeepONet_Tests}
        \centering
        \includegraphics[width=3.2in]{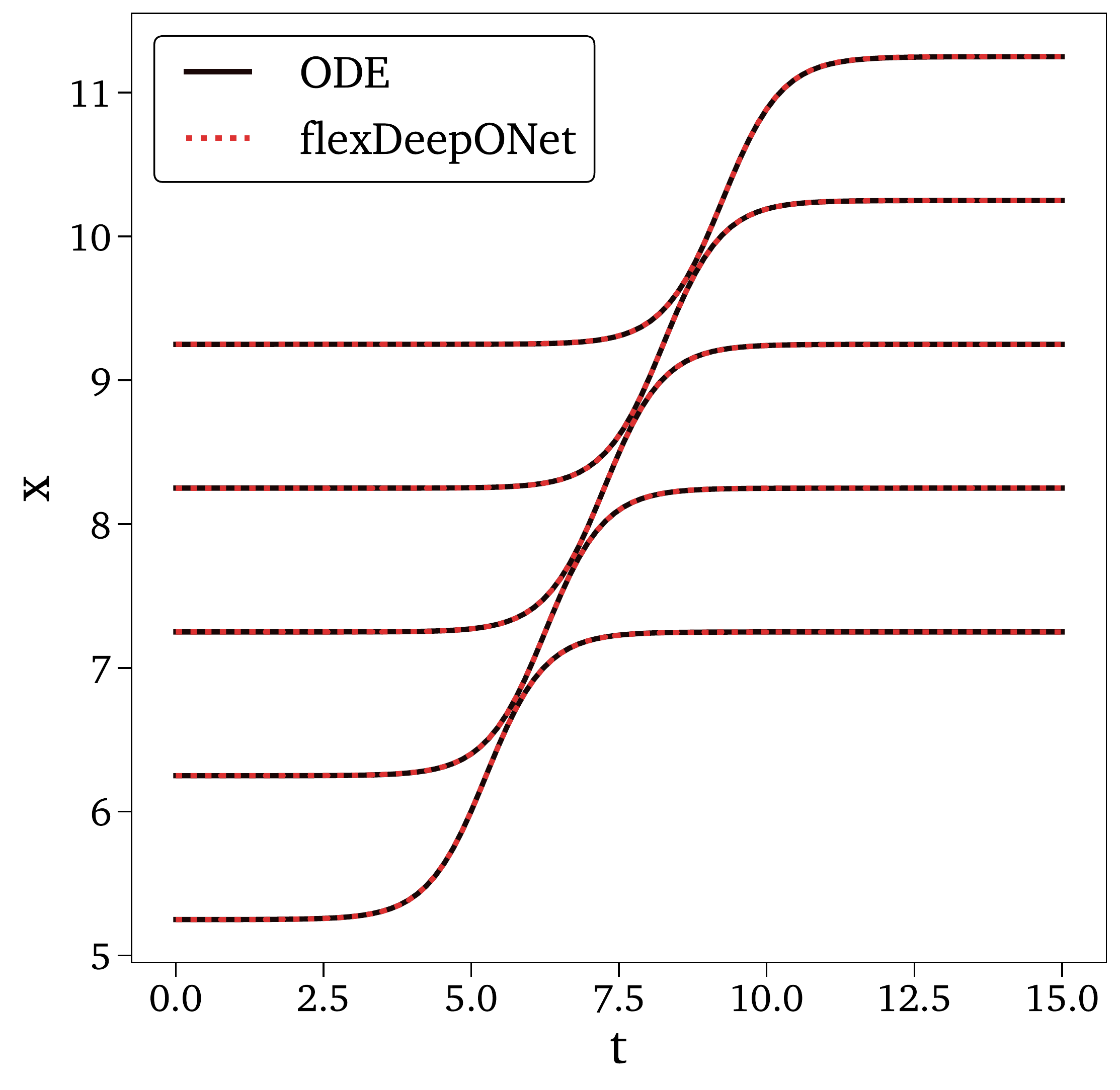}
    \end{subfigure}
    \caption{\textbf{FlexDeepONet applied to the shifting hyperbolic function test case}. (\textbf{A}): Time-integrated $x$ for the test scenarios as a function of the simulation time (black lines) and the simulation time corrected by the shift net (red lines). (\textbf{B}): $x$ at five test scenarios as the results of the ODE integration (solid black lines) and as predicted by flexDeepONet (red dotted lines)  with structure as in Fig.~\ref{fig:Tanh_flexDeepONet}.}
    \label{fig:Tanh_flexDeepONet_Results}
\end{figure}
Fig.~\ref{fig:Tanh_flexDeepONet} represents the flexDeepONet architecture for the shifting hyperbolic function test case, in which the generic Pre-Net is simplified to a shifting block with a single output. The depths and widths of trunk and branch nets are extremely decreased compared to the vanilla DeepONet employed above, and, in this new configuration, the trunk net has only one output. Two outputs are instead used for the branch net, one of which represents the centering coefficient, $c$, assumed to be a function of the initial conditions to improve the generalization capabilities of the architecture. As explained while discussing the first test case, the scaling $d$ is implicitly learned by the vanilla DeepONet as part of the $b_i$ outputs, and it is not considered here to avoid redundancies.
Fig.~\ref{fig:Tanh_flexDeepONet_Shift} reports the evolution of $x$ as a function of the original system of coordinates, $t$, and of the moving reference frame (i.e., $t$ translated by the output of the shift net, $\bar{t}$). The figure attests that the new block is capable of aligning the different scenarios. Because of this improvement, the flexDeepONet is efficient in projecting the dynamics on a one-dimensional basis (i.e., $p=1$), and it can predict the time evolution of $x$ with excellent accuracy (Fig.~\ref{fig:Tanh_flexDeepONet_Tests}) despite the remarkable fact that the number of trainable parameters is reduced by 99.5\% (i.e., from 4,880 to 22). \\
Finally, after having discussed the application of flexDeepONet to the test case under analysis, it is important to mention that the SVD's deficiency in processing translated, rotated, and shifted data is also passed on to the derived techniques, as for POD-based methods~\cite{SymmPOD_Brunton_2019}. For instance, researchers have extensively worked on developing suitable remedies for the challenges resulting from the symmetries in transport-dominated flows~\cite{SymmPOD_Cagniart_2016,SymmPOD_Mojgani_2017,SymmPOD_Rim_2018,SymmPOD_Reiss_2018,SymmPOD_Nair_2019,SymmPOD_Taddei_2020,SymmPOD_Papapicco_2022}. Some of these techniques share strong similarities with the pre-transformation network presented in this work. In particular, it should be mentioned the analogy with the Neural Network shifted-POD (NNsPOD) recently proposed by Papapicco \textit{et al.}~\cite{SymmPOD_Papapicco_2022}, which extends the work of Reiss \textit{et al.}~\cite{SymmPOD_Reiss_2018,SymmPOD_Reiss_2_2018} on constructing continuous shift operators for POD by automatizing the detection of transport velocities.


\subsection{Test Case 3: Combustion Chemistry in an Isobaric Reactor}

\begin{figure}
    \centering
    \includegraphics[width=3.2in]{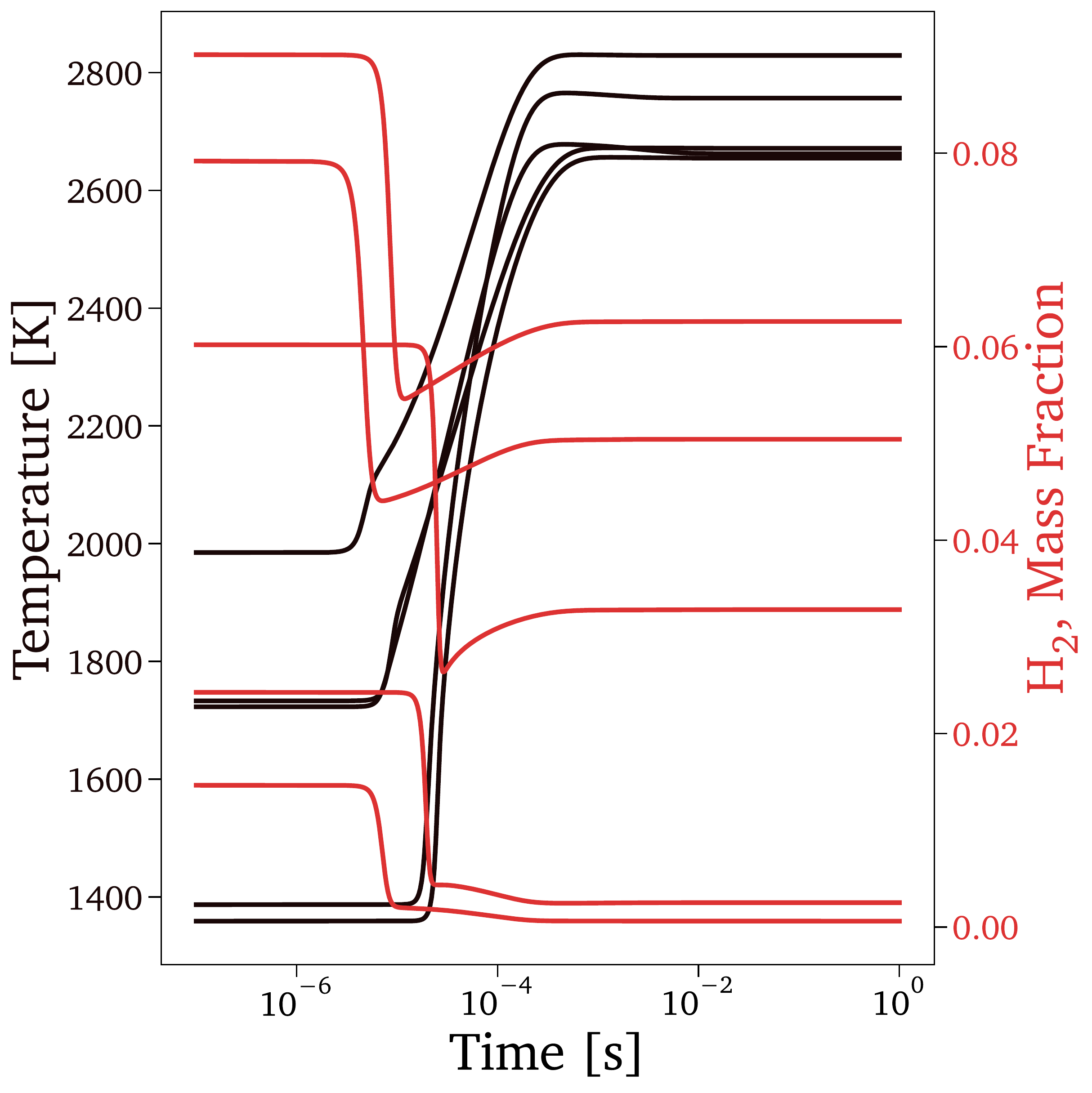}
    \caption{\textbf{Training data for the combustion chemistry test case}. Five examples of training scenarios showing the evolution of temperature and H$_2$ mass fraction. See Fig.~\ref{fig:0DReact_Train_HandNH3} for the mass fractions of additional species.}
    \label{fig:0DReact_Train_TandH2}
\end{figure}

\begin{figure}[!htb]
    \begin{subfigure}{0.49\textwidth}
        \centering
        \caption{}
        \label{fig:0DReact_DeepONet_20_T}
        \includegraphics[width=3.2in]{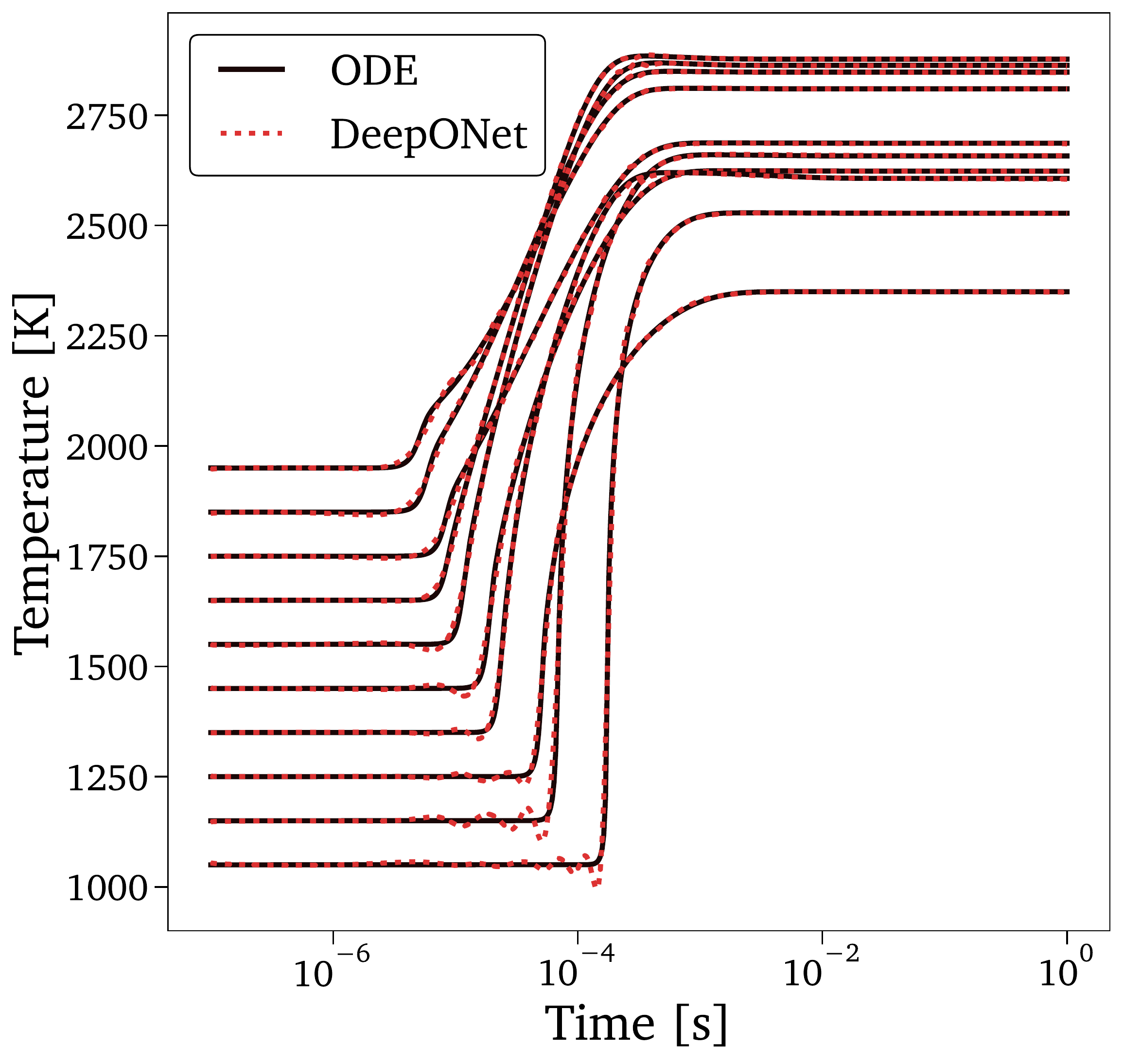}
    \end{subfigure}
    \begin{subfigure}{0.49\textwidth}
        \centering
        \caption{}
        \label{fig:0DReact_DeepONet_20_H2}
        \includegraphics[width=3.2in]{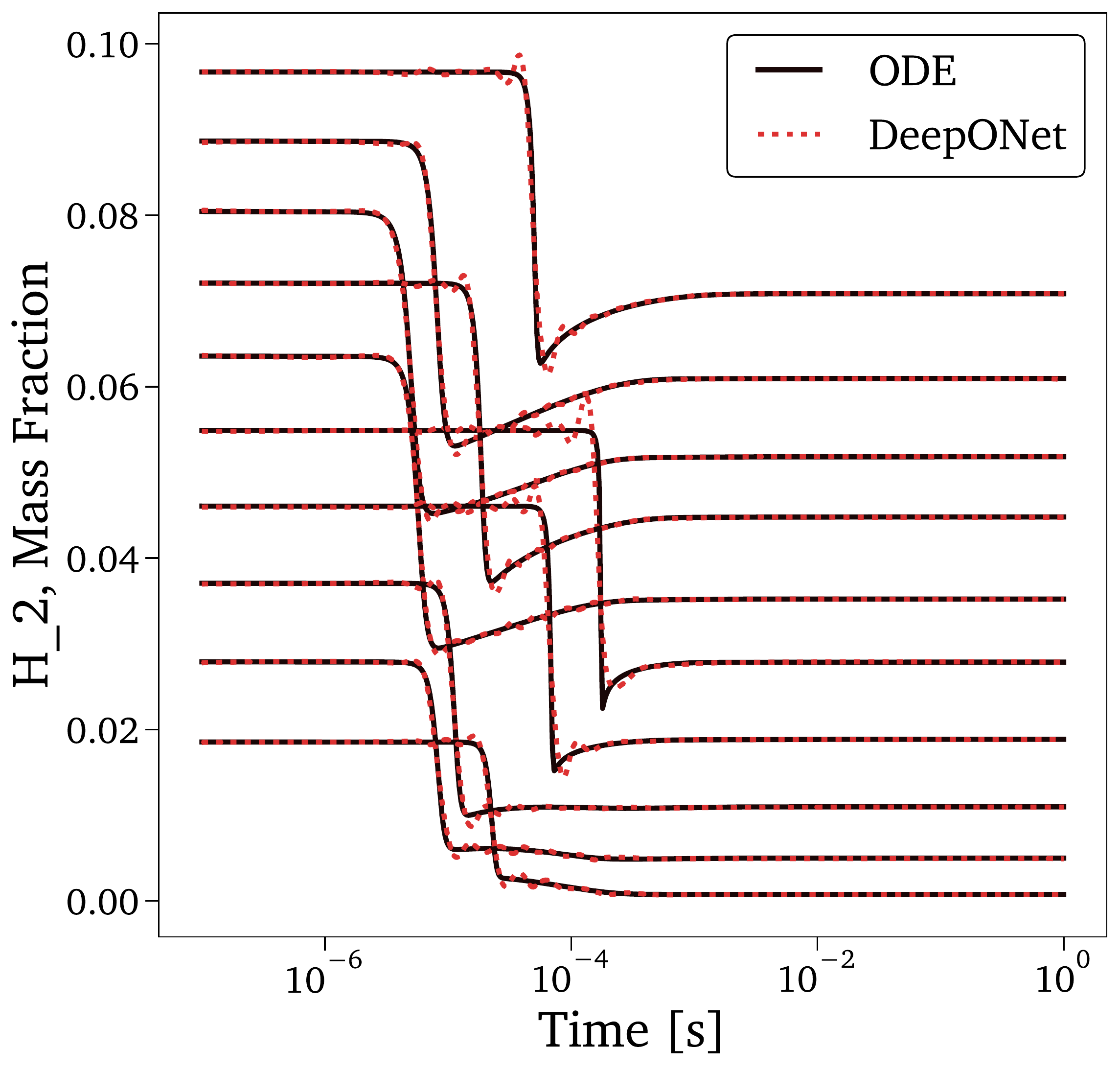}
    \end{subfigure}
    \begin{subfigure}{0.49\textwidth}
        \centering
        \caption{}
        \label{fig:0DReact_DeepONet_20_H}
        \includegraphics[width=3.2in]{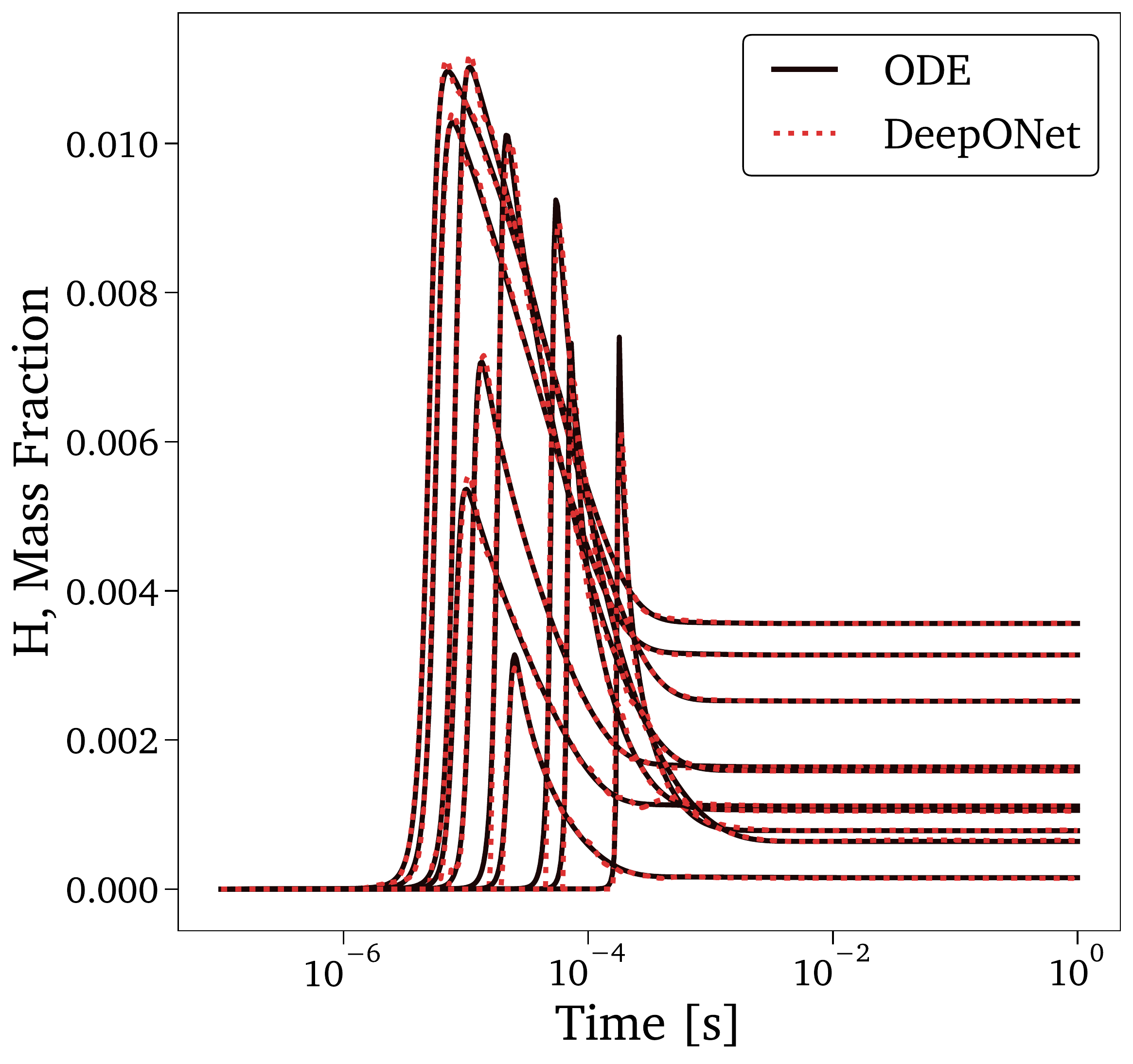}
    \end{subfigure}
    \begin{subfigure}{0.49\textwidth}
        \centering
        \caption{}
        \label{fig:0DReact_DeepONet_20_OH}
        \includegraphics[width=3.2in]{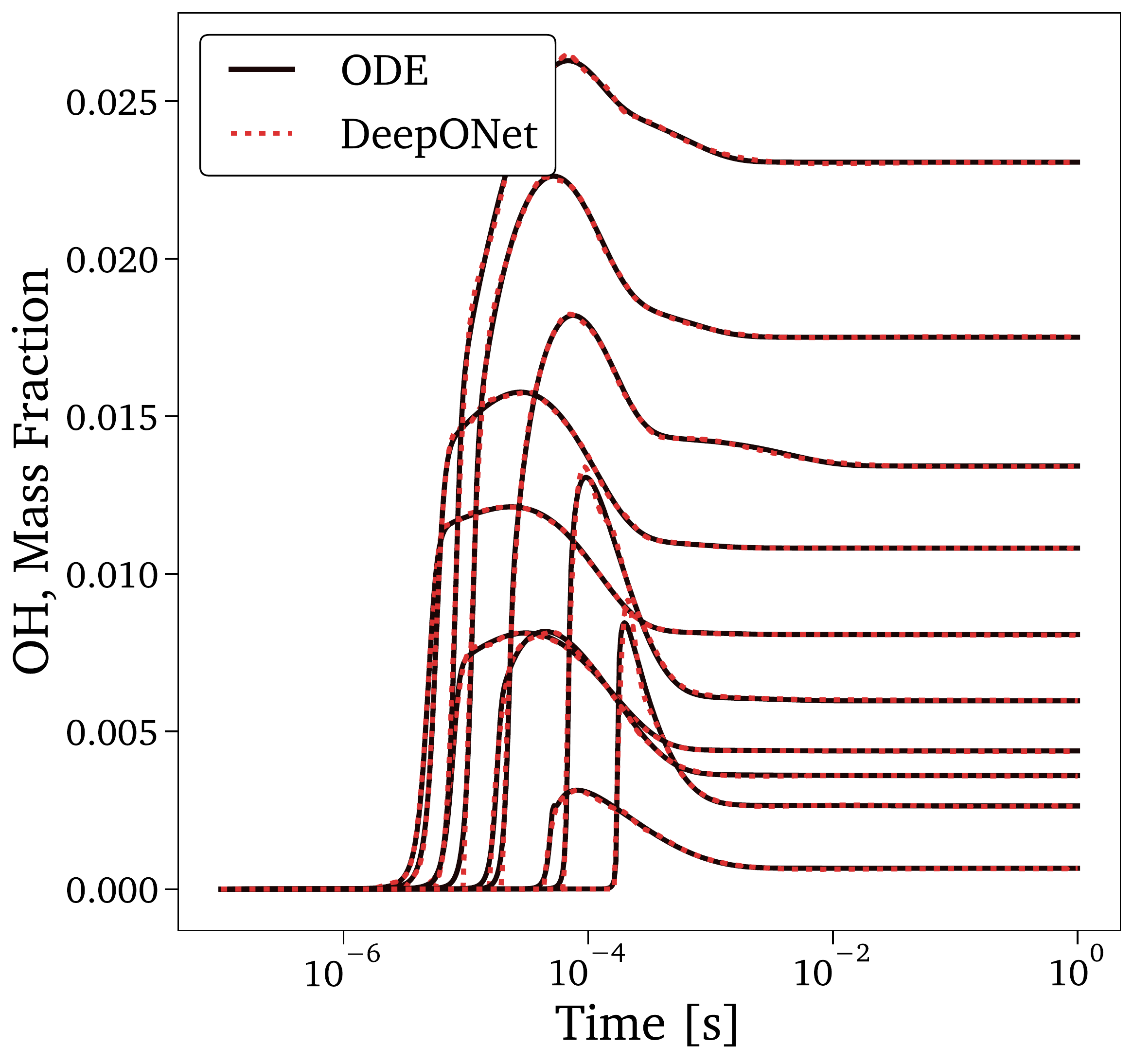}
    \end{subfigure}
    \caption{\textbf{Predictions of the thermodynamic variables from the Vanilla DeepONet with $p=20$}. Temperature (\textbf{A}), and mass fractions of H$_2$ (\textbf{B}), H (\textbf{C}), and OH (\textbf{D}) for ten test scenarios as the results of the ODE integration (solid black lines) and as predicted by the Vanilla DeepONet with 20 trunk outputs (red dotted lines).}
    \label{fig:0DReact_DeepONet_y_20}
\end{figure}

\begin{figure}[!htb]
    \begin{subfigure}{0.49\textwidth}
        \centering
        \caption{}
        \label{fig:0DReact_DeepONet_T}
        \includegraphics[width=3.2in]{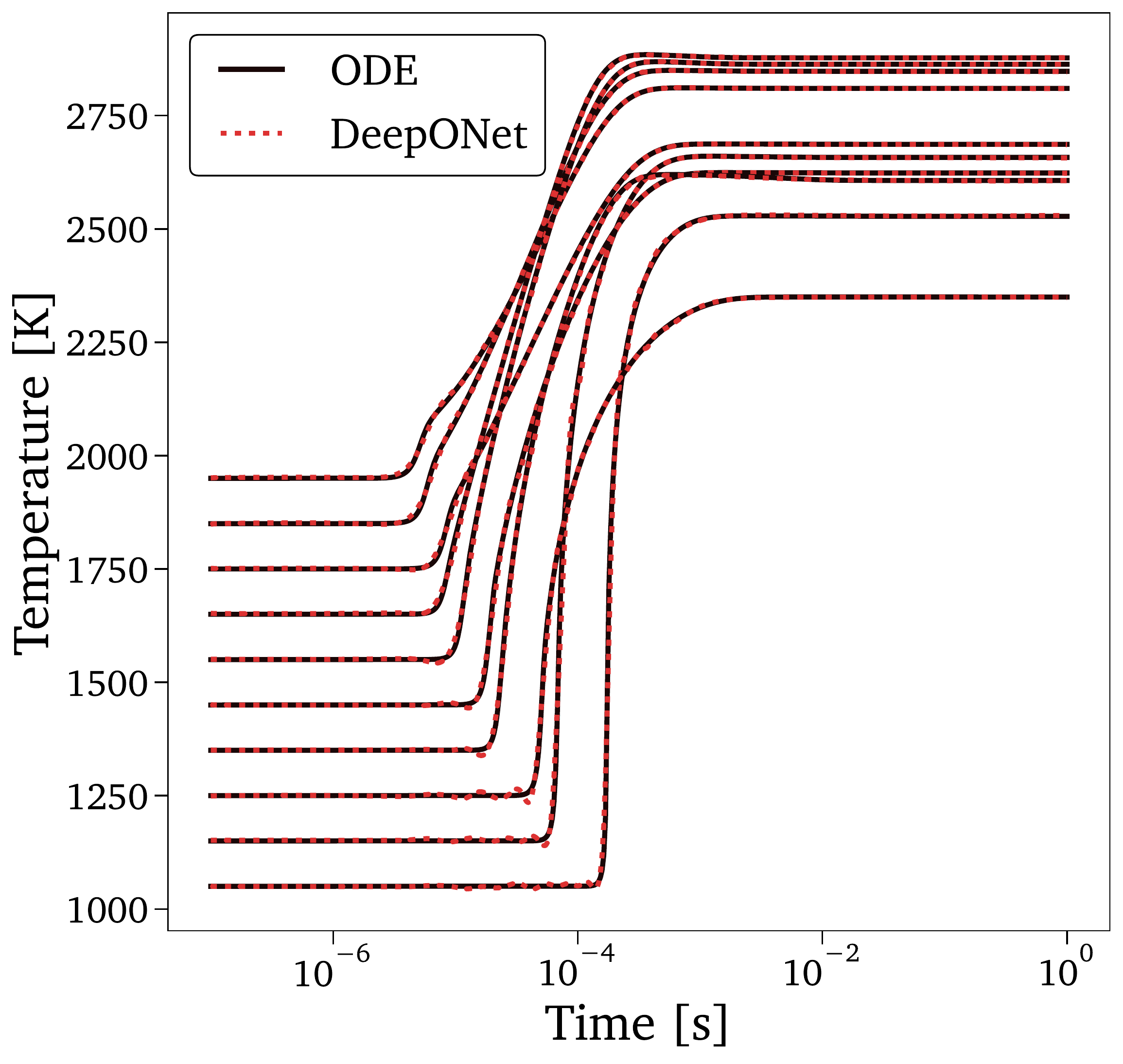}
    \end{subfigure}
    \begin{subfigure}{0.49\textwidth}
        \centering
        \caption{}
        \label{fig:0DReact_DeepONet_H2}
        \includegraphics[width=3.2in]{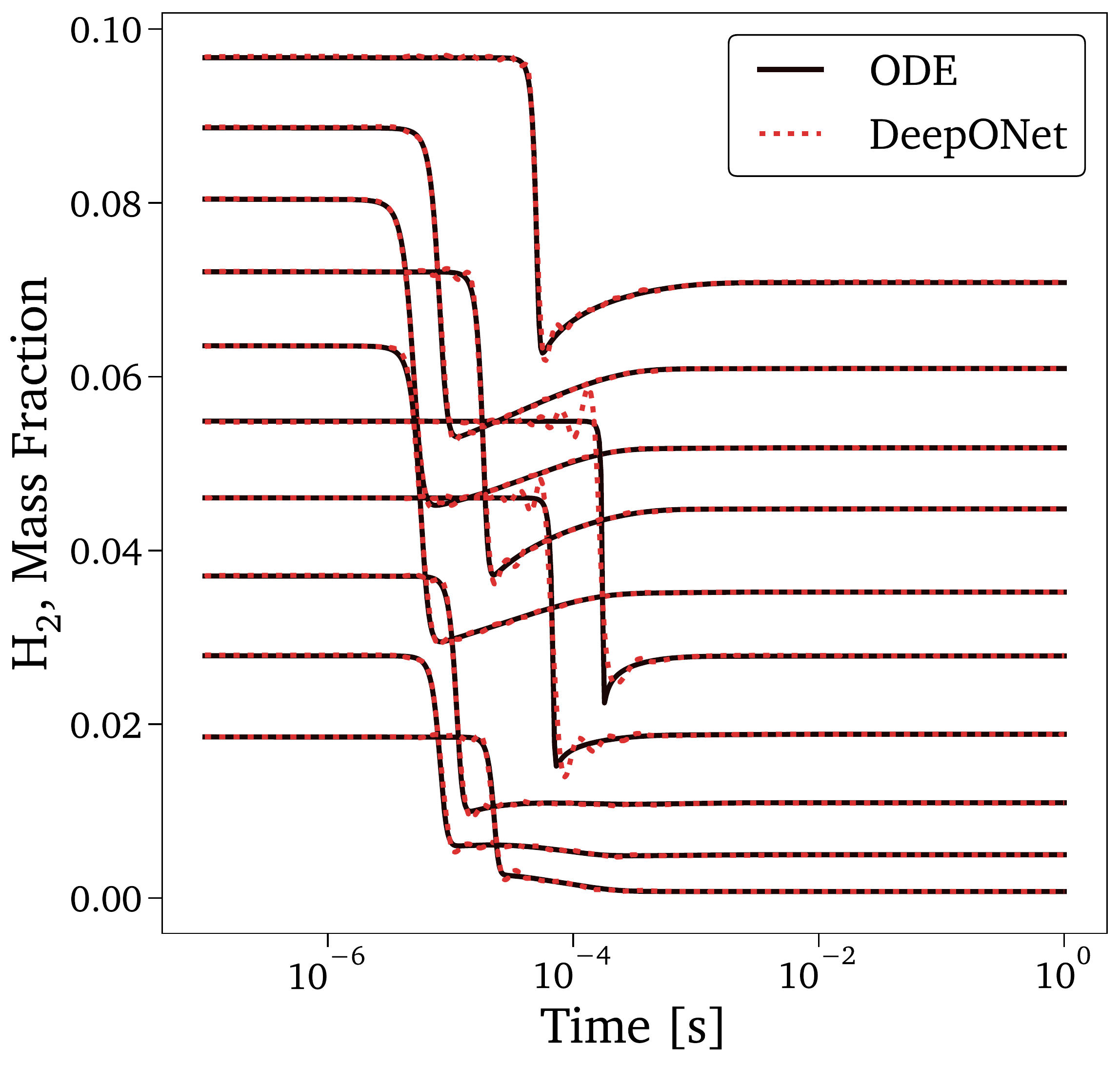}
    \end{subfigure}
    \begin{subfigure}{0.49\textwidth}
        \centering
        \caption{}
        \label{fig:0DReact_DeepONet_H}
        \includegraphics[width=3.2in]{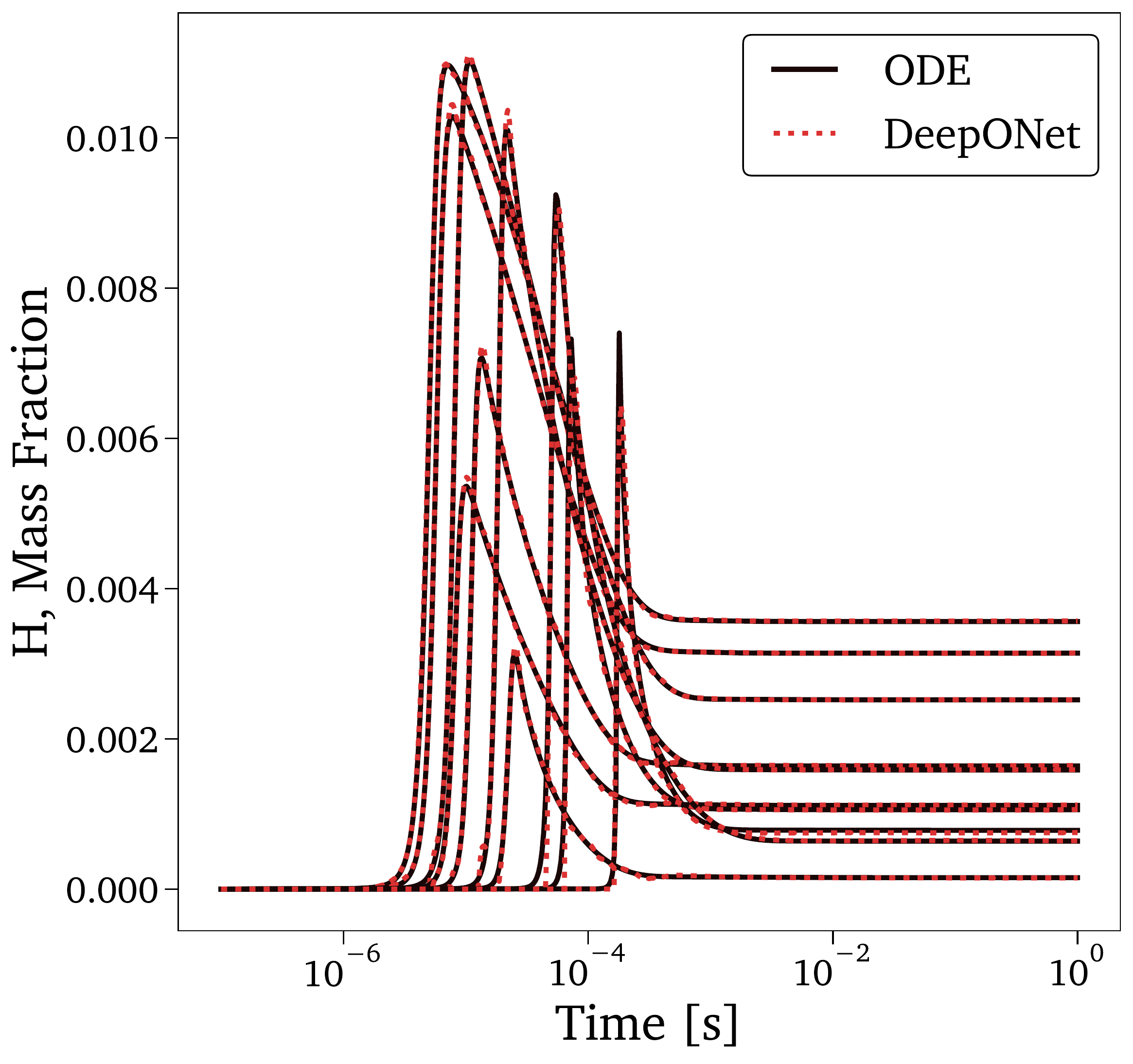}
    \end{subfigure}
    \begin{subfigure}{0.49\textwidth}
        \centering
        \caption{}
        \label{fig:0DReact_DeepONet_OH}
        \includegraphics[width=3.2in]{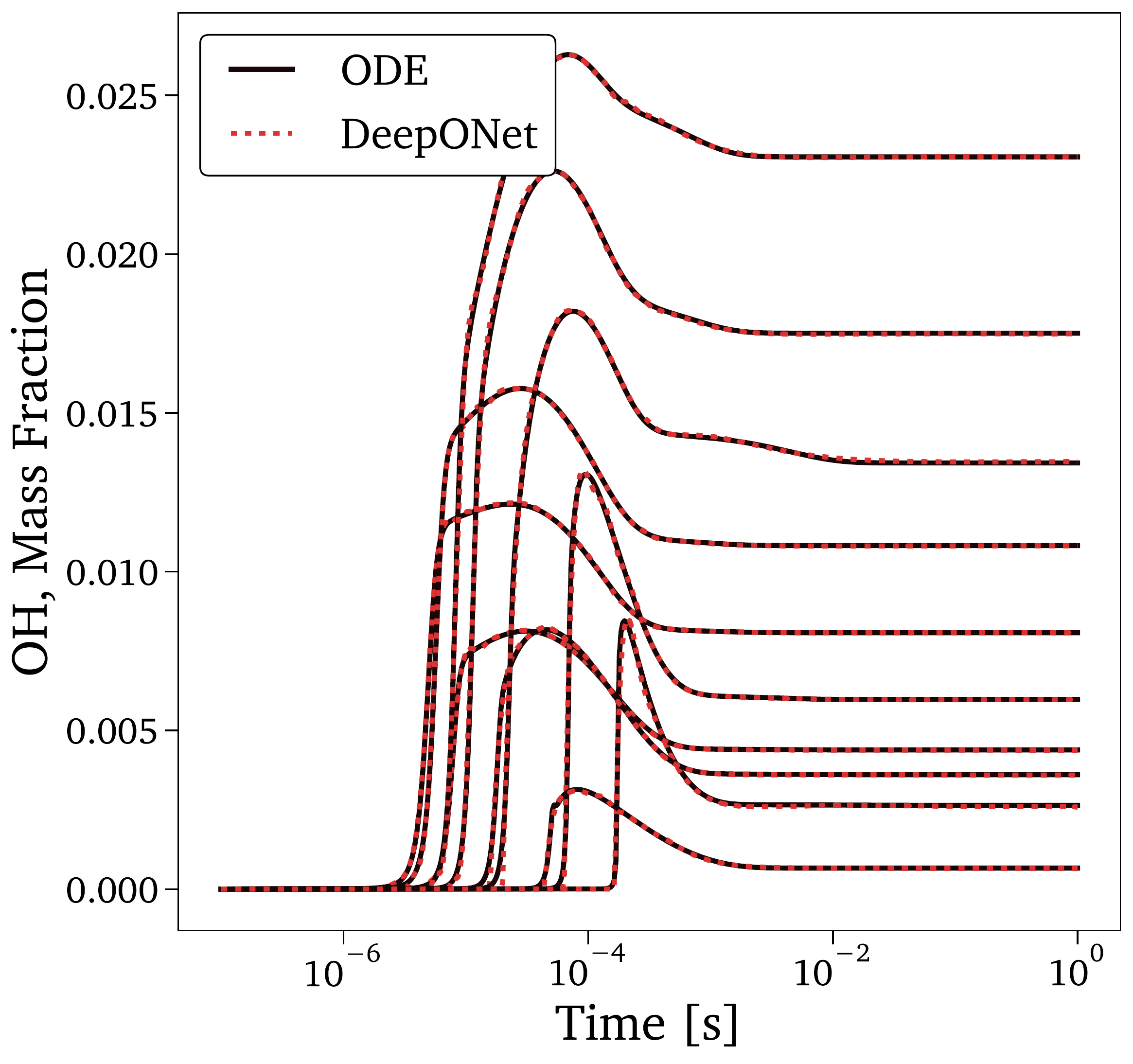}
    \end{subfigure}
    \caption{\textbf{Predictions of the thermodynamic variables from the vanilla DeepONet with $p=32$}. Temperature (\textbf{A}), and mass fractions of H$_2$ (\textbf{B}), H (\textbf{C}), and OH (\textbf{D}) for ten test scenarios as the results of the ODE integration (solid black lines) and as predicted by vanilla DeepONet with 32 trunk outputs (red dotted lines).}
    \label{fig:0DReact_DeepONet_y_32}
\end{figure}

\begin{figure}[!htb]
    \begin{subfigure}{0.49\textwidth}
        \centering
        \caption{}
        \label{fig:0DReact_CumEnergy_T}
        \includegraphics[width=3.2in]{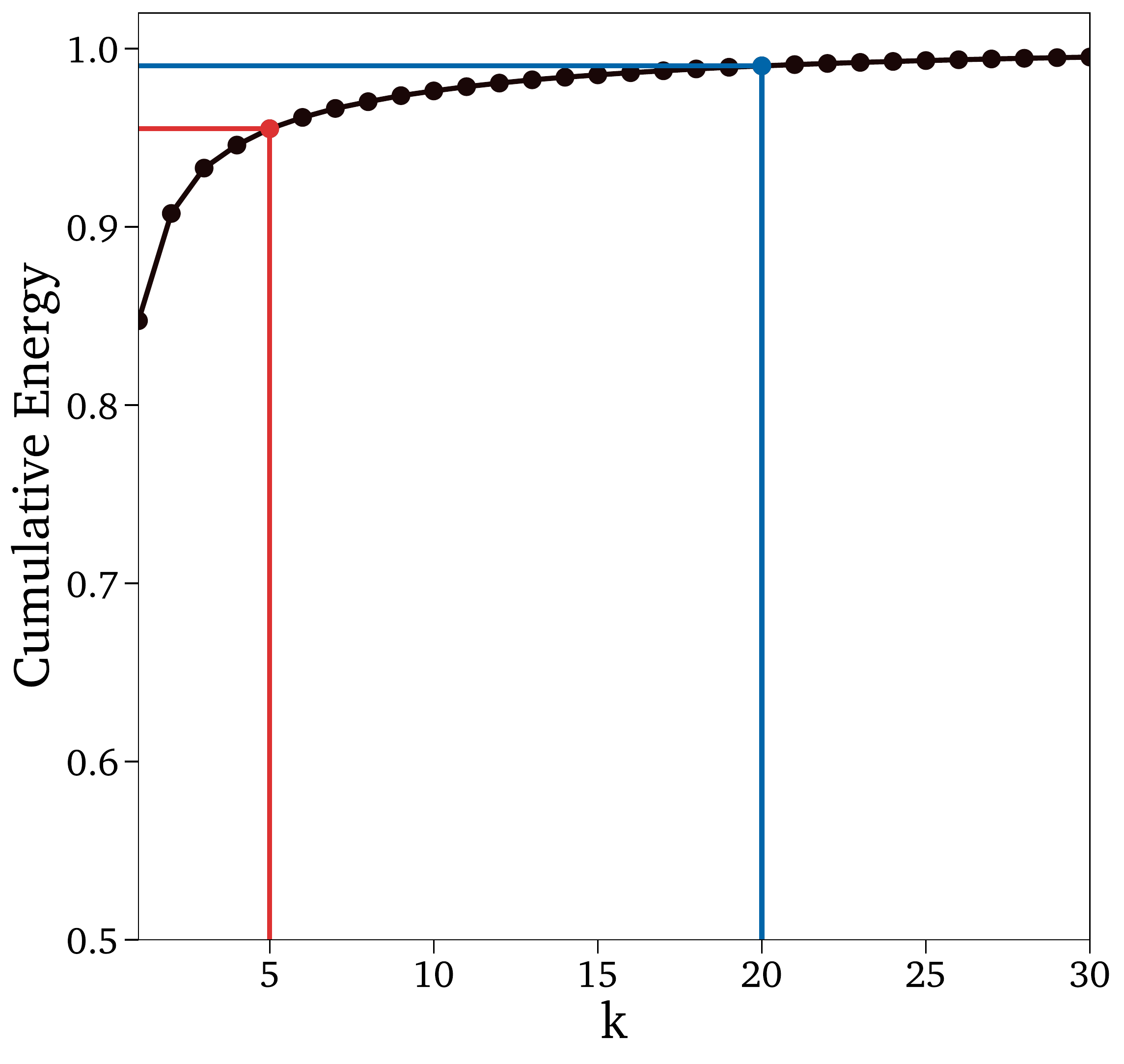}
    \end{subfigure}
    \begin{subfigure}{0.49\textwidth}
        \centering
        \caption{}
        \label{fig:0DReact_CumEnergy_H2}
        \includegraphics[width=3.2in]{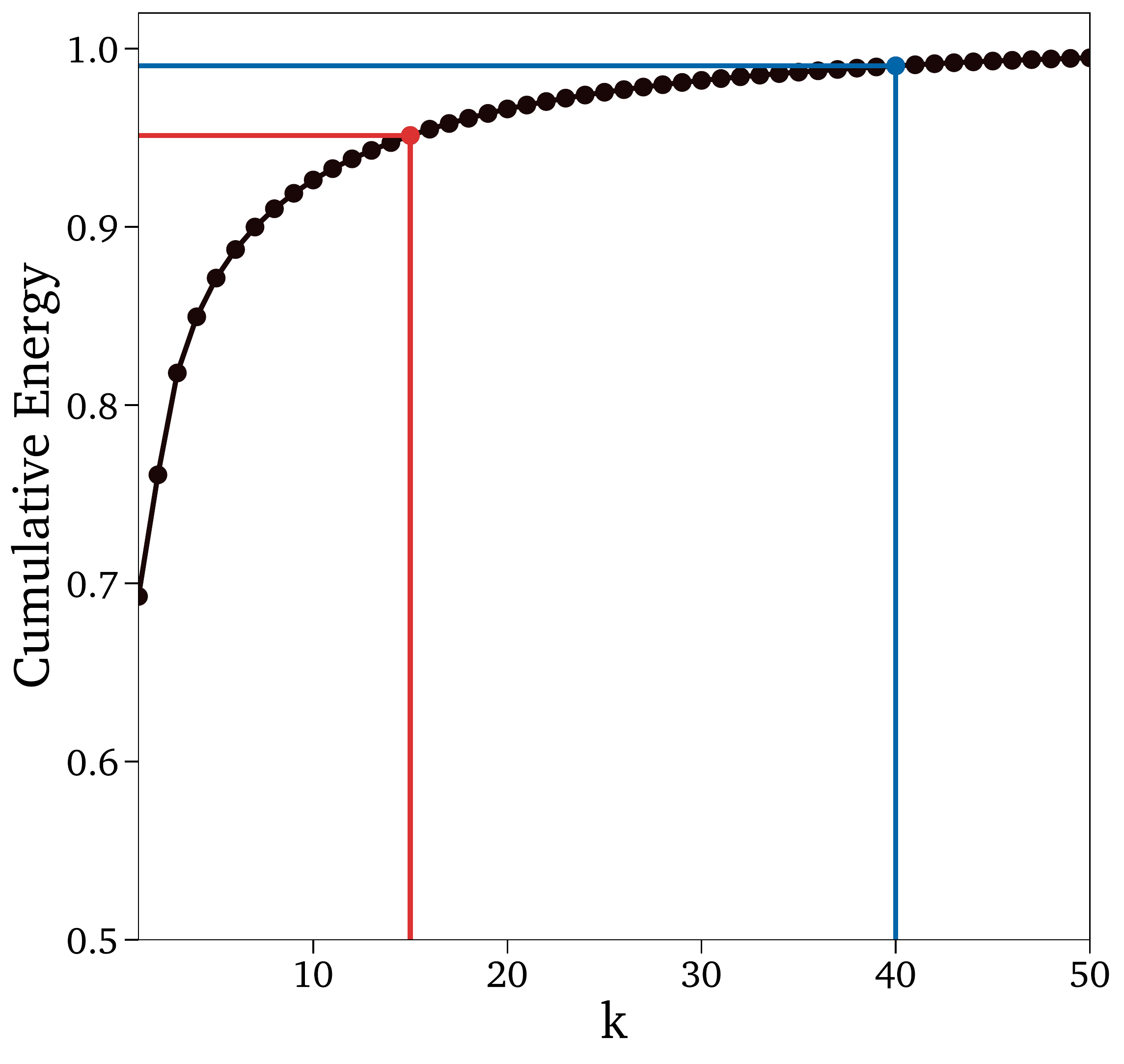}
    \end{subfigure}
    \begin{subfigure}{0.49\textwidth}
        \centering
        \caption{}
        \label{fig:0DReact_CumEnergy_H}
        \includegraphics[width=3.2in]{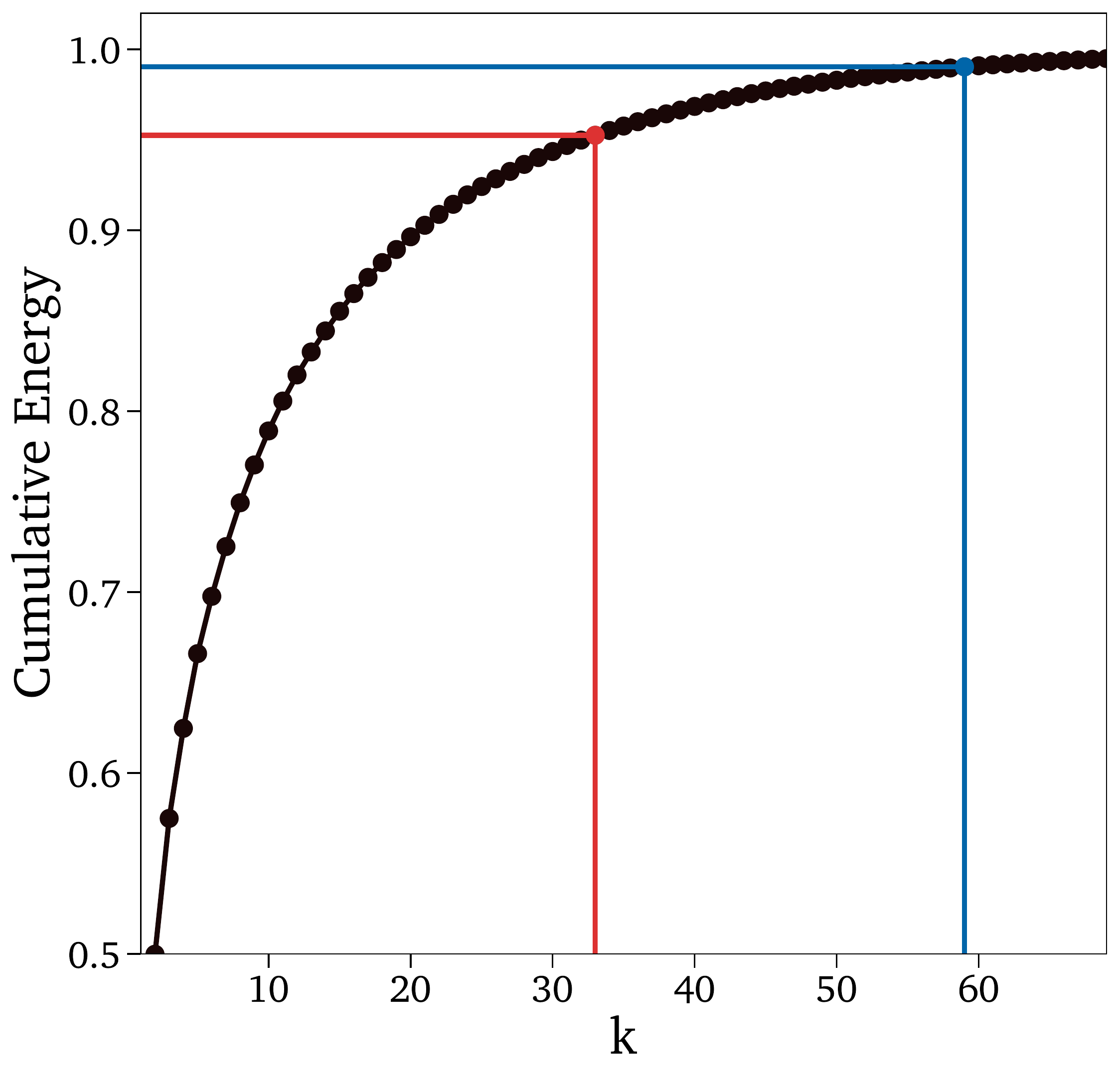}
    \end{subfigure}
    \begin{subfigure}{0.49\textwidth}
        \centering
        \caption{}
        \label{fig:0DReact_CumEnergy_OH}
        \includegraphics[width=3.2in]{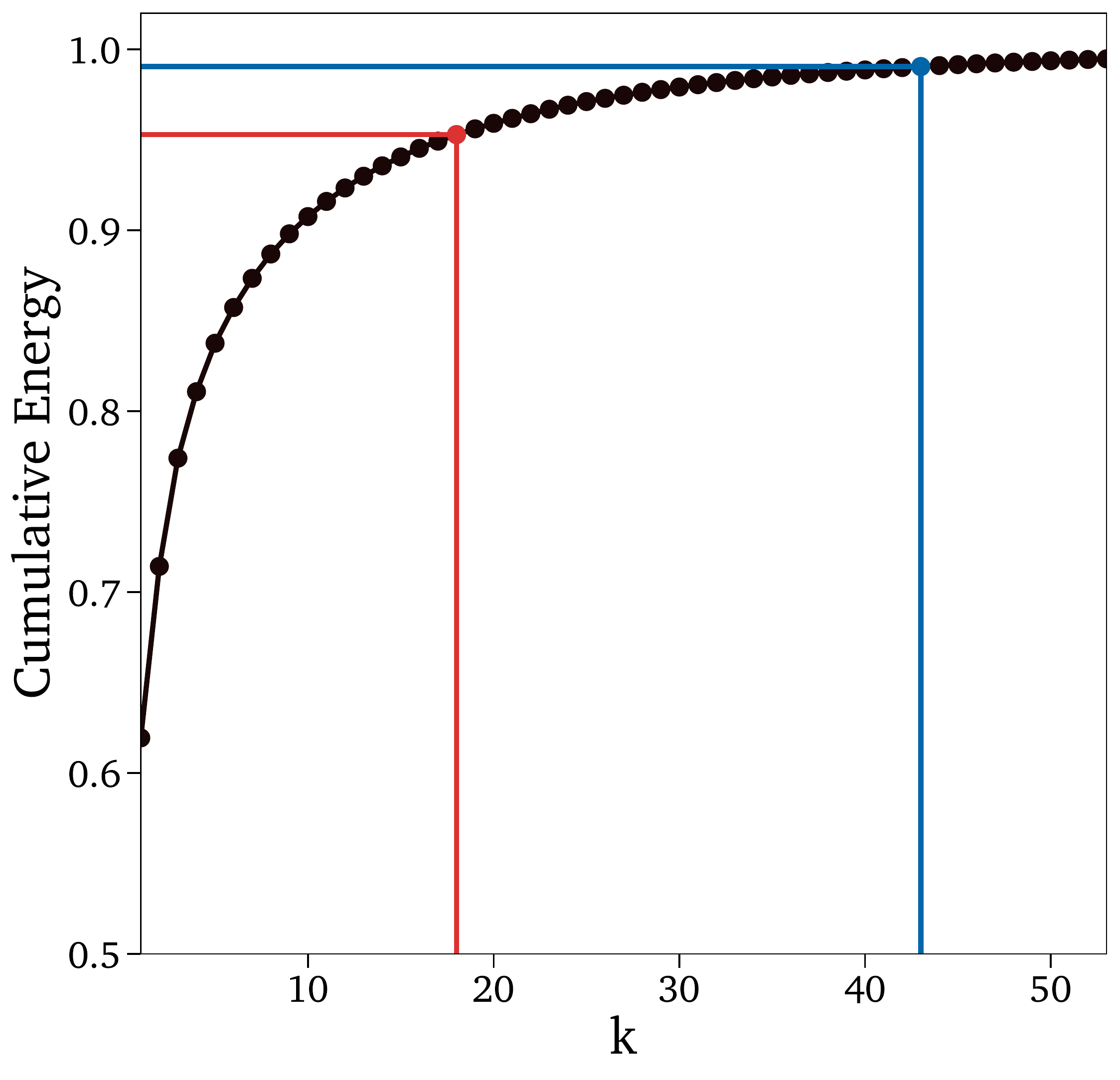}
    \end{subfigure}
    \caption{\textbf{Cumulative energies of the scenario-aggregated data matrices from the combustion chemistry test case}. Cumulative energies contained in the first k singular values of the matrices for the temperature (\textbf{A}) and the mass fractions of H$_2$ (\textbf{B}), H (\textbf{C}), and OH (\textbf{D}). The red and blue dots identify cumulative energies corresponding to 95\% and 99\%, respectively.}
    \label{fig:0DReact_CumEnergies}
\end{figure}

\begin{figure}[!htb]
    \begin{subfigure}{0.49\textwidth}
        \centering
        \caption{}
        \label{fig:0DReact_Reconstructed_T_20}
        \includegraphics[width=3.2in]{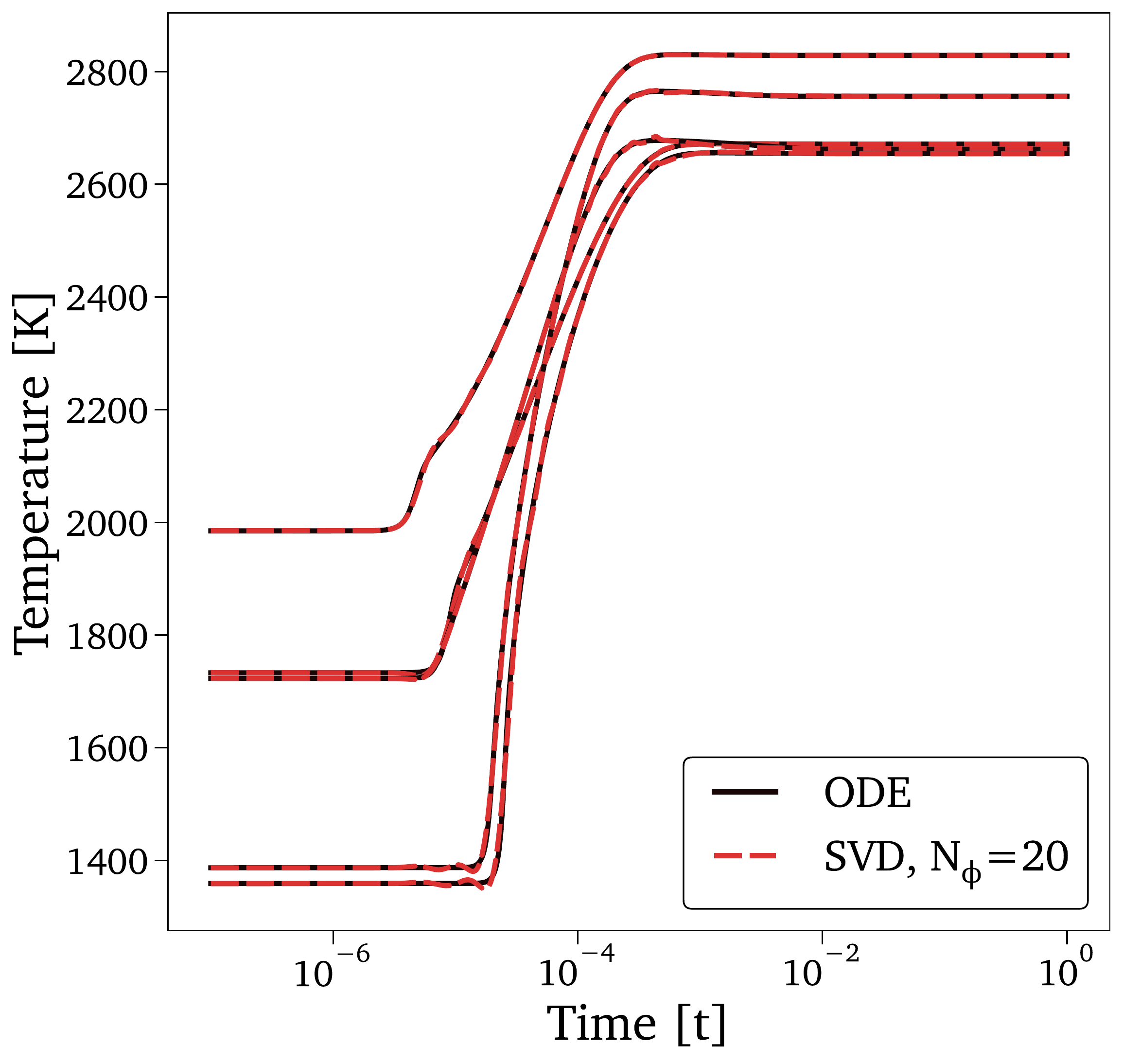}
    \end{subfigure}
    \begin{subfigure}{0.49\textwidth}
        \centering
        \caption{}
        \label{fig:0DReact_Reconstructed_H2_20}
        \includegraphics[width=3.2in]{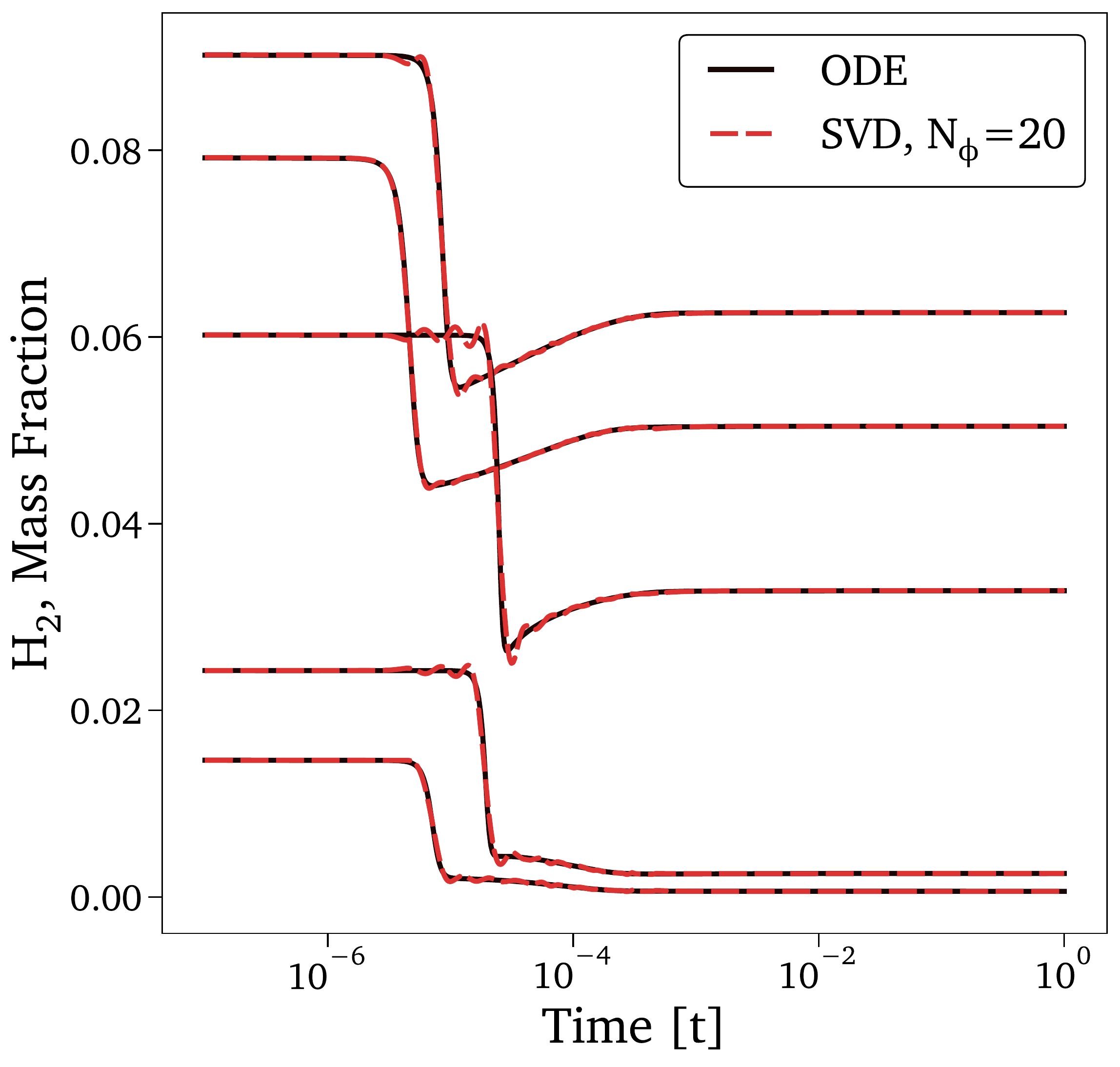}
    \end{subfigure}
    \begin{subfigure}{0.49\textwidth}
        \centering
        \caption{}
        \label{fig:0DReact_Reconstructed_H_20}
        \includegraphics[width=3.2in]{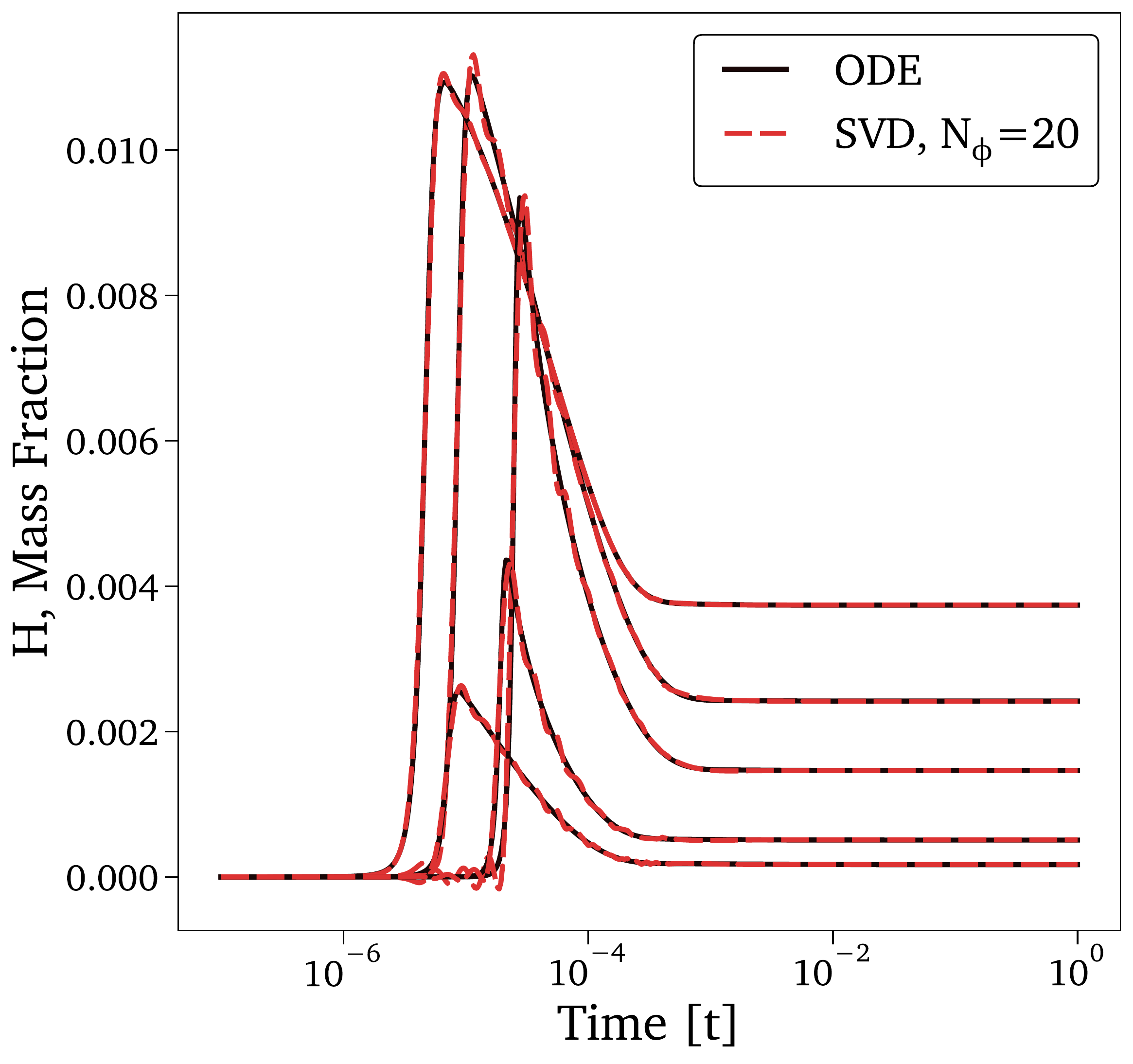}
    \end{subfigure}
    \begin{subfigure}{0.49\textwidth}
        \centering
        \caption{}
        \label{fig:0DReact_Reconstructed_OH_20}
        \includegraphics[width=3.2in]{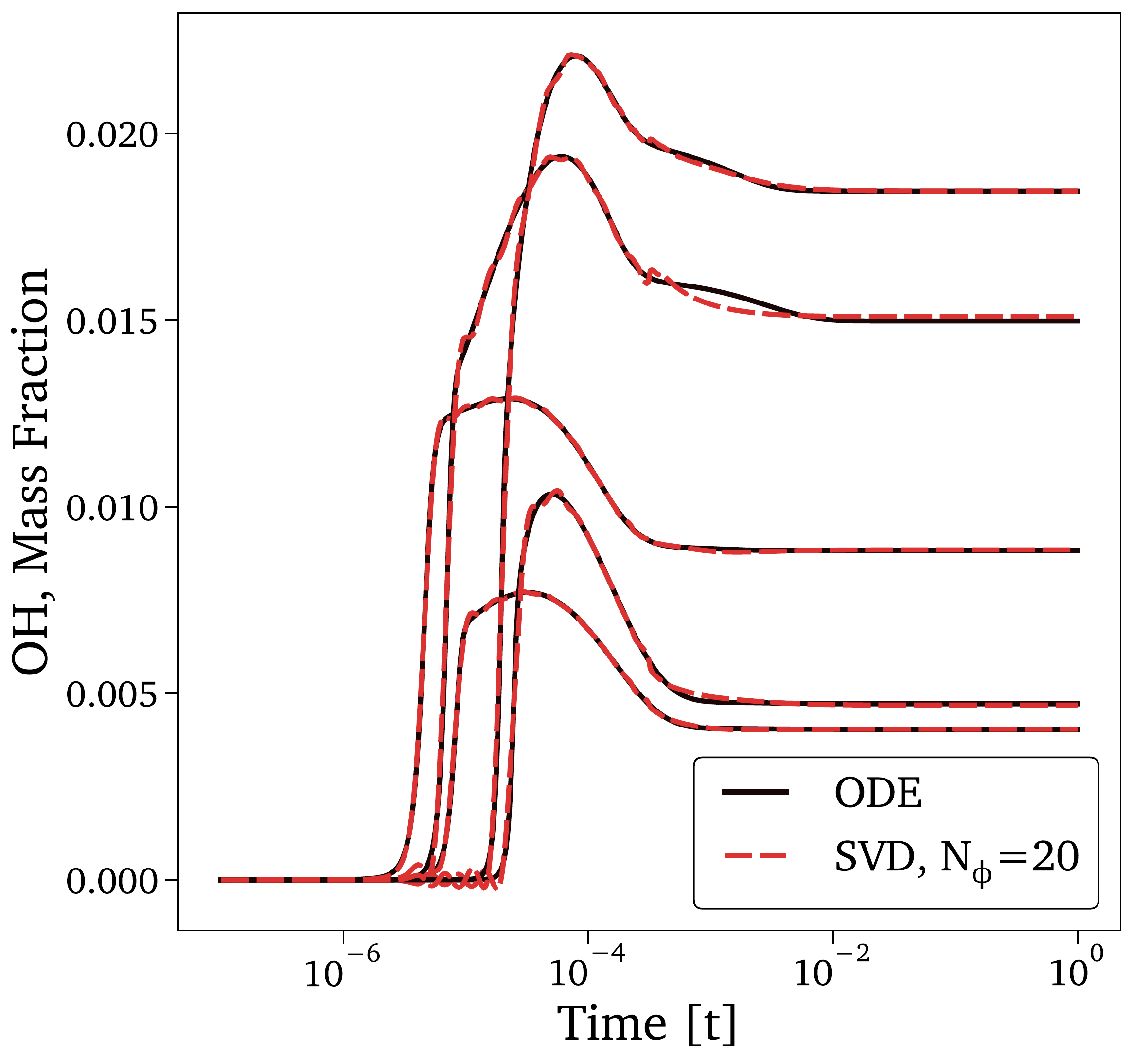}
    \end{subfigure}
    \caption{\textbf{Training scenarios decomposed and reconstructed via SVD}. Five of the training scenarios from the ODE integration (black solid lines) and after being encoded-decoded based on SVD's first twenty singular values (red dotted lines).}
    \label{fig:0DReact_Reconstructed_20}
\end{figure}

\begin{figure}[!htb]
    \begin{subfigure}{0.49\textwidth}
        \centering
        \caption{}
        \label{fig:0DReact_Reconstructed_T_32}
        \includegraphics[width=3.2in]{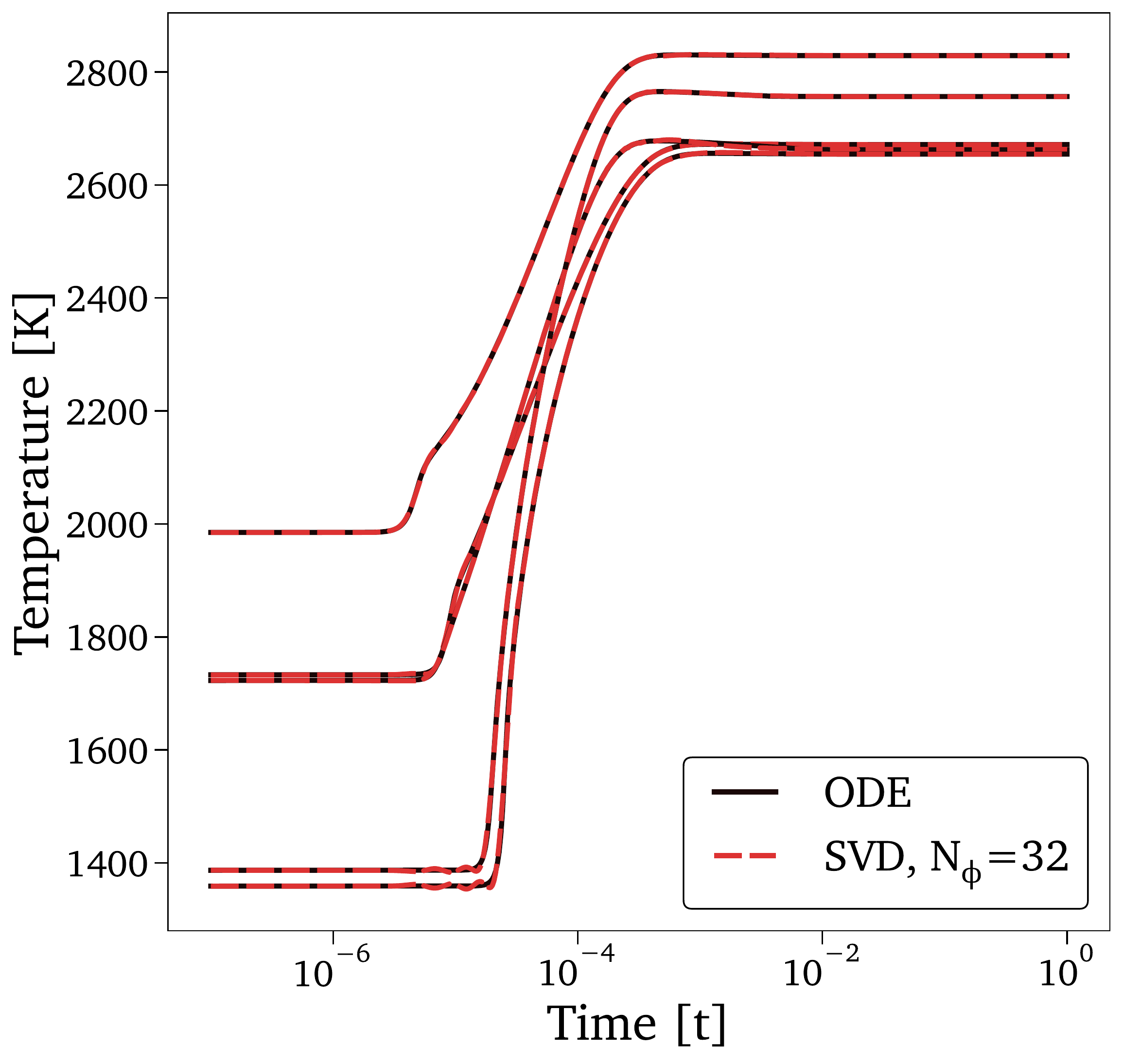}
    \end{subfigure}
    \begin{subfigure}{0.49\textwidth}
        \centering
        \caption{}
        \label{fig:0DReact_Reconstructed_H2_32}
        \includegraphics[width=3.2in]{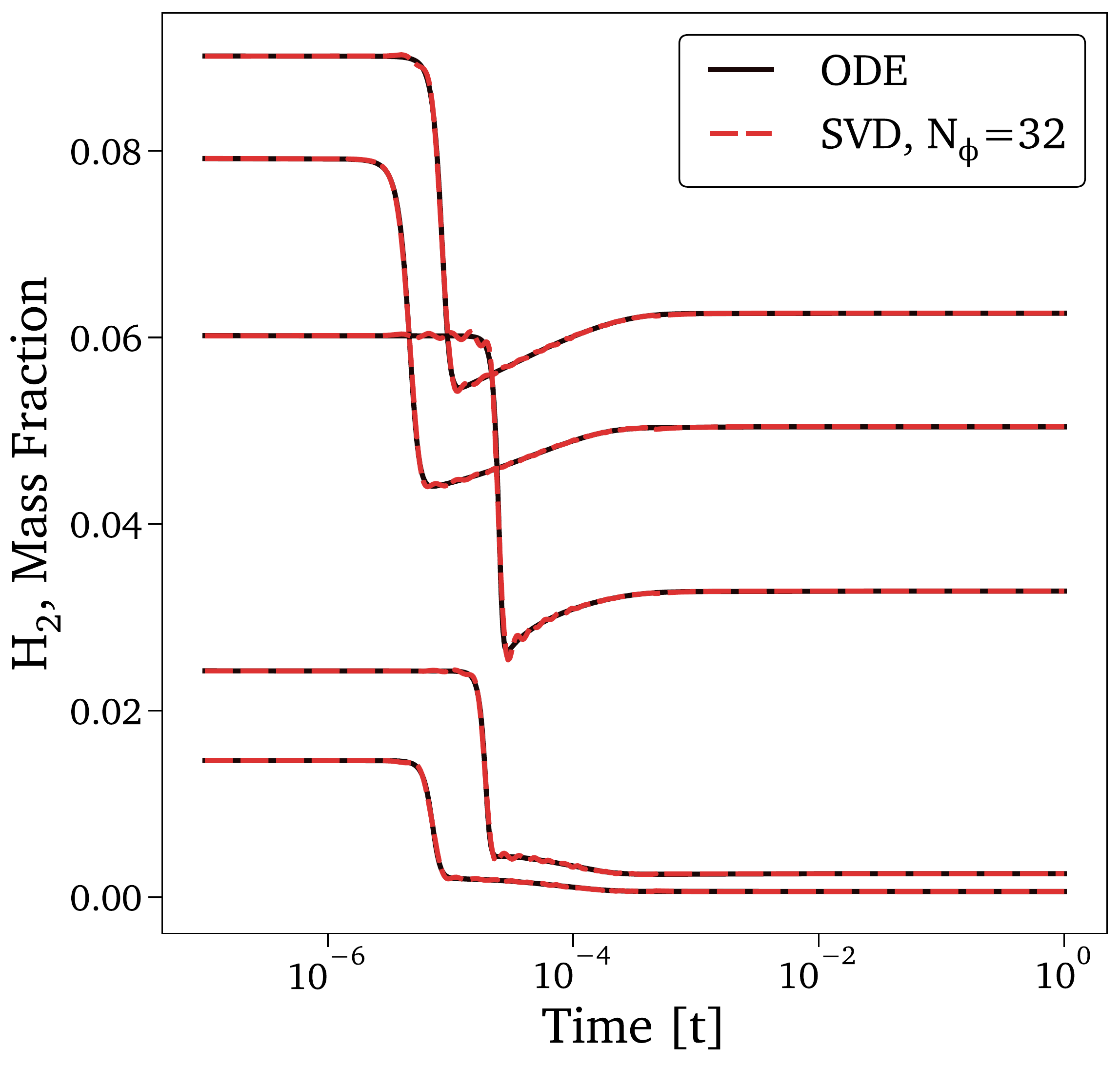}
    \end{subfigure}
    \begin{subfigure}{0.49\textwidth}
        \centering
        \caption{}
        \label{fig:0DReact_Reconstructed_H_32}
        \includegraphics[width=3.2in]{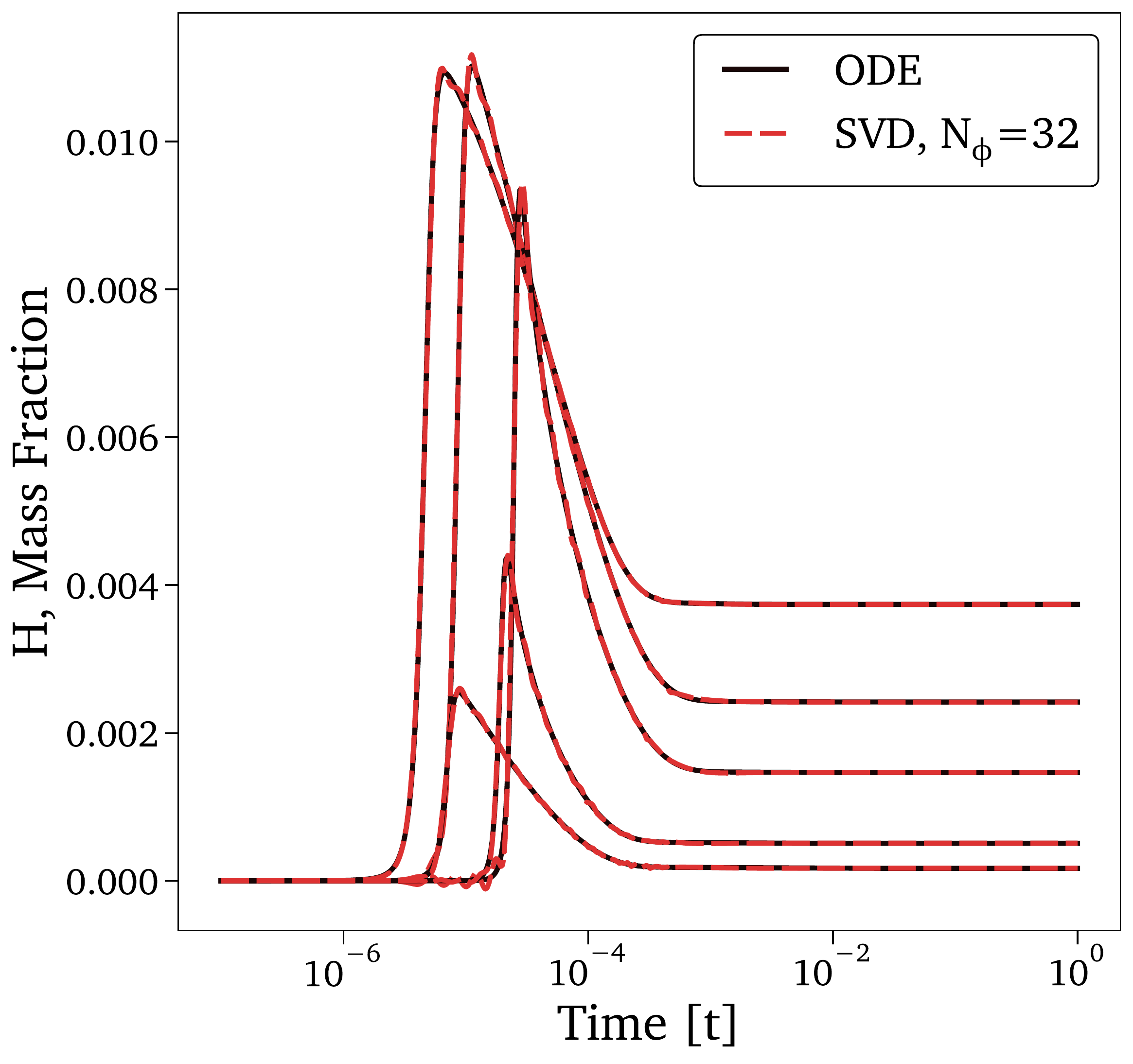}
    \end{subfigure}
    \begin{subfigure}{0.49\textwidth}
        \centering
        \caption{}
        \label{fig:0DReact_Reconstructed_OH_32}
        \includegraphics[width=3.2in]{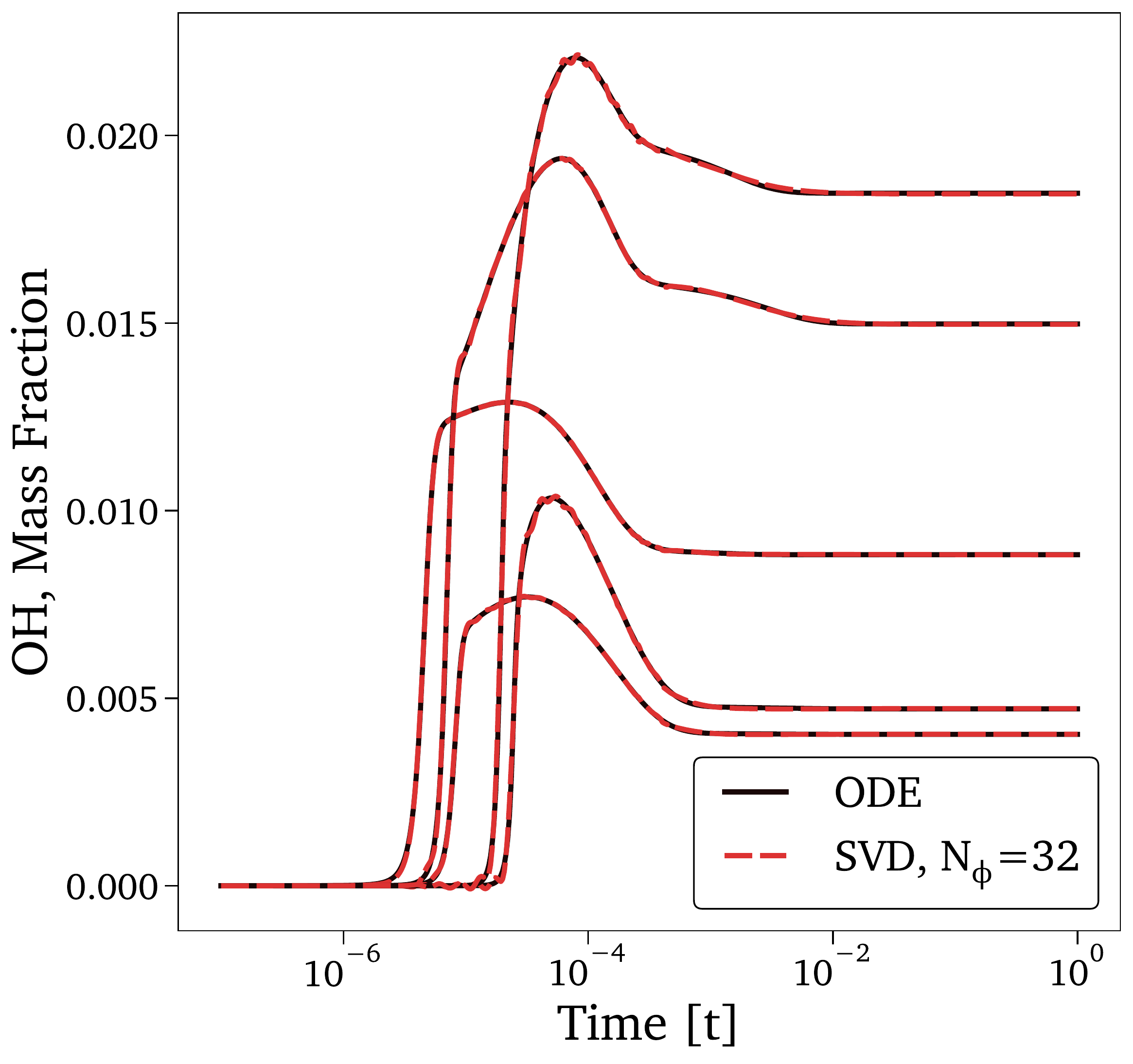}
    \end{subfigure}
    \caption{\textbf{Training scenarios decomposed and reconstructed via SVD}. Five of the training scenarios from the ODE integration (black solid lines) and after being encoded-decoded based on SVD's first thirty two singular values (red dotted lines).}
    \label{fig:0DReact_Reconstructed_32}
\end{figure}

The third test case focuses on a chemical dynamical system of practical relevance, involving the analysis of combustion chemistry dynamics taking place in a high-temperature zero-dimensional isobaric reactor. The chemical system is initially assumed to contain only \ce{H2} and air (assumed to be composed of \ce{N2} and \ce{O2} with respective mole fractions 80\% and 20\%) at initial temperature $T_0$ and pressure $P_0$, and with relative mole fractions specified by an initial equivalence ratio, ${ER}_0$, defined as~\cite{Combust_Poinsot_2005}:
\begin{equation}\label{ER}
    ER_0 = \dfrac{X_{H_2}/X_{Air}}{(X_{H_2}/X_{Air})_{st}},
\end{equation}
where $X_i$ represents the mole fraction of the i-th (macro-)species, and the $st$ subfix identifies the stoichiometric value. The system is then evolved in time, and the overall dynamics of the temperature and the $N_s$ species are described by the following ODE:
\begin{equation}
    \begin{cases}
        \dfrac{d \rho y_i}{d t} = \dot{\omega}_i \\[3pt]
        \rho C_P \dfrac{d T}{d t} = \dot{\omega}_T^{\prime}
    \end{cases}
\end{equation}
where $\rho$ identifies the mixture density, $y_i$ denotes the mass fraction of the i-th species with $i=1,...,N_s$, and $C_p = C_p(T)$ represents the isobaric specific heat capacity. The rate of species production/depletion, $\dot{\omega}_i$ is defined as:
\begin{equation}
    \dot{\omega}_i = \sum_{j=1}^{N_s} \dot{\omega}_{i,j}(y_i,y_j), 
\end{equation}
while the heat release, $\dot{\omega}_T^{\prime}$, is given by:
\begin{equation}
    \dot{\omega}_T^{\prime} = - \sum_{i=1}^{N_s} h_i \dot{\omega}_i(y_i,y_j)
\end{equation}
with $h_i$ being the partial specific enthalpies. For details on the analytical dependencies of the reaction rates, $\dot{\omega}_{i,j}(y_i,y_j)$ on $y_i$ and $y_j$, the reader should refer to~\cite{Combust_Poinsot_2005}. In this work, we employ a sub-component of the GRI-Mech 3.0 reaction scheme \cite{Combust_GREMech3_1999} to define the \ce{H2} chemistry. The full mechanism would include 325 reactions and 53 species based on the five elements O, H, C, N, and Ar. However, since none of the argon- and carbon-containing chemistry is active, the effective number of reactions produced by the combustion of hydrogen with air is significantly lower, and only eighteen species are considered. This brings to 19 the total number of thermodynamic variables, $\boldsymbol{x}$, for the problem under analysis (i.e., $\boldsymbol{x}=\{T,\boldsymbol{y}\}$). The system of equations is finally closed by the ideal gas law:
\begin{equation}
    P = \rho \dfrac{R}{M} T,
\end{equation}
where $R$ and $M$ respectively represent the ideal gas constant and the mixture molecular weight.\\
By relying on Cantera~\cite{Combust_Cantera_2021}, an open-source suite of tools for problems involving chemical kinetics, thermodynamics, and transport processes, we generate 500 simulations for as many sets of initial conditions. Each of those is obtained by fixing the reactor pressure at $P_0=1$ [atm] and by selecting initial temperatures and initial residence times through Latin hypercube sampling of the space $(T_0,ER_0) \in (1000$ [K]$,2000$ [K]$)\times(0.5,4)$. For each simulation, the solutions are collected at 500 time instants, which altogether form a set of 250,000 data points for the ODE solution $\boldsymbol{x}:[(T_0,\boldsymbol{y_0}),t] \mapsto (T,\boldsymbol{y})$. Temperature and H${}_2$ mass fractions from five training scenarios are shown in Figs.~\ref{fig:0DReact_Train_TandH2}. Given the different order of magnitudes that characterize the thermodynamic variables, feature scaling based on min-max normalization is applied to the data before training all the operator surrogates for this physical system. In analogy with the previous test case and because of what will be discussed later, it is important to highlight that (logarithmic) time shifts are noticeable components of the overall transformations affecting the evolution profiles as functions of the scenarios~\cite{Combust_Lemke_2015,Combust_Lemke_2019}. It should also be noted that shifts in a logarithmic scale correspond to stretches in a linear scale.
\begin{figure}
    \centering
    \includegraphics[width=6.0in]{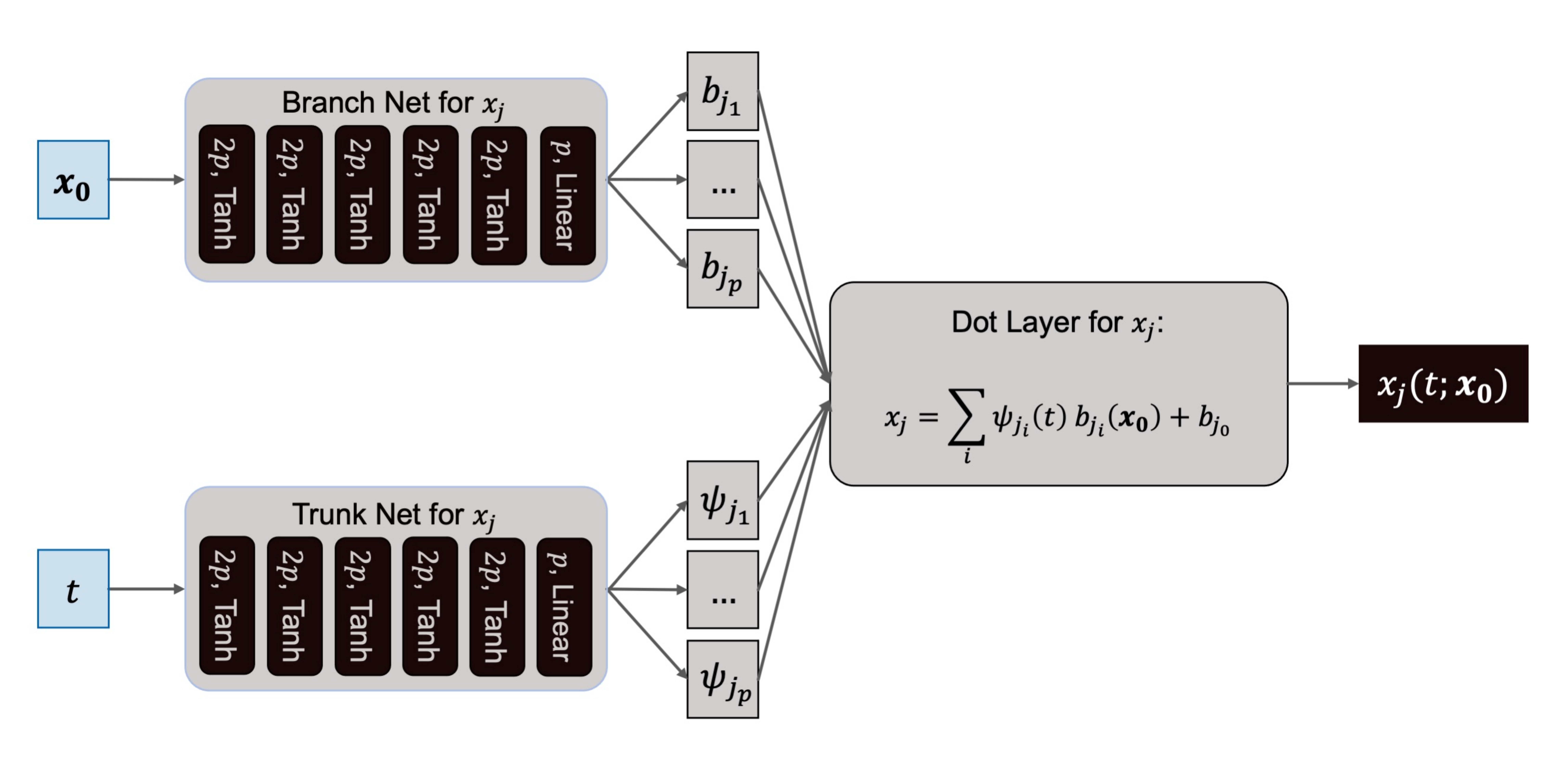}
    \caption{\textbf{Architecture of the vanilla DeepONet for one of the 19 thermodynamic variables}. $j=1,...,19$, and each thermodynamic variable has its own trunk and branch nets. We study two separate vanilla DeepONets, respectively characterized by $p=20$ and $p=32$.}
    \label{fig:0DReact_DeepONet}
\end{figure}
\begin{figure}
    \begin{subfigure}{0.49\textwidth}
        \centering
        \caption{}
        \label{fig:0DReact_H2Modes_32}
        \includegraphics[width=3.18in]{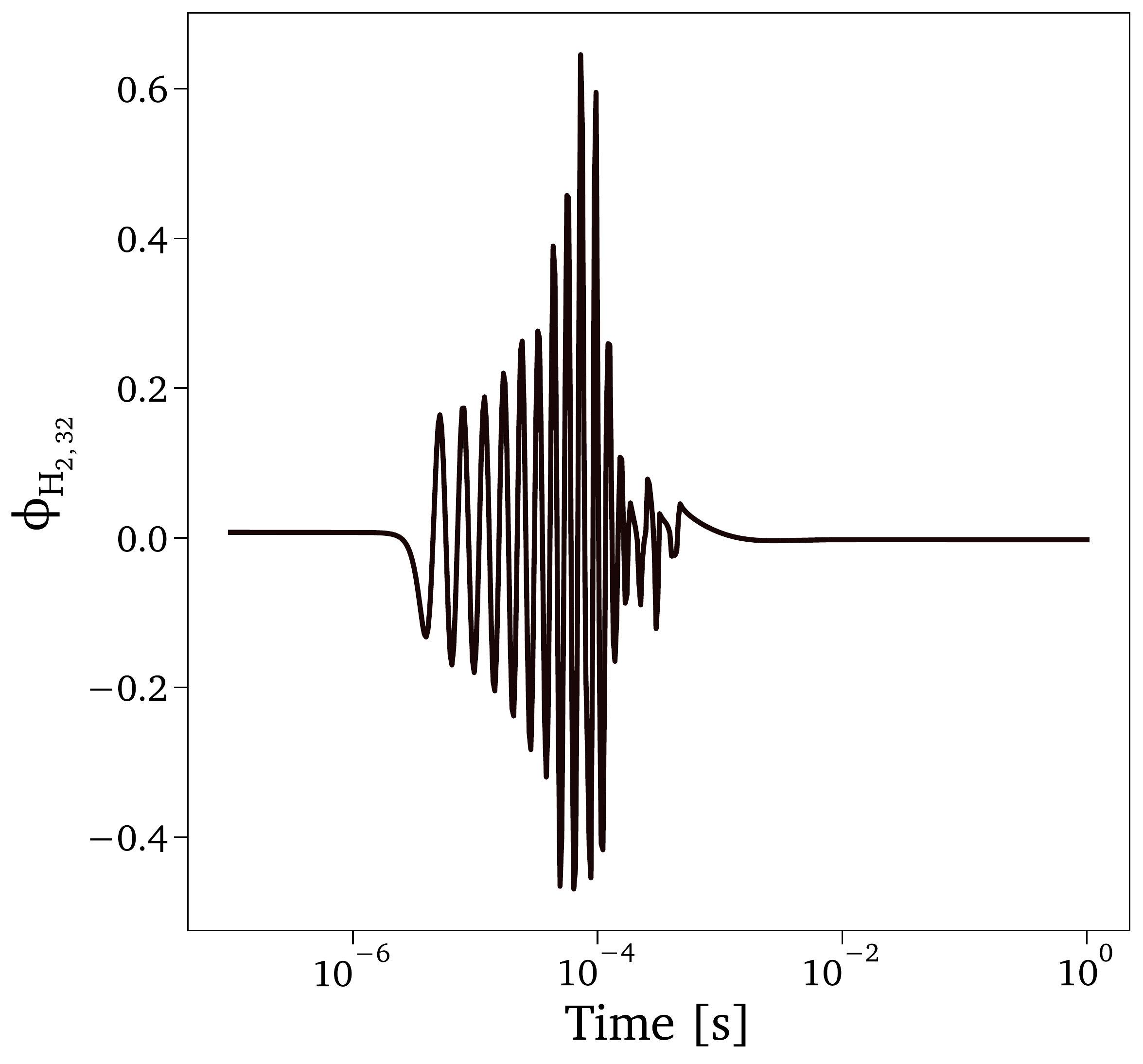}
    \end{subfigure}
    \begin{subfigure}{0.49\textwidth}
        \centering
        \caption{}
        \label{fig:0DReact_H2Modes_64}
        \includegraphics[width=3.22in]{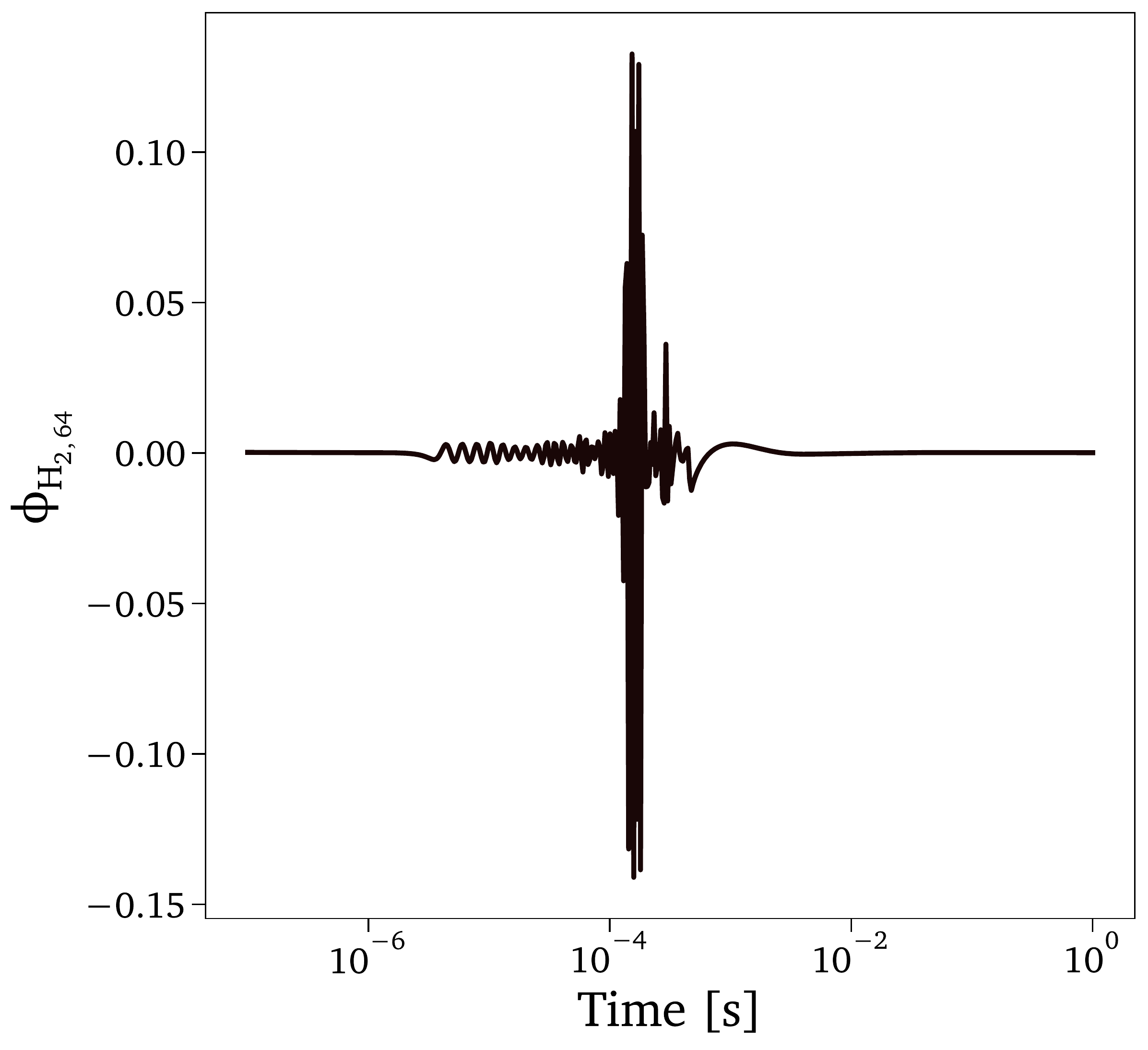}
    \end{subfigure}
    \caption{\textbf{Principal components of the data matrix constructed with the H$_2$ scenarios}. Principal components associated with the thirty second (\textbf{A}) and sixty fourth (\textbf{B}) singular values.}
    \label{fig:0DReact_H2Modes}
\end{figure}
With the resulting data points, we train two vanilla DeepONets that have one branch and one trunk for each of the $x_j$ thermodynamic variables, with $j=1,...,19$. These blocks are characterized by six hidden layers and $p$ outputs each, with $p=20$ for the first DeepONet and $p=32$ for the second. The structures of the two surrogates are shown in Fig.~\ref{fig:0DReact_DeepONet}, and they respectively contain 297,160 and 738,131 trainable parameters. Predictions on test scenarios are presented in Fig.s.~\ref{fig:0DReact_DeepONet_y_20} and~\ref{fig:0DReact_DeepONet_y_32}, and they show significant oscillations in the time instants that immediately preceded and succeed sharp gradients, especially for $T$ and $y_{H_2}$.\\
In light of what was observed for the previous {\it tanh} test case, we proceed by investigating the results from the SVD perspective. As before, we create a scenario-aggregated snapshot matrix with dimensions $[500 \times 500]$ for each of the thermodynamic variables. We then center each of its columns, normalize them based on their standard deviations, and perform the SVD. Two interesting results emerge from this procedure. Firstly, in order to preserve 95\% of the matrices' cumulative energy content, the number of singular values that must be retained ranges between 5 and 32, depending on the thermodynamic variable under analysis, and at least 20 to 60 must be kept to reach 99\% (Fig.~\ref{fig:0DReact_CumEnergies}). The consequences of retaining only 20 and 32 singular values are respectively shown in Fig.~\ref{fig:0DReact_Reconstructed_20} and Fig.~\ref{fig:0DReact_Reconstructed_32} in terms of SVD's encoding-decoding inaccuracies. After being projected to the reduced basis, the reconstruction of the dynamics results in artificial oscillations similar to those that characterize the vanilla DeepONet's predictions. Secondly, in analogy with what has been observed for the shifting hyperbolic tangent test case, lower singular values are associated with principal components and directions characterized by higher frequencies. As some examples, Fig.~\ref{fig:0DReact_H2Modes} shows the principal components corresponding to the 32nd and 64th singular values of H$_2$. This should be taken into account while selecting the number of scenarios and time instants to be simulated for generating DeepONet's training data to avoid aliasing-related issues.\\
Based on these two findings, we deduce that an increase in the number of trunk outputs reduces DeepONet's approximation error but also complicates the trunk and branch fitting tasks. As a consequence, the sub-block regression errors would rise unless their network expressivities are significantly improved.
To summarize, the following aspects should be considered altogether:  
\begin{itemize}
    \item The vanilla DeepONet autonomously performs the projection by means of the dot-layers;
    \item In the case of an $\ell_2$-optimal projection, the information content retained by only 32 singular values is insufficient for retrieving with high accuracy some of the evolving thermodynamic variables;
    \item In the case of an $\ell_2$-optimal projection, some of the trunk and branch nets are asked to reproduce highly oscillatory modes.
\end{itemize}
As discussed concerning the previous test case, by representing the dynamics on a coordinate system that moves as a function of the initial conditions, it is possible to take into account the geometrical similarities between the evolution profiles for different scenarios and improve the projection to the reduced basis. Following this intuition, Lemke \textit{et. al.}~\cite{Combust_Lemke_2015} developed a reduced-order model for a 0-D isobaric reactor. Their test case was similar to the one we are proposing, but i) had methane as the fuel, ii) the initial mass fractions were not changed while varying scenarios, and iii) the initial temperatures were selected from a smaller range (i.e., 1020 - 1080 [K]). Lemke \textit{et. al.}'s strategy for constructing a data-driven surrogate capable of predicting the reactor's thermodynamic state for unseen initial conditions relied on the following steps: i) time-shifting the scenarios so that their maximum temperature gradients matched in time, ii) performing SVD, and iii) interpolating the coefficients of the projection matrix. As their data alignment allowed the SVD to improve the singular values' rate of decay, an autonomously-discovered shifting reference frame would empower the DeepONet to project the operator dynamics on a lower-dimensional basis. In this sense, the flexDeepONet architecture has the capability of transforming the Lemke \textit{et. al.} approach to a fully AI-operated surrogate construction.
\begin{figure}[!hbt]
    \centering
    \includegraphics[width=6.2in]{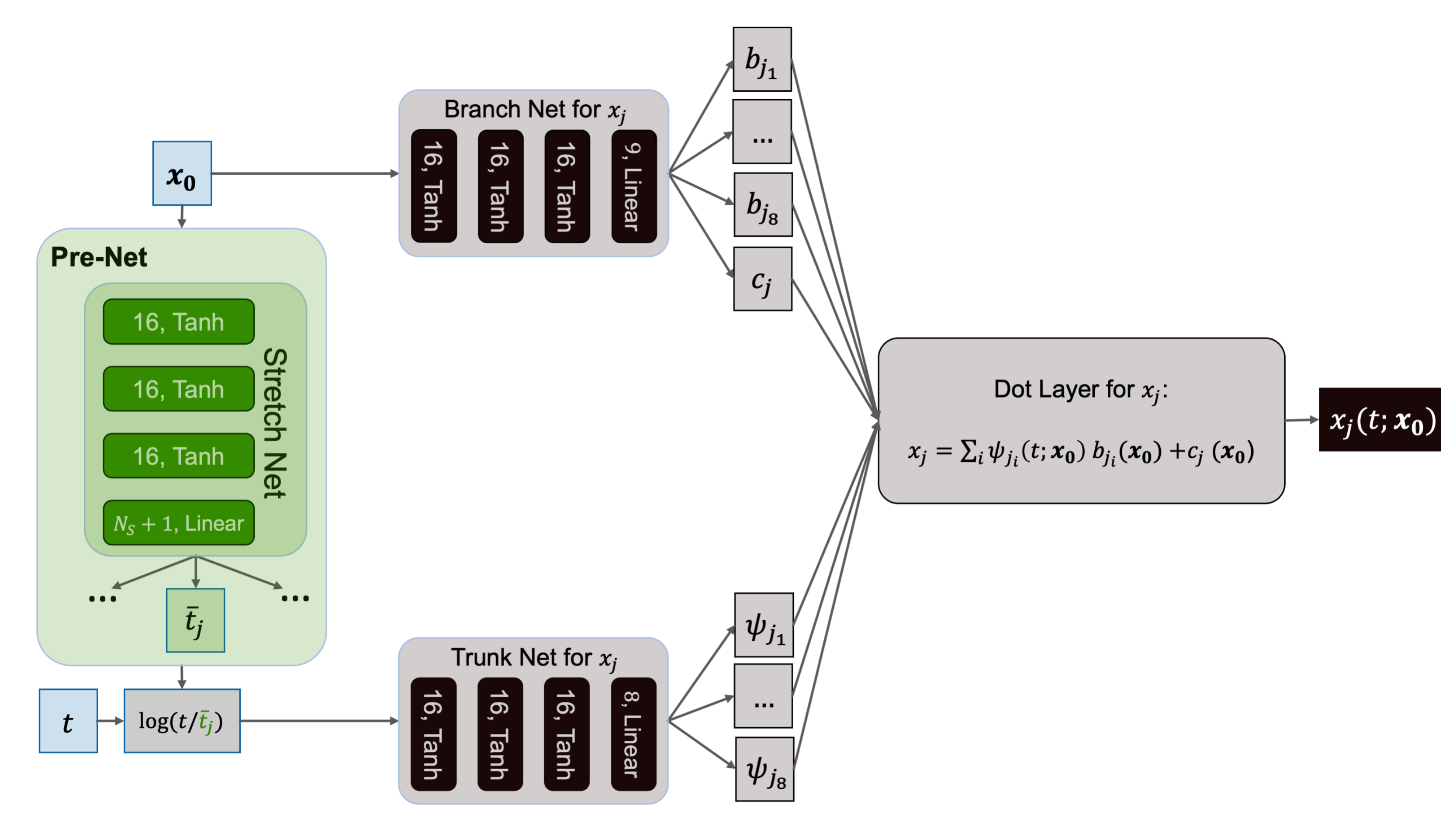}
    \caption{\textbf{FlexDeepONet architecture for the j-th thermodynamic state variable of the combustion chemistry test case}. The DeepONet is composed of $N_s+1$ trunks, $N_s+1$ branches, and a single Pre-Net made of one stretching FNN, where $N_s=18$ represents the number of chemical species. The stretching block is introduced to guarantee the alignment of the different scenarios, and it outputs $N_s+1$ factors, one for each of the trunk nets. An additional neuron is added to the output layer of each branch net to permit scenario-specific centering.}
    \label{fig:0DReact_flexDeepONet}
\end{figure}
We now build a flexDeepONet for the present test case. Compared to the vanilla architecture, the surrogate prepends a single pre-transformation network. This block is shared by all the thermodynamic variables, and it outputs one quantity for each of them, $\bar{t}_j$, that can be either seen as a stretching factor operating in a linear scale or as a shifting bias in a logarithmic scale. Additionally, we add one output to each branch net to learn scenario-dependent centering. These two simple improvements permit to decrease the number of DeepONet's projection bases (i.e., $p$). As less expressivity is required from the trunks and branches, the depths and widths of these nets can be significantly reduced. The flexDeepONet structure used for this test case is reported in Fig.~\ref{fig:0DReact_flexDeepONet}. Overall, the number of trainable parameters is lowered to 34,038. 
\begin{figure}[!tb]
    \begin{subfigure}{0.49\textwidth}
        \centering
        \caption{}
        \label{fig:0DReact_Shifts_T0}
        \includegraphics[width=3.2in]{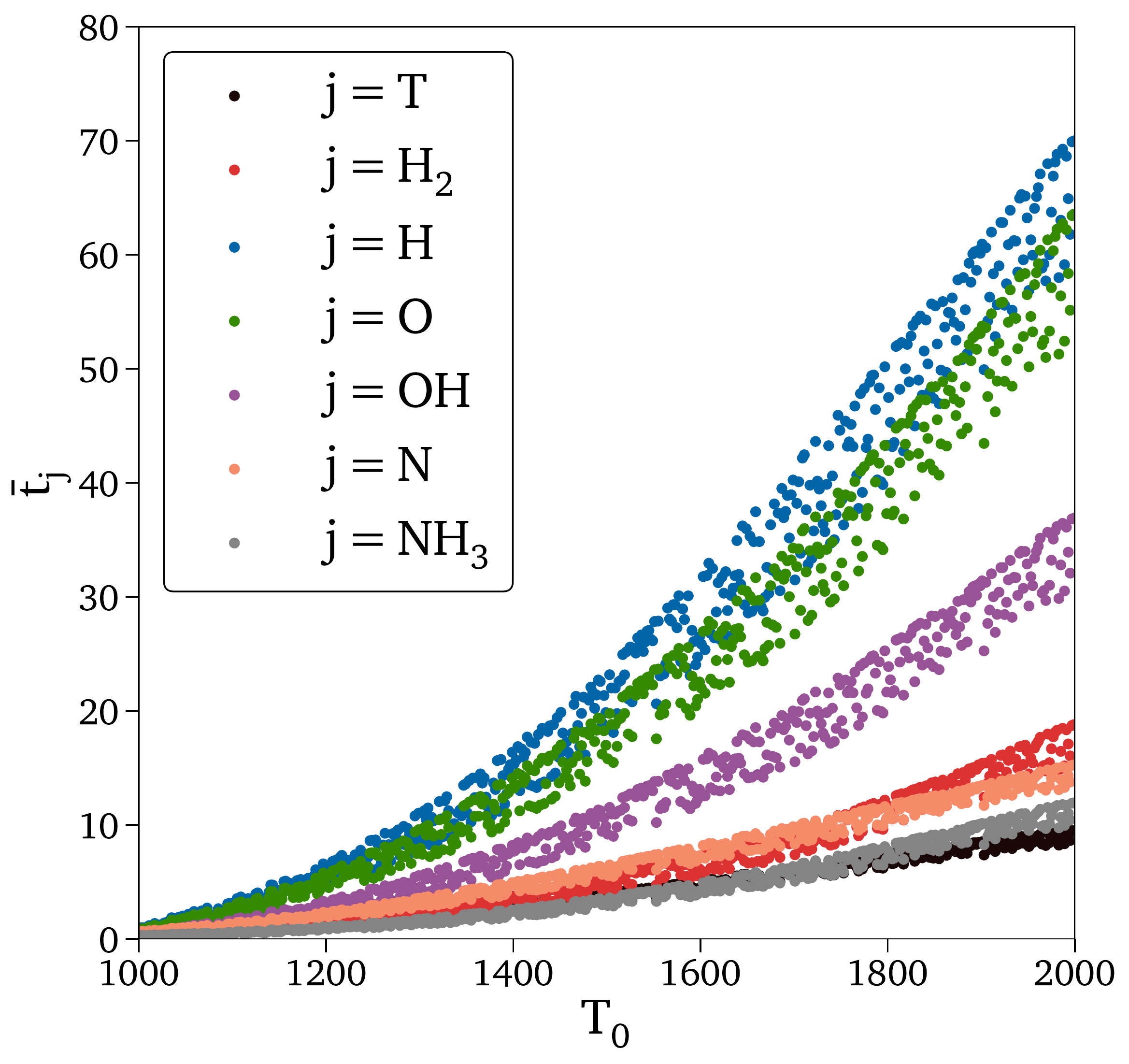}
    \end{subfigure}
    \begin{subfigure}{0.49\textwidth}
        \centering
        \caption{}
        \label{fig:0DReact_Shifts_y}
        \includegraphics[width=3.2in]{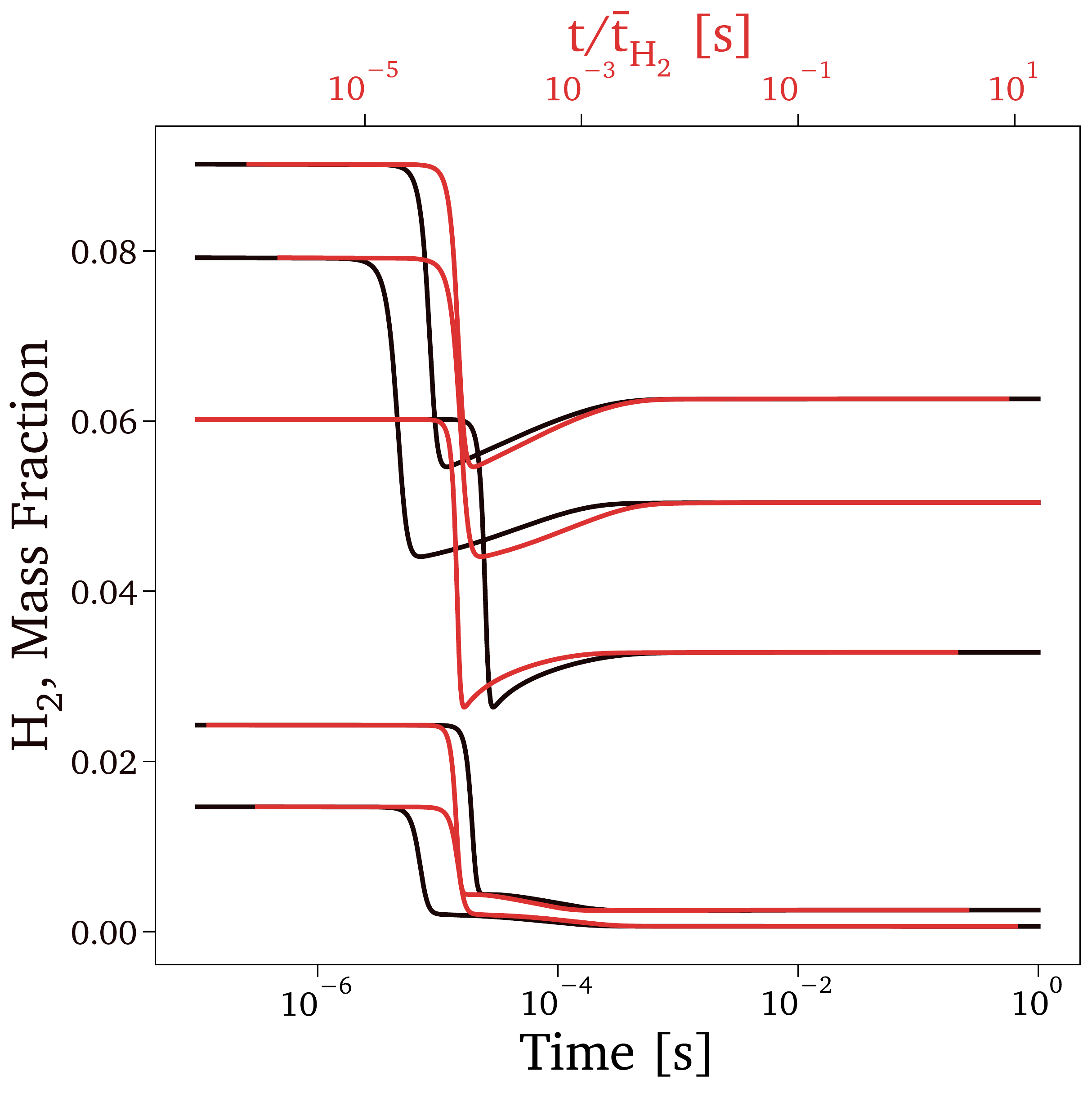}
    \end{subfigure}
    \caption{\textbf{Contributions of the shift net to the scenario alignment}. (\textbf{A}): Some outputs of the stretch net, corresponding to the scaling factors of some thermodynamic states. The predictions are plotted for all the training scenario as function of their initial temperatures. (\textbf{B}): Time-integrated H$_2$ mass fractions for some of the test scenarios as functions of the simulation time (black lines) and of the simulation time corrected by the stretch net (red lines). See Fig.~\ref{fig:0DReact_Shifted} for additional mass fractions, and Fig.~\ref{fig:0DReact_flexDeepONet} for details about the flexDeepONet's structure.}
    \label{fig:0DReact_Shifts}
\end{figure}
Fig.~\ref{fig:0DReact_Shifts_T0} shows the values of the scaling factors discovered by the stretch net for some thermodynamic variables as functions of the initial temperature. Thanks to these quantities, the scenario-dependent dynamics are well aligned with respect to the scaled system of coordinates, as testified by Fig.~\ref{fig:0DReact_Shifts_y}. The effectiveness of DeepONet's projections increases, and the oscillations in the predictions are removed (Fig.~\ref{fig:0DReact_flex_y_Main}).
Despite the $95\%$ reduction in the number of trainable parameters compared to the vanilla DeepONet and POD-DeepONet architectures, both with $p=32$, flexDeepONet appreciably enhances the accuracy in predicting temperature and mass fractions, as reported in Table~\ref{table:0DReact_TestErrors}. It should be noted that the few species that are not improved by the flexDeepONet are also the ones that are not remarkably affected by stretches in time and that can effectively be compressed to a small number of modes. As already stated, the architecture enhancement is not more beneficial than the vanilla configuration in emulating dynamics in which rigid motion is negligible. For these variables, reduced accuracies result from the decreased expressivity in the respective FNN blocks (from 18,848 parameters each in the vanilla/POD-DeepONet, to 712 in the flexDeepONet).
\begin{figure}[!bt]
    \begin{subfigure}{0.49\textwidth}
        \centering
        \caption{}
        \label{fig:0DReact_flex_T_}
        \includegraphics[width=3.2in]{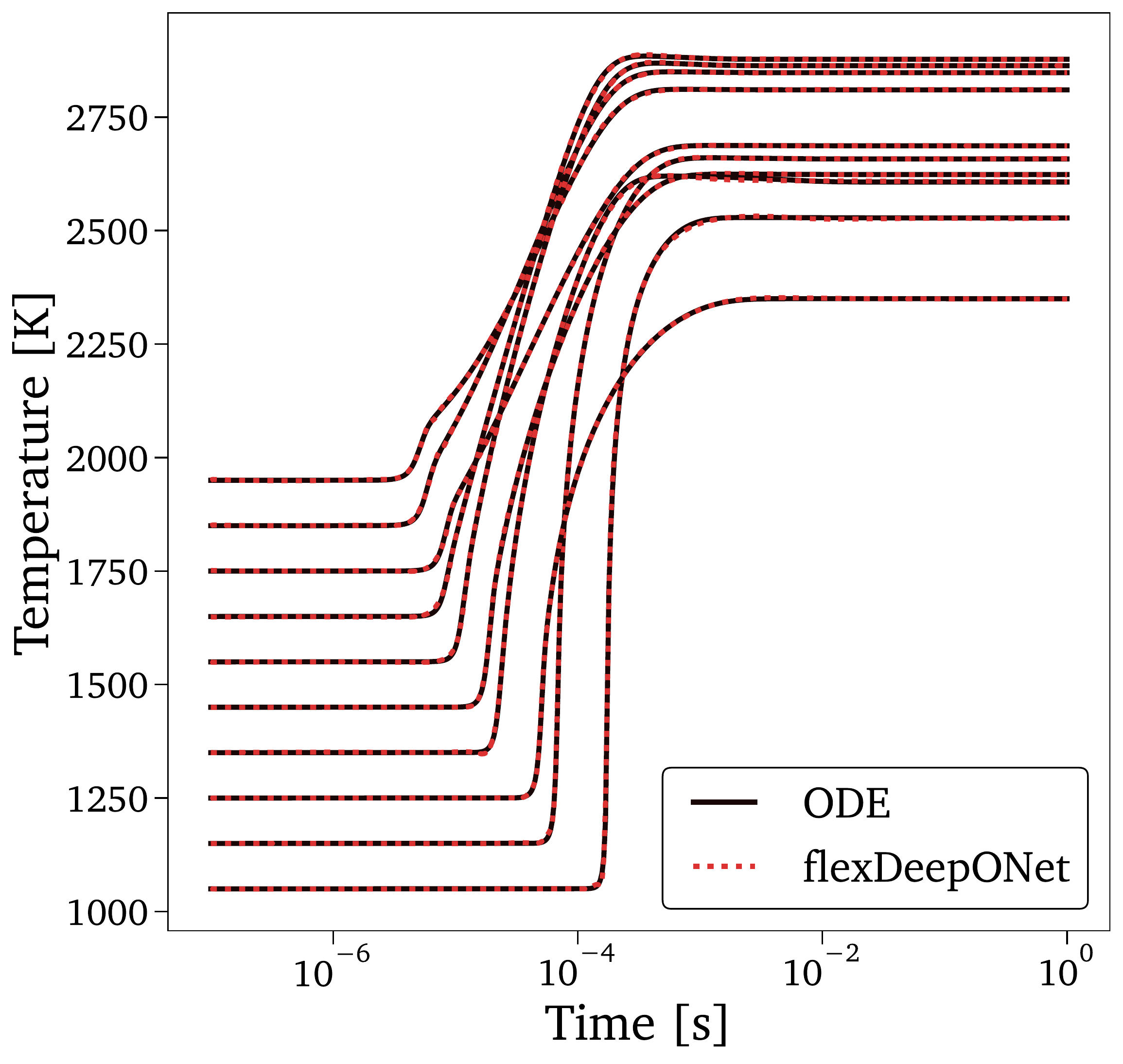}
    \end{subfigure}
    \begin{subfigure}{0.49\textwidth}
        \centering
        \caption{}\label{fig:0DReact_flex_H2}
        \includegraphics[width=3.2in]{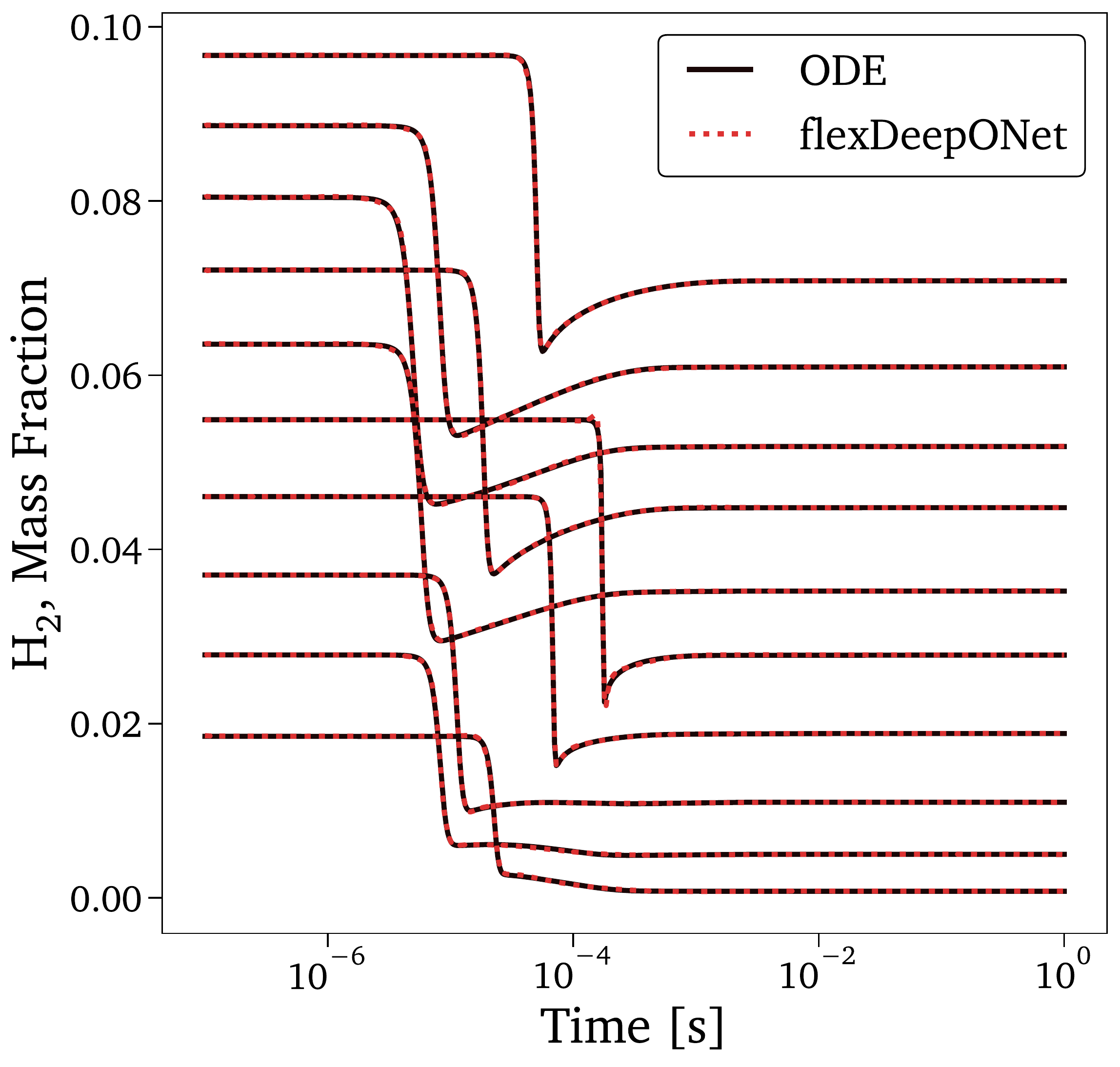}
    \end{subfigure}
    \caption{\textbf{FlexDeepONet predictions of the thermodynamic variables}. Temperature (\textbf{A}) and H$_2$ mass fraction (\textbf{B}) for ten test scenarios as the results of the ODE integration (solid black lines) and as predicted by flexDeepONet (red dotted lines) with structure as in Fig.~\ref{fig:0DReact_flexDeepONet}. See Fig.~\ref{fig:0DReact_flex_y} for additional mass fractions.}
    \label{fig:0DReact_flex_y_Main}
\end{figure}
\FloatBarrier


\subsection{Test Case 4: POD and DeepONet Analysis of a Rotating-Translating-Stretching Rigid Body} \label{sec:AppendixI}

\begin{figure}[!htb]
    \centering
    \includegraphics[width=3.2in]{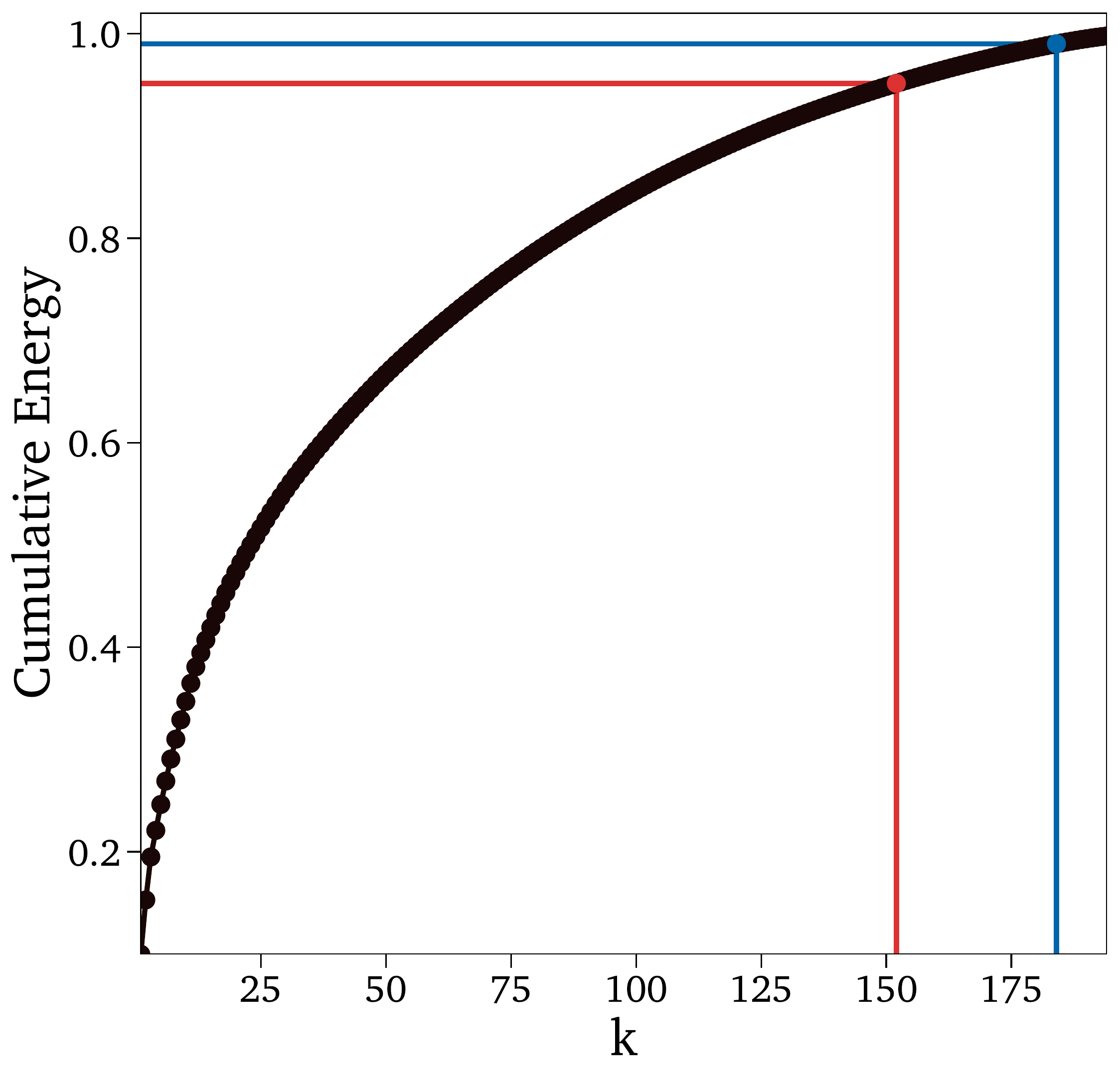}
    \caption{\textbf{Cumulative energies of the $[400,000 \times 200]$ time-aggregated snapshot matrix from the rotating-translating-stretching rigid body test case}. Cumulative energies contained in the first k singular values. The red and blue dots identify cumulative energies corresponding to 95\% and 99\%, respectively.}
    \label{fig:Rect_CumEnergy_All}
\end{figure}

\begin{figure}[!htp] 
    \centering
    \includegraphics[width=6.2in]{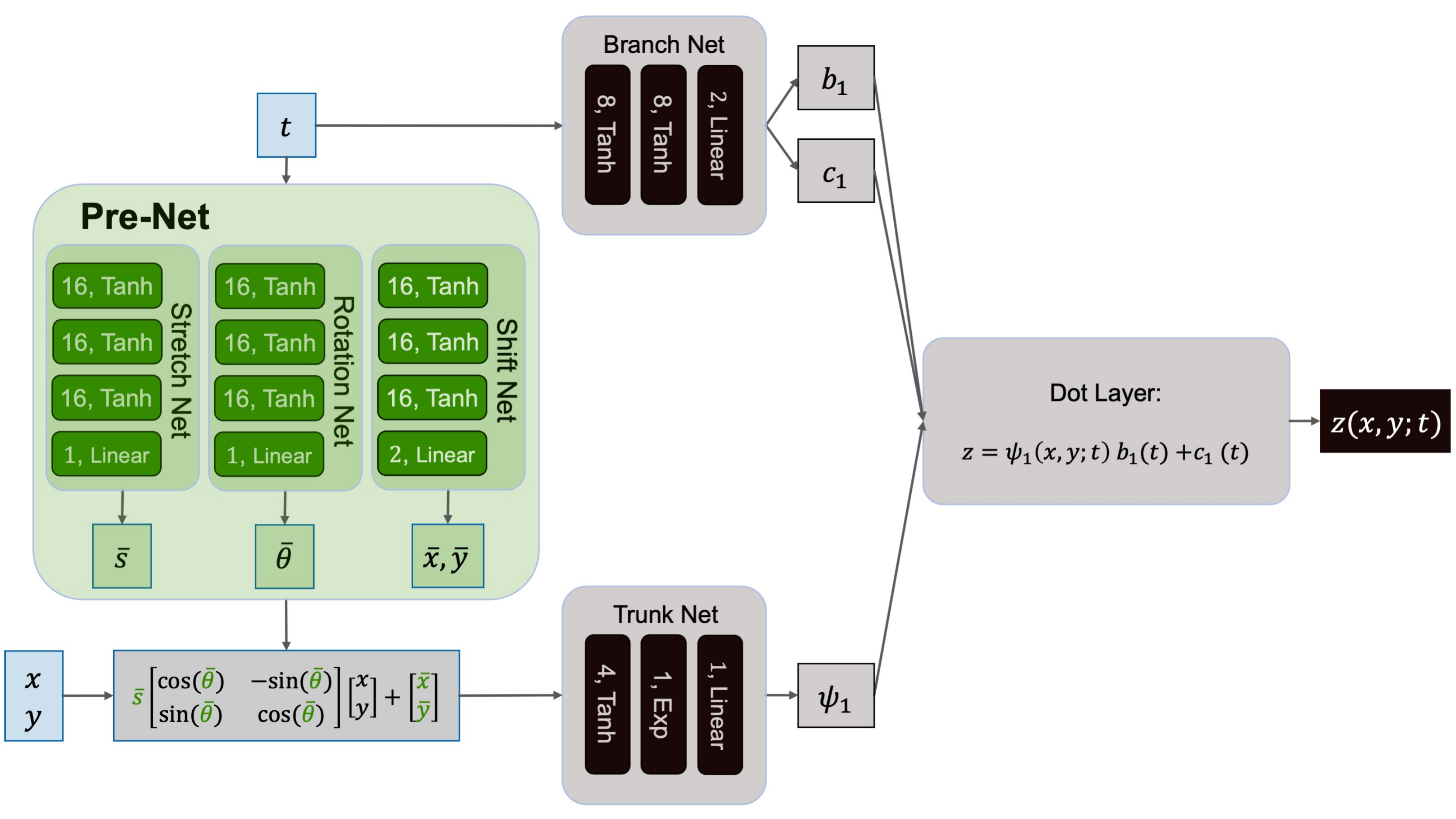}
    \caption{\textbf{FlexDeepONet architecture for the rotating-translating-stretching rigid body test case}. A Pre-Net composed of three separate blocks is introduced to guarantee the alignment of the different time snapshots. An additional neuron is added to the output layer of the branch net to permit time-dependent centering. Note: the combination of hyperbolic tangent and exponential activation functions is here employed consistently with the test cases' generative equations.}
    \label{fig:Rect_DeepONet_2}
\end{figure}

The three test cases above showed how SVD perspectives on the DeepONet can improve the technique's flexibility and interpretability. However, we believe that the links between the two methodologies are bidirectional and that some of DeepONet's peculiarities can significantly facilitate and extend applications of SVD-based methods, such as PCA, DMD, and POD. This sub-section briefly proposes an example of such a promising research direction.
\begin{figure}[b!]
    \begin{subfigure}{0.24\textwidth}
        \centering
        \caption{}
        \label{fig:Rect_Orig_1}
        \includegraphics[width=1.5in]{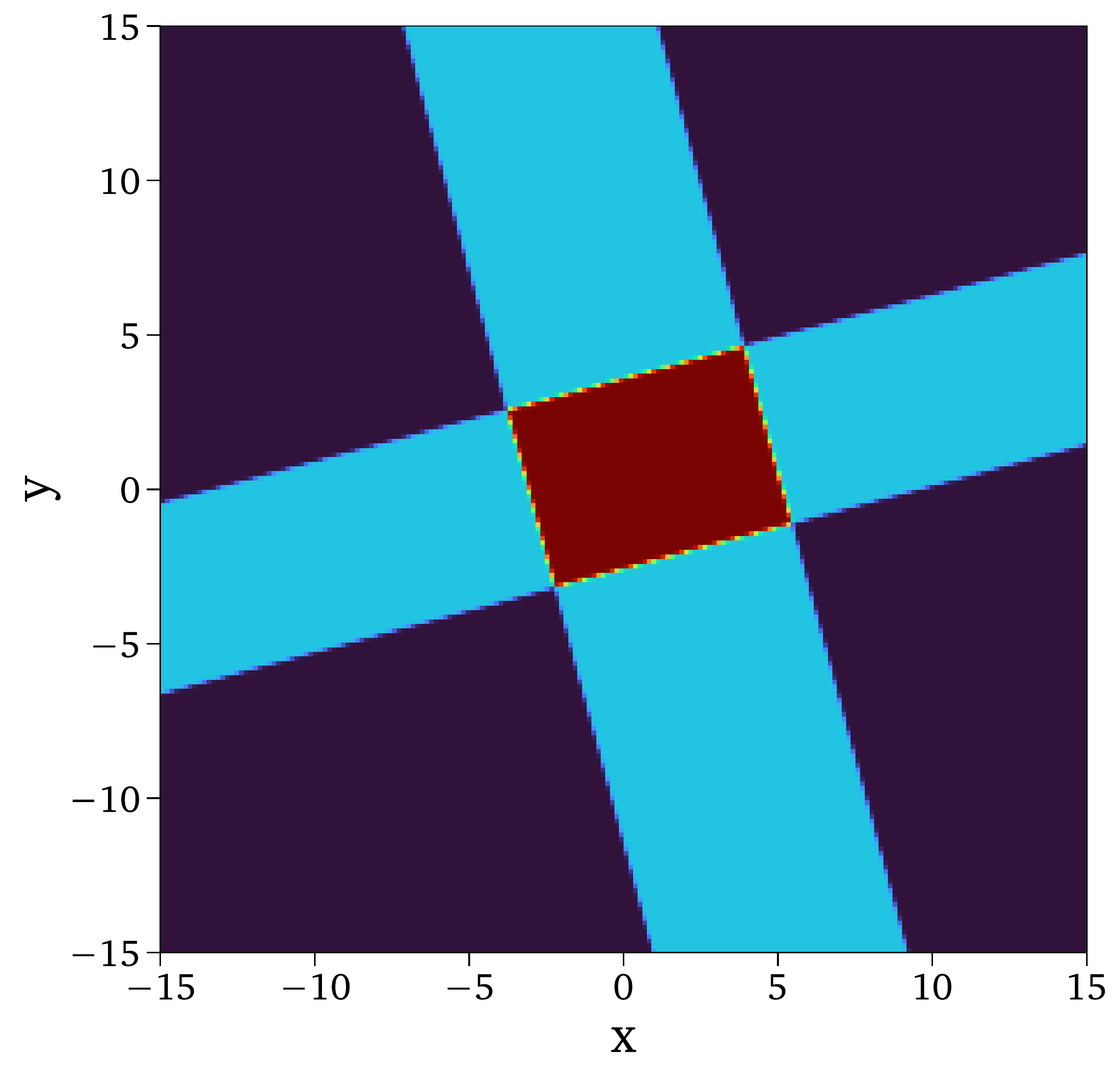}
    \end{subfigure}
    \begin{subfigure}{0.24\textwidth}
        \centering
        \caption{}
        \label{fig:Rect_Orig_2}
        \includegraphics[width=1.5in]{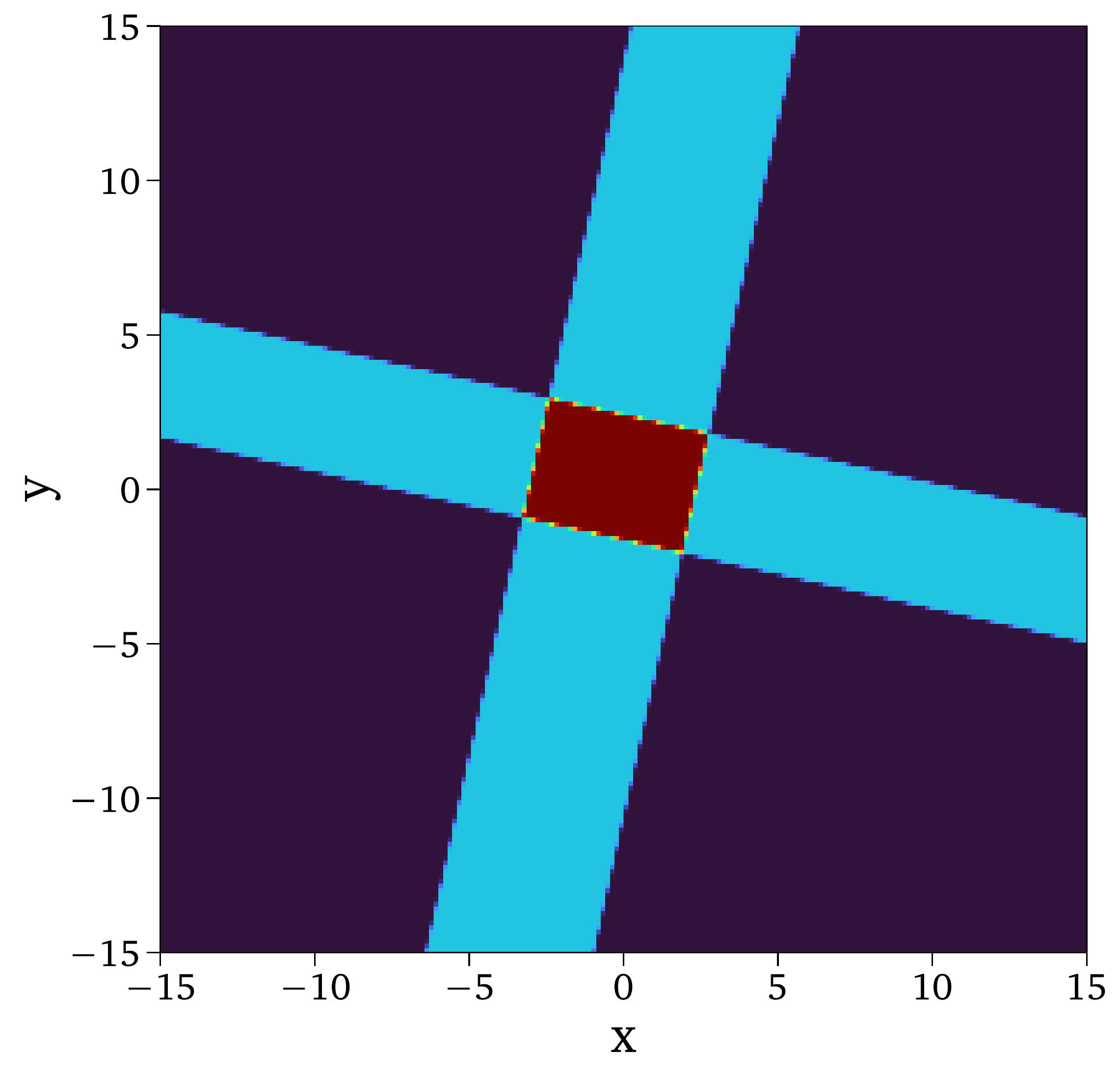}
    \end{subfigure}
    \begin{subfigure}{0.24\textwidth}
        \centering
        \caption{}
        \label{fig:Rect_Orig_3}
        \includegraphics[width=1.5in]{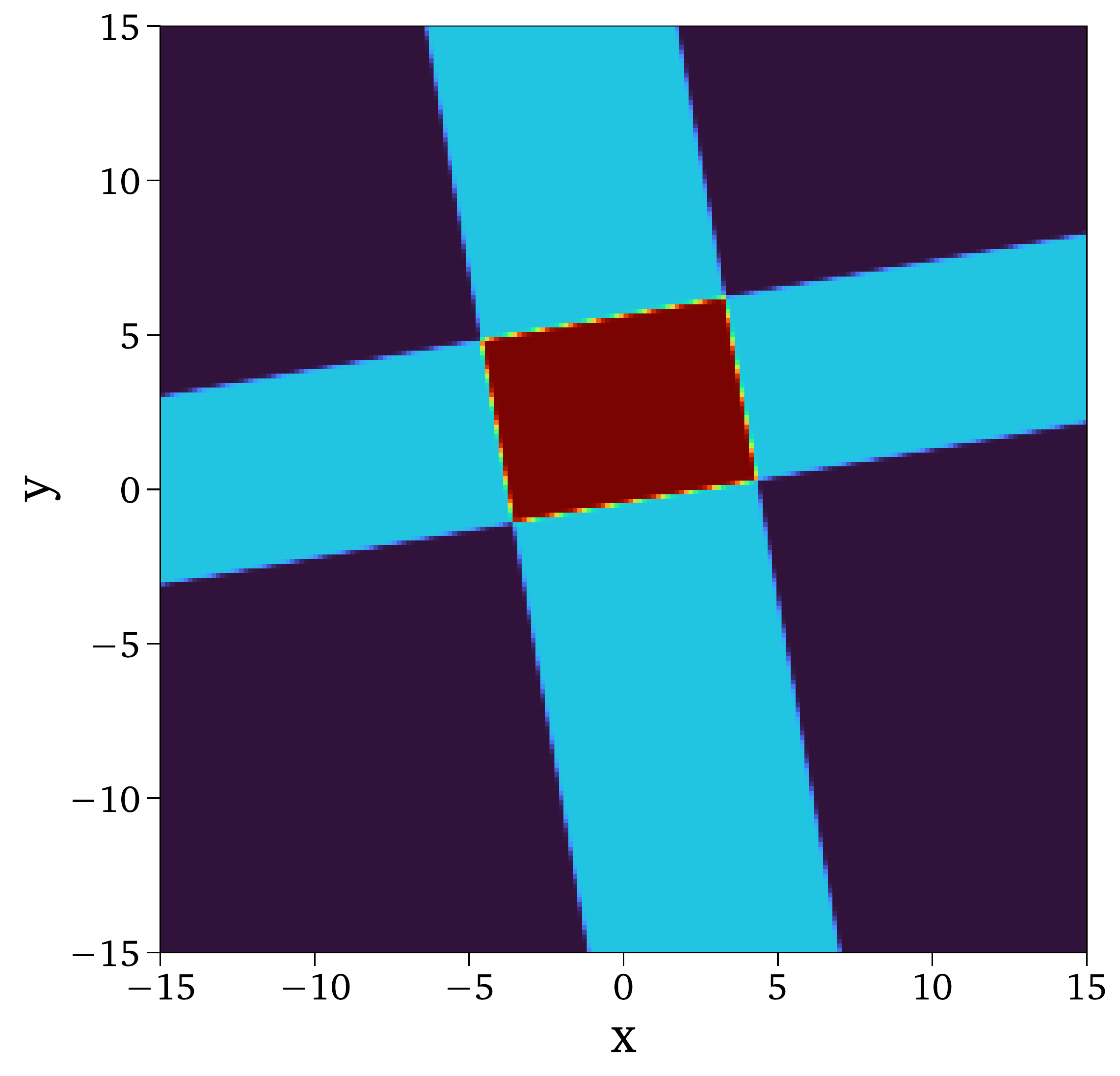}
    \end{subfigure}
    \begin{subfigure}{0.24\textwidth}
        \centering
        \caption{}
        \label{fig:Rect_Orig_4}
        \includegraphics[width=1.5in]{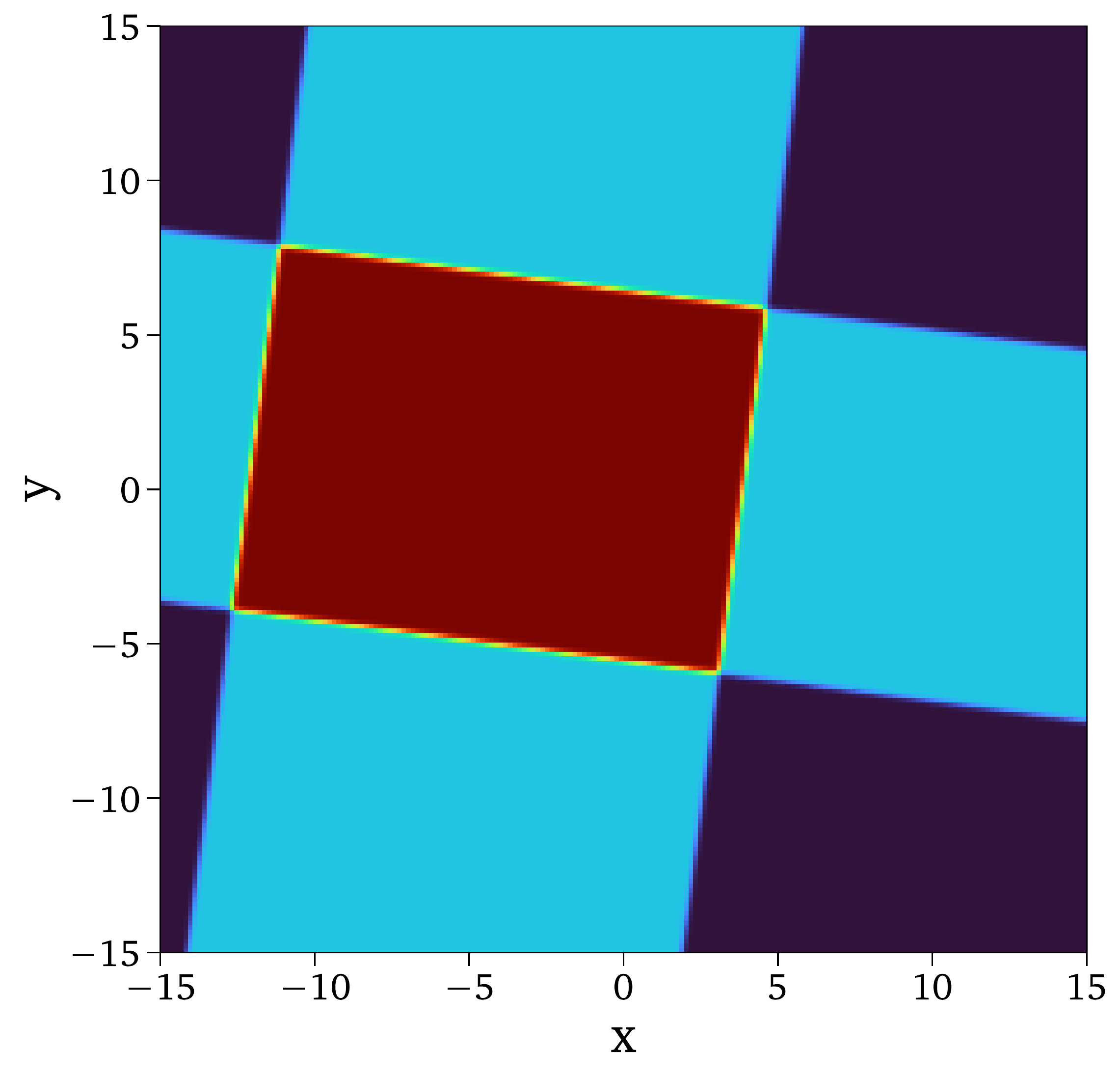}
    \end{subfigure}
    \caption{\textbf{Four time snapshots for the dynamics of the rotating-translating-stretching rigid body}. The snapshots are produced at $t=\{0,3.33,6.67,10\}$ [s]. See Fig.~\ref{fig:Rect_Orig_Supp} for additional time instants.}
    \label{fig:Rect_Orig}
\end{figure}
\begin{figure}
    \begin{subfigure}{0.24\textwidth}
        \centering
        \caption{}
        \label{fig:Rect_Orig_1_}
        \includegraphics[width=1.5in]{./Figures/Rect_Orig_1.pdf}
    \end{subfigure}
    \begin{subfigure}{0.24\textwidth}
        \centering
        \caption{}
        \label{fig:Rect_128PC_1}
        \includegraphics[width=1.5in]{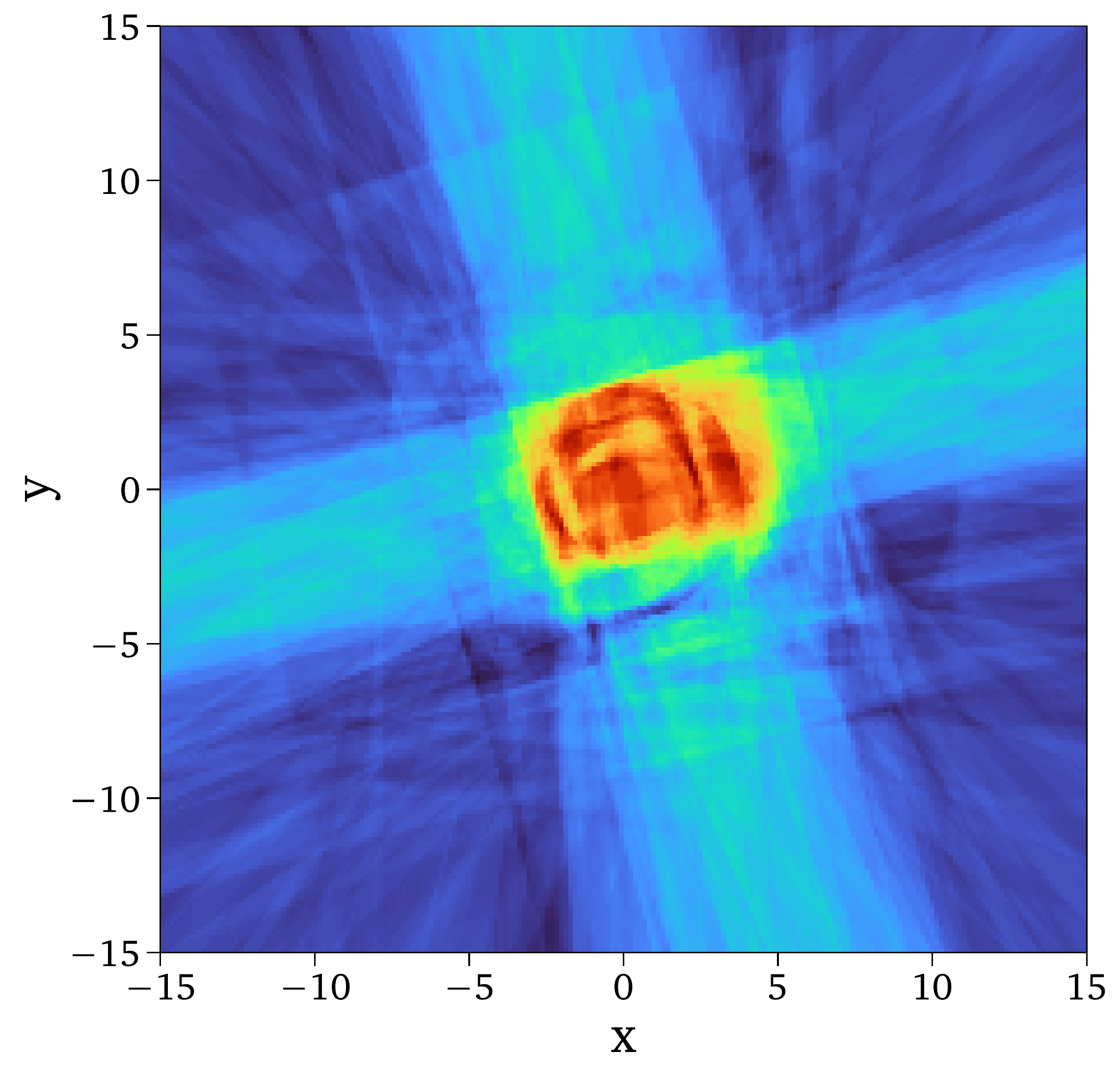}
    \end{subfigure}
    \begin{subfigure}{0.24\textwidth}
        \centering
        \caption{}
        \label{fig:Rect_64PC_1}
        \includegraphics[width=1.5in]{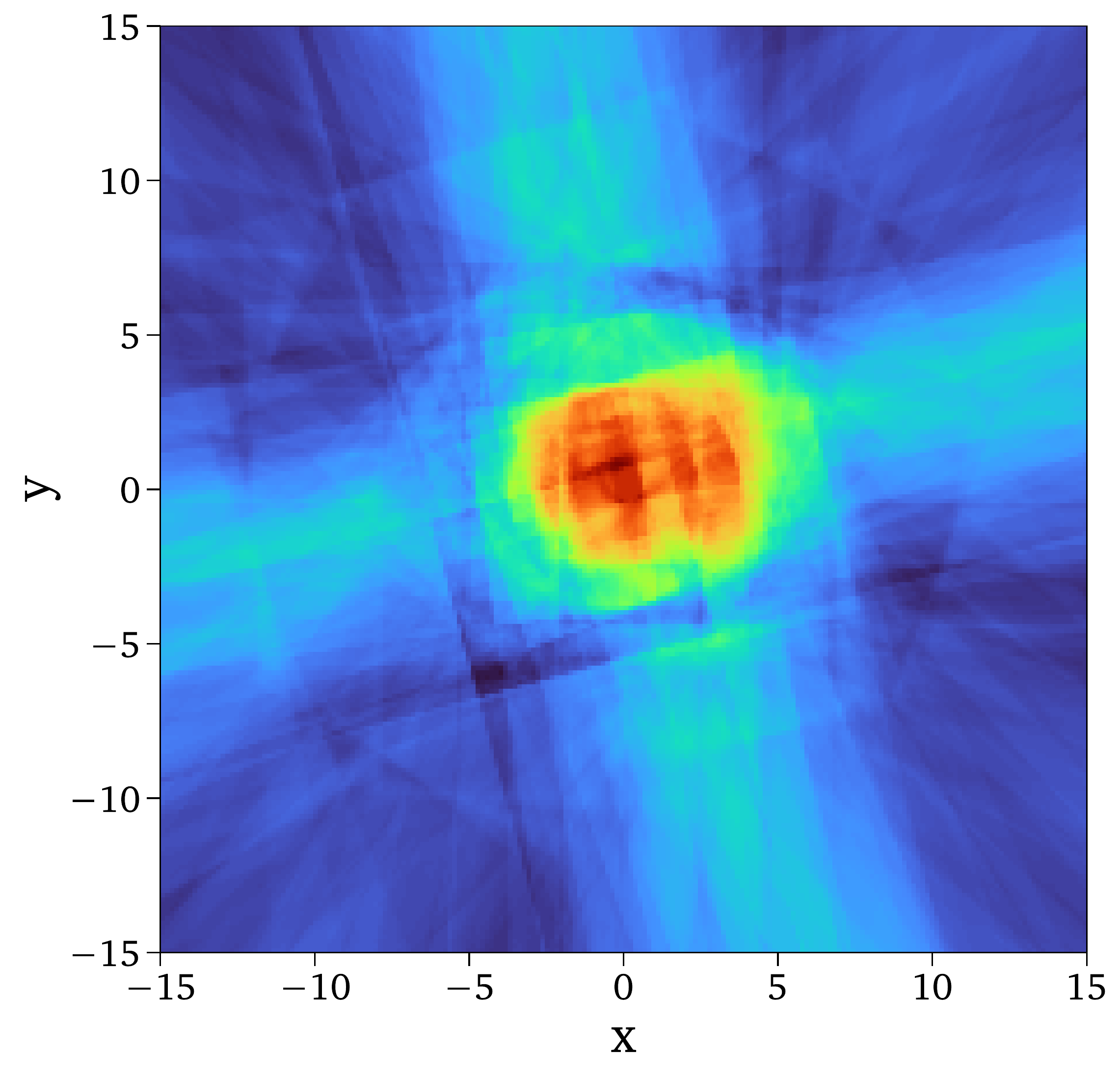}
    \end{subfigure}
    \begin{subfigure}{0.24\textwidth}
        \centering
        \caption{}
        \label{fig:Rect_32PC_1}
        \includegraphics[width=1.5in]{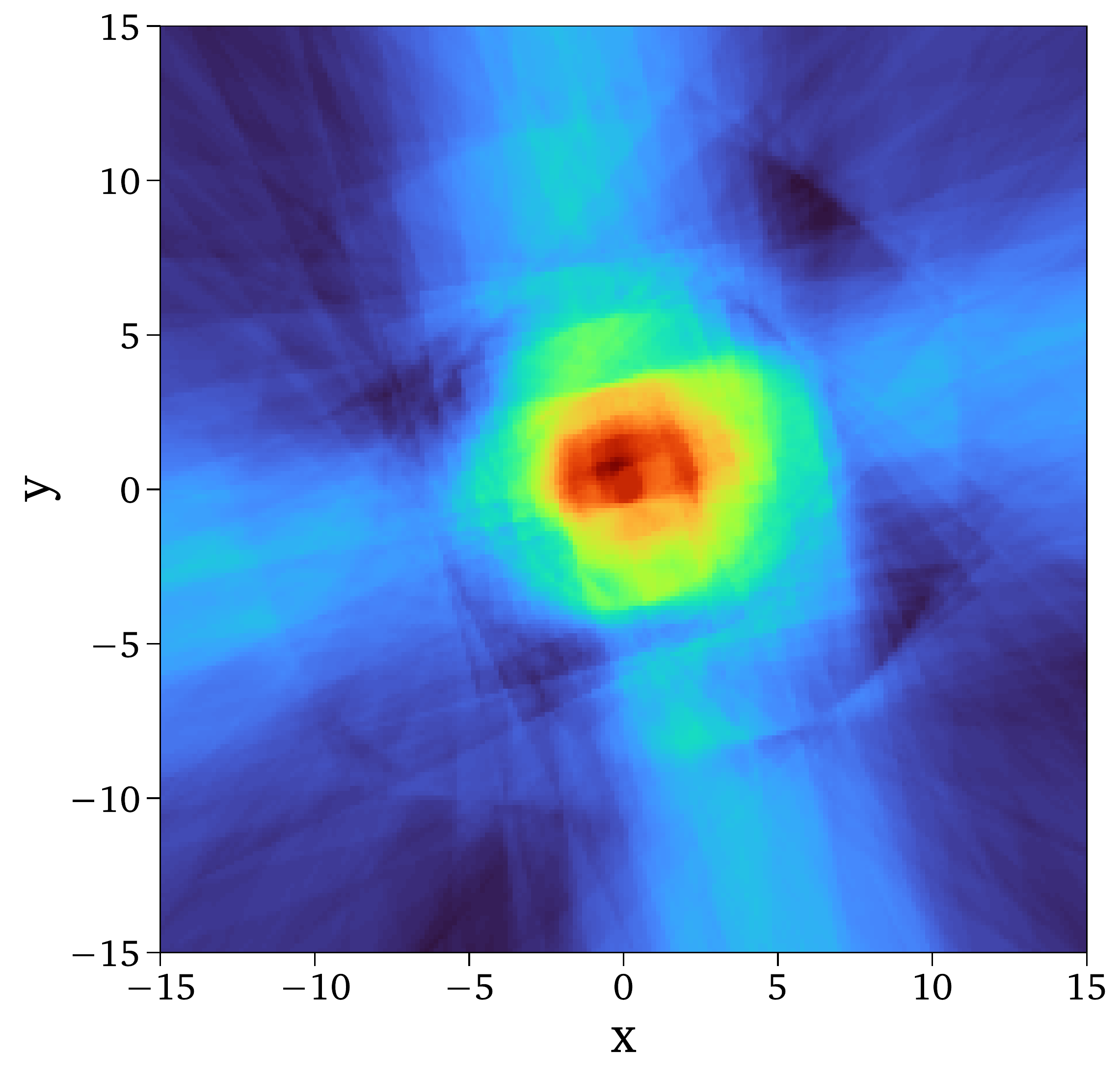}
    \end{subfigure}
    \caption{\textbf{Time snapshot decomposed and reconstructed via POD}. Time snapshot at $t=0$ [s] as generated by the original equations (first column from the left) and after being encoded-decoded based on POD's first 128, 64, and 32 modes (second, third, and fourth columns, respectively). See Fig.~\ref{fig:Rect_PC_Supp} for additional encoded-decoded time snapshots.}
    \label{fig:Rect_PC_1}
\end{figure}
\begin{figure}[!b]
    \begin{subfigure}{0.24\textwidth}
        \centering
        \caption{}
        \label{fig:Rect_flexDeepONet_1}
        \includegraphics[width=1.5in]{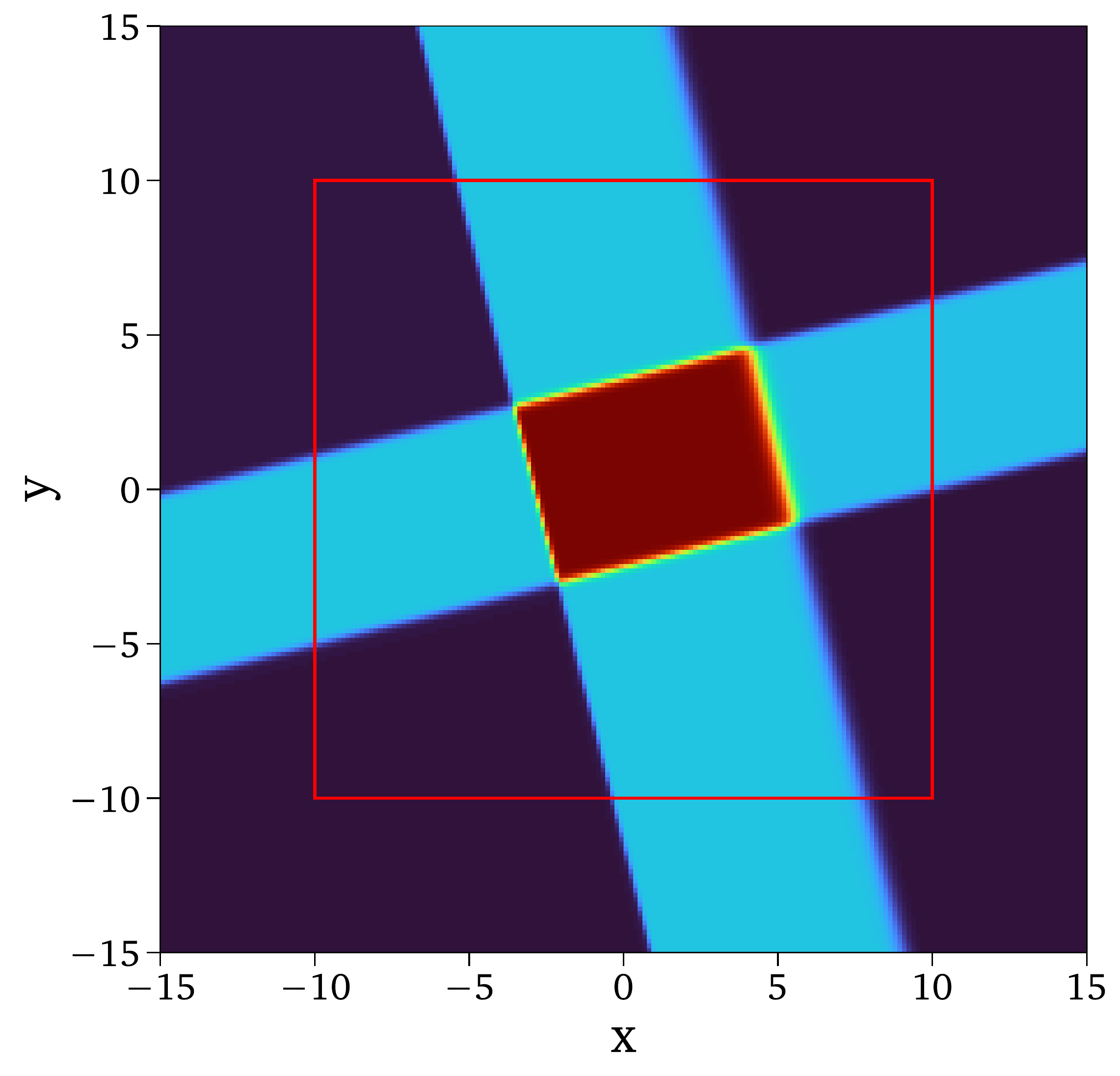}
    \end{subfigure}
    \begin{subfigure}{0.24\textwidth}
        \centering
        \caption{}
        \label{fig:Rect_flexDeepONet_2}
        \includegraphics[width=1.5in]{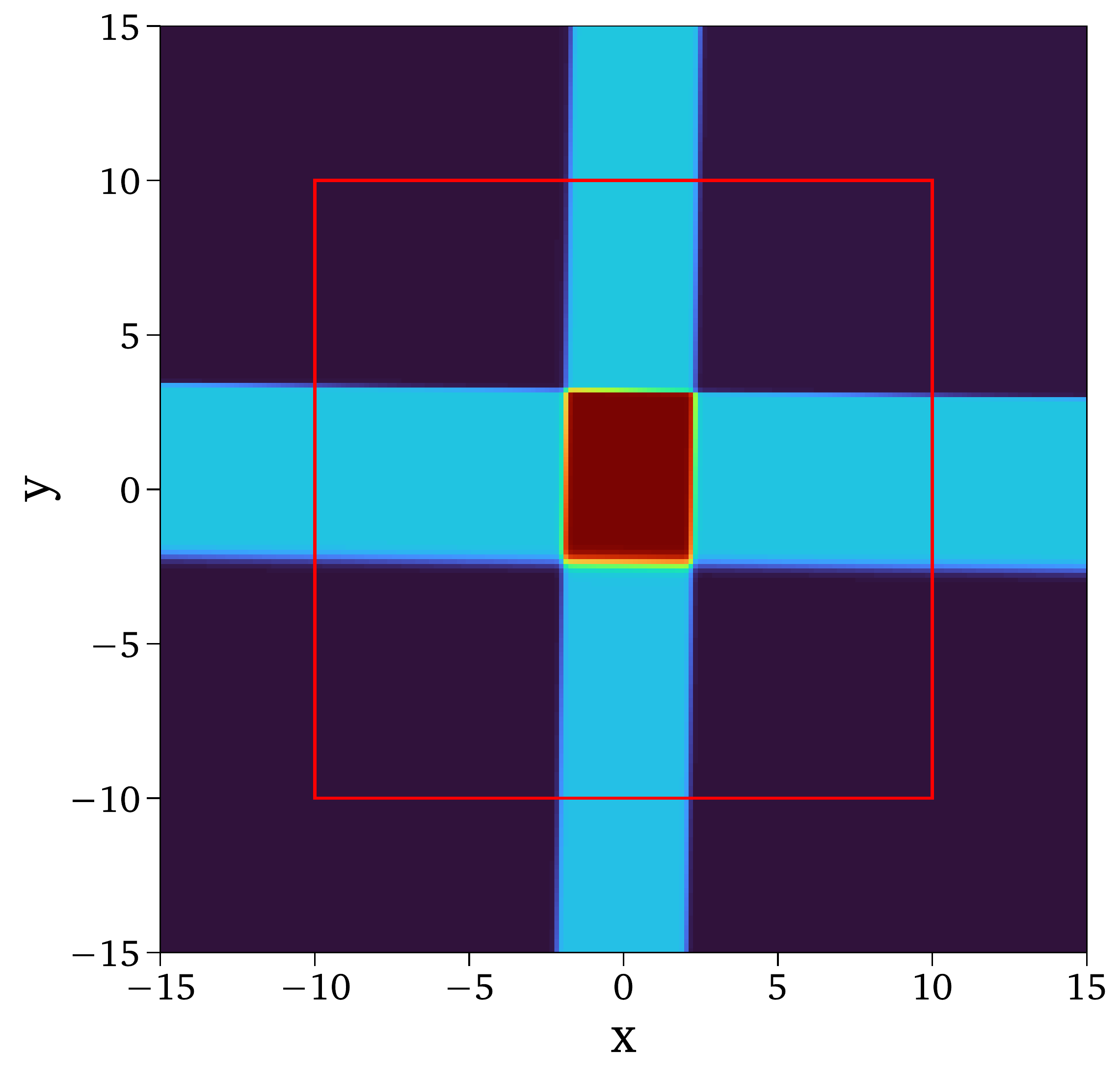}
    \end{subfigure}
    \begin{subfigure}{0.24\textwidth}
        \centering
        \caption{}
        \label{fig:Rect_flexDeepONet_3}
        \includegraphics[width=1.5in]{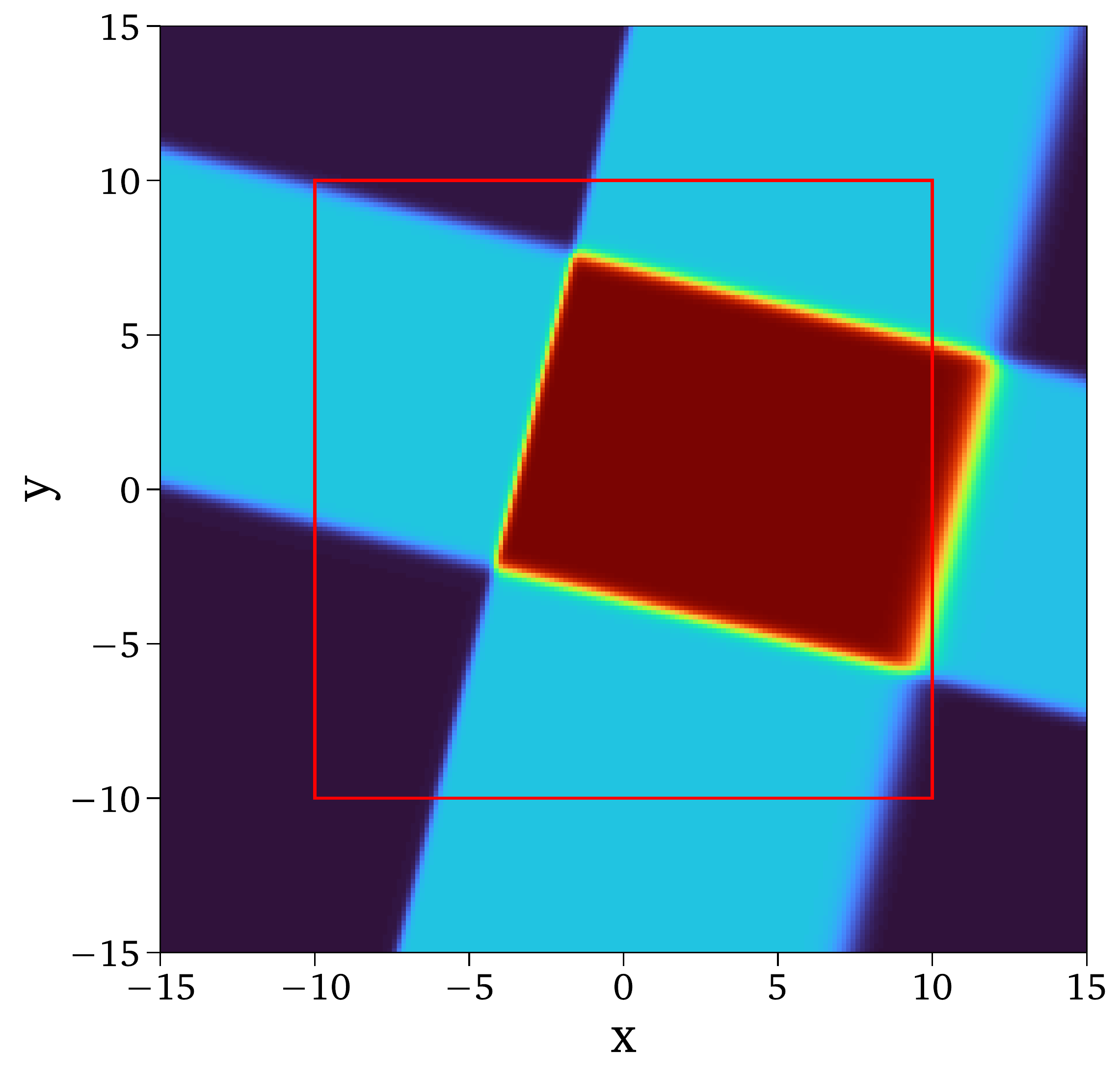}
    \end{subfigure}
    \begin{subfigure}{0.24\textwidth}
        \centering
        \caption{}
        \label{fig:Rect_flexDeepONet_4}
        \includegraphics[width=1.5in]{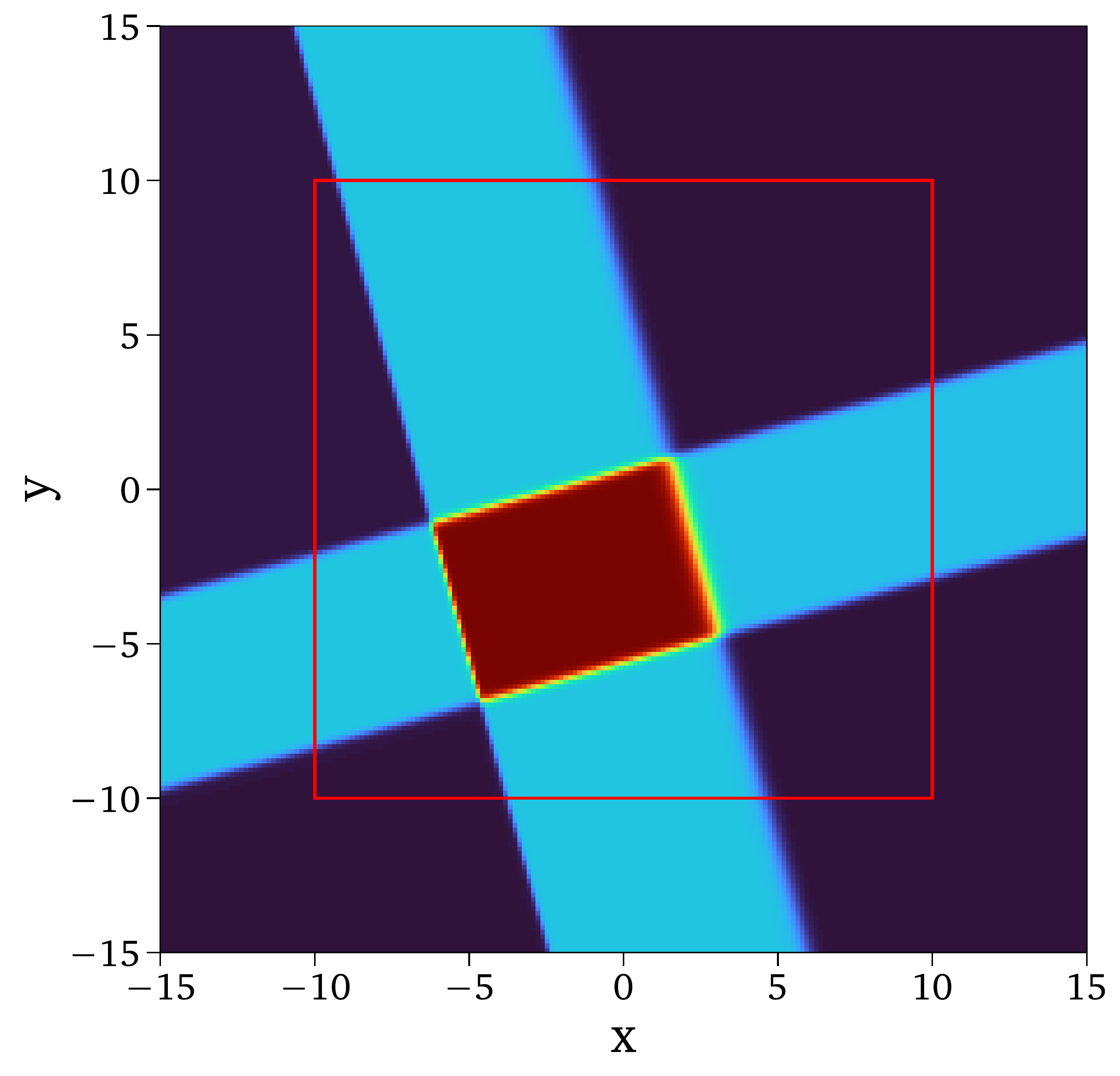}
    \end{subfigure}
    \caption{\textbf{FlexDeepONet predictions of the rigid body's dynamics}. The snapshots are produced at $t=\{0,3.33,6.67,10\}$ [s]. Note: FlexDeepONet's training relied only on data randomly-selected inside the $x-y$ domain represented by the red squares. Predictions outside the red squares are extrapolations. See Fig.~\ref{fig:Rect_DeepONet_2} for FlexDeepONet's structure and Fig.~\ref{fig:Rect_flexDeepONet_Supp} for additional time snapshots.}
    \label{fig:Rect_flexDeepONet}
\end{figure}
This last test case's objective is to reproduce 2D rigid body dynamics, the rotation, translation, and stretch of which are non-linear functions of time (a cosine, a spiral, and a sine, respectively). The equations used for producing the test case can be found in Sec.~S\ref{Sec:Supp_Test4} of the supplementary material, and some time-snapshots of the dynamics are reported in Fig.~\ref{fig:Rect_Orig}. In the following, we assume that we have no knowledge of the PDE that governs the motion and are forced to reconstruct the dynamics fully from the data. At this point, it is important to highlight the conceptual differences between this problem and the three test cases previously analyzed. In the examples above, we studied the full operators for a wide range of scenarios. Here, instead, we are interested in a particular instance of the dynamics based on a single (i.e., predefined) initial condition. As a consequence, the analysis of the problem from an SVD-based perspective involves the construction of a time-aggregated snapshot matrix consistently with the POD methodology~\cite{SymmPOD_Brunton_2019}. Such a matrix is generated by selecting 200 time instants in the dynamics between $t=0$ [s] and $t=10$ [s] and by populating the columns with the solution corresponding to a $[200 \times 200]$ $x-y$ grid at each specific time, appropriately reshaped to a $[40,000 \times 1]$ vector. Due to the motion being dominated by symmetries, the SVD fails to discover an accurate low-rank decomposition, and the application of the classical POD method requires more than 150 modes to recover 95\% of the dynamics' cumulative energy (Fig.~\ref{fig:Rect_CumEnergy_All}). As demonstrated in Fig.~\ref{fig:Rect_PC_1}, the encoding-decoding errors generated by the compression of the time-aggregated snapshot matrix into the dominant POD modes are visible even when 128 singular solutions are retained. \\
With the aim of learning the body's dynamics, we test two separate DeepONet architectures. In contrast to the three cases above, the $t$ (i.e., time) variable is used as the branch's input, while the $x$ and $y$ space coordinates are passed to the trunk. This unconventional application of the DeepONet architecture should be emphasized: instead of employing it as a neural operator, it is here used for the automatic generation of a reduced-order surrogate of the single scenario. Moreover, this type of approach should not be confused with the POD-DeepONet technique by Lu \textit{et al.}~\cite{PODDeepONet_Lu_2021}, still applied to the neural operator construction and in which the trunk net is replaced by the modes of the scenario-aggregated data matrix.\\
In order to further demonstrate DeepONet's capabilities, we train the surrogates with a data set that has two remarkable differences from the one employed in the POD analysis above:
\begin{itemize}
    \item Instead of using the 8,000,000 data points distributed in two hundred equally-spaced time snapshots with the same grid of $[200 \times 200]$ $x$ and $y$ nodes, we randomly sample 1,000,000 data points in the $(x,y;t)$ space. 80\% of them are used for training, and the remaining 20\% for validation.
    \item Rather than generating training data in a $(x,y) \in (-15,15)\times(-15,15)$ domain, we restrict the sampling of the spatial coordinates to the $(x,y) \in (-10,10)\times(-10,10)$ subset region. Time, instead, is sampled over the full $t \in (0,10)$ domain.
\end{itemize}
We firstly train a relatively large vanilla DeepONet, in which branch and trunk nets rely on 128 outputs and are composed of six hidden layers and 128 neurons each, for a total of 165,761 parameters. Consistent with the extremely slow decay of the singular values and with what we observed in the previous test cases, the vanilla DeepONet is noticeably inaccurate, as reported in the supplementary material at sixteen testing times. The surrogate is tested for predictions at locations contained in the $(x,y) \in (-10,10)\times(-10,10)$ training domain but also for extrapolations outside the square. The body's shape is highly distorted by the ineffective compression to the 128 bases and by the difficulties of the sub-nets in retrieving the modes and the related coefficients. The erroneous patterns are even accentuated outside the training domain, as expected.\\
We then apply the flexDeepONet to the same regression task. Given the different types of symmetries involved in the dynamics, the Pre-Net employed in this test case outputs four different variables: a rotation angle, $\bar{\theta}$, a stretching factor, $\bar{s}$, and two shifting parameters, $\bar{x}$ and $\bar{y}$. These four quantities are here autonomously regressed from data through three separate blocks (i.e., rotation net, stretch net, and shift net, as reported in Fig.~\ref{fig:Rect_DeepONet_2}), as an alternative to using a single FNN. The coefficients are then passed to a transformation layer, which combines them with the spatial coordinates via a scaled rotation matrix, based on $\bar{s}$ and $\bar{\theta}$, and a shifting vector, formed as $[\bar{x}, \bar{y}]^T$. The two quantities outputed by this transformation layer are finally passed to the trunk net as inputs. The flexDeepONet's branch and trunk nets are significantly smaller than the vanilla's corresponding sub-nets, as we impose $p=1$. In this way, the trunk net is asked to retrieve only one mode from the data. However, because this basis is defined in the moving reference frame discovered by the Pre-Net, the single mode that the trunk sees and aims to reconstruct is the full rigid body itself, rather than one of the dynamics' low-energy-content components. This key aspect has vital consequences on the surrogate's generalizing capabilities. As shown in Fig~\ref{fig:Rect_flexDeepONet}, flexDeepONet remarkably improves the predictions/extrapolations of the rigid body's dynamics for $x,y$ coordinates both inside and outside the training window at unseen time instants. Furthermore, these advancements are achieved by relying on only 1,921 trainable parameters, which corresponds to a 98.8\% reduction compared to the vanilla architecture. \\



\section{Summary and Discussion}\label{sec:Conclusions}

Recent in-depth analysis and applications to real-world problems in many fields have demonstrated the unprecedented capabilities of DeepONet as a neural operator surrogate approach.
The effectiveness and efficiency of the technique are in part attributable to the projection component of the architecture, which is implemented as dot-product layers.
Some unique characteristics make DeepONet significantly more flexible than the classical fully data-driven projection-based methods, as it overcomes many limitations of the latter. For instance, it allows the construction of physics-informed surrogates by also relying on sparse and multifidelity datasets. In this paper, we have shown that there are also shortcomings that DeepONet inherits from the projection-based attribute, and we propose suitable remedies that can extend its flexibility. \\
We initially applied DeepONet to learn the operators underlying three dynamical systems, all described by relatively simple ODEs. The first test case was based on a mass-spring-damper model and allowed us to propose a novel approach to DeepONet training: SVD-DeepONet. While the applicability of this technique to real-world problems might be limited to cases in which the data is available on scenario-spatio-temporal grids (i.e., cases in which the data allows us to perform the SVD), it has the instructive purpose of drawing parallelisms between data decompositions methods and the DeepONet architecture. In fact, we showed that its trunk and branch nets regress the modes and the related coefficients discovered by the projection. Moreover, if SVD can be performed on subsets of DeepONet’s training data, the SVD- approach can be employed in an exploratory phase, in which the number of trunk outputs and the capacity of the subnets (e.g., depths and widths) need to be decided. \\
From the analogy with projection-based methods, we also developed a shared-trunk formulation. This modification can be seen as a dimensionality reduction of the full DeepONet architecture, as multiple state variables share the same trunk net under the assumption of underlying similarities in their modes. In future work, we will analyze other strategies for combining dimensionality reduction techniques to multi-output DeepONets, such as projecting the state variables into lower-dimensional spaces or grouping them into macro-variables~\cite{ DeepONetApps_Zanardi_2022}. These approaches will include PCA, non-linear manifold learning, and coarse-graining.
Furthermore, in light of the connections between low-rank decompositions and DeepONet, an additional research direction is the analysis of multi-input DeepONets~\cite{MIDeepONet_Jin_2022,MIDeepONet_Tan_2022} under the perspective of tensor decomposition~\cite{TensordDecomposition_Bernardi_2013}.\\
The second and third test cases revealed DeepONet’s inefficiencies in surrogating dynamics characterized by symmetries. Scenario-aggregated SVDs need many modes of increasing complexity for accurately projecting operators affected by translations, rotations, and/or scalings. Similarly, the vanilla DeepONet architecture requires several trunk outputs and remarkably expressive subnets for the same applications. This significantly increases computational costs and challenges of both the training and prediction phases. Therefore, we employed a simple but effective modification to the original structure. We added a pre-transformation block that acted as an artificial intelligence (AI)-operated data alignment aimed at symmetry removal. This sub-net automatically discovered a moving frame of reference with respect to which the multiple scenarios were efficiently compressible in fewer modes.
We also included an additional branch output to learn the underlying scenario-based centering. The resulting architecture, called flexDeepONet, drastically reduced the size of the surrogate without compromising its accuracy. In particular, flexDeepONet’s trainable parameters for emulating the combustion chemistry test case (19 state variables spanning many orders of magnitudes) were about 95\% fewer than those of vanilla DeepONets characterized by larger generalization errors. Thanks to the resulting speed-ups, we believe that flexDeepONet can benefit the surrogation of operators with latent symmetries, particularly in real-time applications and digital twins. \\
The present comparative analysis of DeepONet and SVD-based methodologies is far from being exhaustive. More work needs to be done in constructing rigorous mathematical definitions of the analogies mentioned above. Together with the extension to PDEs, this will be a core focus of future work. Overall, this paper specifically aims to interpret the DeepONet structure and suggest a promising research direction. In light of these parallelisms, multiple concepts behind refined SVD-based methods can be adapted and introduced to DeepONet. For example, underlying ideas can be recovered from robust PCA to work with corrupted data and probabilistic PCA to represent the uncertainties caused by limited information.\\
We also believe that the contributions between DeepONet and SVD-based methods can be mutual, and the flexibility of the former can extend some of the PCA, DMD, and POD applications. For this purpose, we included a fourth test case that applied the DeepONet to the surrogation of a two-dimensional rigid body's dynamics characterized by a combination of rotations, translations, and stretching constructed as non-linear functions of time. Our employment of the architecture was by design inconsistent with the neural operator formulation, and the resulting surrogate was only meant to reproduce a single scenario of the dynamics. However, the purpose of this last test case was to show that the DeepONet has the potential to extend the applicability of POD's frameworks in two main directions: 
\begin{itemize}
    \item The training data is not required to be aligned on spatial and temporal grids
    \item Predictions can be obtained at generic spatial and temporal coordinates
\end{itemize}
While we do not expect the flexDeepONet to significantly increase the performance of the vanilla architecture in applications in which rigid body motions are negligible, we found that the proposed extension adds significant advantages for problems that involve symmetries and for which POD-based strategies require a large number of modes:
\begin{itemize}
    \item It increases the surrogate's interpretability, as the dynamics are learned on an automatically-discovered moving reference frame. This allows the isolation of the rigid features and the a-posterior analysis of the frame's coordinates
    \item It augments the generalization capabilities as a direct consequence of the previous point. In the last test case, we showed that the flexDeepONet is able to accurately reconstruct the rigid body even outside the borders of the spatial training window
    \item It drastically increases the efficiency of the surrogate, as it improves the effectiveness of the projection. For the last test case in which POD needed 150 modes to retain 95\% of the cumulative energy, a flexDeepONet was constructed with only 1,921 trainable parameters, which is 98.8\% fewer than that required by the vanilla architecture, which still resulted in extremely low accuracy 
\end{itemize}


\section*{Acknowledgments}

This research was supported by the Exascale Computing Project (17-SC-20-SC), a collaborative effort of the U.S. Department of Energy Office of Science and the National Nuclear Security Administration. Sandia National Laboratories is a multimission laboratory managed and operated by National Technology and Engineering Solutions of Sandia, LLC., a wholly owned subsidiary of Honeywell International, Inc., for the U.S. Department of Energy’s National Nuclear Security Administration under contract DE-NA-0003525. This paper describes objective technical results and analysis. Any subjective views or opinions that might be expressed in the paper do not necessarily represent the views of the U.S. Department of Energy or the United States Government.

This article has been authored by an employee of National Technology \& Engineering Solutions of Sandia, LLC under
Contract No. DE-NA0003525 with the U.S. Department of Energy (DOE). The employee owns all right, title and interest 
in and to the article and is solely responsible for its contents. The United States Government retains and the
publisher, by accepting the article for publication, acknowledges that the United States Government retains a 
non-exclusive, paid-up, irrevocable, world-wide license to publish or reproduce the published form of this article 
or allow others to do so, for United States Government purposes. The DOE will provide public access to these
results of federally sponsored research in accordance with the DOE Public Access Plan 
https://www.energy.gov/downloads/doe-public-access-plan.




\printbibliography 


\titleformat{\section}[hang]{\bfseries}{S\thesection.\ }{5pt}{}
\titleformat{\subsection}[hang]{\itshape}{S\thesubsection.\ }{5pt}{}
\setcounter{section}{0}

\beginsupplement

\clearpage
\section{Supplementary Material for Test Case 1}

\subsection{Training and Test Scenarios}

\begin{figure}[!htb]
    \begin{subfigure}{0.49\textwidth}
        \centering
        \caption{}
        \label{fig:MSD_Data_Ics}
        \includegraphics[width=3.1in]{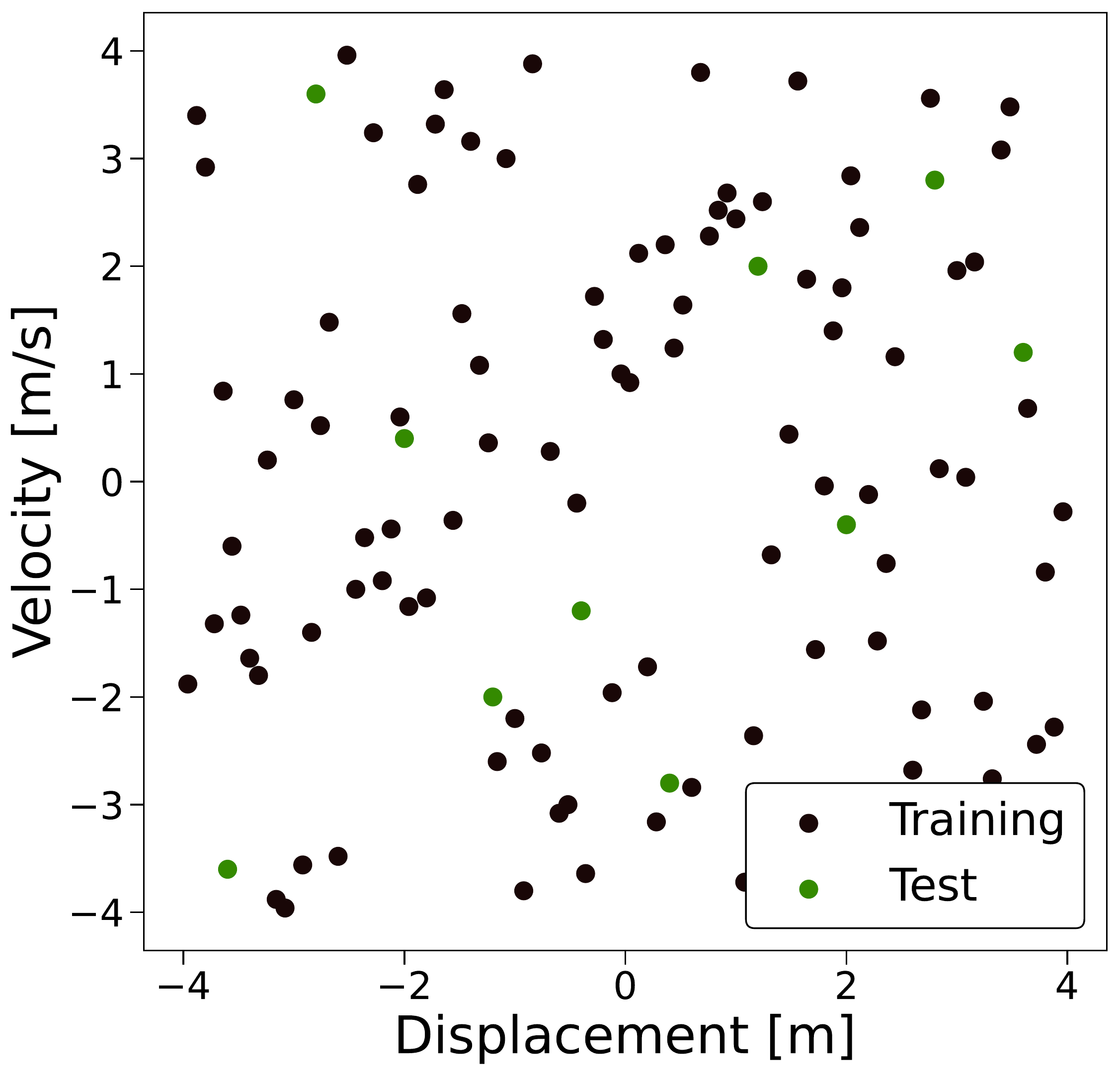}
    \end{subfigure}
    \begin{subfigure}{0.49\textwidth}
        \centering
        \caption{}
        \label{fig:MSD_Data_Train}
        \includegraphics[width=3.05in]{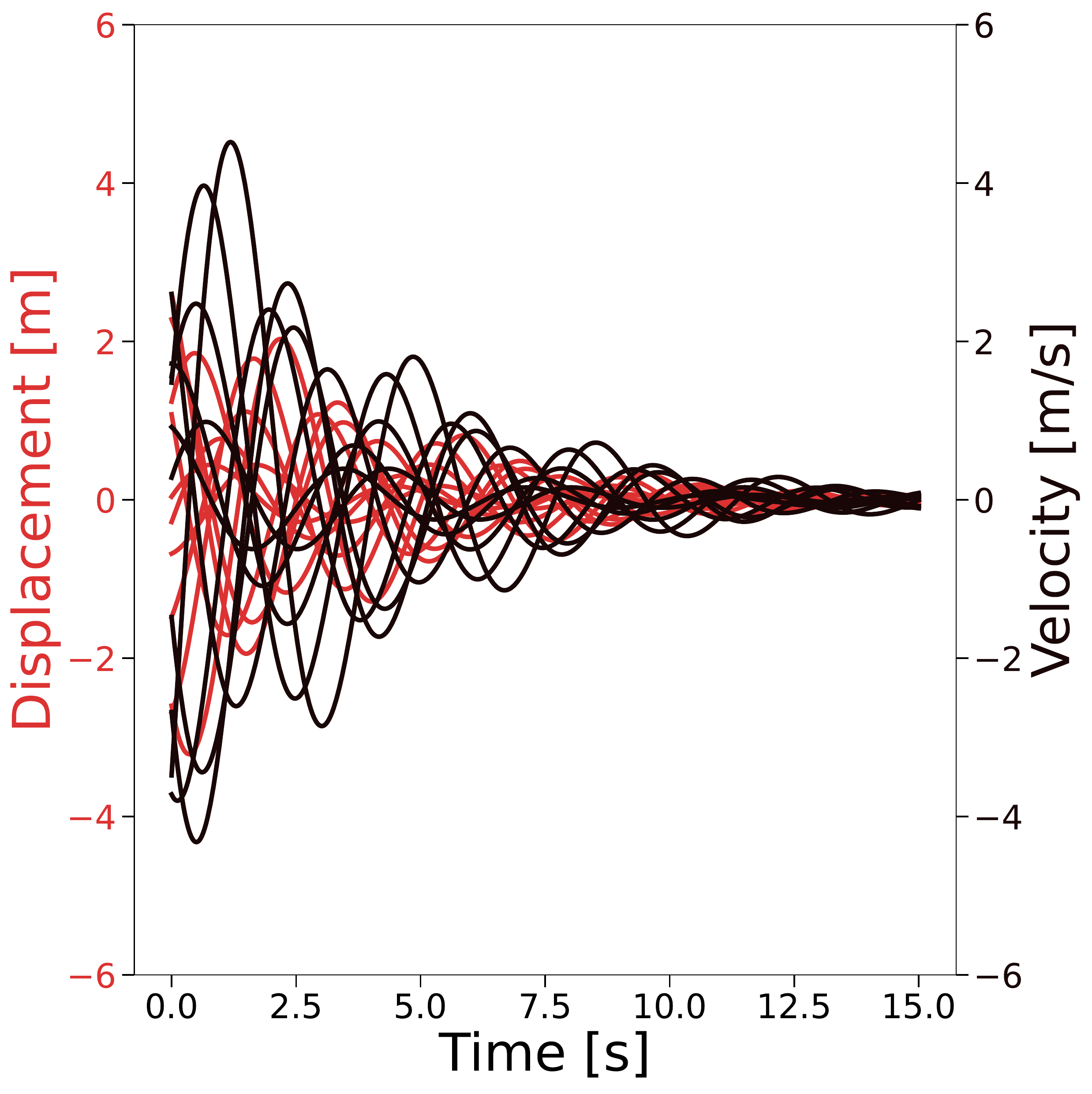}
    \end{subfigure}
    \caption{\textbf{Training and test data for the mass-spring-damper test case}. (\textbf{A}): Initial conditions, randomly selected based on Latin hypercube sampling. (\textbf{B}): Ten examples of training scenarios.}
    \label{fig:MSD_Data}
\end{figure}

\clearpage
\subsection{Vanilla Deep Operator Network (DeepONet)}

\clearpage
\subsection{Singular Value Decomposition Deep Operator Network (SVD-DeepONet)}

\begin{figure}[!htb]
    \begin{subfigure}{0.49\textwidth}
        \centering
        \caption{}
        \label{fig:MSD_CumEnergy_x}
        \includegraphics[width=3.2in]{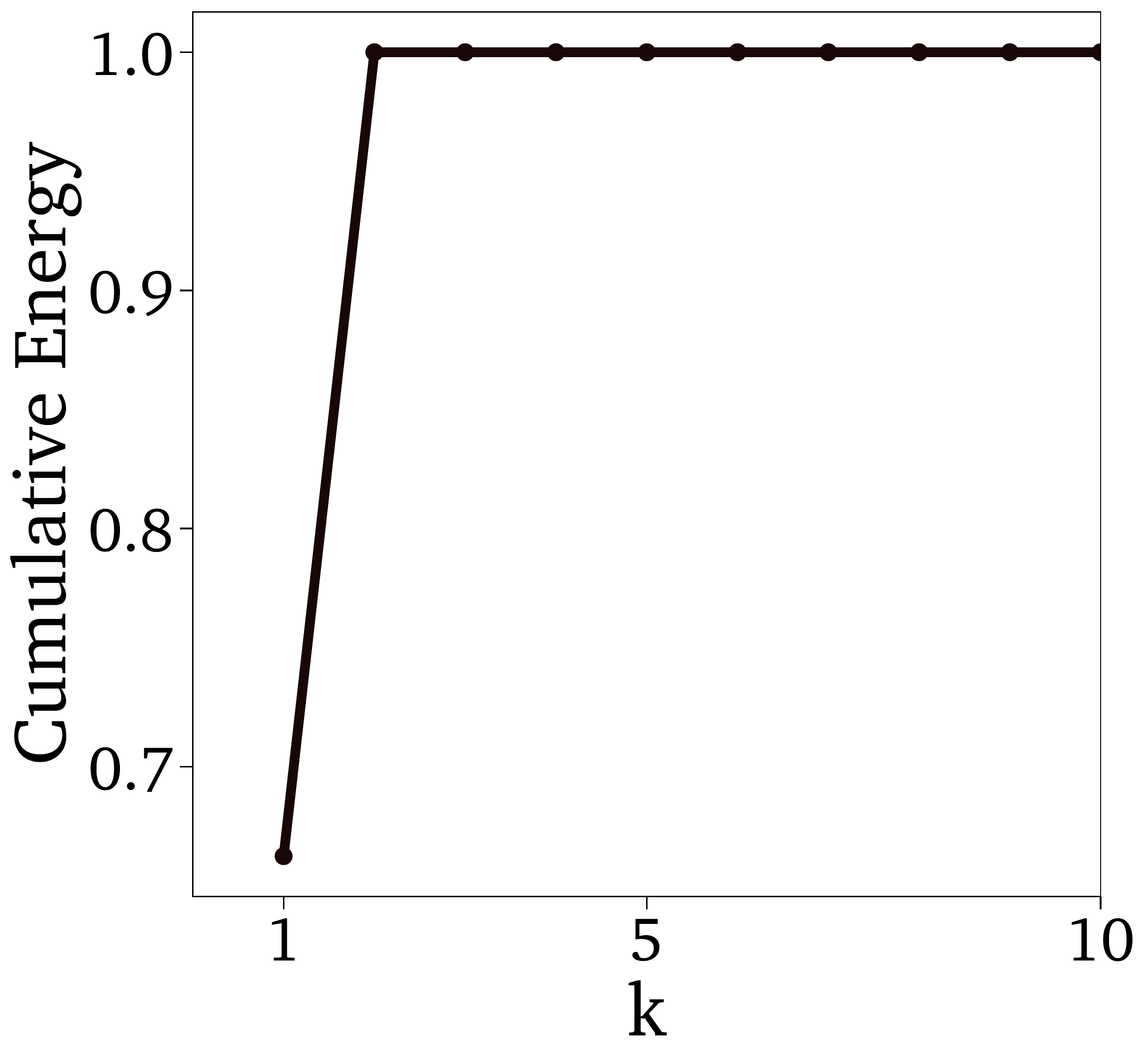}
    \end{subfigure}
    \begin{subfigure}{0.49\textwidth}
        \centering
        \caption{}
        \label{fig:MSD_CumEnergy_v}
        \includegraphics[width=3.2in]{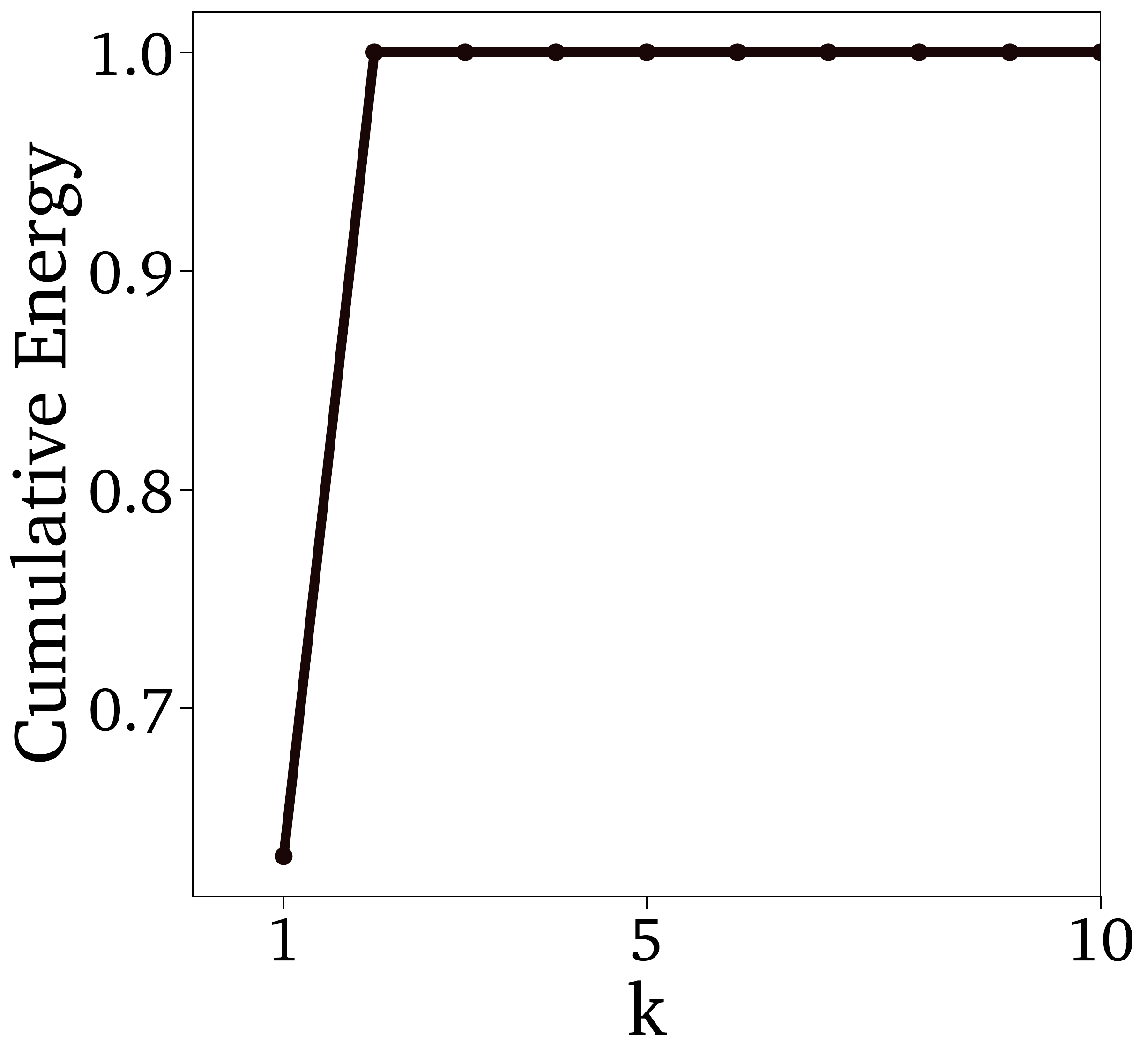}
    \end{subfigure}
    \caption{\textbf{Cumulative energies of the $\mathbf{X}$ and $\mathbf{V}$ matrices from the mass-spring-damper test case}. Cumulative energies contained in the first k singular values of $\mathbf{X}$ (\textbf{A}) and $\mathbf{V}$ (\textbf{B}).}
    \label{fig:MSD_CumEnergies}
\end{figure}

\clearpage
\subsection{Shared-Trunk SVD-DeepONet}

\begin{figure}[!htb]
    \centering
    \includegraphics[width=3.2in]{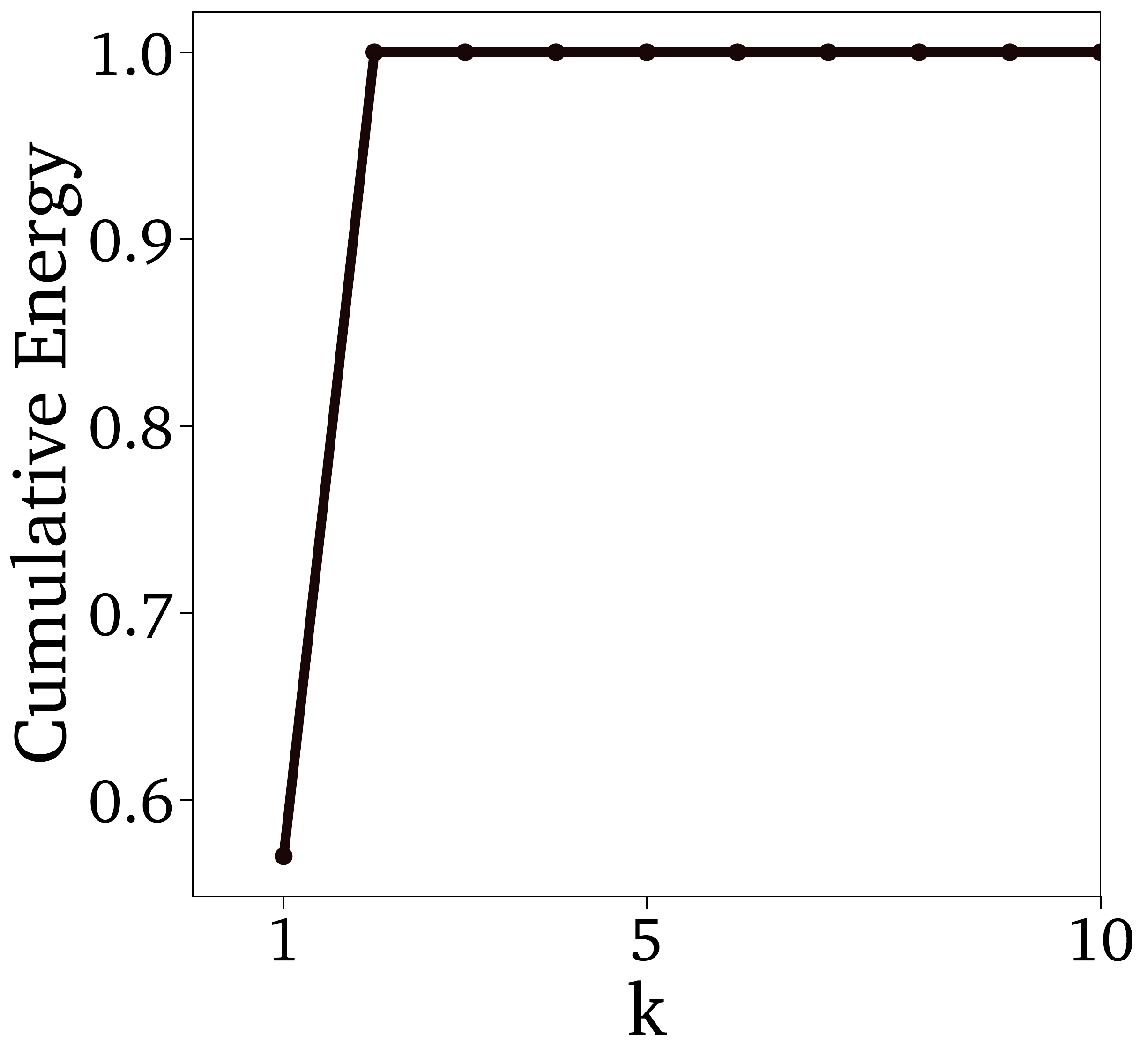}
    \caption{\textbf{Cumulative energies of the displacement-velocity combined data matrix,} $\mathbf{Z}$ \textbf{, from the mass-spring-damper test case}. Cumulative energies contained in the first k singular values of $\mathbf{Z}$.}
    \label{fig:MSD_CumEnergy_All}
\end{figure}

\subsection{Architectures Comparison}

\begin{table}[!htp] 
    \centering
    \begin{tabular}{||l|c|c|c||} 
        \hline
         & Vanilla DeepONet & POD-DeepONet & SVD-DeepONet \\[1.2ex] 
         & (Fig.~\ref{fig:MSD_DeepONet_1}) & & (Fig.~\ref{fig:MSD_DeepONet_2}) \\[1.2ex] 
        \hline
        $p$ & 2 & 2 & 2 \\[1.8ex]
        \hline
        No. of Parameters & 9,034 & 9,034 & 9,098 \\[1.8ex]
        \hline
        \hline
        $x$, RMSE & 0.00102264 & 0.00174439 & 0.0013894 \\[1.2ex] 
        \hline
        $v$, RMSE & 0.00179907 & 0.00105212 & 0.0012283 \\[1.2ex] 
        \hline
        \hline
        Training Time & 17,528 & $(6,848)_{T} \times 2$ + 12,511 & $(6,848)_{T} \times 2$ + $(9,080)_{B}\times 2$ \\[1.2ex] 
                      & = 17,528 [s] & = 26,207 [s] & = 31,856 [s] \\[1.2ex] 
        \hline
        Prediction Time & 48 [ms] & 48 [ms] & 48 [ms] \\[1.2ex] 
        \hline
    \end{tabular}
    \vspace{0.5cm}
    \caption{\textbf{Comparison between architectures}. Comparisons between the number of trunks' outputs ($p$), the number of trainable parameters, the root mean squared errors (RMSEs) at the ten test scenarios, the training times, and the prediction times. Training has been performed via Tensorflow 2.8. For the training times, ${(...)}_T$ indicates the time necessary for training each of the trunk blocks, while ${(...)}_B$ indicates the time required for each branch. It should be noted, however, that the two trunks of POD-/SVD-DeepONet, as well as the two branches of the SVD-DeepONet, can be independently trained in parallel. Prediction times refer to the overall time necessary for the architectures to predict $x$ and $v$ at 10,000 times and initial conditions, and they have been computed by relying on architectures' implementations based on the NumPy 1.22.3 library.}\label{table:MSD_TestErrors}
\end{table}


\clearpage
\section{Supplementary Material for Test Case 2}

\subsection{Training Scenarios}

\begin{figure}[!htb]
    \centering
    \includegraphics[width=3.2in]{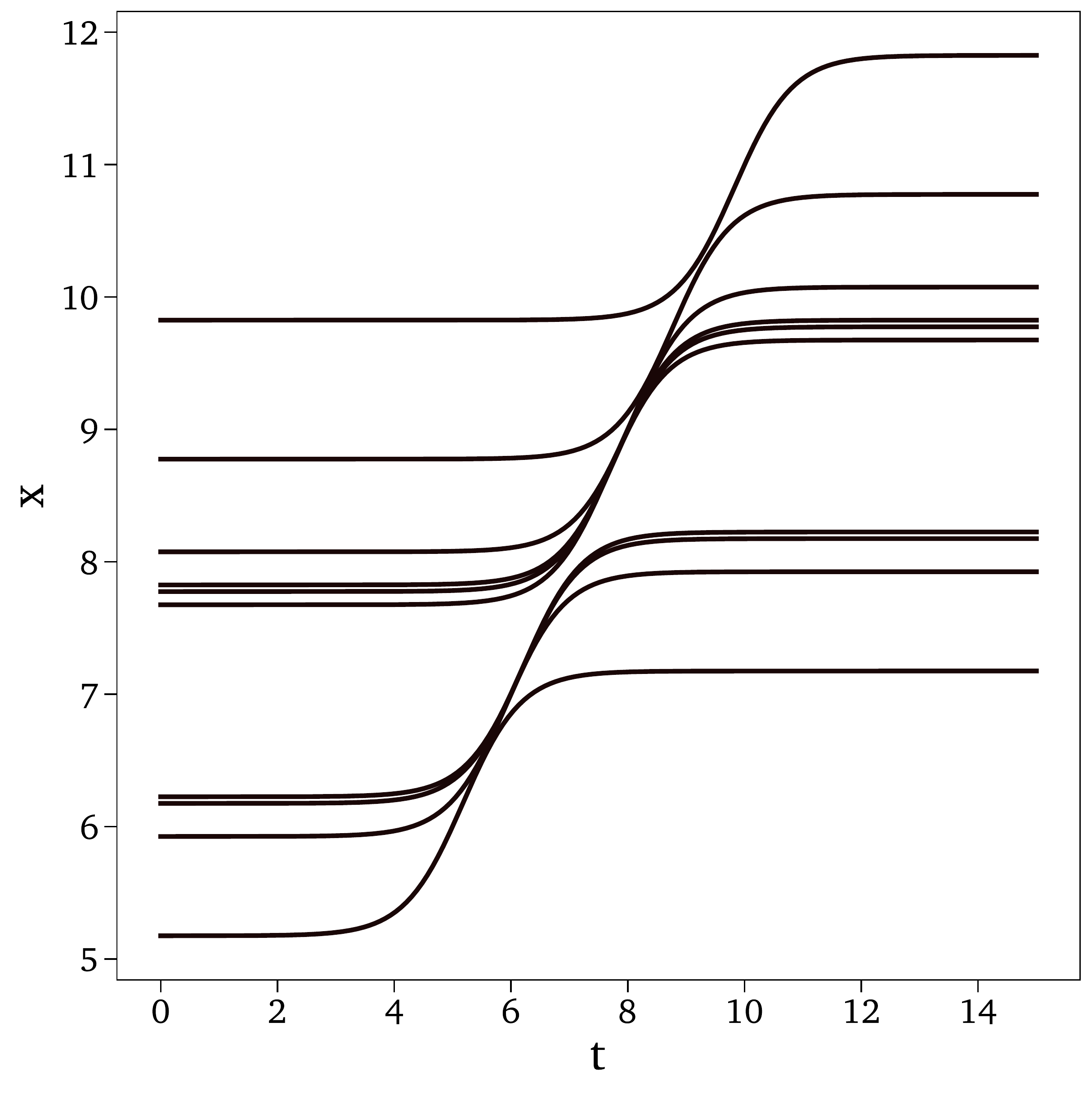}
    \caption{\textbf{Training data for the shifting hyperbolic function test case}. Ten examples of training scenarios.}
    \label{fig:Tanh_TrainData}
\end{figure}

\subsection{Vanilla Deep Operator Network (DeepONet)}

\clearpage
\subsection{Singular Value Decomposition (SVD)}

\clearpage
\subsection{Flexible Deep Operator Network (flexDeepONet)}

\begin{figure}[!htb]
    \centering
    \includegraphics[width=3.2in]{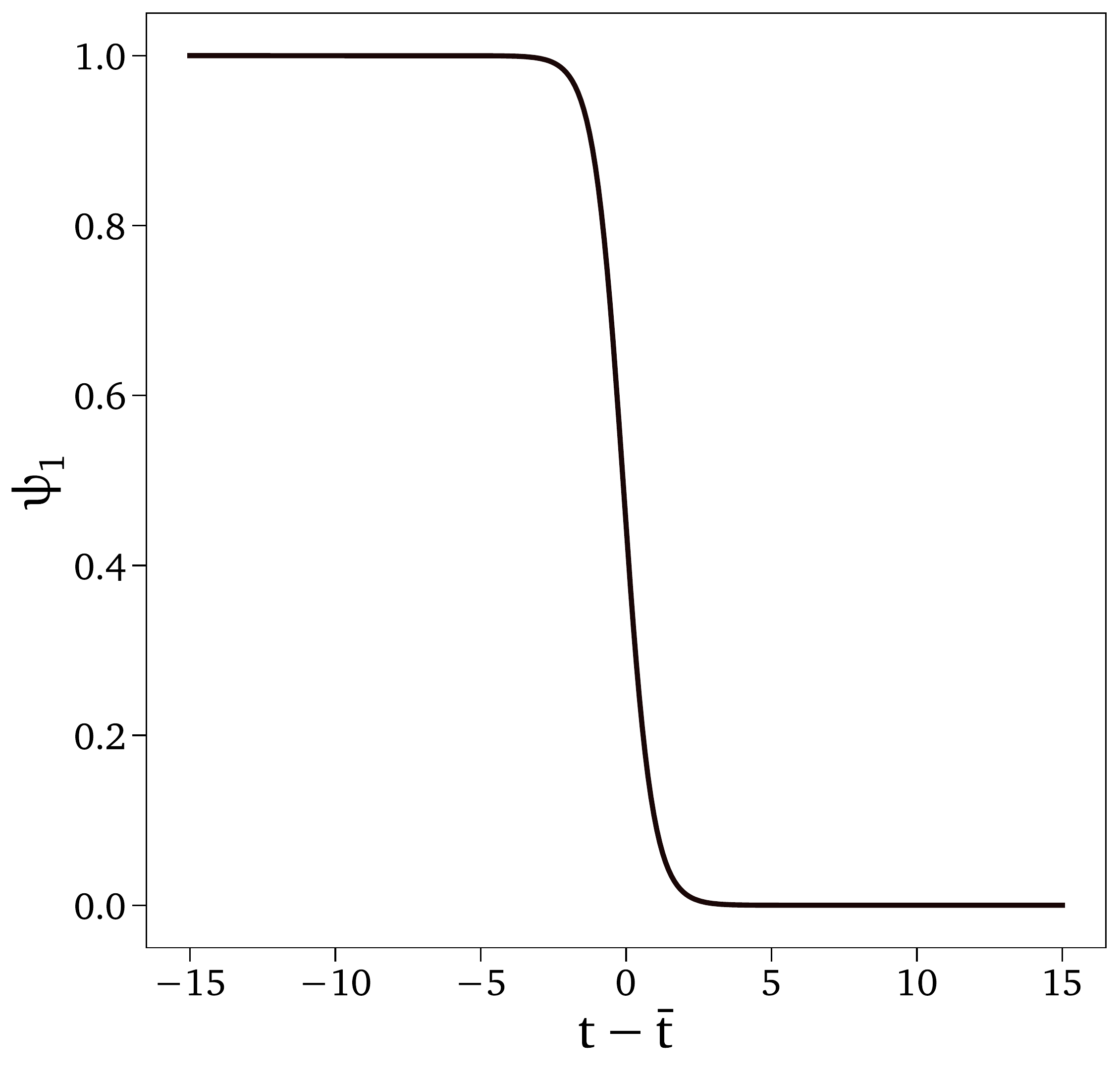}
    \caption{\textbf{Output of flexDeepONet's trunk for the shifting hyperbolic function test case}. Output of the flexDeepONet's trunk net as function of its input (i.e., as function of the time processed by the shifting block).}
    \label{fig:Tanh_SDeepONet_Modes}
\end{figure}

\subsection{Architectures Comparison}

\begin{table}[!htp] 
    \centering
    \begin{tabular}{||l|c|c|c||} 
        \hline
        & Vanilla DeepONet & POD-DeepONet & flexDeepONet \\[1.2ex] 
        & (Fig.~\ref{fig:Tanh_DeepONet_1}) & & (Fig.~\ref{fig:Tanh_flexDeepONet}) \\[1.2ex] 
        \hline
        $p$ & 8 & 8 & 1 \\[1.2ex] 
        \hline
        No. of Parameters & 4,881 & 4,881 & 18 \\[1.8ex]
        \hline
        \hline
        RMSE & 0.00147976 & 0.00144003 & 0.00028490 \\[1.2ex] 
        \hline
        \hline
        Training Time & 1,013 & $(1,709)_T$ + 692 & 361 \\[1.2ex] 
                      & = 1,013 [s] & = 2,401 [s] & = 361 [s] \\[1.2ex] 
        \hline
        Prediction Time & 24 [ms] & 24 [ms] & 0.45 [ms] \\[1.2ex] 
        \hline
    \end{tabular}
    \vspace{0.5cm}
    \caption{\textbf{Comparison between architectures}. Comparisons between the number of trunks' outputs ($p$), the number of trainable parameters, the root mean squared errors (RMSEs) at the ten test scenarios, the training times, and the prediction times. Training has been performed via Tensorflow 2.8. For the training times, ${(...)}_T$ indicates the time necessary for training each trunk block. Prediction times refer to the overall time necessary for the architectures to predict $x$ at 10,000 times and initial conditions, and they have been computed by relying on architectures' implementations based on the NumPy 1.22.3 library.}\label{table:Tanh_TestErrors}
\end{table}


\newpage
\section{Supplementary Material for Test Case 3}

\subsection{Training and Test Scenarios}

\begin{figure}[!htb]
    \begin{subfigure}{0.49\textwidth}
        \centering
        \caption{}
        \label{fig:0DReact_ICs}
        \includegraphics[width=3.2in]{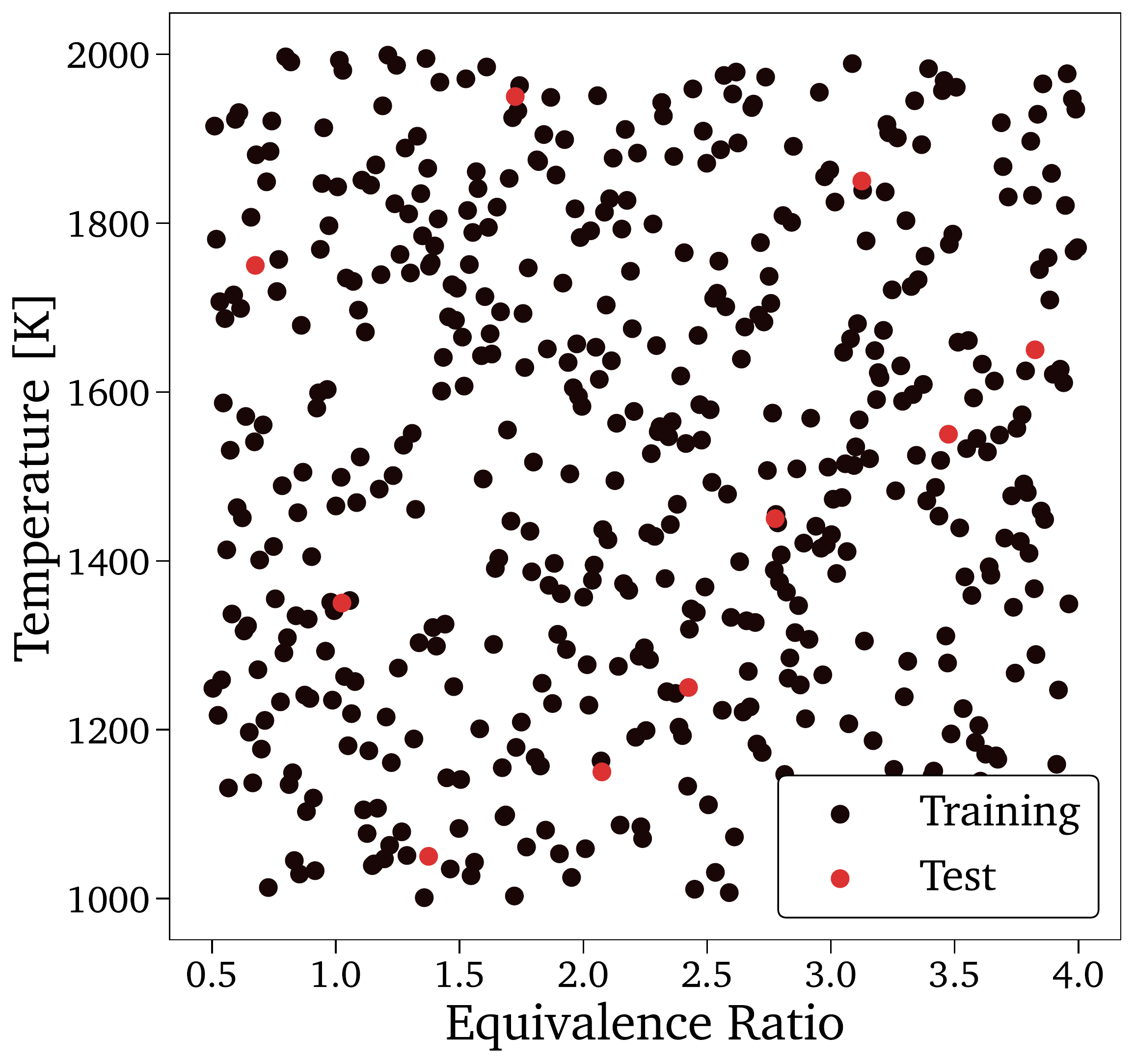}
    \end{subfigure}
    \begin{subfigure}{0.49\textwidth}
        \centering
        \caption{}
        \label{fig:0DReact_Train_HandNH3}
        \includegraphics[width=3.2in]{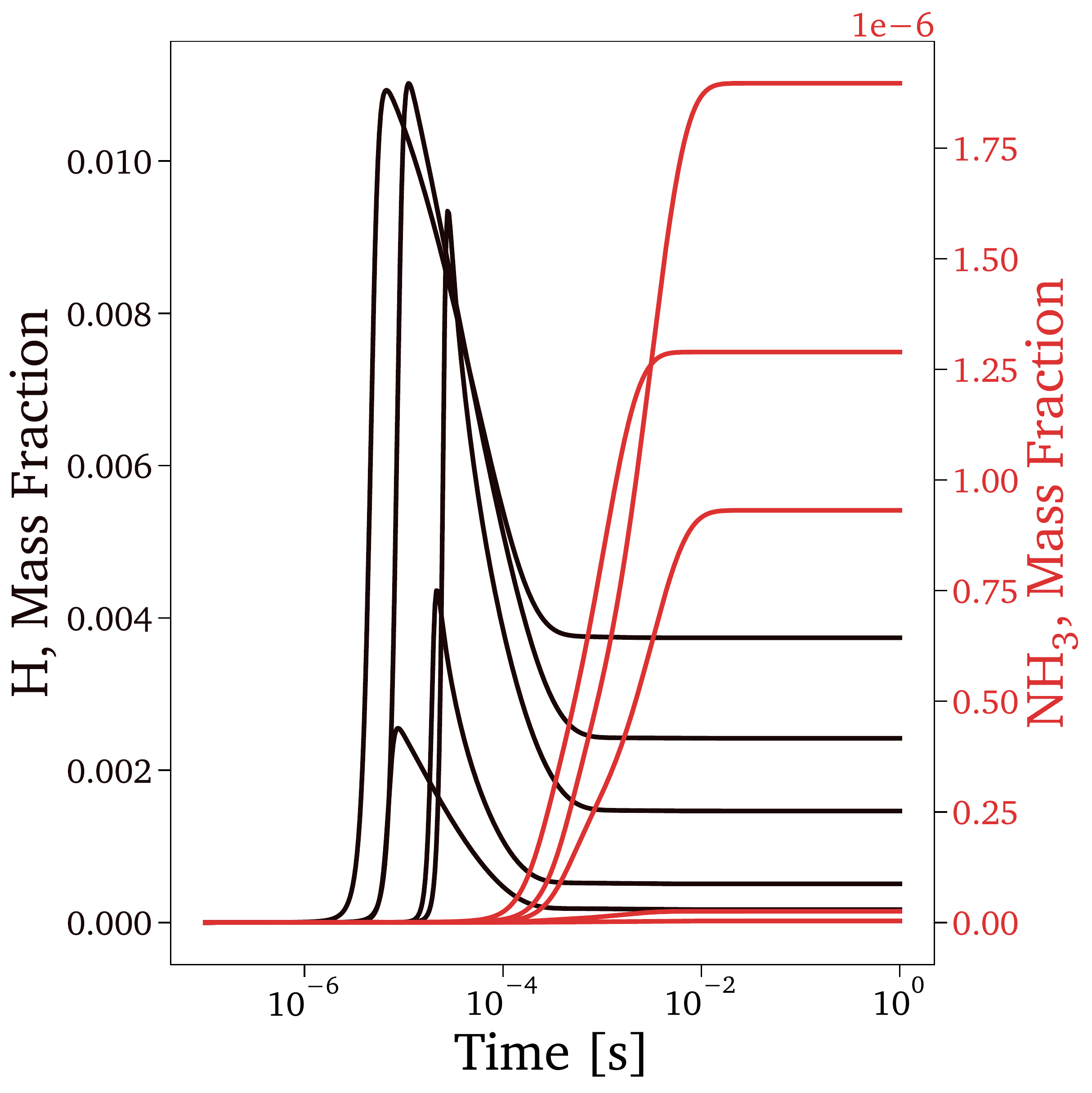}
    \end{subfigure}
    \caption{\textbf{Training and test data for the combustion chemistry test case}. (\textbf{A}): Initial conditions, randomly selected based on Latin hypercube sampling. (\textbf{B}): Five examples of training scenarios for H and NH$_3$ mass fractions.}
    \label{fig:0DReact_Train_}
\end{figure}

\clearpage
\subsection{Vanilla DeepONet}

\begin{figure}[!htb]
    \centering
    \includegraphics[width=6.0in]{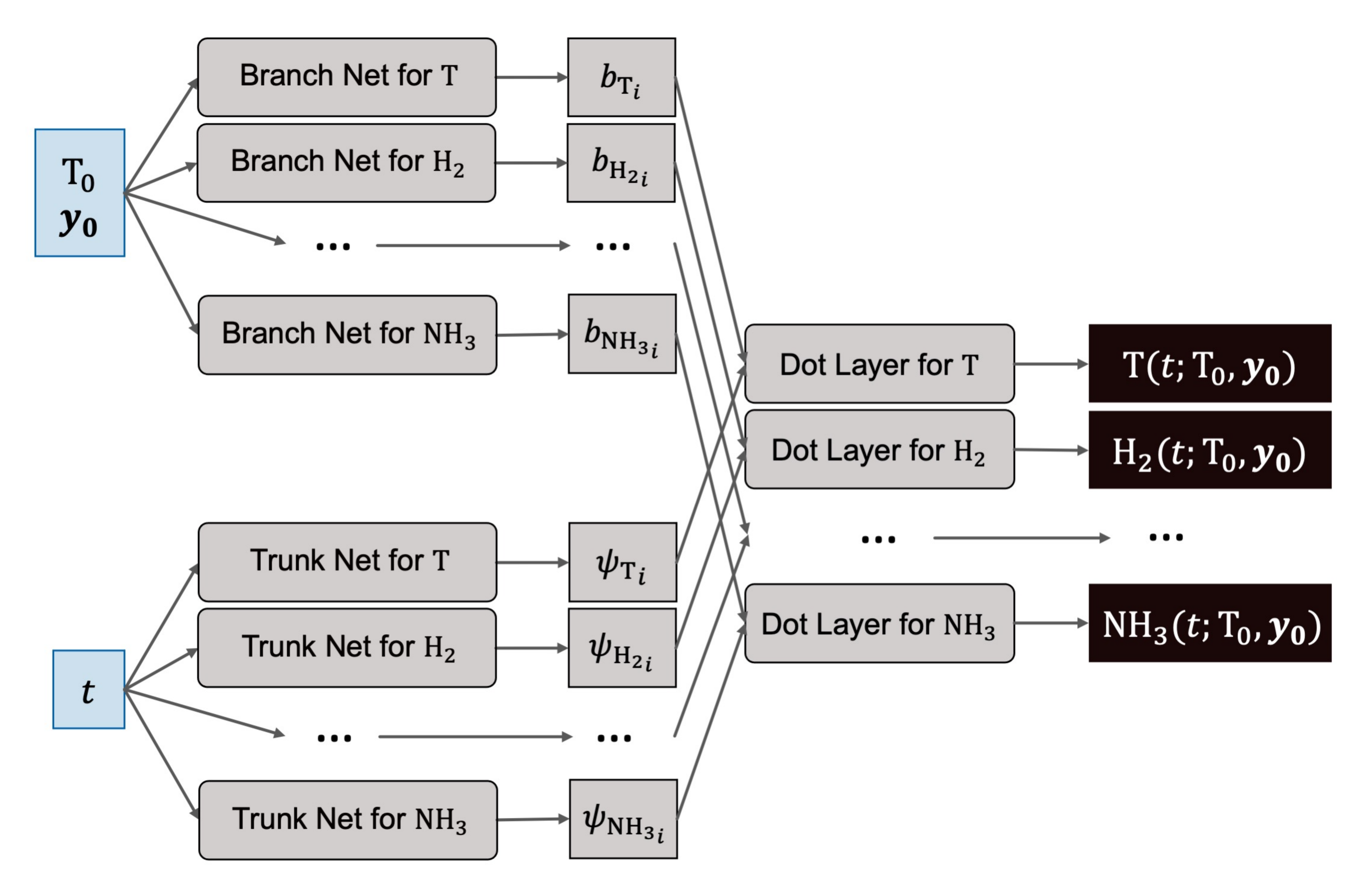}
    \caption{\textbf{FlexDeepONet architecture for the combustion chemistry test case}. Each of the 19 thermodynamic state variables has its own trunk and branch nets.}
    \label{fig:0DReact_DeepONet_All}
\end{figure}

\subsection{Singular Value Decomposition (SVD)}

\subsection{flexDeepONet}

\begin{figure}[!htb]
    \begin{subfigure}{0.49\textwidth}
        \centering
        \caption{}
        \label{fig:0DReact_flex_T}
        \includegraphics[width=3.2in]{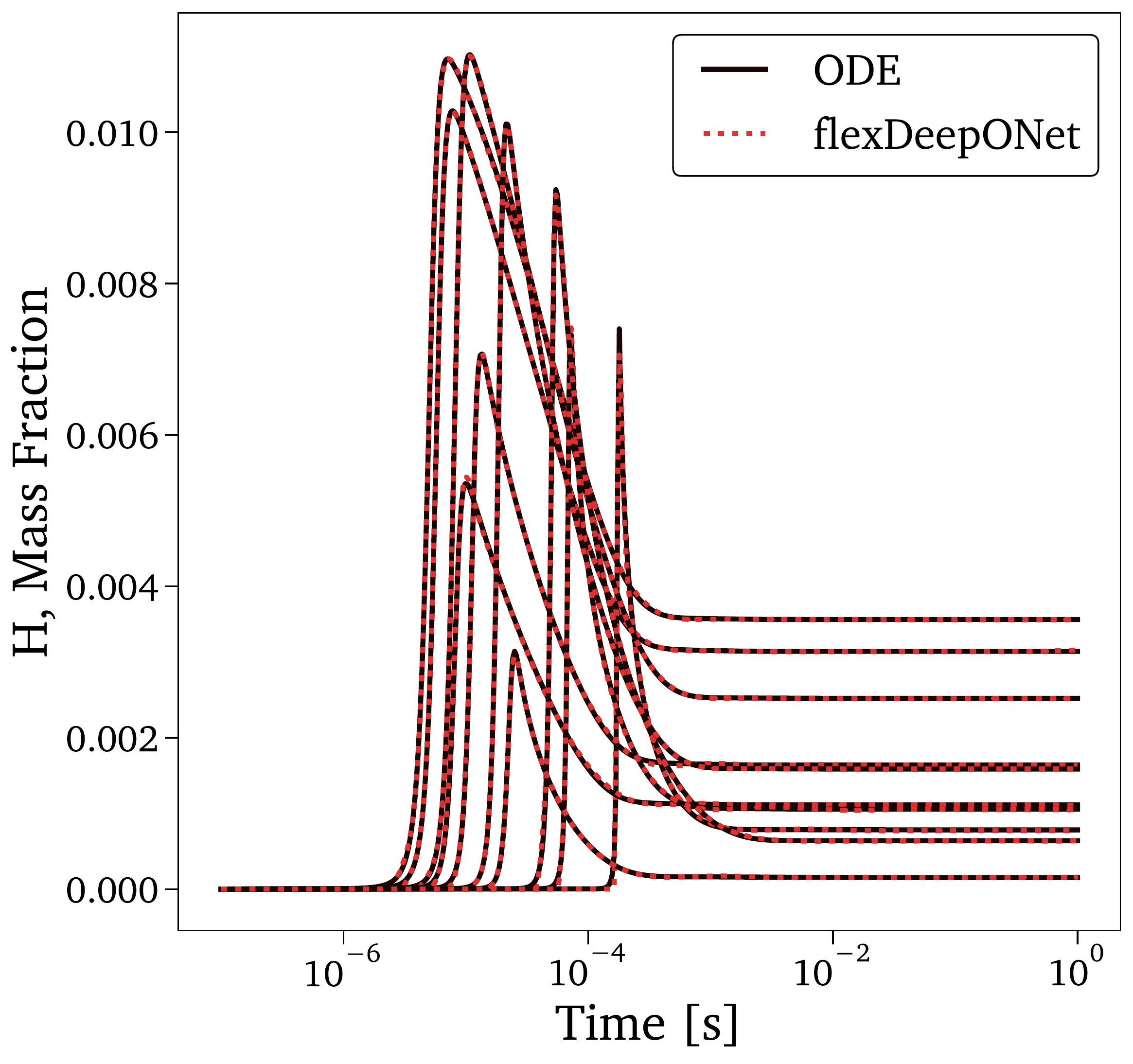}
    \end{subfigure}
    \begin{subfigure}{0.49\textwidth}
        \centering
        \caption{}
        \label{fig:0DReact_flex_OH}
        \includegraphics[width=3.2in]{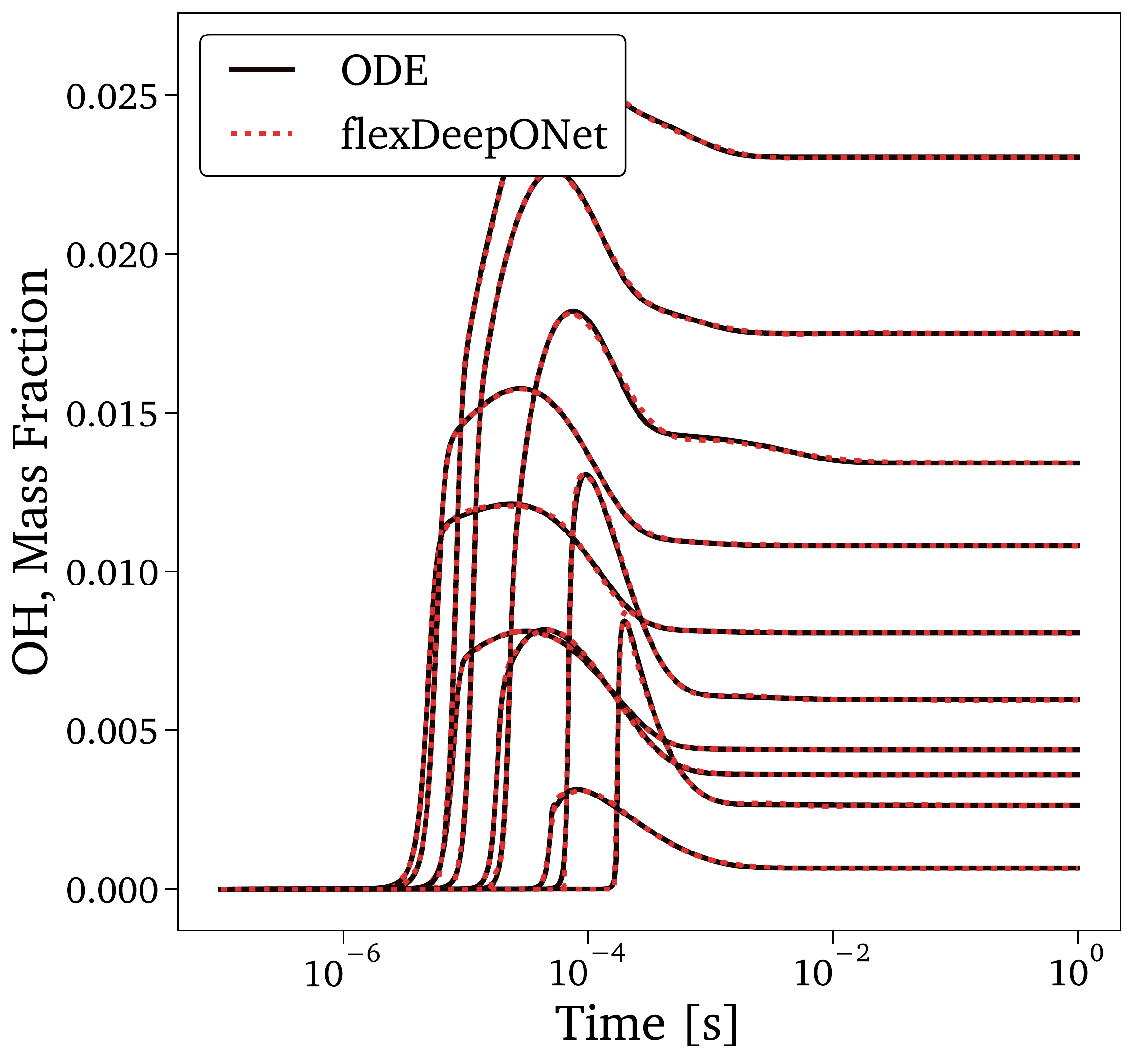}
    \end{subfigure}
    \begin{subfigure}{0.49\textwidth}
        \centering
        \caption{}
        \label{fig:0DReact_flex_H2O}
        \includegraphics[width=3.2in]{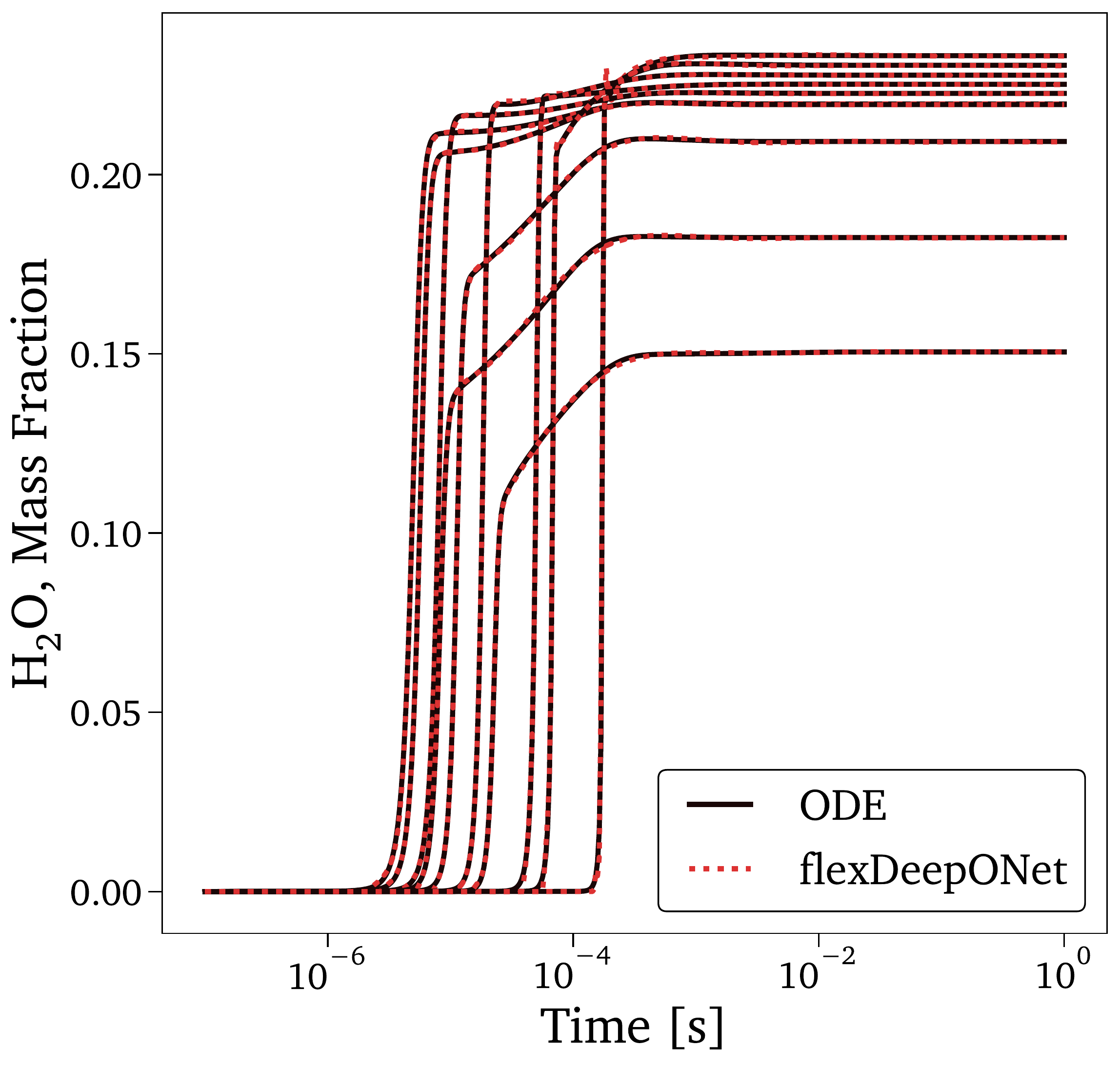}
    \end{subfigure}
    \begin{subfigure}{0.49\textwidth}
        \centering
        \caption{}
        \label{fig:0DReact_flex_NH3}
        \includegraphics[width=3.2in]{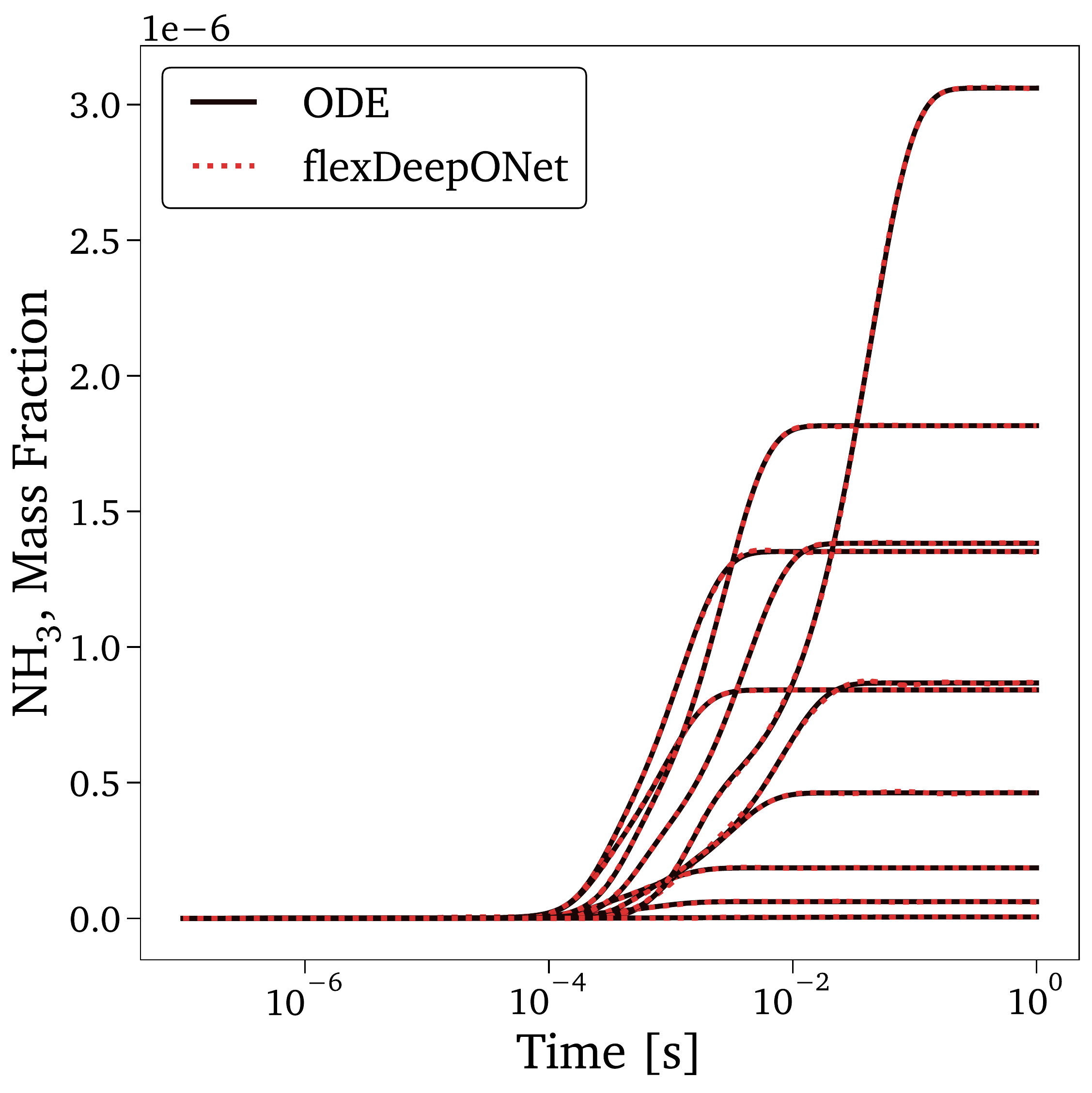}
    \end{subfigure}
    \caption{\textbf{FlexDeepONet predictions of the thermodynamic variables}. H (\textbf{A}), OH (\textbf{B}), H$_2$O (\textbf{C}), and NH$_3$ (\textbf{D}) mass fractions for ten test scenarios as the results of the ODE integration (solid black lines) and as predicted by flexDeepONet (red dotted lines).}
    \label{fig:0DReact_flex_y}
\end{figure}

\begin{figure}[!htb]
    \begin{subfigure}{0.49\textwidth}
        \centering
        \caption{}
        \label{fig:0DReact_T_Shifted}
        \includegraphics[width=3.2in]{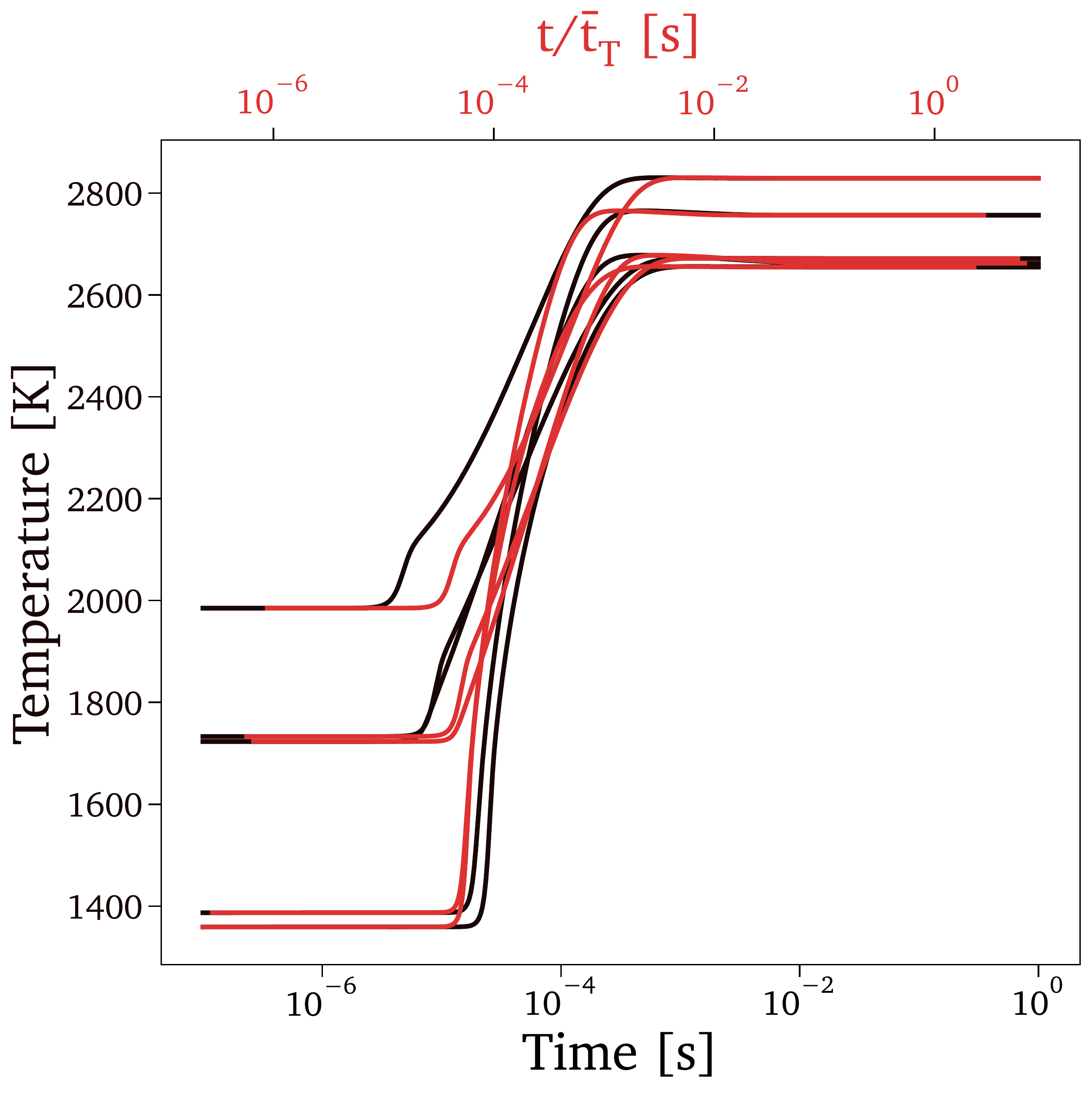}
    \end{subfigure}
    \begin{subfigure}{0.49\textwidth}
        \centering
        \caption{}
        \label{fig:0DReact_H_Shifted}
        \includegraphics[width=3.2in]{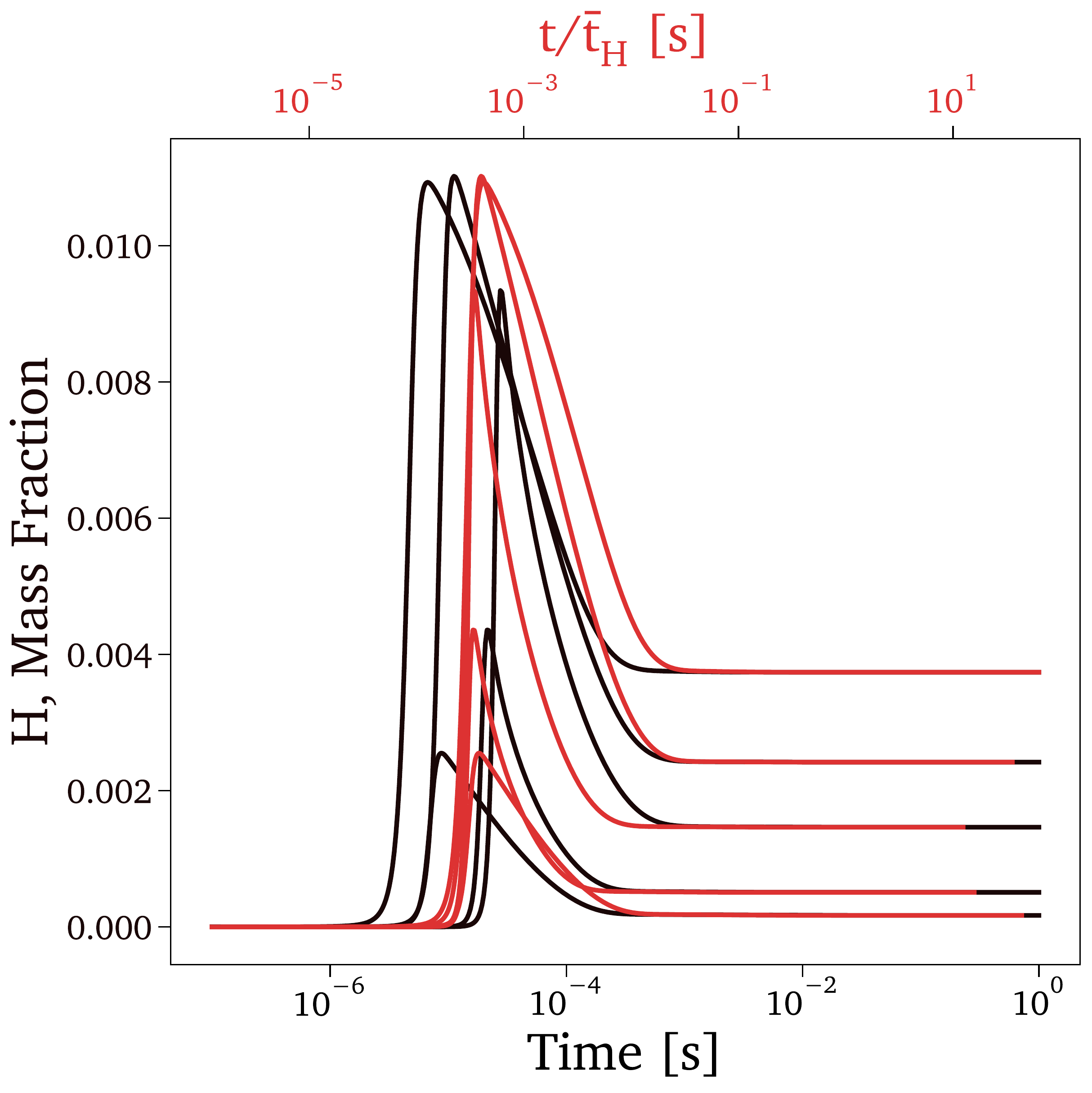}
    \end{subfigure}
    \begin{subfigure}{0.49\textwidth}
        \centering
        \caption{}
        \label{fig:0DReact_OH_Shifted}
        \includegraphics[width=3.2in]{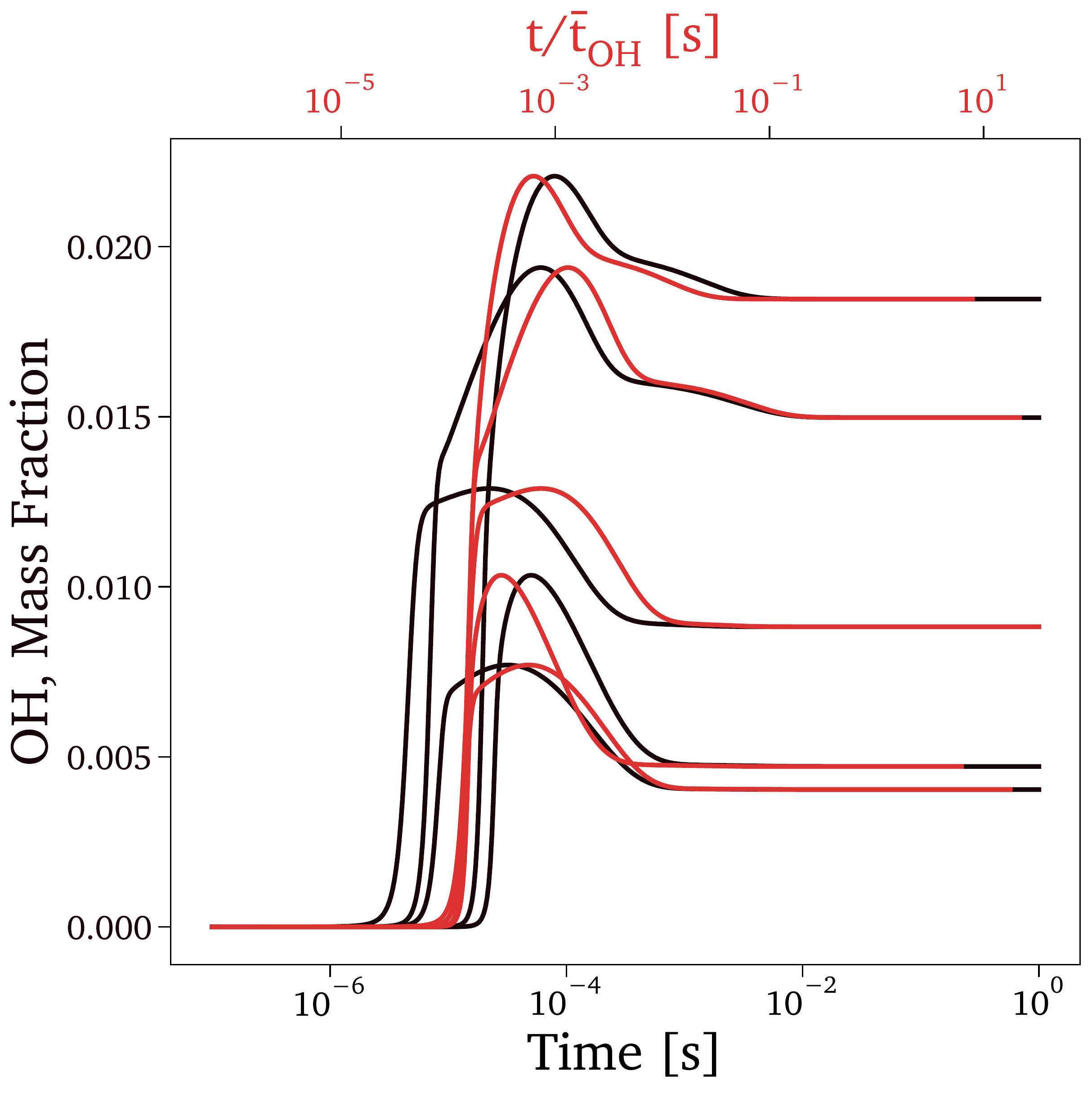}
    \end{subfigure}
    \begin{subfigure}{0.49\textwidth}
        \centering
        \caption{}
        \label{fig:0DReact_NH3_Shifted}
        \includegraphics[width=3.2in]{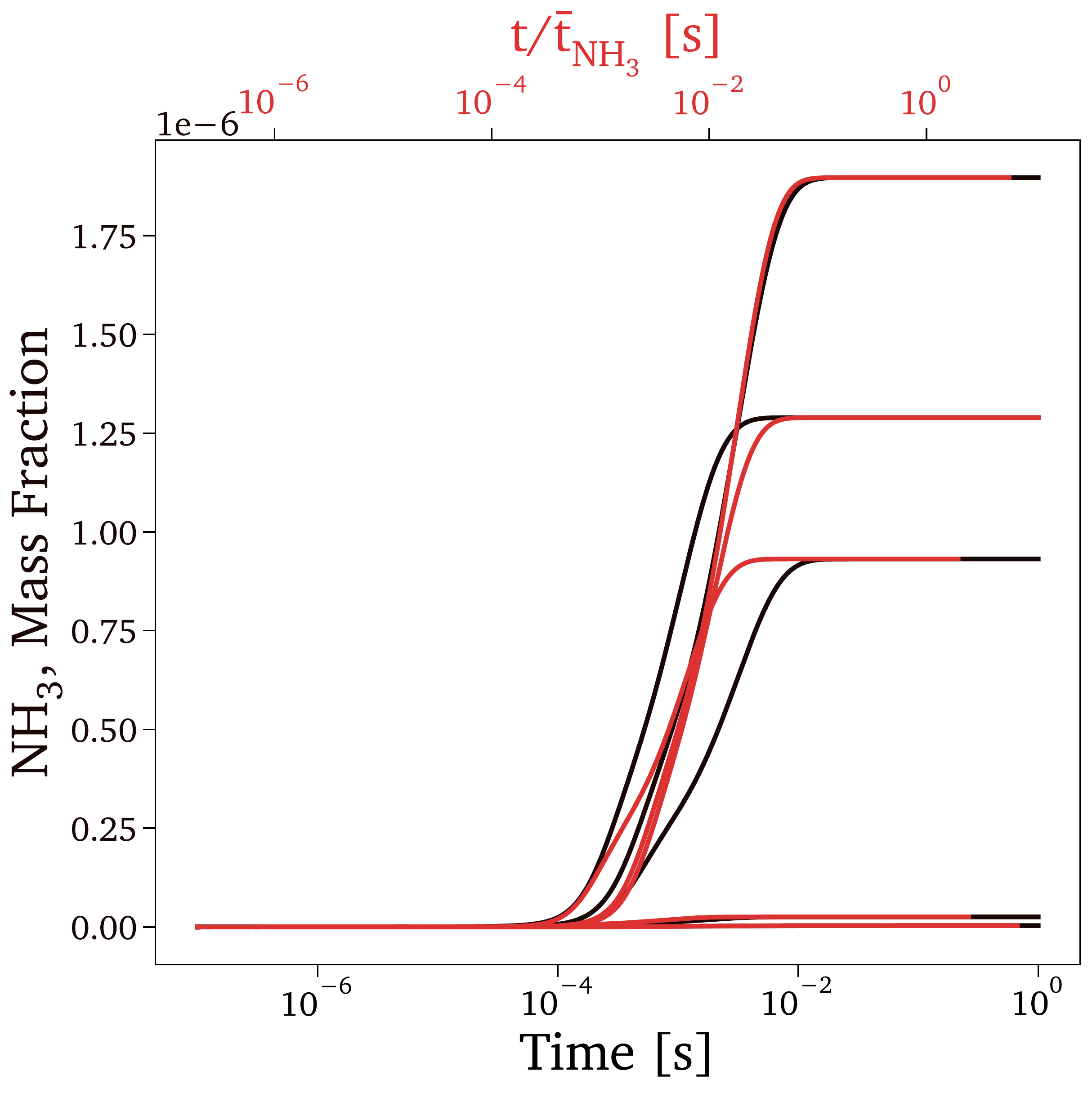}
    \end{subfigure}
    \caption{\textbf{Contributions of the shift net to the scenario alignment}. Time-integrated temperature (\textbf{A}) and  H (\textbf{B}), OH (\textbf{C}), and NH$_3$ (\textbf{D}) mass fractions for some of the test scenarios as functions of the simulation time (black lines) and of the simulation time corrected by the stretch net (red lines).}
    \label{fig:0DReact_Shifted}
\end{figure}

\clearpage
\subsection{Comparisons of Test Errors}

\begin{table}[!htp] 
    \begin{tabular}{||l | c | c | c || r} 
        \hline
        & Vanilla DeepONet & POD-DeepONet & flexDeepONet & $k$ for 99 \\[1.2ex] 
        \hline
        $p$ & 32 & 32 & 8 & \\[1.2ex]
        \hline
        No. of Parameters & 738,131 & 738,131 & 34,038 \\[1.8ex]
        \hline\hline
        T                             & 3.8105     & 2.8923     & 1.3629+00 & 20\\[1.2ex]
        \hline                                                                    
        H${}_2$, Mass Fraction        & 3.7857e-04 & 3.5192e-04 & 7.4989-05 & 40\\[1.2ex]
        \hline                                                                  
        H, Mass Fraction              & 8.1168e-05 & 6.4134e-05 & 2.1747-05 & 59\\[1.2ex]
        \hline                                                                  
        O, Mass Fraction              & 5.7953e-05 & 6.5780e-05 & 2.7380-05 & 59\\[1.2ex]
        \hline                                                                  
        O${}_2$, Mass Fraction         & 1.2515e-03 & 1.2202e-03 &3.6907-04  & 50\\[1.2ex]
        \hline                                                                  
        OH, Mass Fraction             & 8.4117e-05 & 6.4716e-05 & 2.8910-05 & 43\\[1.2ex]
        \hline                                                                  
        H${}_2$O, Mass Fraction       & 1.4060e-03 & 1.1086e-03 & 3.3034-04 & 45\\[1.2ex]
        \hline                                                                  
        HO${}_2$, Mass Fraction       & 1.0812e-06 & 1.0737e-06 & 3.0370-07 & 68\\[1.2ex]
        \hline                                                                  
        H${}_2$O${}_2$, Mass Fraction & 5.1295e-08 & 4.1965e-08 & 2.0288-08 & 73\\[1.2ex]
        \hline                                                                  
        N, Mass Fraction              & 4.1960e-09 & 4.0213e-09 & 2.5091-09 & 21\\[1.2ex]
        \hline                                                                  
        NH, Mass Fraction             & 7.4437e-10 & 5.5512e-10 & 7.1590-10 & 12\\[1.2ex]
        \hline                                                                  
        NH${}_2$, Mass Fraction       & 1.1905e-10 & 1.5052e-10 & 3.0020-10 & 6 \\[1.2ex]
        \hline                                                                  
        NH${}_3$, Mass Fraction       & 5.6947e-10 & 7.0749e-10 & 1.4949-09 & 7 \\[1.2ex]
        \hline                                                                  
        NNH, Mass Fraction            & 1.7792e-09 & 1.0353e-09 & 4.6500-10 & 55\\[1.2ex]
        \hline                                                                  
        NO, Mass Fraction             & 2.7464e-06 & 4.6540e-06 & 8.5739-06 & 5 \\[1.2ex]
        \hline                                                                  
        NO${}_2$, Mass Fraction       & 1.1931e-09 & 1.7855e-09 & 3.3697-09 & 5 \\[1.2ex]
        \hline                                                                  
        N${}_2$O, Mass Fraction       & 1.2605e-09 & 1.0142e-09 & 1.5176-09 & 25\\[1.2ex]
        \hline                                                                  
        HNO, Mass Fraction            & 2.4558e-10 & 2.6315e-10 & 5.7490-10 & 5 \\[1.2ex]
        \hline                                                                  
        N${}_2$, Mass Fraction        & 8.1701e-06 & 2.8235e-05 & 2.4410-05  & 1 \\[1.2ex]
        \hline                                                                             
        \hline
        Training Time & & $(10,117)_{T} \times 19$ + 458,222 & \\[1.2ex] 
        & 932,341 [s] & = 650,445 [s] & 55,293 [s] \\[1.2ex] 
        \hline
        Prediction Time & 1,080 [ms] & 1,080 [ms] & 248 [ms/scenario] & \\[1.2ex] 
        \hline
    \end{tabular}
    \vspace{5mm}
    \caption{\textbf{Comparison between architectures}. Comparisons between the number of trunks' outputs ($p$), the number of trainable parameters, the root mean squared errors (RMSEs) at the ten test scenarios, the training times, and the prediction times. Training has been performed via Tensorflow 2.8. For the training times, ${(...)}_T$ indicates the time necessary for training each trunk block. It should be noted, however, that the trunks of POD-DeepONet can be independently trained in parallel. Prediction times refer to the overall time necessary for the architectures to predict all the thermodynamic variables at 10,000 times and initial conditions, and they have been computed by relying on architectures' implementations based on the NumPy 1.22.3 library. The last column on the right contains the number of singular values to preserve 99\% of the training data's cumulative energy.}\label{table:0DReact_TestErrors}
\end{table}



\clearpage
\section{Supplementary Material for Test Case 4}\label{Sec:Supp_Test4}

\subsection{Equations of Motion and Data}

The fourth test case is created based on the following equations in space coordinates ($x$ and $y$) and time ($t$):
\begin{equation}
    z(x,y;t) = \exp \Big( z_x(x,y;t) + z_y(x,y;t) \Big) z_s(t),
\end{equation}
where:
\begin{equation}
    z_s(t) = z_0 + v_z t.
\end{equation}
$z_x(x,y;t)$ are $z_y(x,y;t)$ are defined as:
\begin{equation}
    \begin{cases}
        z_x(x,y;t) = \frac{1}{2}\tanh{ \Big( a_x (x_{I}+l_{x}) \Big) } + \frac{1}{2}\tanh{ \Big( b_x(x_{I}-l_{x}) \Big) } \\[3pt]
        z_y(x,y;t) = \frac{1}{2}\tanh{ \Big( a_y (y_{I}+l_{y}) \Big) } + \frac{1}{2}\tanh{ \Big( b_y(y_{I}-l_{y}) \Big) }.
    \end{cases}
\end{equation}
The $x_{I}(x,y;t)$ and $y_{I}(x,y;t)$ (i.e., the coordinates in the rotated-shifted-translated reference frame) are obtained from $x$ and $y$ (i.e., the original reference frame's coordinates) via:
\begin{equation}\label{eq3}
    \renewcommand{\arraystretch}{2.0}
    \begin{bmatrix}
        x_{I}(x,y;t)\\
        y_{I}(x,y;t)
    \end{bmatrix}
    =
    s(t)
    \left[
        \begin{array}{cc}
        \cos{\theta(t)} & -\sin{\theta(t)} \\
        \sin{\theta(t)} & \cos{\theta(t)}
        \end{array}
    \right]
    \begin{bmatrix}
        x \\
        v
    \end{bmatrix}
    +
    \begin{bmatrix}
        x_c(t)\\
        y_c(t)
    \end{bmatrix},
\end{equation}
where the rotation angle, $\theta(t)$, is defined as:
\begin{equation}
    \theta(t) = \theta_0 + v_{\theta} \cos{\big( \omega_{\theta} t +\phi_{\theta} \big)}.
\end{equation}
Additionally, the stretching factor, $s(t)$, is given by:
\begin{equation}
    s(t) = s_0 + v_{s} \sin{\big( \omega_{s} t +\phi_{s} \big)}.
\end{equation}
Finally, the shifting coefficients, $x_c(t)$ and $y_c(t)$, are constructed as:
\begin{equation}
    \begin{cases}
        x_c(t) = x_{c_0} + v_{\xi} \xi(t) * \cos{(\xi)} \\[3pt]
        y_c(t) = y_{c_0} + v_{\xi} \xi(t) * \sin{(\xi)},
    \end{cases}
\end{equation}
where $\xi(t) = \omega_{\xi} t$. The values of the equation parameters are reported in Table~\ref{table:Rect_EqParameters}.


\begin{table}[!htp] 
    \centering
    \begin{tabular}{||c | c ||} 
        \hline
        Name & Value \\[1.2ex] 
        \hline\hline
        $z_0$ & 1 \\[1.2ex] 
        \hline
        $v_z$ & 0.2 \\[1.2ex] 
        \hline
        $l_{x}$ & 8 \\[1.2ex] 
        \hline
        $l_{y}$ & 6 \\[1.2ex] 
        \hline
        $a_x$ & 10 \\[1.2ex] 
        \hline
        $b_x$ & 10 \\[1.2ex] 
        \hline
        $a_y$ & 10 \\[1.2ex] 
        \hline
        $b_y$ & 10 \\[1.2ex] 
        \hline
        $\theta_0$ & $0$ \\[1.2ex] 
        \hline
        $v_{\theta}$ & $\pi$ \\[1.2ex] 
        \hline
        $\omega_{\theta}$ & $\frac{2\pi}{t_{End}}\pi$ \\[1.2ex] 
        \hline
        $t_{End}$ & $10$ \\[1.2ex] 
        \hline
        $\phi_{\theta}$ & $2$ \\[1.2ex] 
        \hline
        $s_0$ & $2$ \\[1.2ex] 
        \hline
        $v_{s}$ & $1$ \\[1.2ex] 
        \hline
        $\omega_{s}$ & $\frac{2\pi}{t_{End}}\pi$ \\[1.2ex] 
        \hline
        $\phi_{s}$ & $0$ \\[1.2ex] 
        \hline
        $x_{c_0}$ & $1$ \\[1.2ex] 
        \hline
        $y_{c_0}$ & $-0.5$ \\[1.2ex] 
        \hline
        $v_{\xi}$ & $0.5$ \\[1.2ex] 
        \hline
        $\omega_{\xi}$ & $\frac{5\pi}{18}$ \\[1.2ex] 
        \hline
        \hline
    \end{tabular}
    \vspace{0.5cm}
    \caption{\textbf{Equation parameters used for generating the fourth test case}.}
    \label{table:Rect_EqParameters}
\end{table}

\begin{figure}[!htb]
    \begin{subfigure}{0.24\textwidth}
        \centering
        \caption{}
        \label{fig:Rect_Orig_Supp_1}
        \includegraphics[width=1.5in]{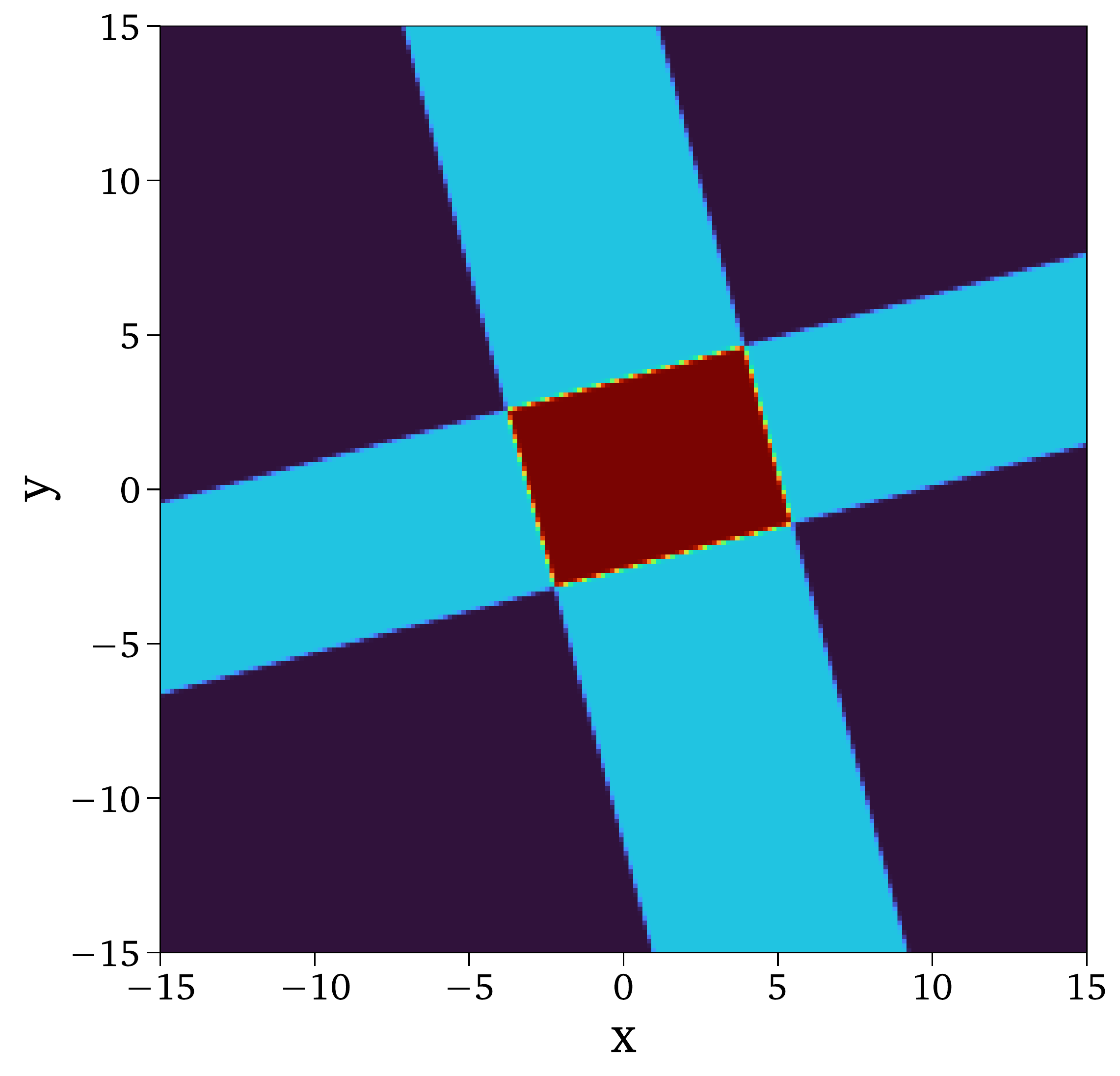}
    \end{subfigure}
    \begin{subfigure}{0.24\textwidth}
        \centering
        \caption{}
        \label{fig:Rect_Orig_Supp_2}
        \includegraphics[width=1.5in]{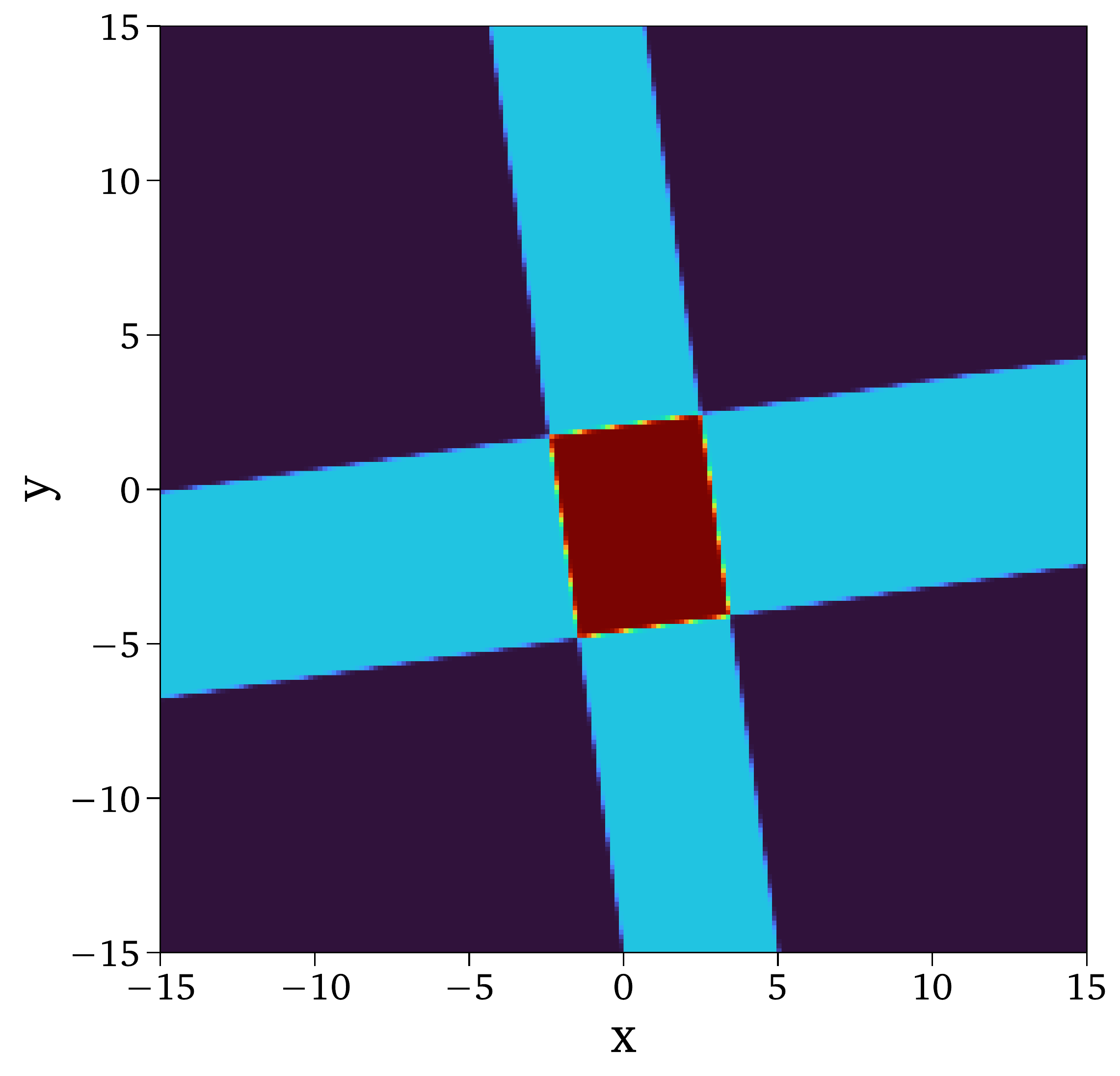}
    \end{subfigure}
    \begin{subfigure}{0.24\textwidth}
        \centering
        \caption{}
        \label{fig:Rect_Orig_Supp_3}
        \includegraphics[width=1.5in]{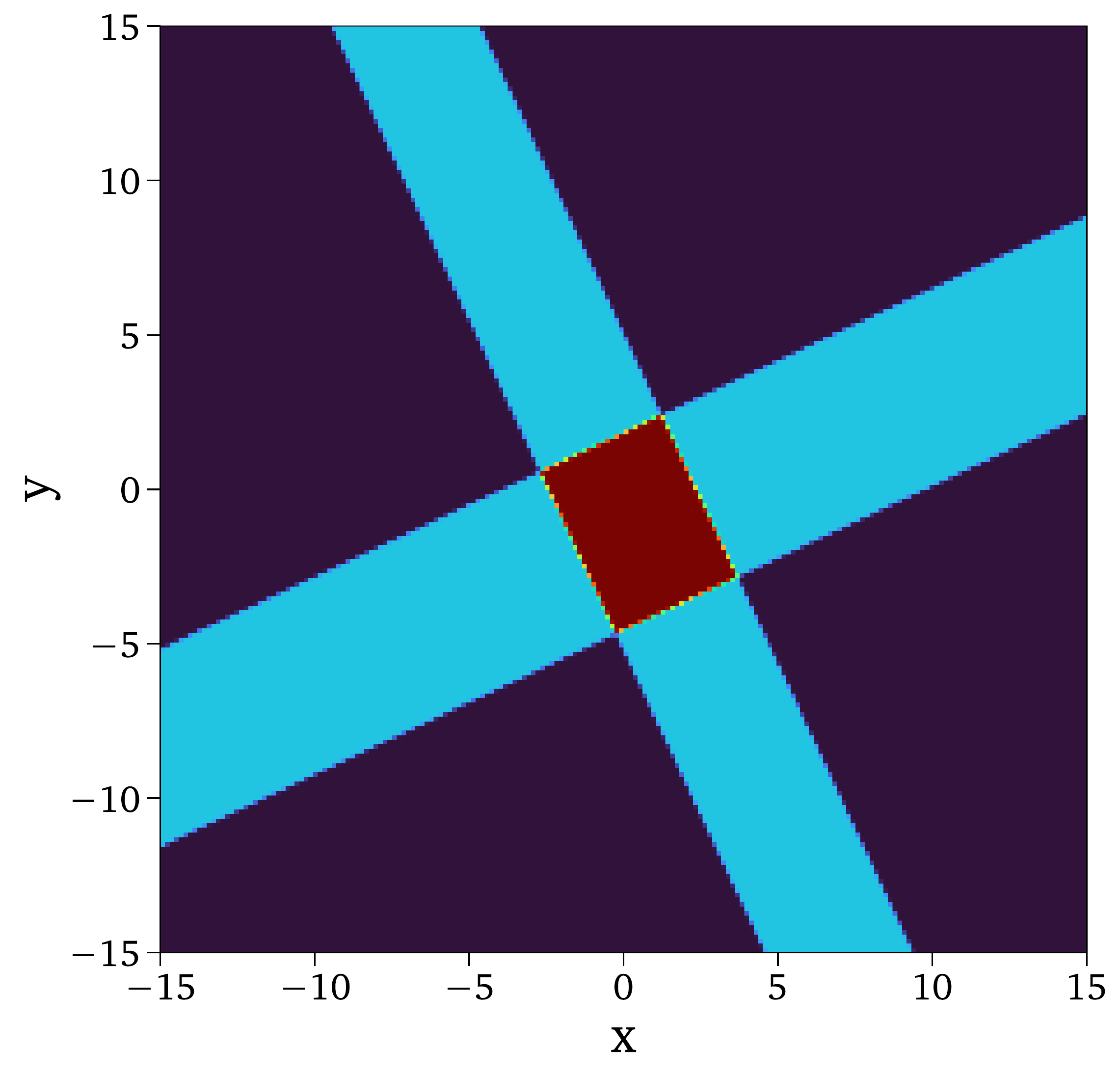}
    \end{subfigure}
    \begin{subfigure}{0.24\textwidth}
        \centering
        \caption{}
        \label{fig:Rect_Orig_Supp_4}
        \includegraphics[width=1.5in]{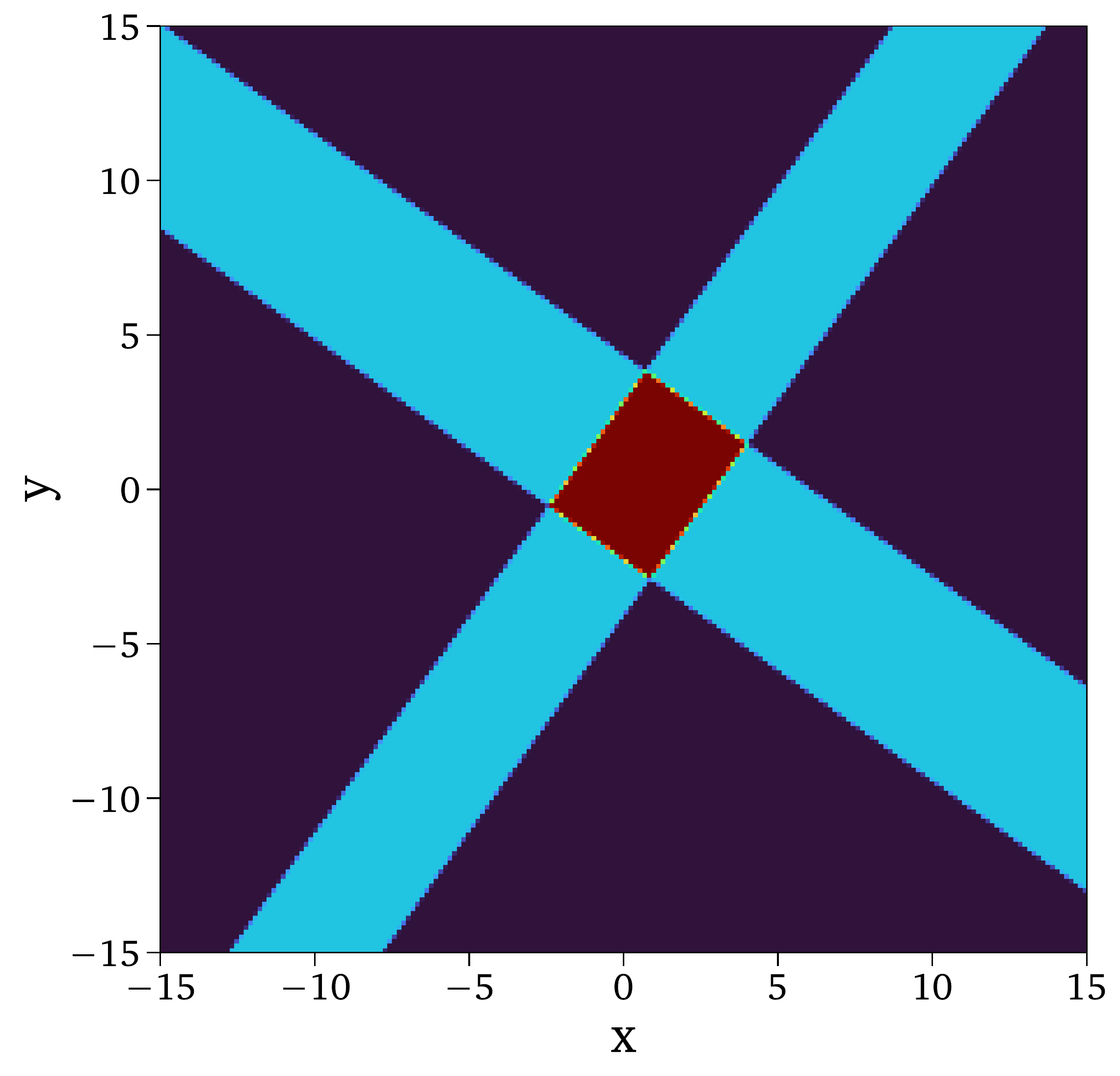}
    \end{subfigure}
    
    \begin{subfigure}{0.24\textwidth}
        \centering
        \caption{}
        \label{fig:Rect_Orig_Supp_5}
        \includegraphics[width=1.5in]{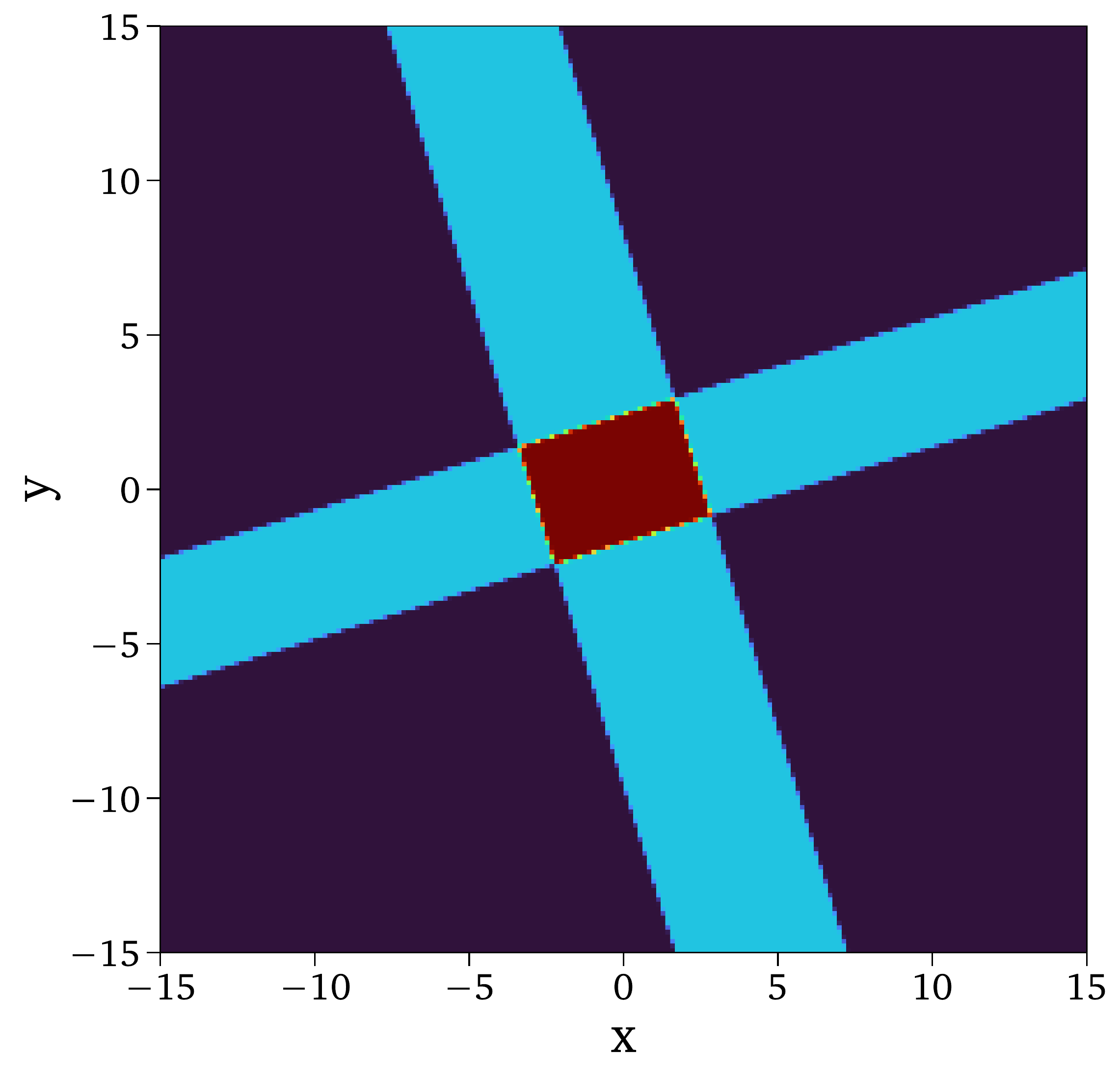}
    \end{subfigure}
    \begin{subfigure}{0.24\textwidth}
        \centering
        \caption{}
        \label{fig:Rect_Orig_Supp_6}
        \includegraphics[width=1.5in]{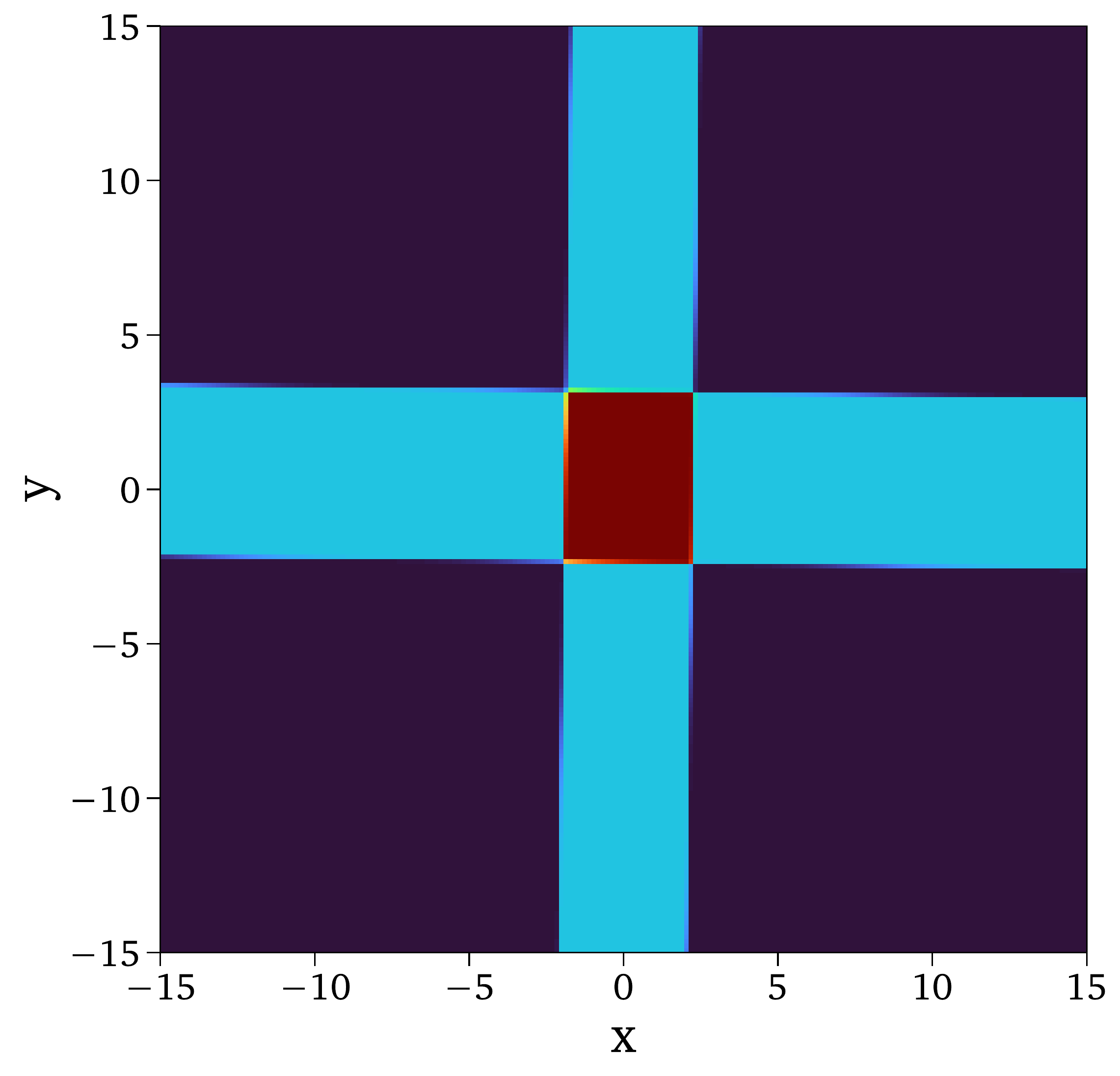}
    \end{subfigure}
    \begin{subfigure}{0.24\textwidth}
        \centering
        \caption{}
        \label{fig:Rect_Orig_Supp_7}
        \includegraphics[width=1.5in]{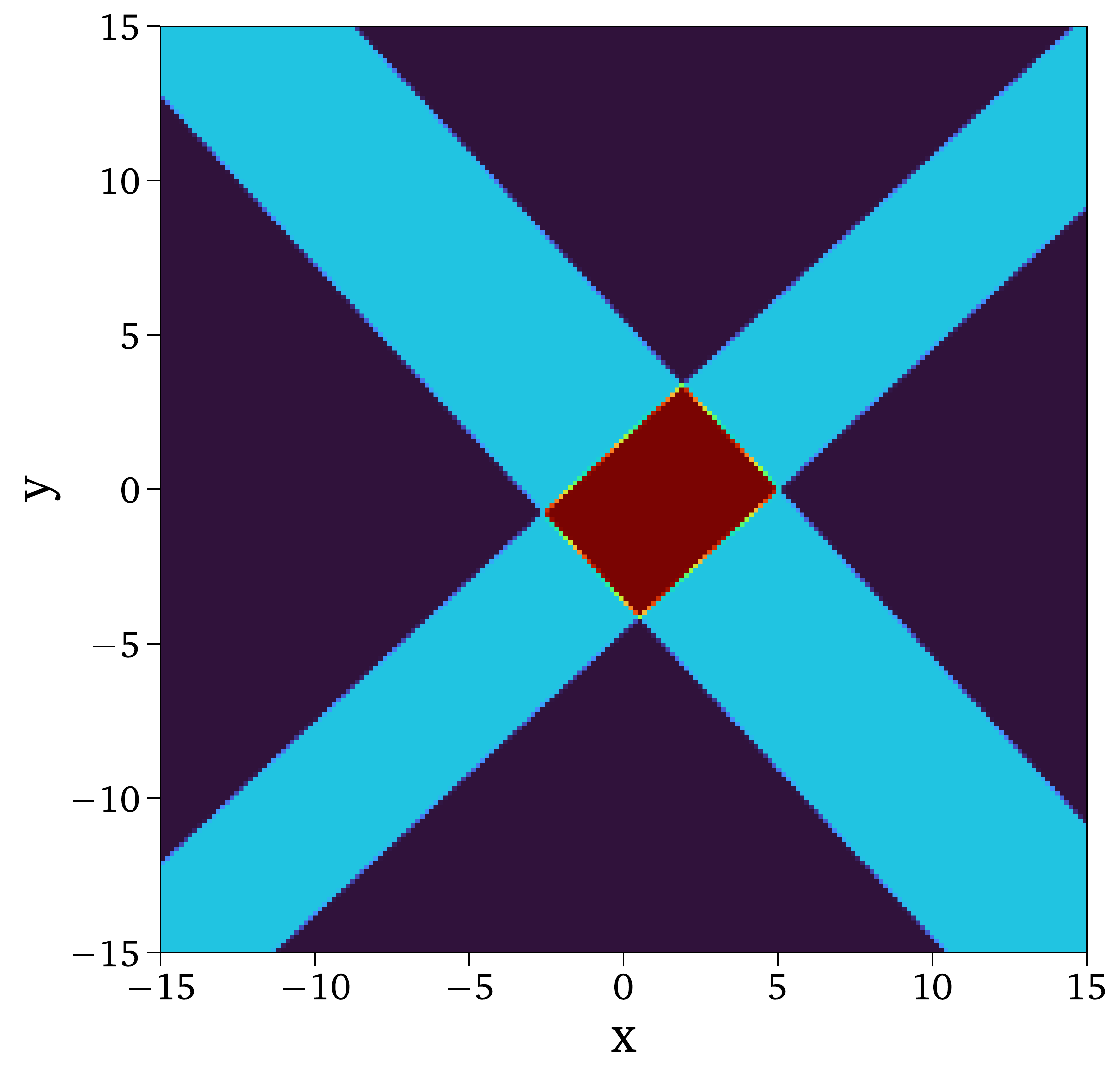}
    \end{subfigure}
    \begin{subfigure}{0.24\textwidth}
        \centering
        \caption{}
        \label{fig:Rect_Orig_Supp_8}
        \includegraphics[width=1.5in]{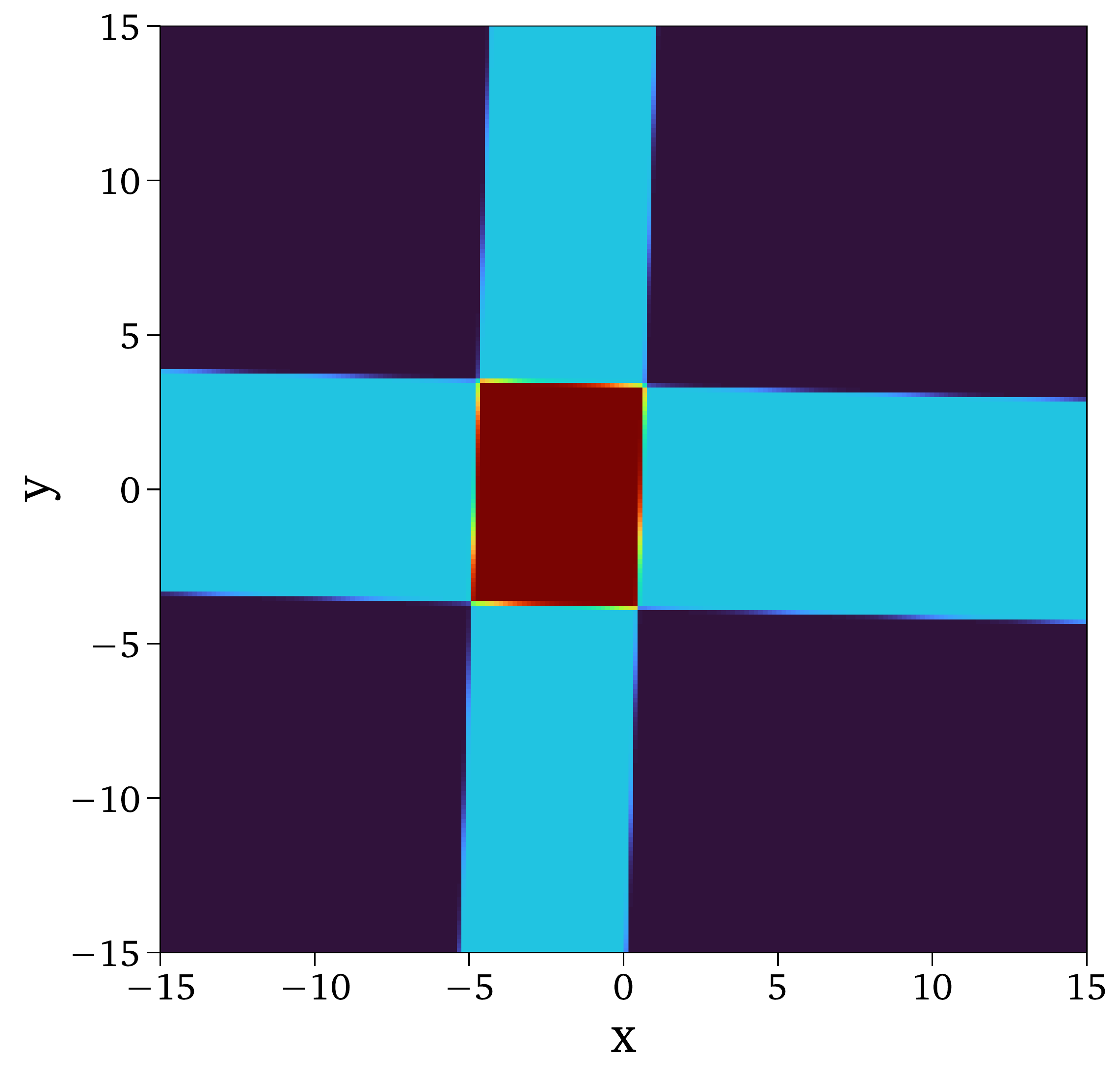}
    \end{subfigure}
    
    \begin{subfigure}{0.24\textwidth}
        \centering
        \caption{}
        \label{fig:Rect_Orig_Supp_9}
        \includegraphics[width=1.5in]{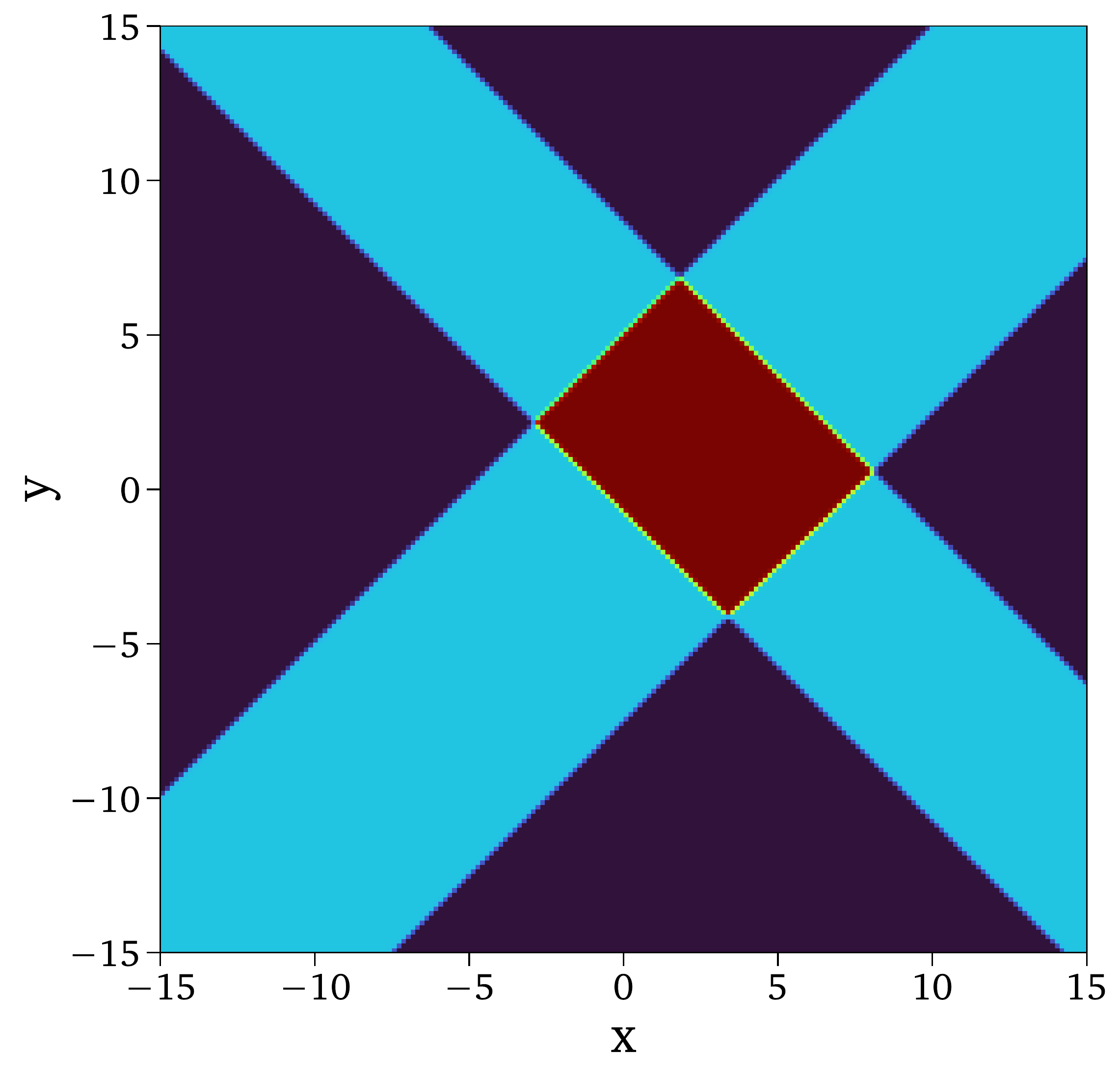}
    \end{subfigure}
    \begin{subfigure}{0.24\textwidth}
        \centering
        \caption{}
        \label{fig:Rect_Orig_Supp_10}
        \includegraphics[width=1.5in]{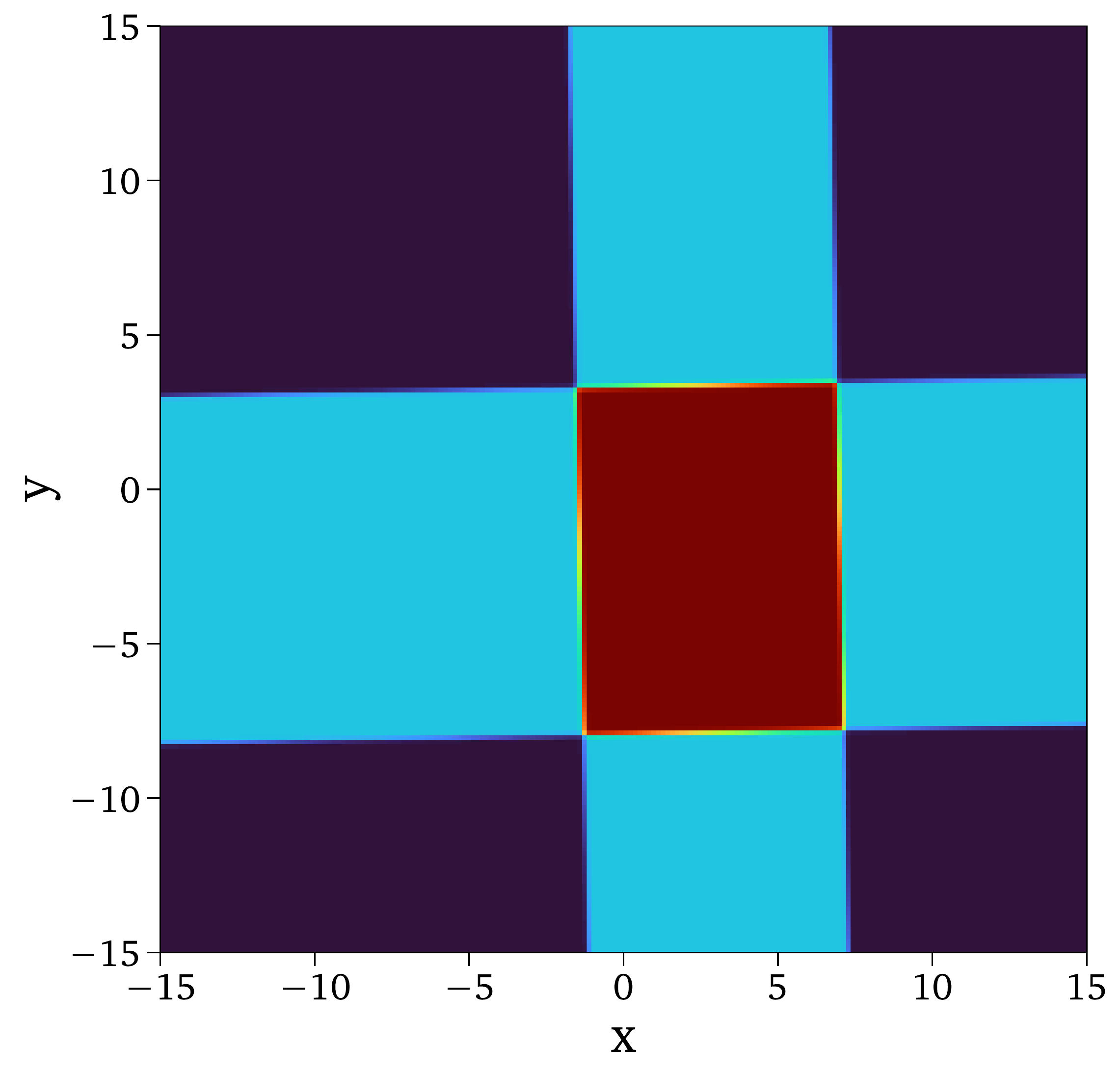}
    \end{subfigure}
    \begin{subfigure}{0.24\textwidth}
        \centering
        \caption{}
        \label{fig:Rect_Orig_Supp_11}
        \includegraphics[width=1.5in]{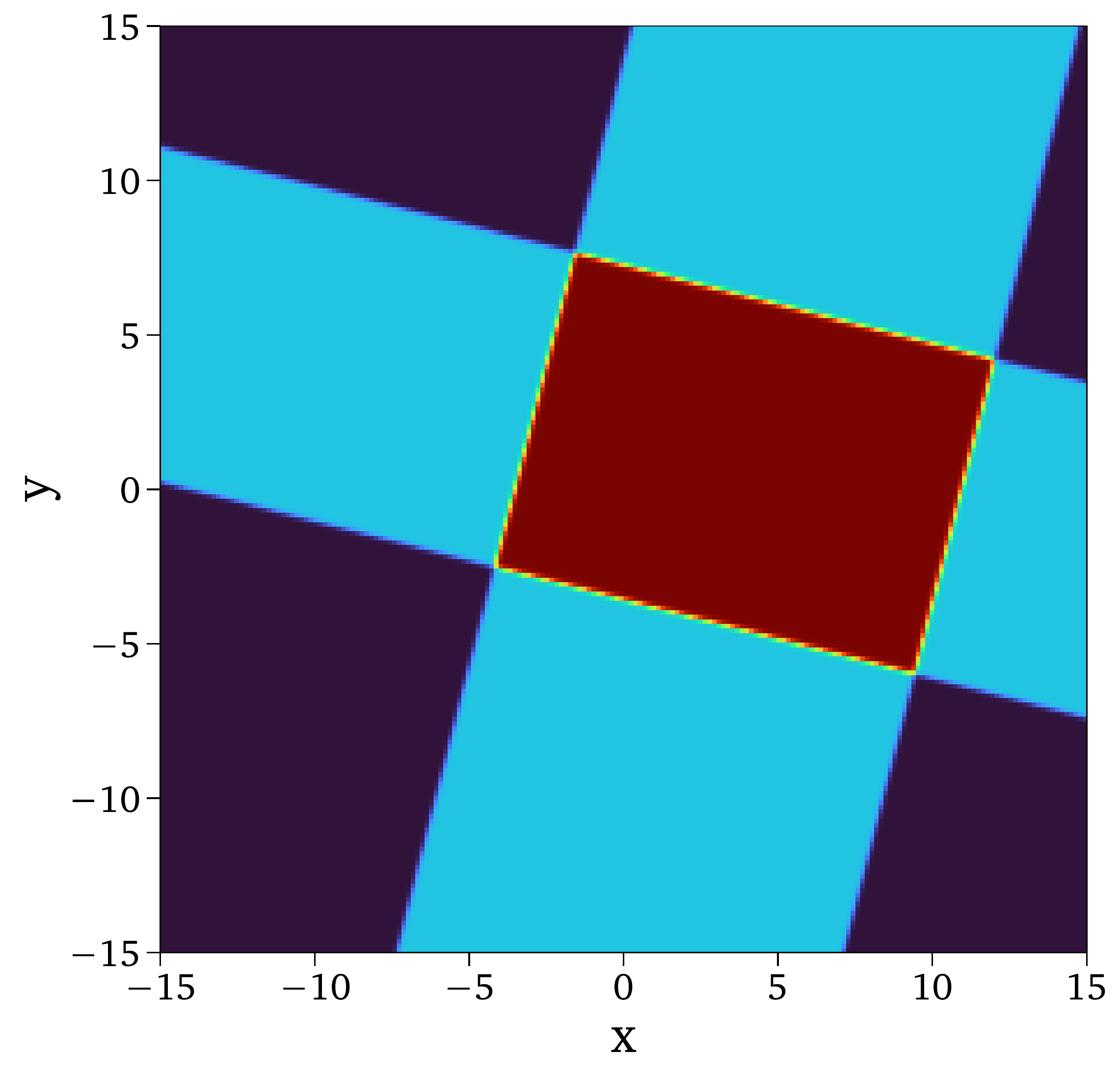}
    \end{subfigure}
    \begin{subfigure}{0.24\textwidth}
        \centering
        \caption{}
        \label{fig:Rect_Orig_Supp_12}
        \includegraphics[width=1.5in]{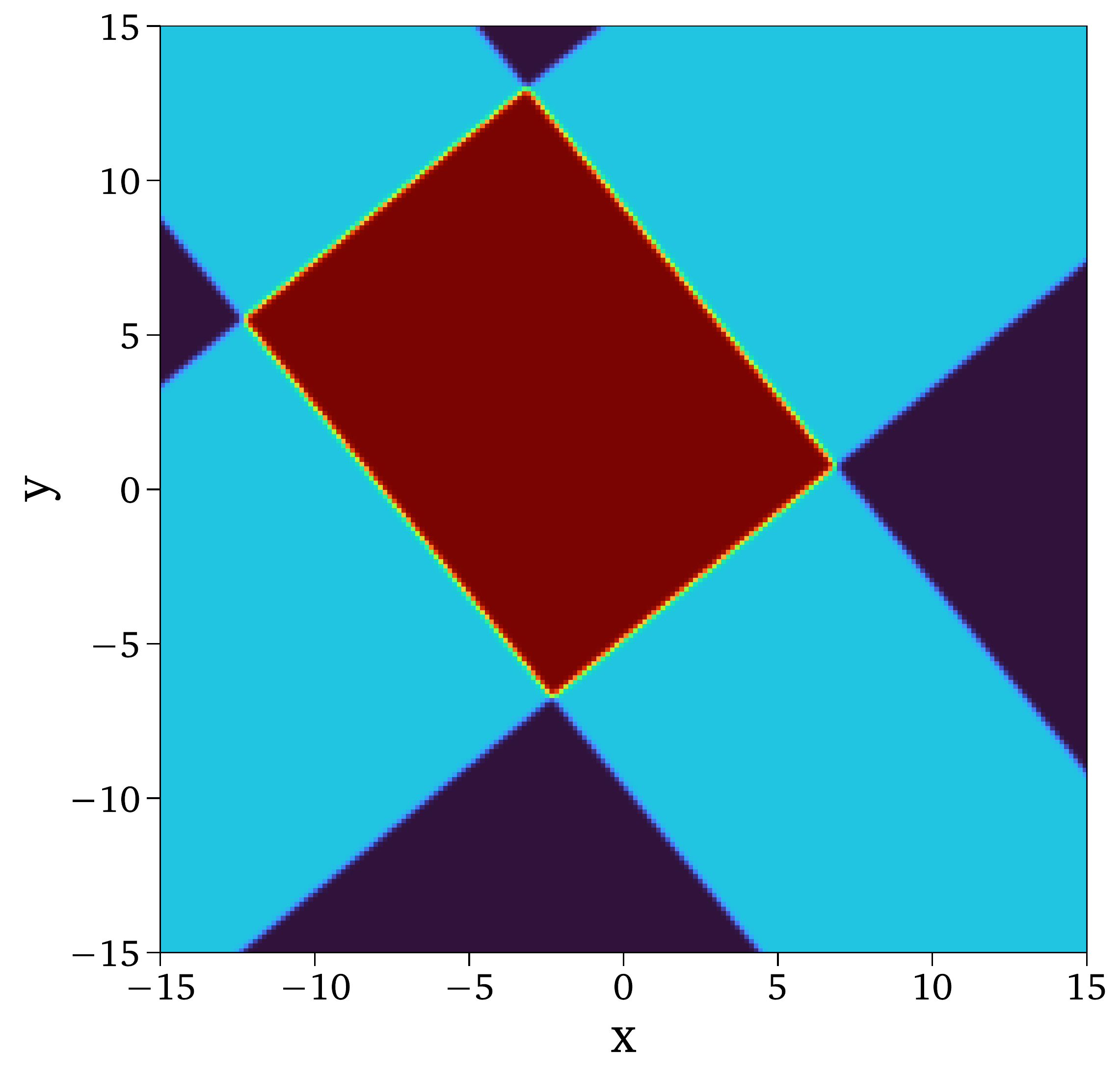}
    \end{subfigure}
    
    \begin{subfigure}{0.24\textwidth}
        \centering
        \caption{}
        \label{fig:Rect_Orig_Supp_13}
        \includegraphics[width=1.5in]{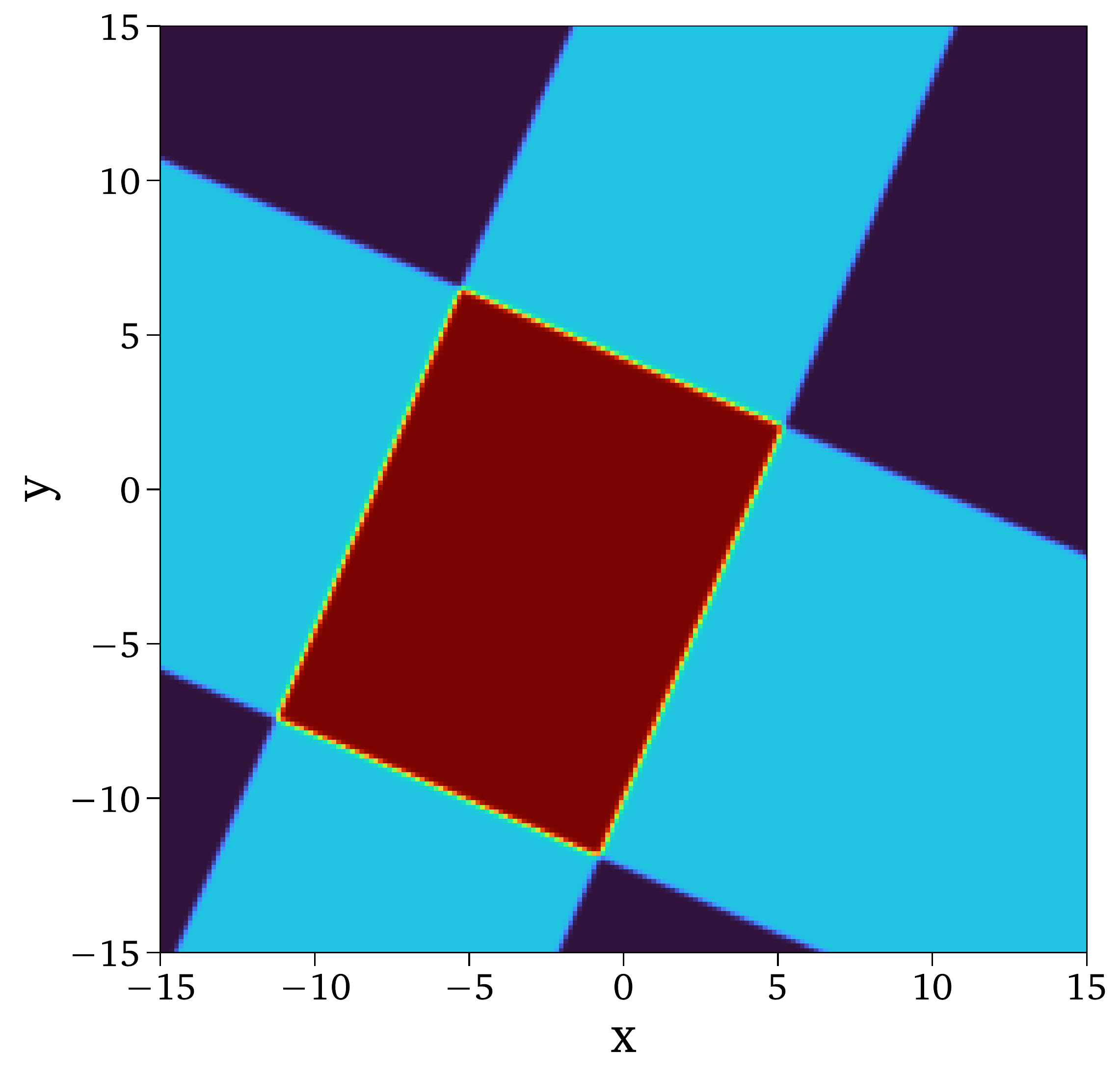}
    \end{subfigure}
    \begin{subfigure}{0.24\textwidth}
        \centering
        \caption{}
        \label{fig:Rect_Orig_Supp_14}
        \includegraphics[width=1.5in]{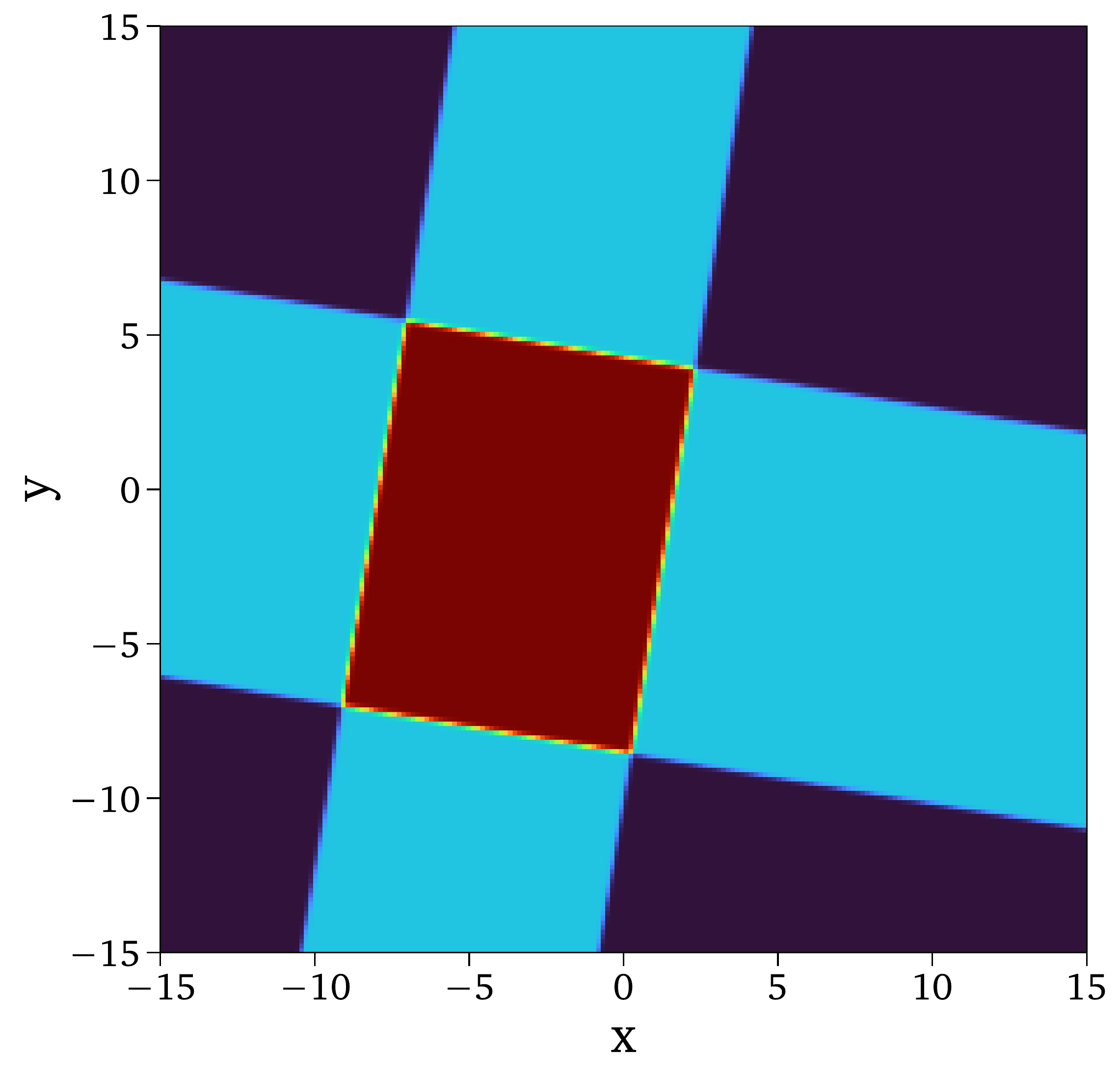}
    \end{subfigure}
    \begin{subfigure}{0.24\textwidth}
        \centering
        \caption{}
        \label{fig:Rect_Orig_Supp_15}
        \includegraphics[width=1.5in]{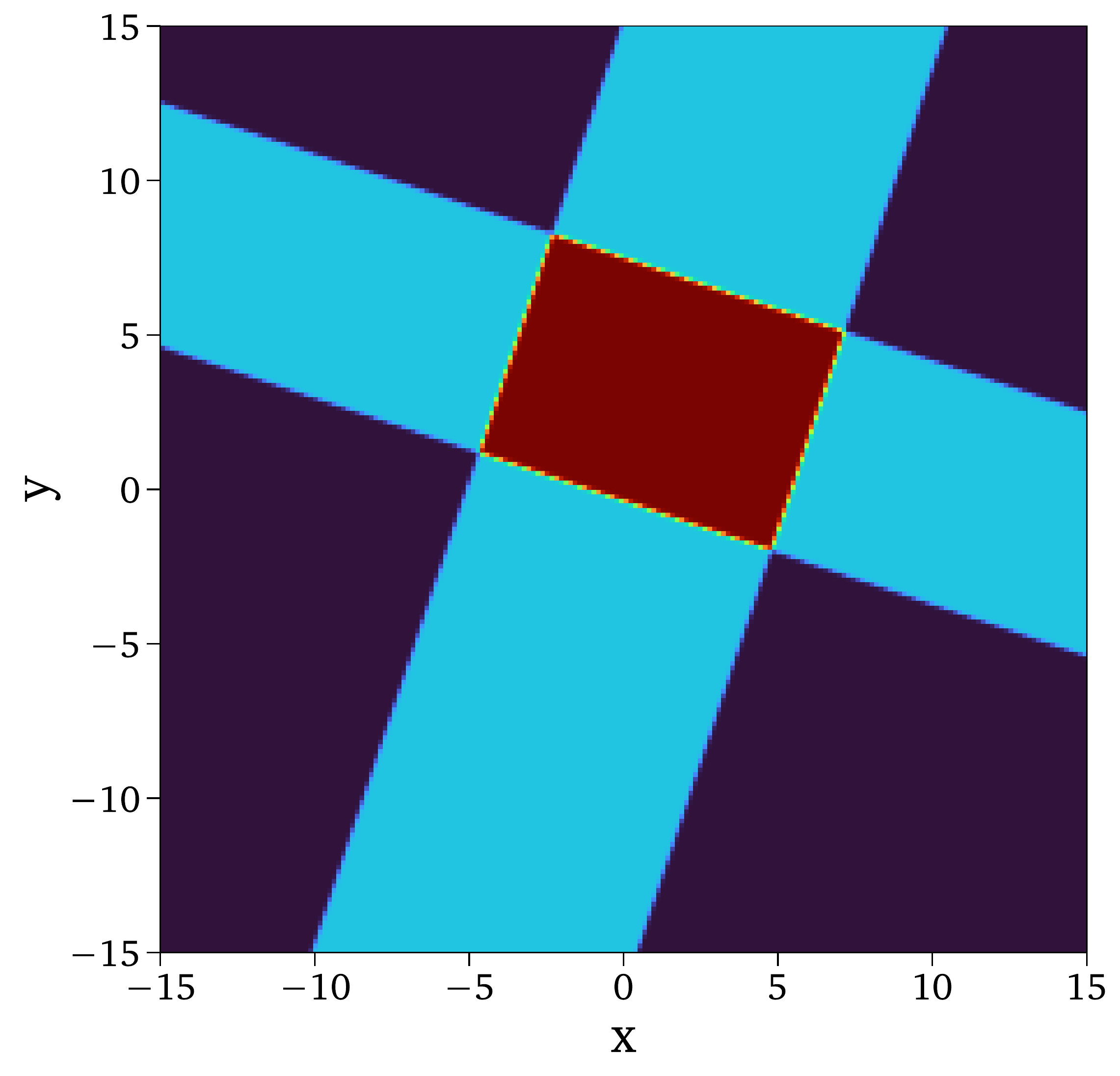}
    \end{subfigure}
    \begin{subfigure}{0.24\textwidth}
        \centering
        \caption{}
        \label{fig:Rect_Orig_Supp_16}
        \includegraphics[width=1.5in]{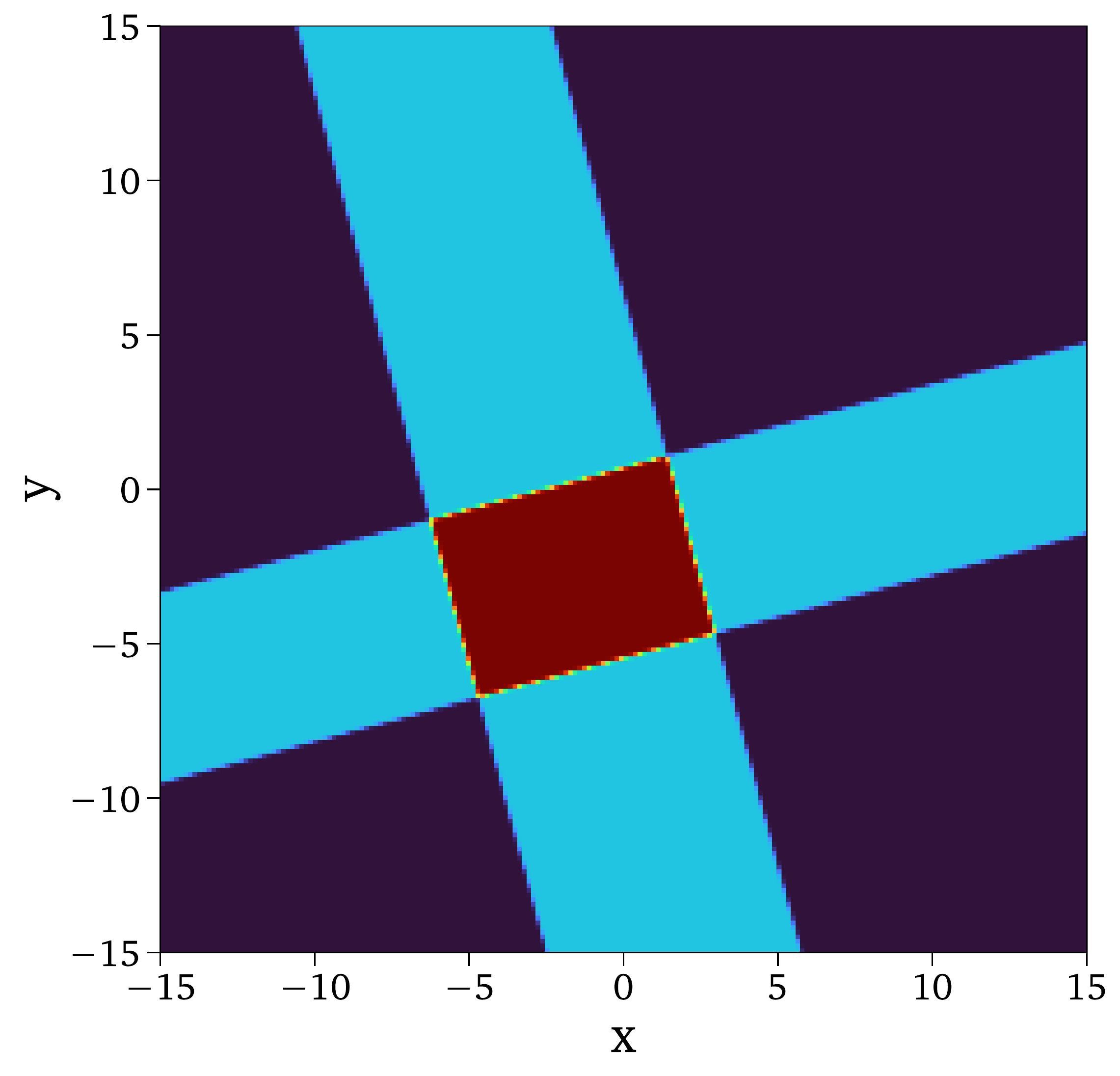}
    \end{subfigure}
    \caption{\textbf{Some of the time snapshots for the dynamics of the rotating-translating-stretching rigid body}. The snapshots are produced at sixteen time instants uniformly spaced between $t=0$ [s] and $t=10$ [s].}
    \label{fig:Rect_Orig_Supp}
\end{figure}

\clearpage
\subsection{Proper Orthogonal Decomposition (POD)}

\begin{figure}[!htb]
    \begin{subfigure}{0.24\textwidth}
        \centering
        \caption{}
        \label{fig:Rect_Orig_2_Supp}
        \includegraphics[width=1.5in]{./Figures/Rect_Orig_2.pdf}
    \end{subfigure}
    \begin{subfigure}{0.24\textwidth}
        \centering
        \caption{}
        \label{fig:Rect_128PC_2}
        \includegraphics[width=1.5in]{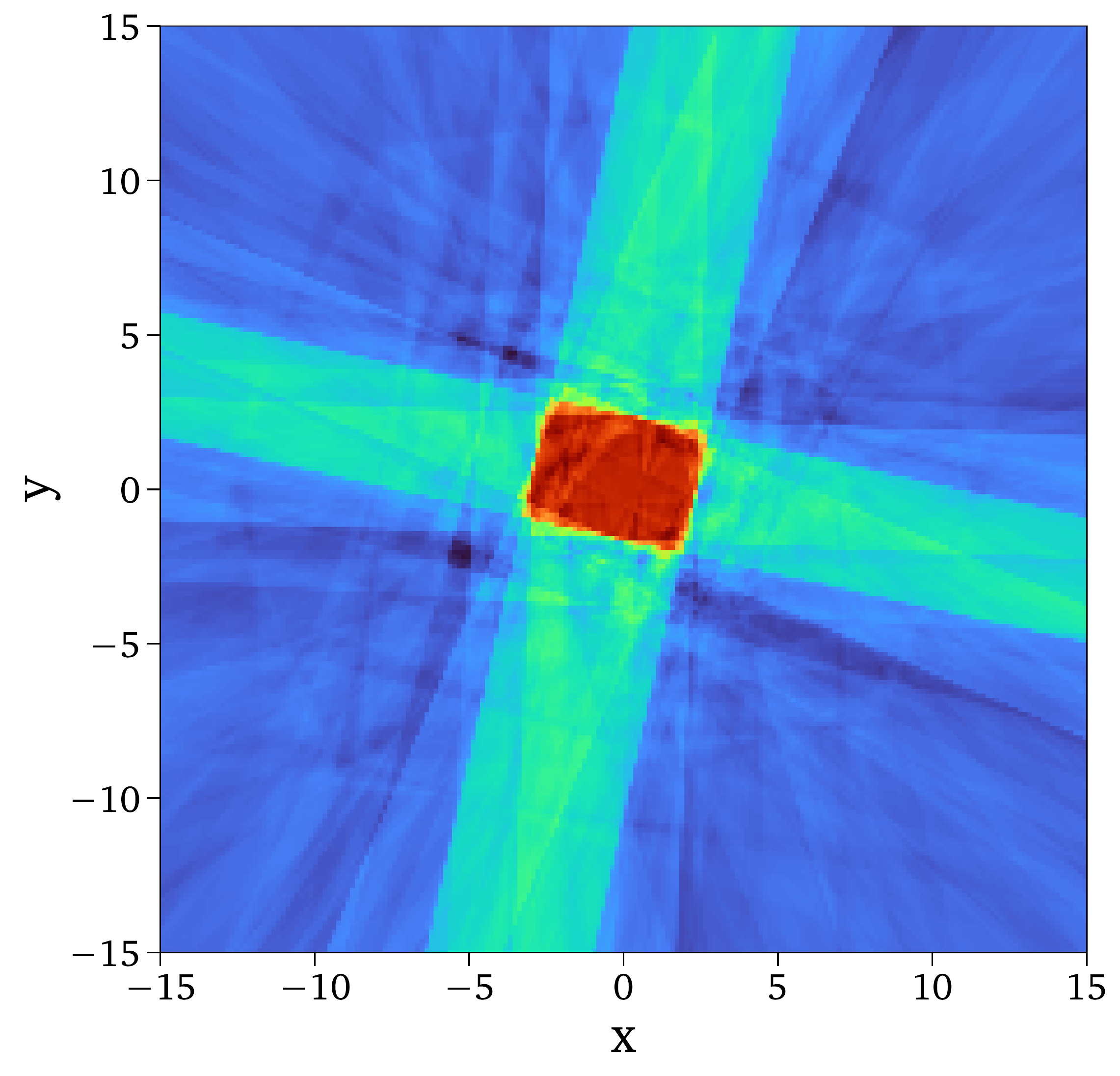}
    \end{subfigure}
    \begin{subfigure}{0.24\textwidth}
        \centering
        \caption{}
        \label{fig:Rect_64PC_2}
        \includegraphics[width=1.5in]{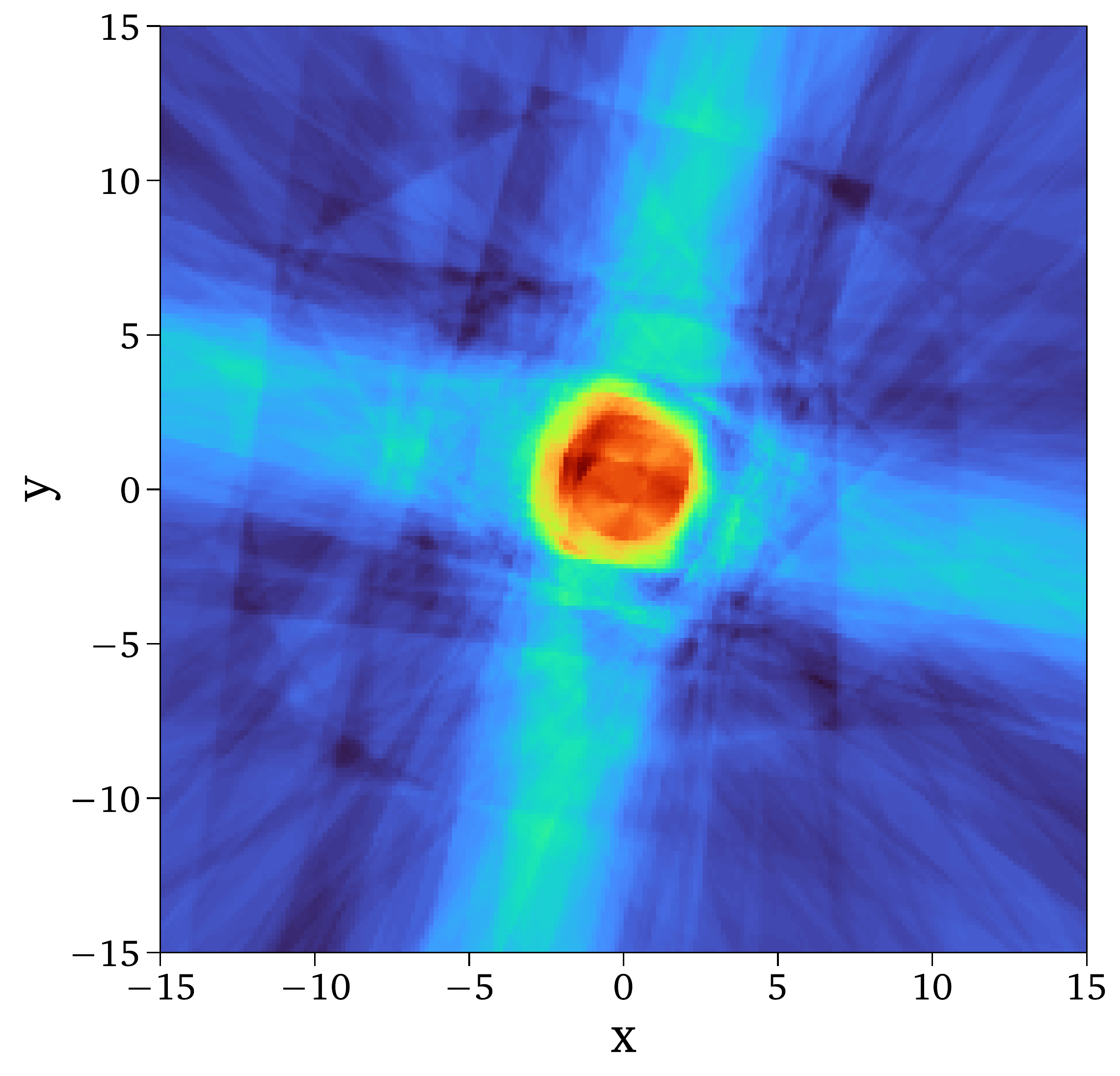}
    \end{subfigure}
    \begin{subfigure}{0.24\textwidth}
        \centering
        \caption{}
        \label{fig:Rect_32PC_2}
        \includegraphics[width=1.5in]{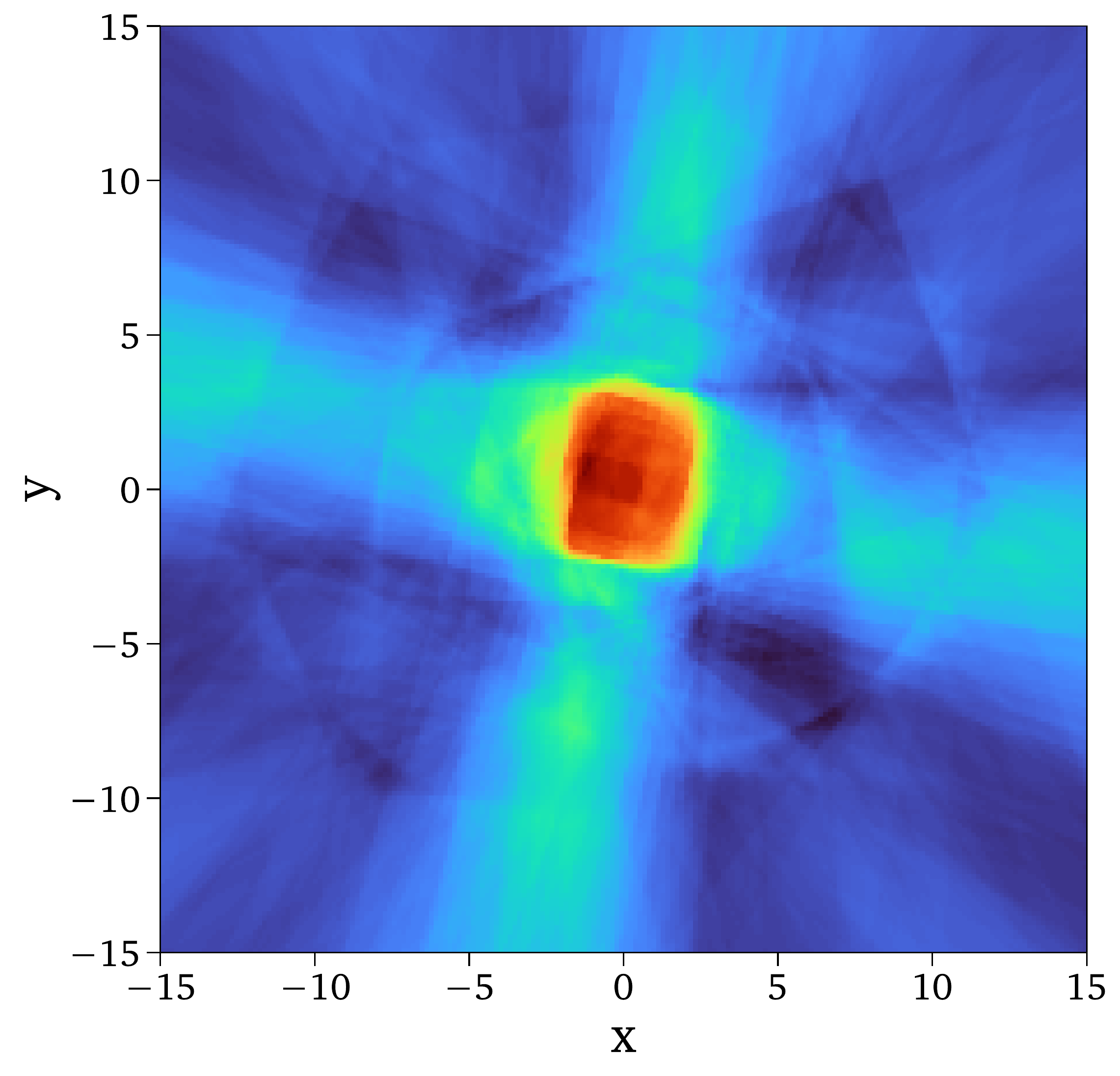}
    \end{subfigure}
    
    \begin{subfigure}{0.24\textwidth}
        \centering
        \caption{}
        \label{fig:Rect_Orig_3_Supp}
        \includegraphics[width=1.5in]{./Figures/Rect_Orig_3.pdf}
    \end{subfigure}
    \begin{subfigure}{0.24\textwidth}
        \centering
        \caption{}
        \label{fig:Rect_128PC_3}
        \includegraphics[width=1.5in]{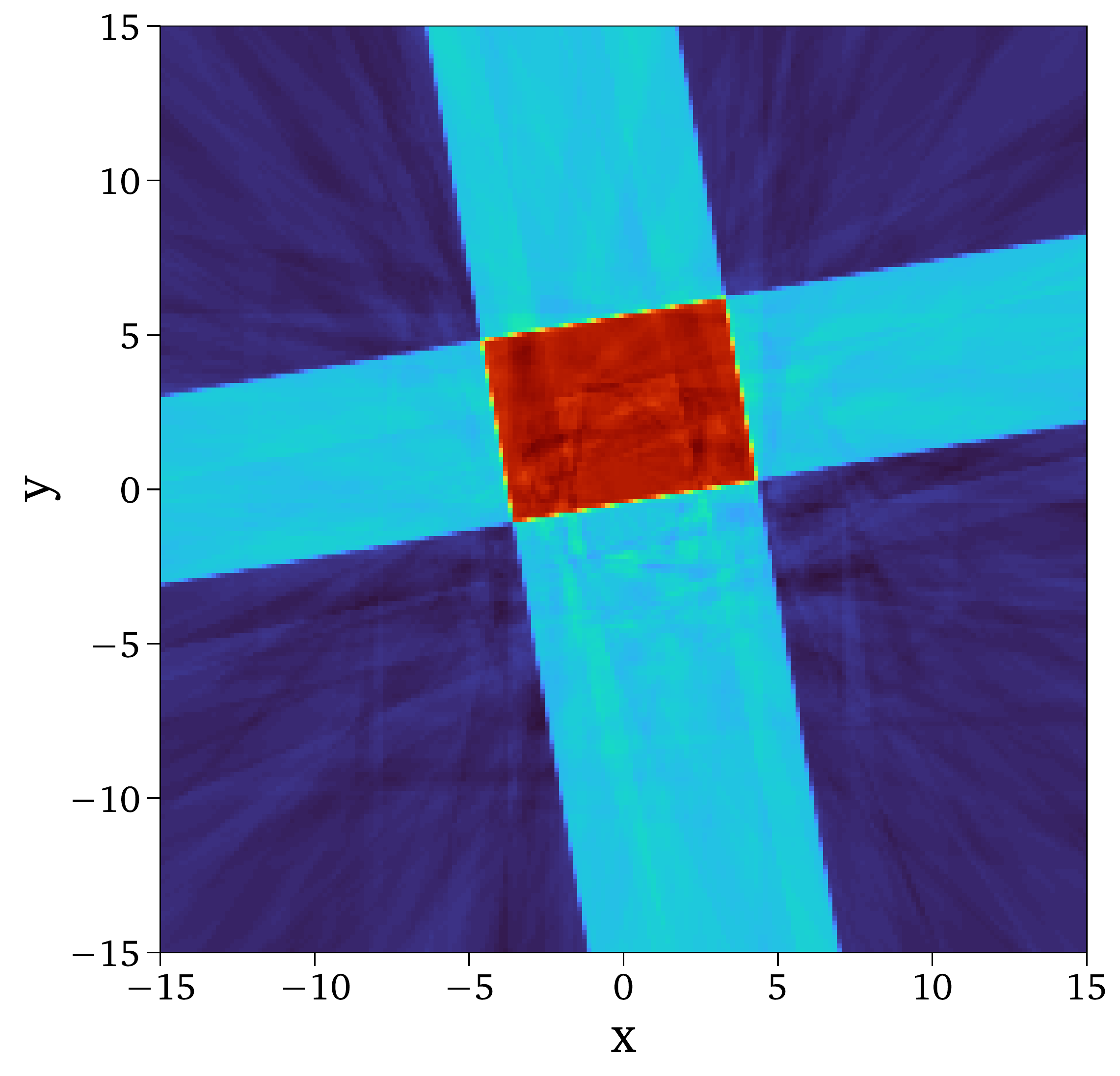}
    \end{subfigure}
    \begin{subfigure}{0.24\textwidth}
        \centering
        \caption{}
        \label{fig:Rect_64PC_3}
        \includegraphics[width=1.5in]{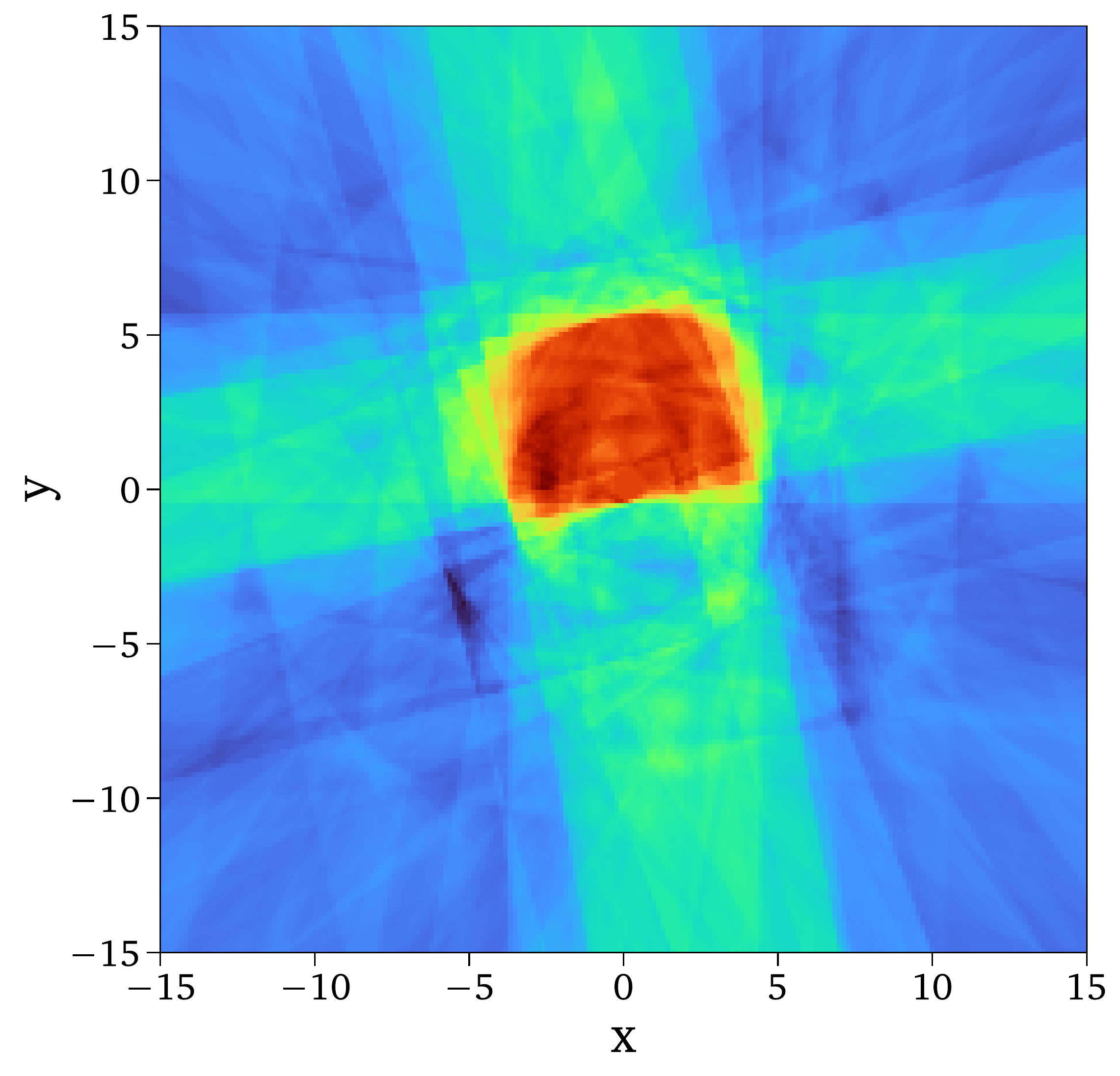}
    \end{subfigure}
    \begin{subfigure}{0.24\textwidth}
        \centering
        \caption{}
        \label{fig:Rect_32PC_3}
        \includegraphics[width=1.5in]{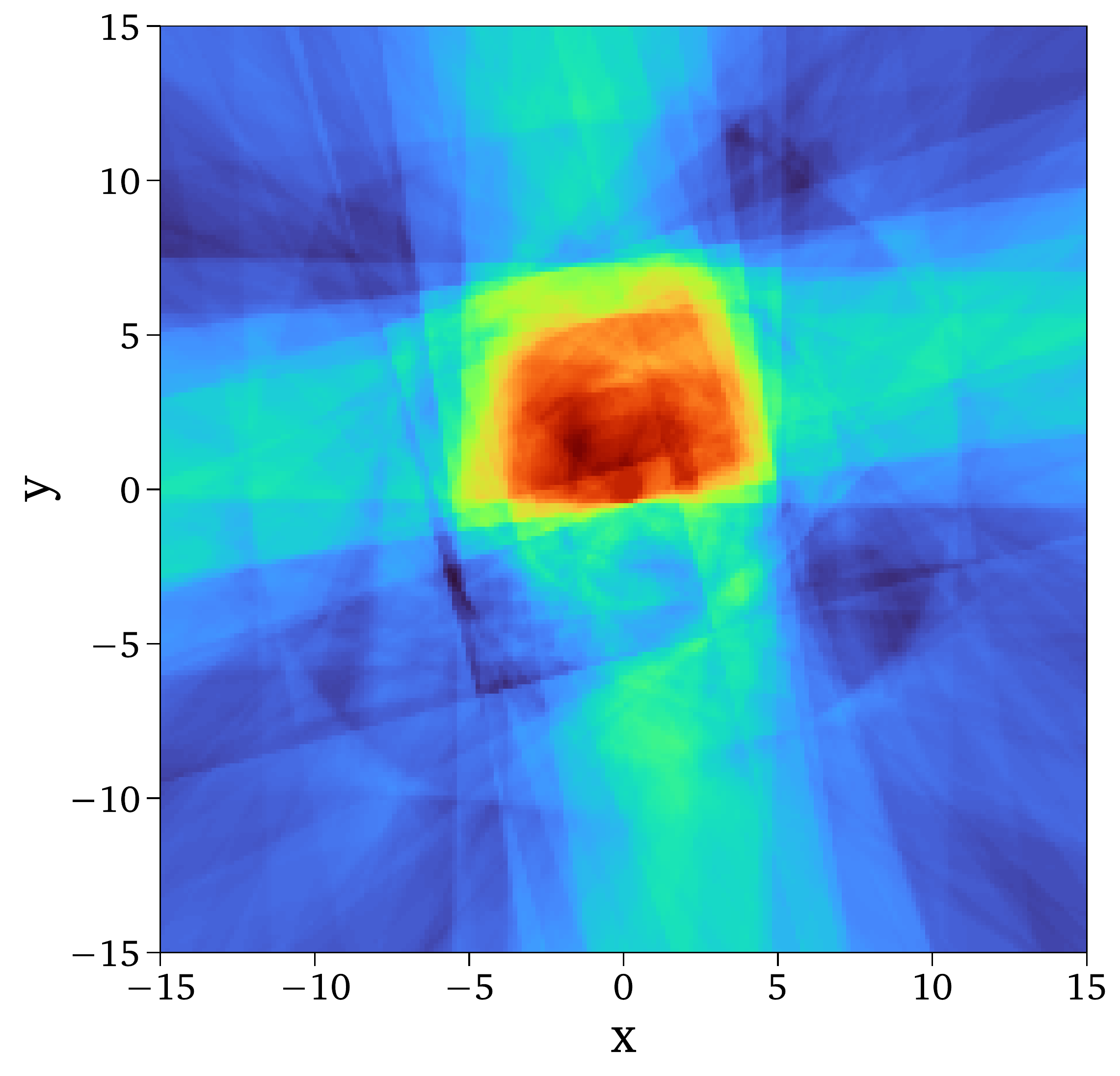}
    \end{subfigure}
    
    \begin{subfigure}{0.24\textwidth}
        \centering
        \caption{}
        \label{fig:Rect_Orig_4_Supp}
        \includegraphics[width=1.5in]{./Figures/Rect_Orig_4.pdf}
    \end{subfigure}
    \begin{subfigure}{0.24\textwidth}
        \centering
        \caption{}
        \label{fig:Rect_128PC_4}
        \includegraphics[width=1.5in]{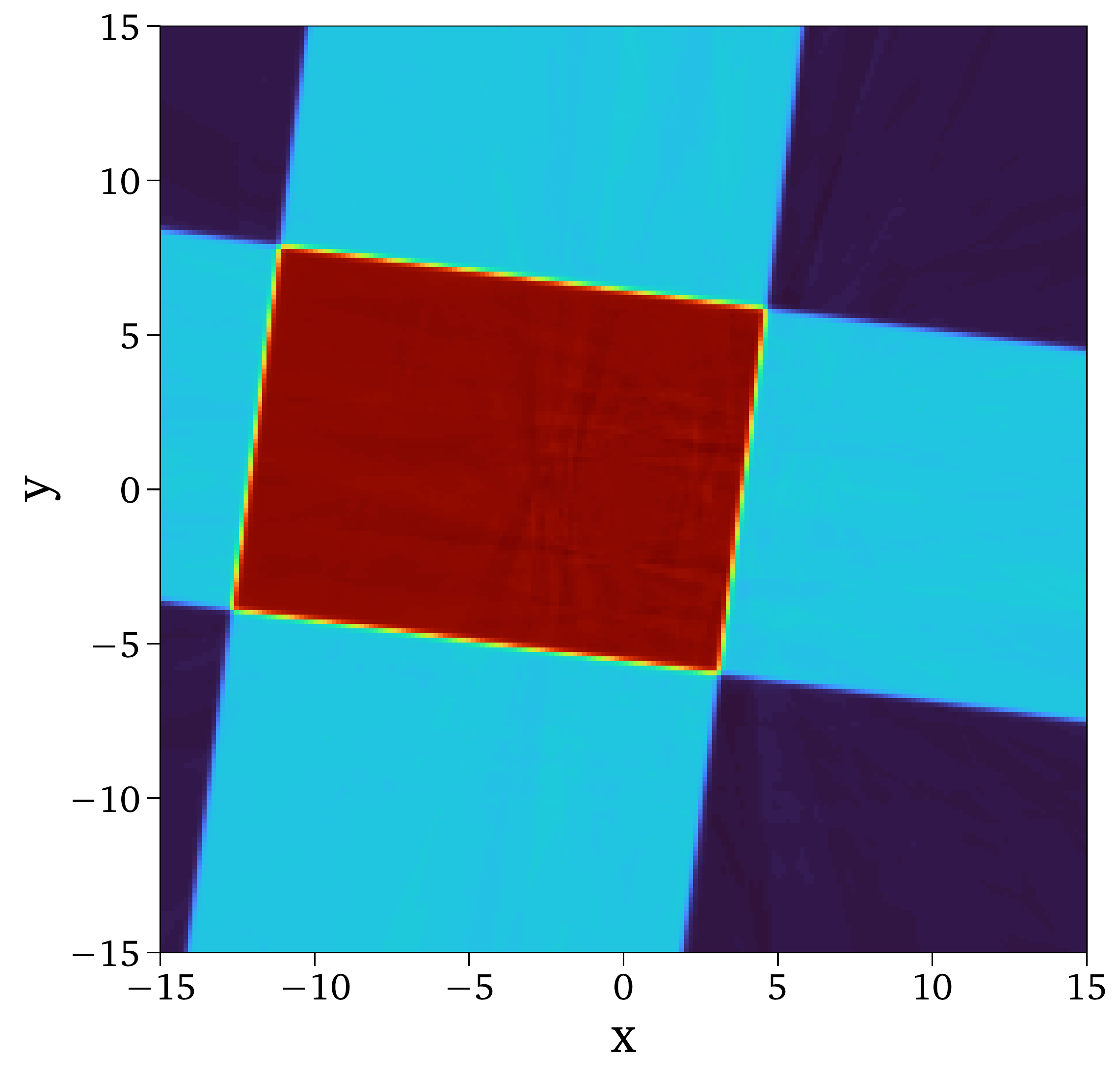}
    \end{subfigure}
    \begin{subfigure}{0.24\textwidth}
        \centering
        \caption{}
        \label{fig:Rect_64PC_4}
        \includegraphics[width=1.5in]{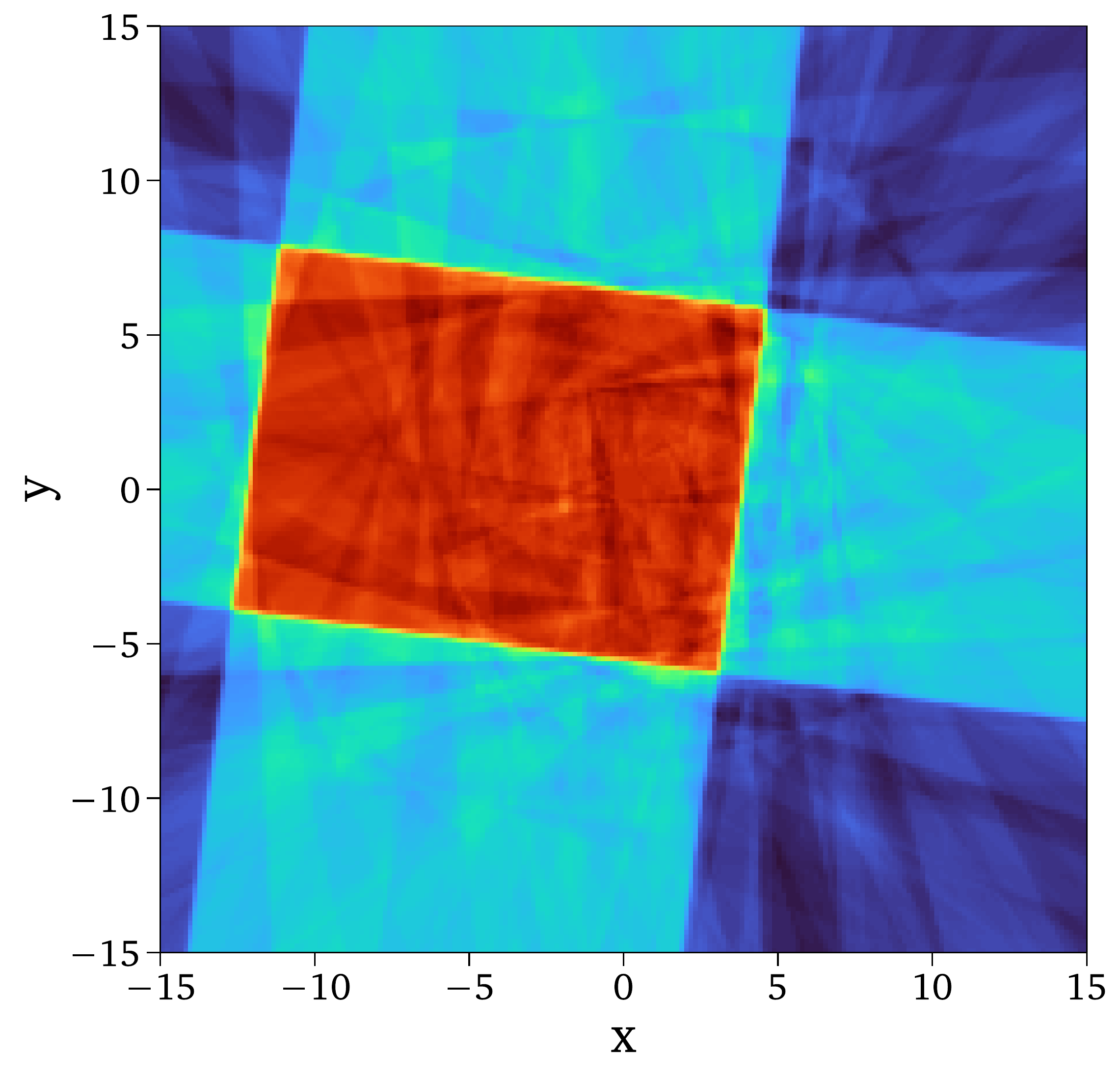}
    \end{subfigure}
    \begin{subfigure}{0.24\textwidth}
        \centering
        \caption{}
        \label{fig:Rect_32PC_4}
        \includegraphics[width=1.5in]{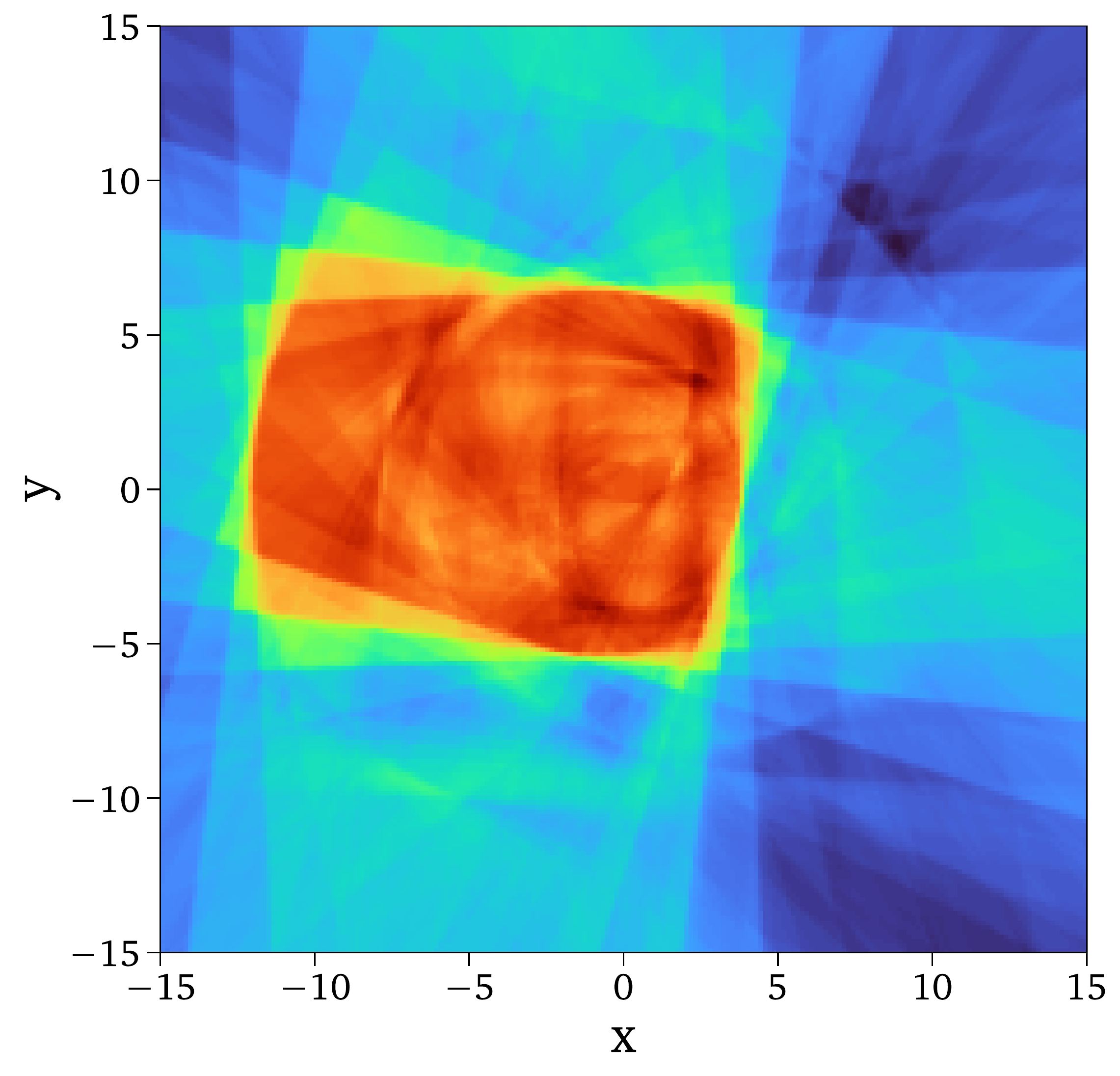}
    \end{subfigure}
    
    \begin{subfigure}{0.24\textwidth}
        \centering
        \caption{}
        \label{fig:Rect_Orig_5}
        \includegraphics[width=1.5in]{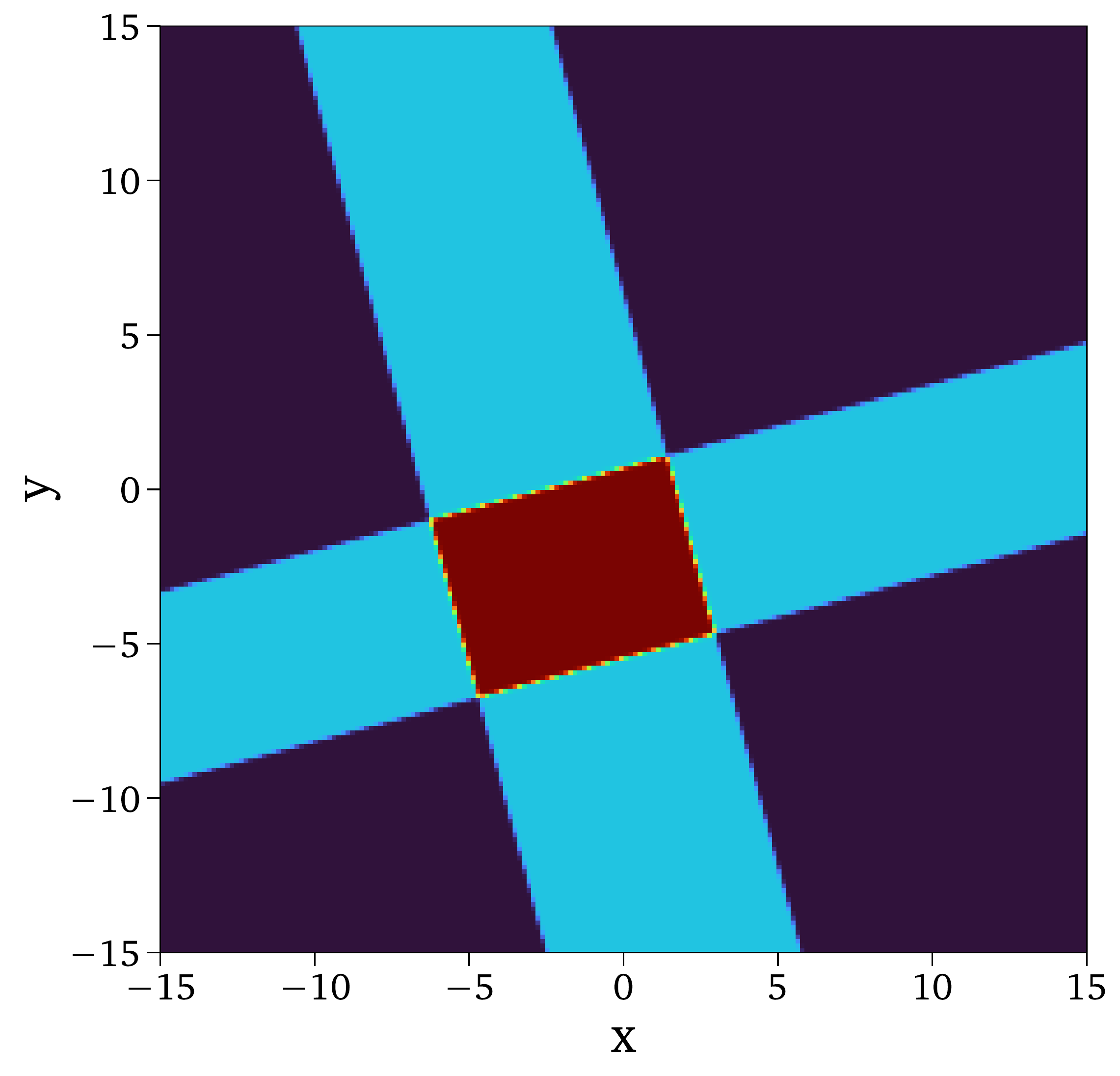}
    \end{subfigure}
    \begin{subfigure}{0.24\textwidth}
        \centering
        \caption{}
        \label{fig:Rect_128PC_5}
        \includegraphics[width=1.5in]{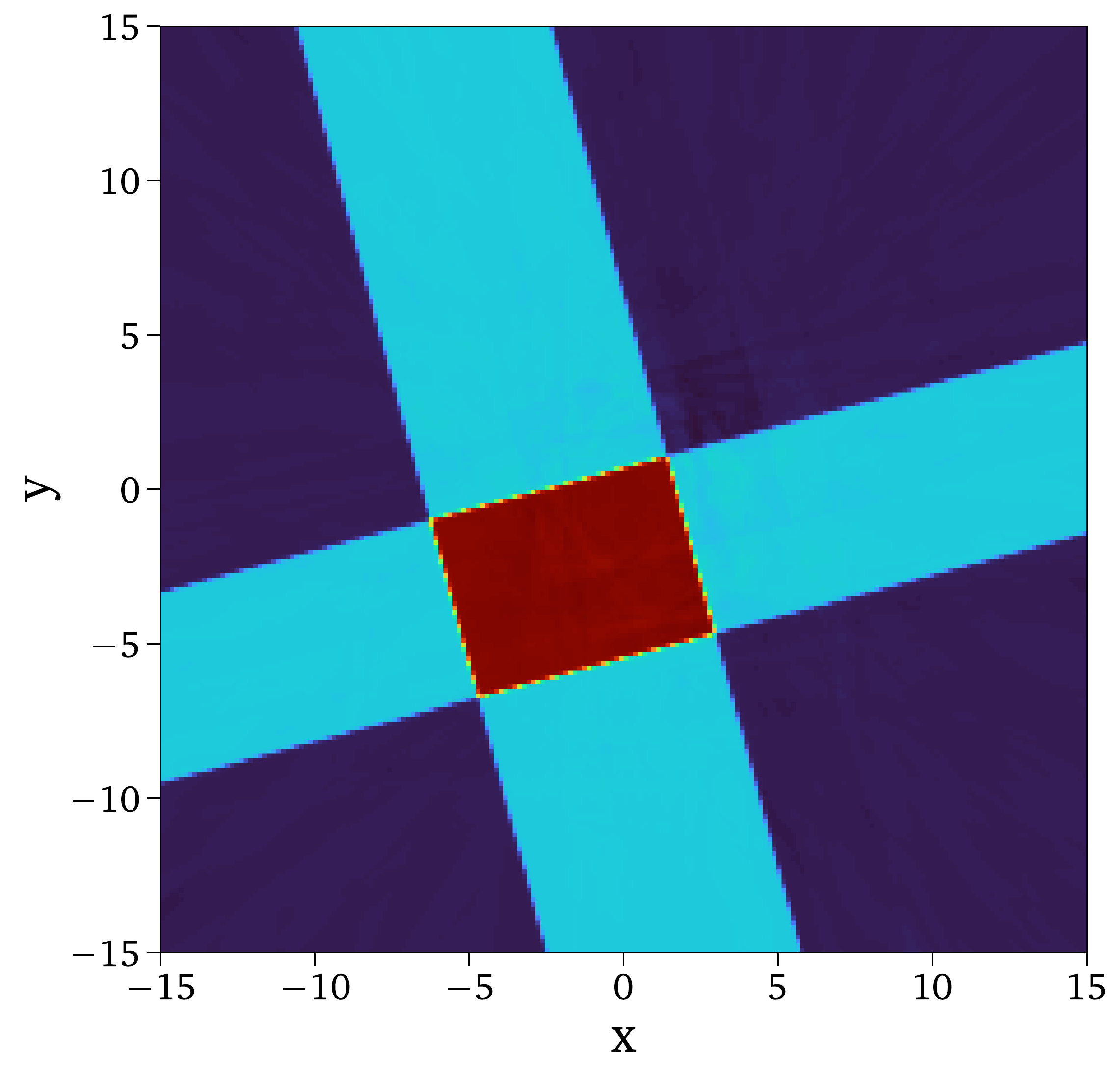}
    \end{subfigure}
    \begin{subfigure}{0.24\textwidth}
        \centering
        \caption{}
        \label{fig:Rect_64PC_5}
        \includegraphics[width=1.5in]{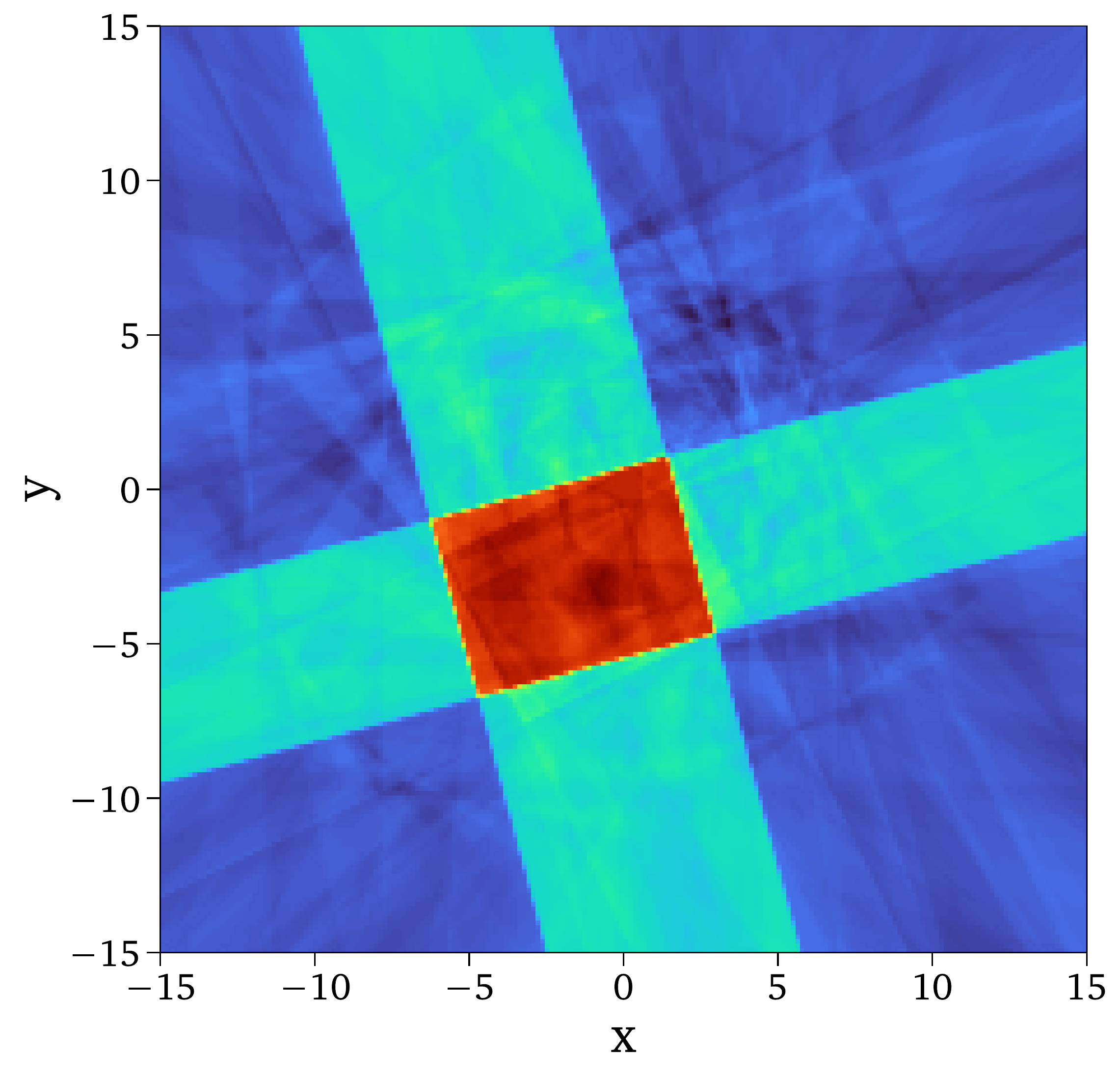}
    \end{subfigure}
    \begin{subfigure}{0.24\textwidth}
        \centering
        \caption{}
        \label{fig:Rect_32PC_5}
        \includegraphics[width=1.5in]{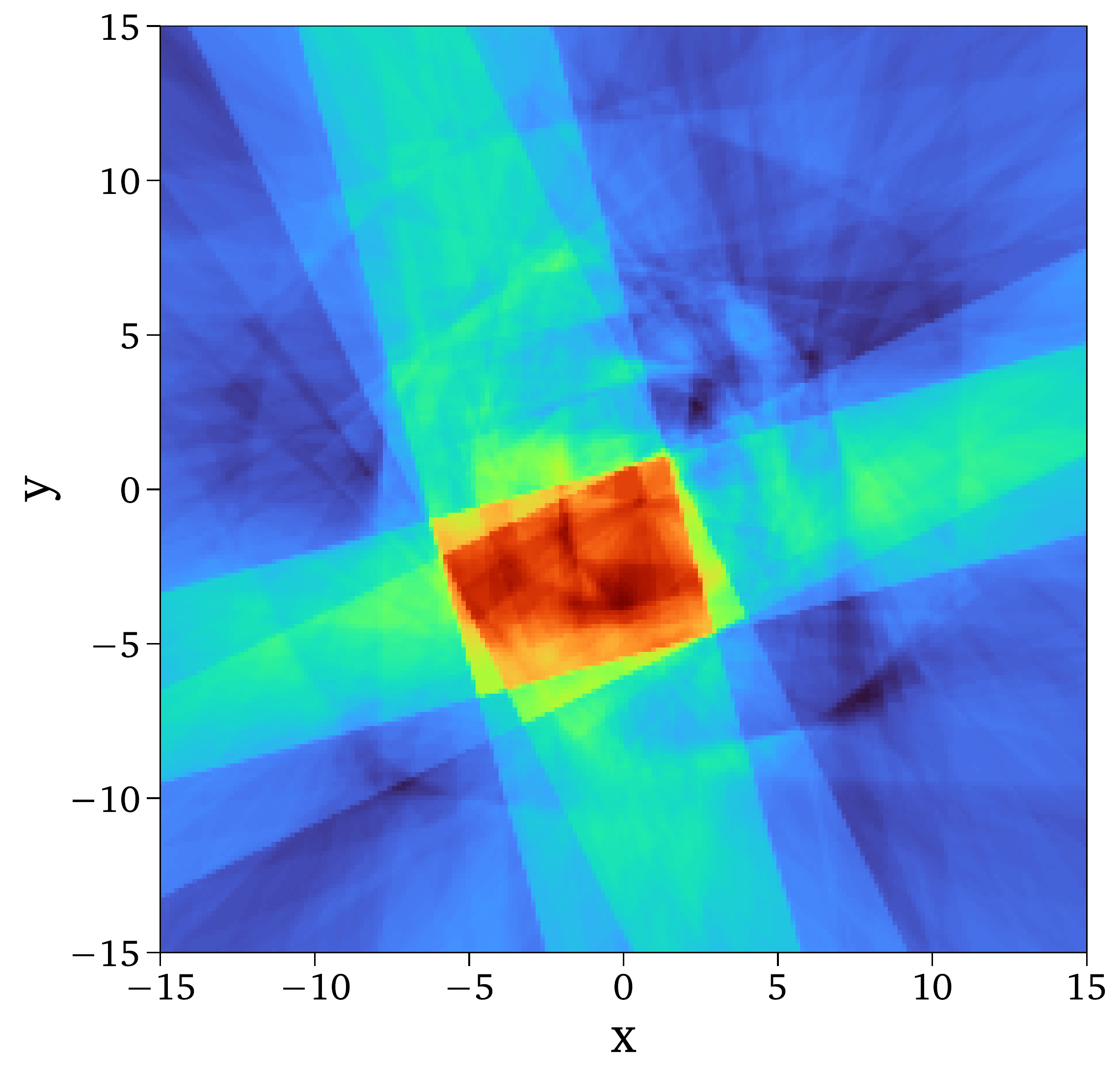}
    \end{subfigure}
    \caption{\textbf{Time snapshots decomposed and reconstructed via POD}. Four of the time snapshots as generated by the original equations (first column from the left) and after being encoded-decoded based on POD's first 128, 64, and 32 modes (second, third, and fourth columns, respectively). The time snapshots are taken at $t=\{2.5,5,7.5,10\}$ [s] (rows from top to bottom).}
    \label{fig:Rect_PC_Supp}
\end{figure}


\clearpage
\subsection{Vanilla DeepONet}

\begin{figure}[!htp] 
    \centering
    \includegraphics[width=6.0in]{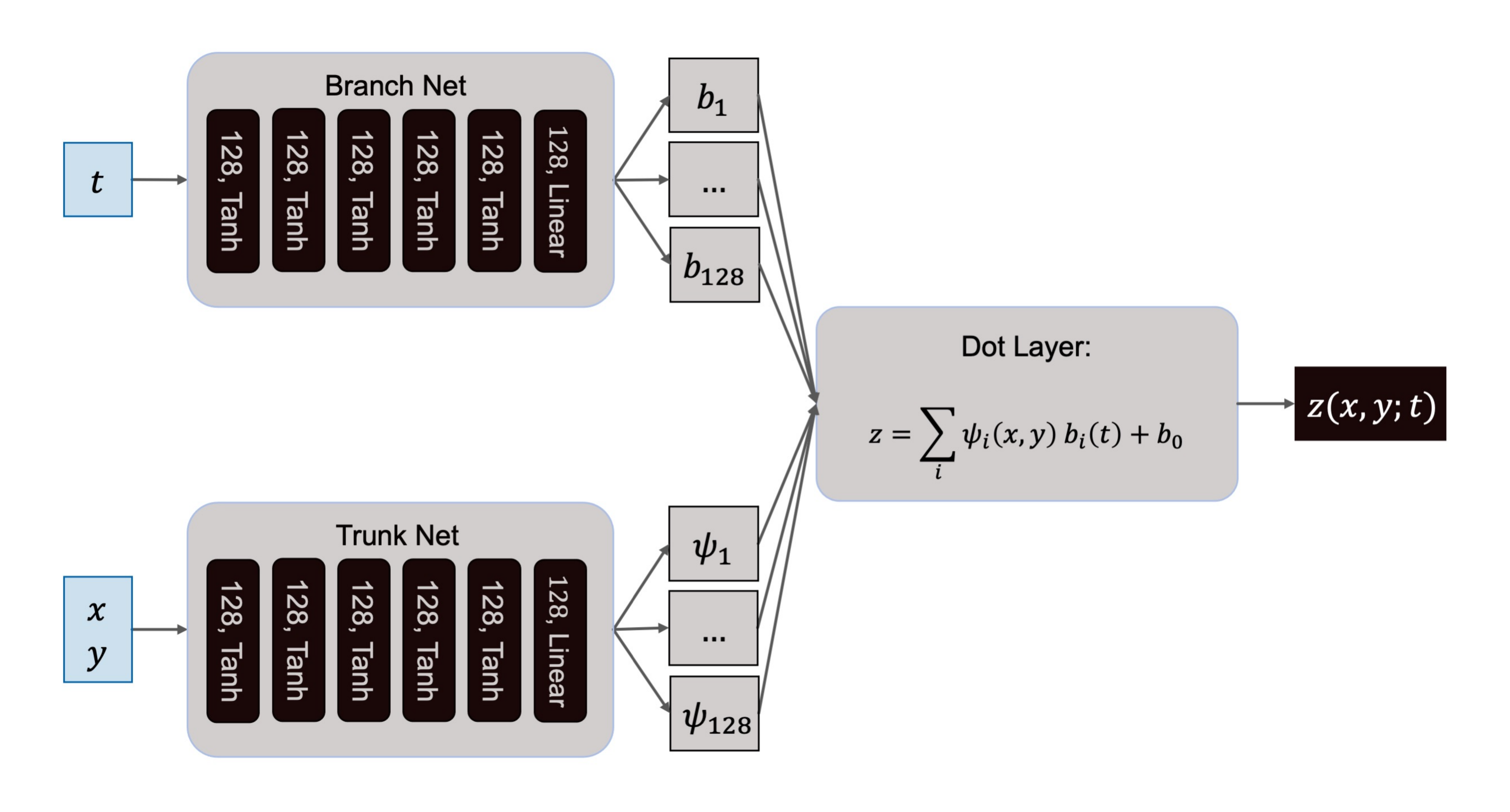}
    \caption{\textbf{Architecture of the vanilla DeepONet for the rotating-translating-stretching rigid body test case}. Note: the combination of hyperbolic tangent and exponential activation functions is here employed consistently with the test cases' generative equations.}
    \label{fig:Rect_DeepONet_1}
\end{figure}

\begin{figure}[!htb]
    \begin{subfigure}{0.24\textwidth}
        \centering
        \caption{}
        \label{fig:Rect_DeepONet_Supp_1}
        \includegraphics[width=1.5in]{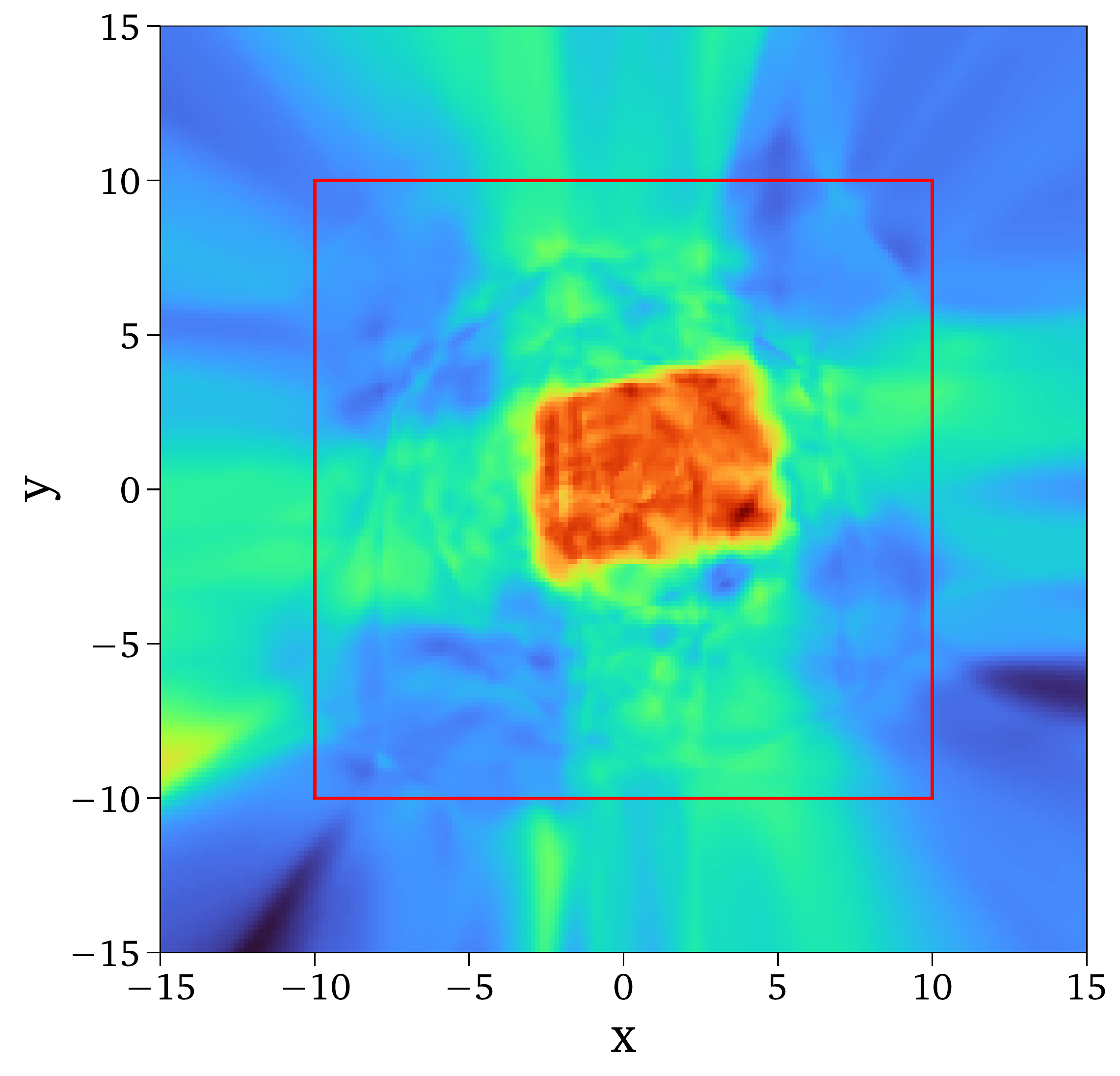}
    \end{subfigure}
    \begin{subfigure}{0.24\textwidth}
        \centering
        \caption{}
        \label{fig:Rect_DeepONet_Supp_2}
        \includegraphics[width=1.5in]{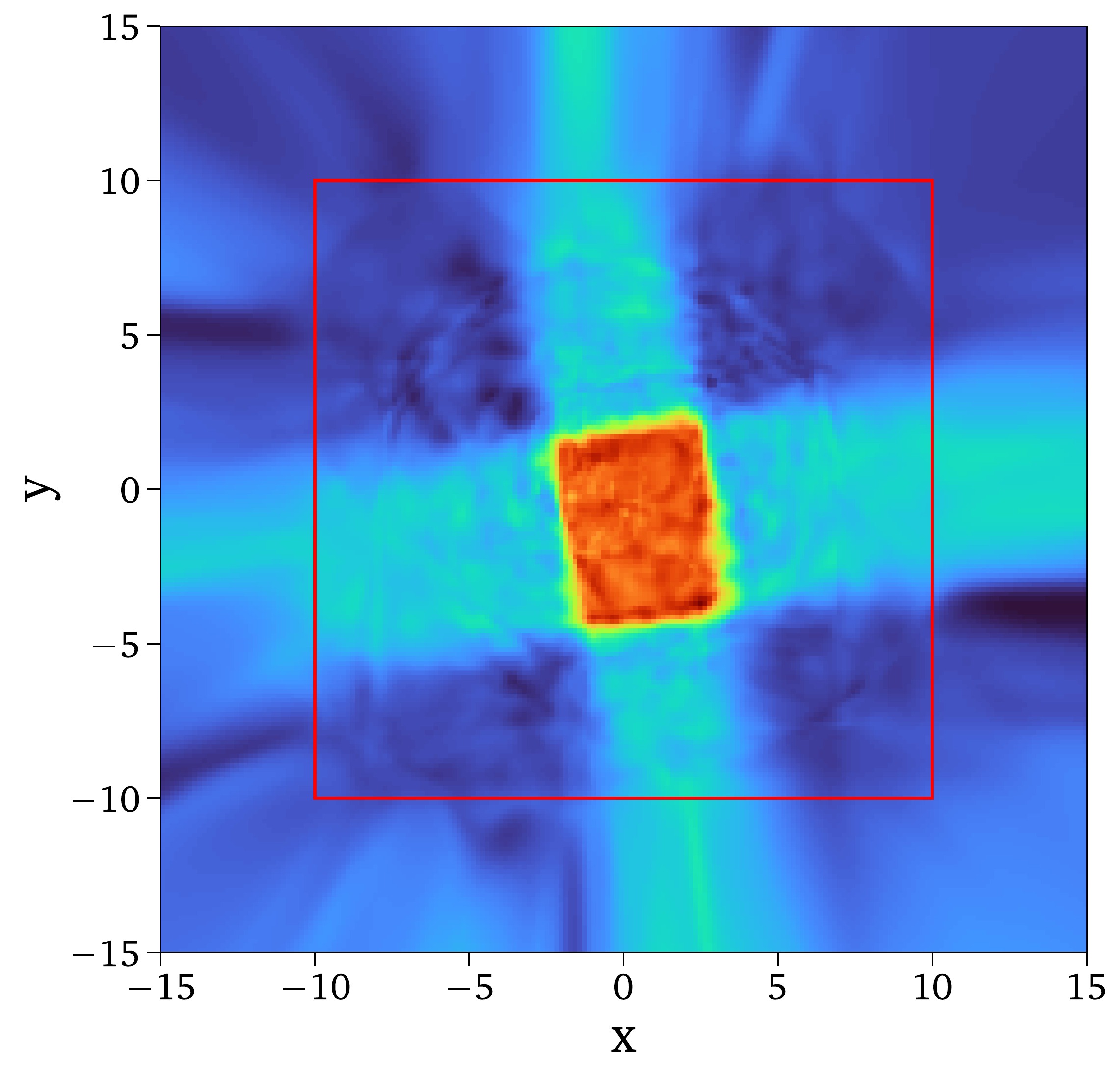}
    \end{subfigure}
    \begin{subfigure}{0.24\textwidth}
        \centering
        \caption{}
        \label{fig:Rect_DeepONet_Supp_3}
        \includegraphics[width=1.5in]{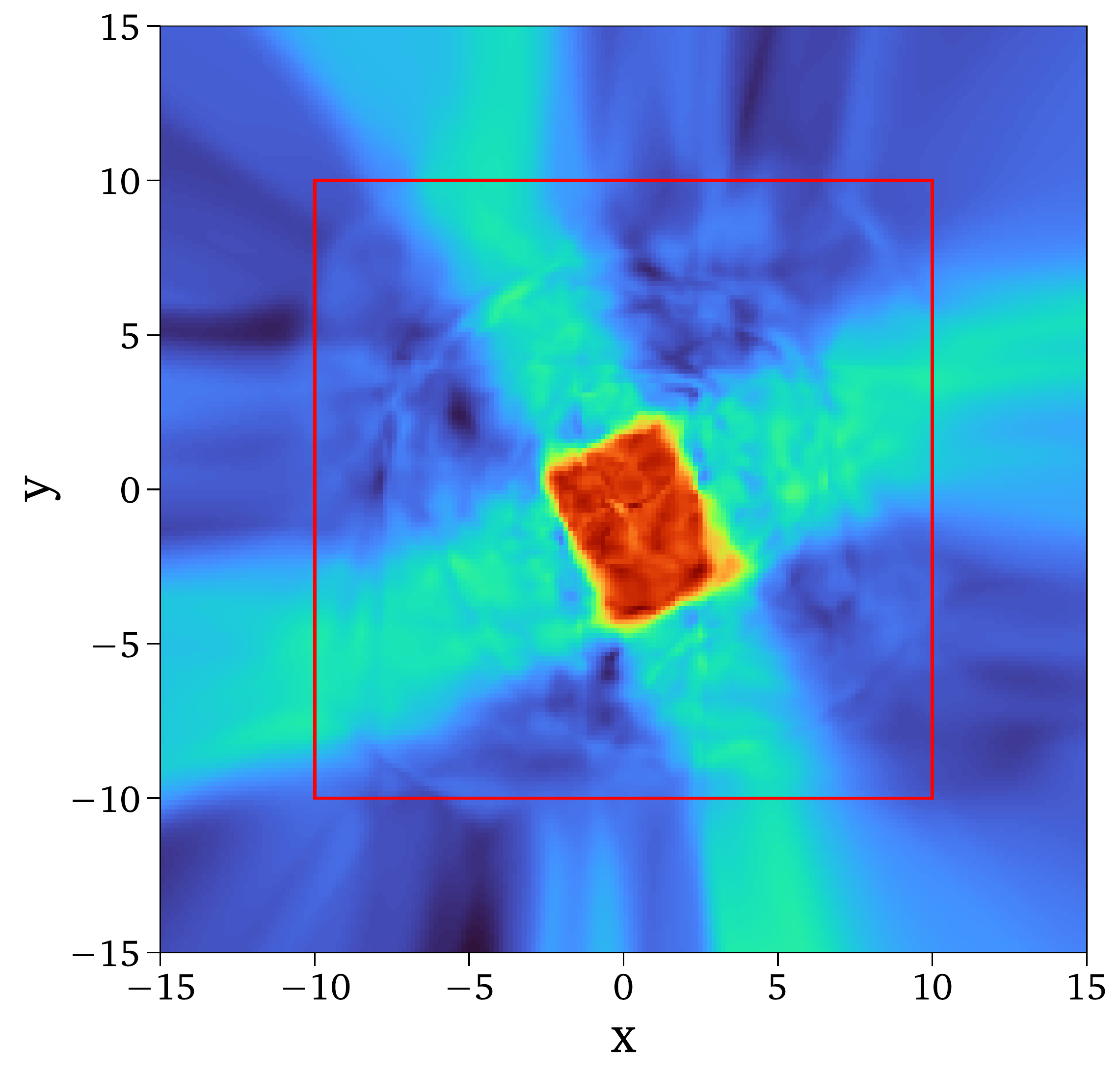}
    \end{subfigure}
    \begin{subfigure}{0.24\textwidth}
        \centering
        \caption{}
        \label{fig:Rect_DeepONet_Supp_4}
        \includegraphics[width=1.5in]{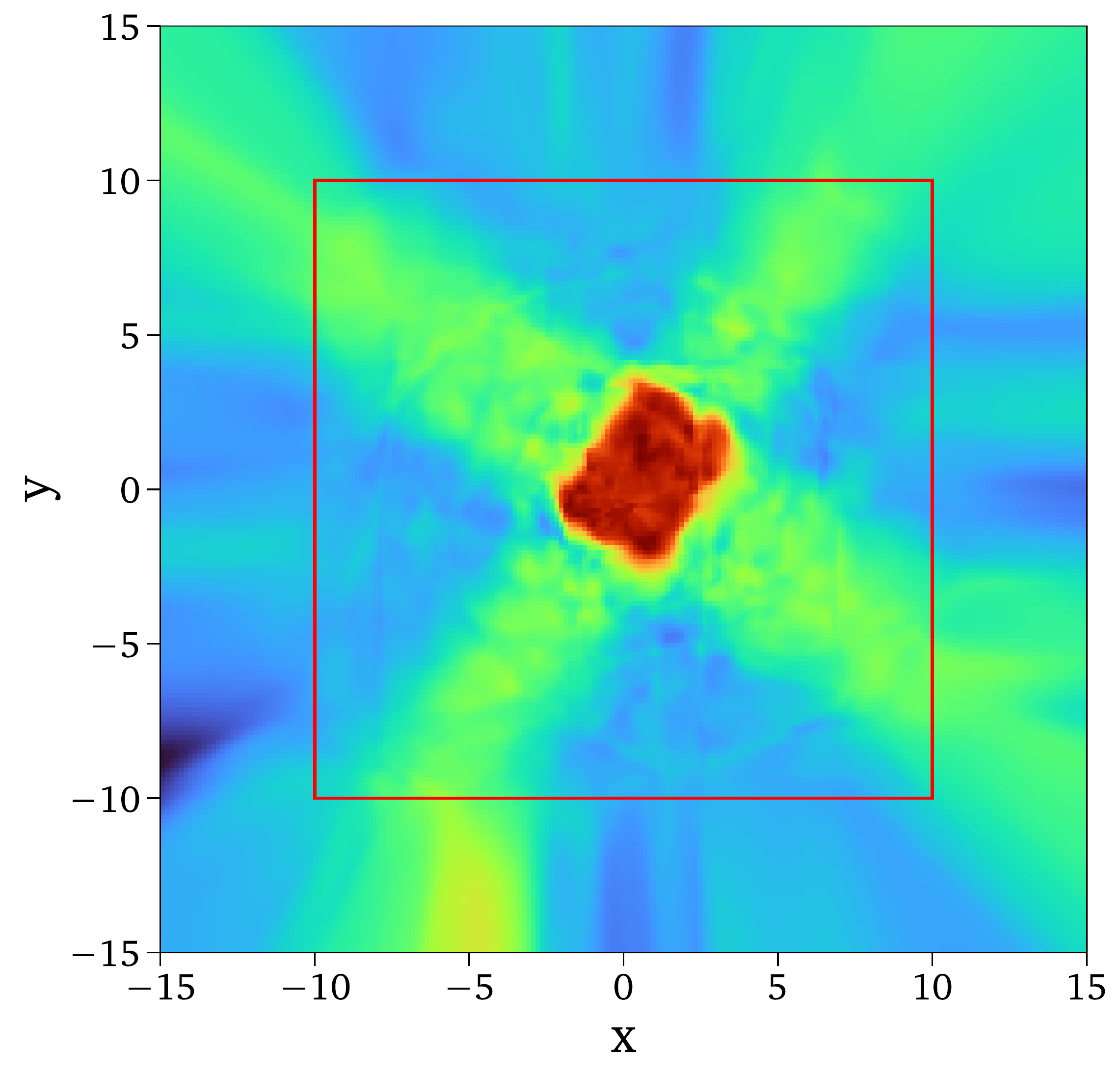}
    \end{subfigure}
    
    \begin{subfigure}{0.24\textwidth}
        \centering
        \caption{}
        \label{fig:Rect_DeepONet_Supp_5}
        \includegraphics[width=1.5in]{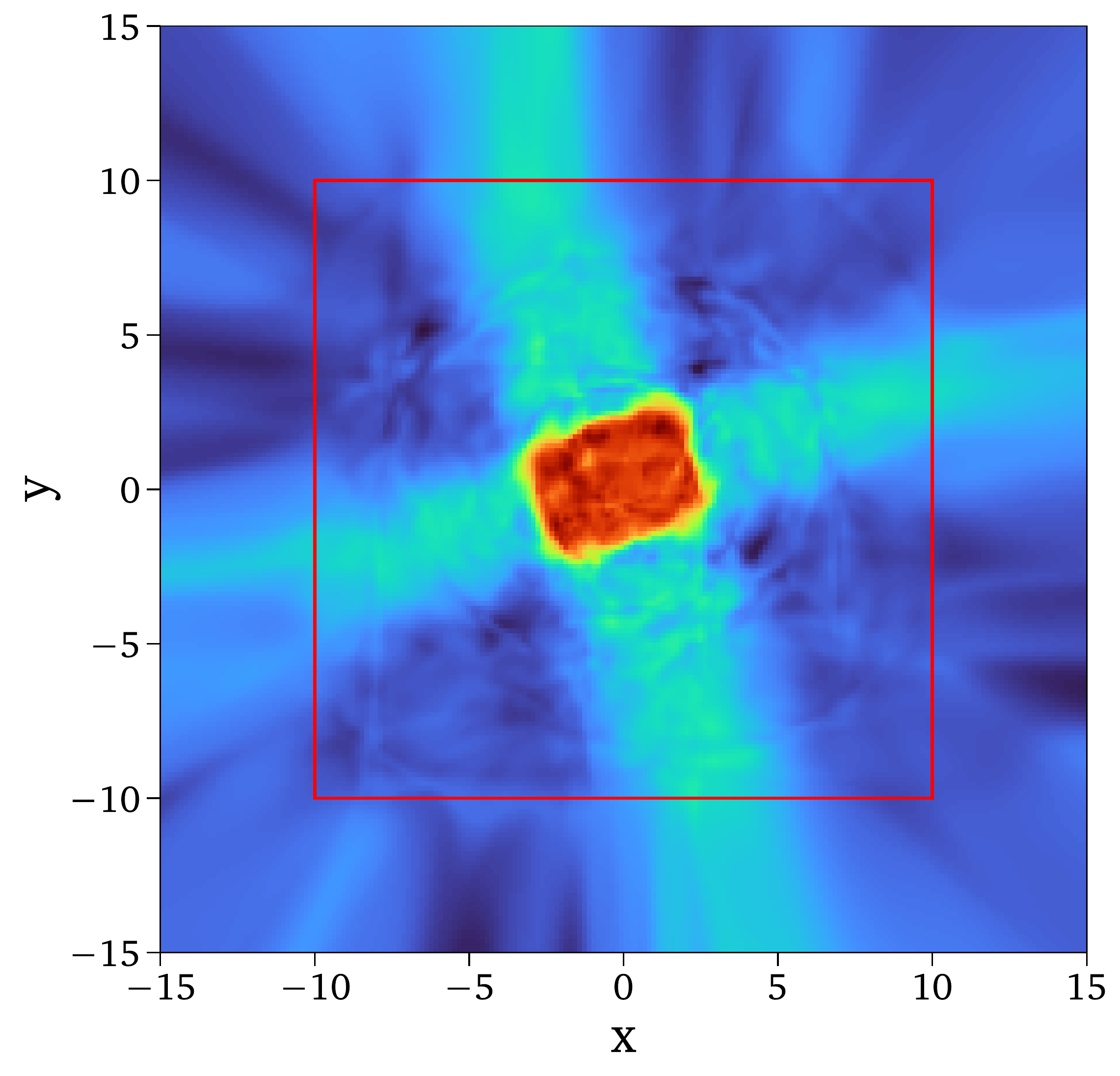}
    \end{subfigure}
    \begin{subfigure}{0.24\textwidth}
        \centering
        \caption{}
        \label{fig:Rect_DeepONet_Supp_6}
        \includegraphics[width=1.5in]{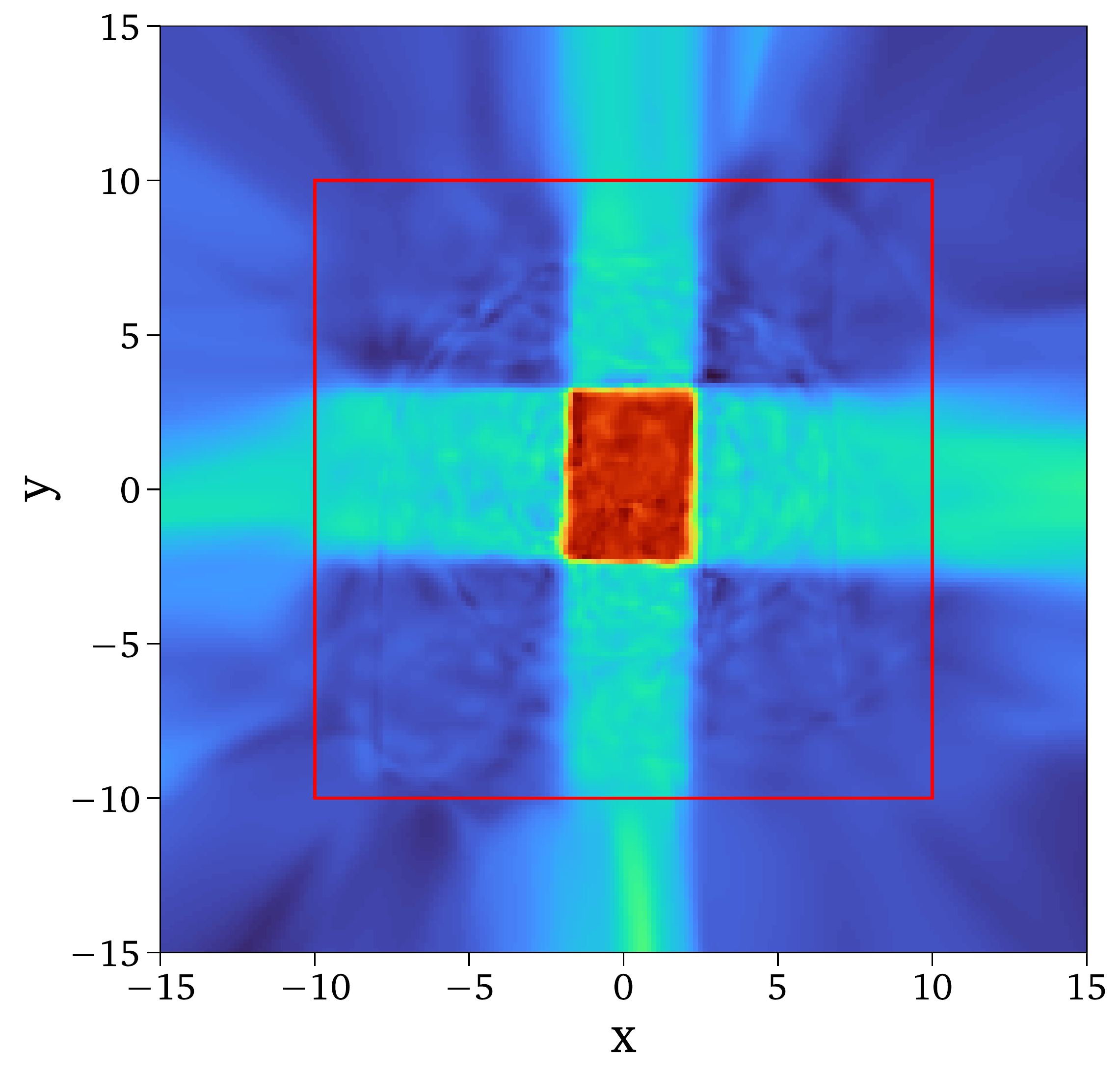}
    \end{subfigure}
    \begin{subfigure}{0.24\textwidth}
        \centering
        \caption{}
        \label{fig:Rect_DeepONet_Supp_7}
        \includegraphics[width=1.5in]{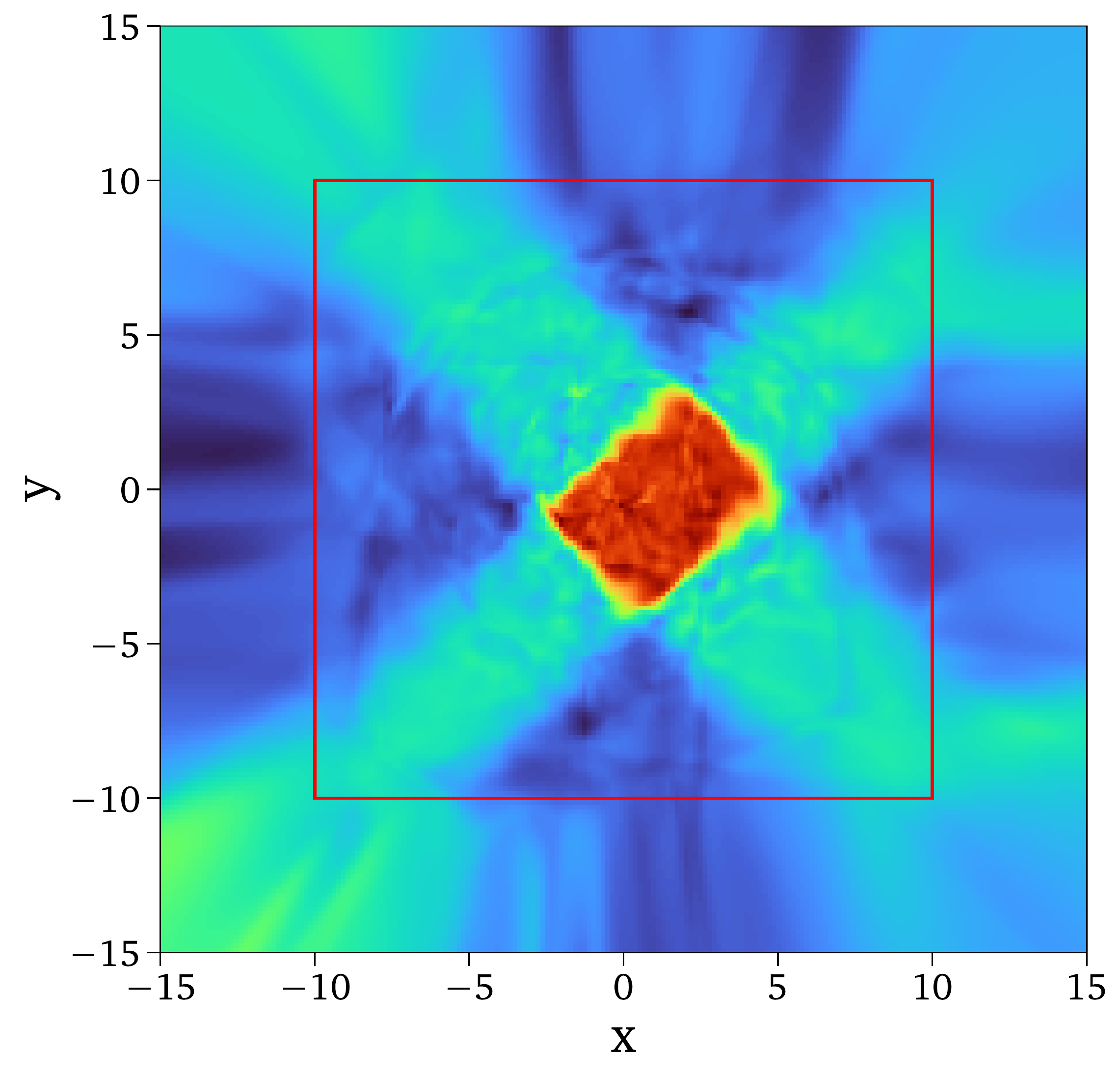}
    \end{subfigure}
    \begin{subfigure}{0.24\textwidth}
        \centering
        \caption{}
        \label{fig:Rect_DeepONet_Supp_8}
        \includegraphics[width=1.5in]{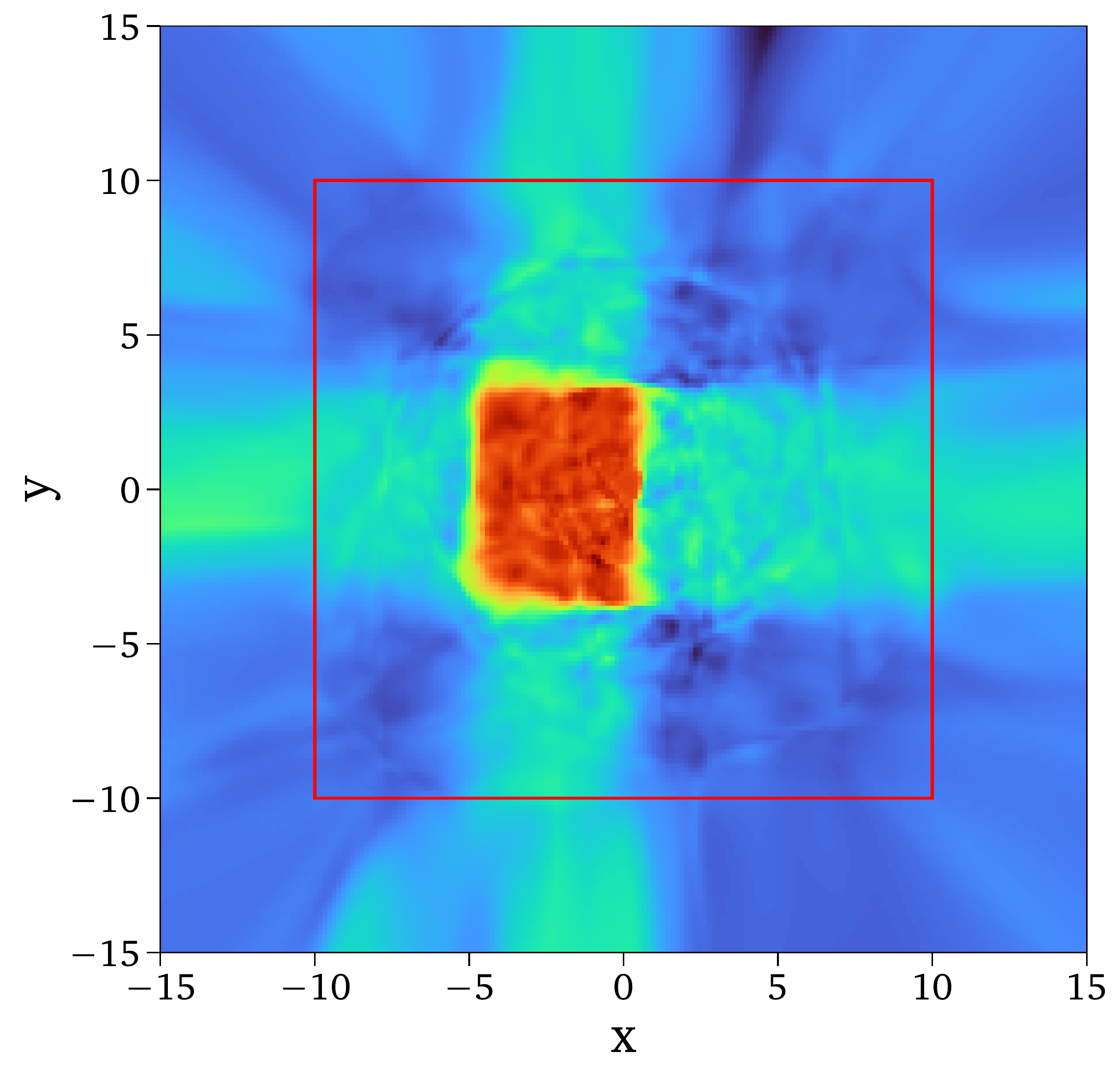}
    \end{subfigure}
    
    \begin{subfigure}{0.24\textwidth}
        \centering
        \caption{}
        \label{fig:Rect_DeepONet_Supp_9}
        \includegraphics[width=1.5in]{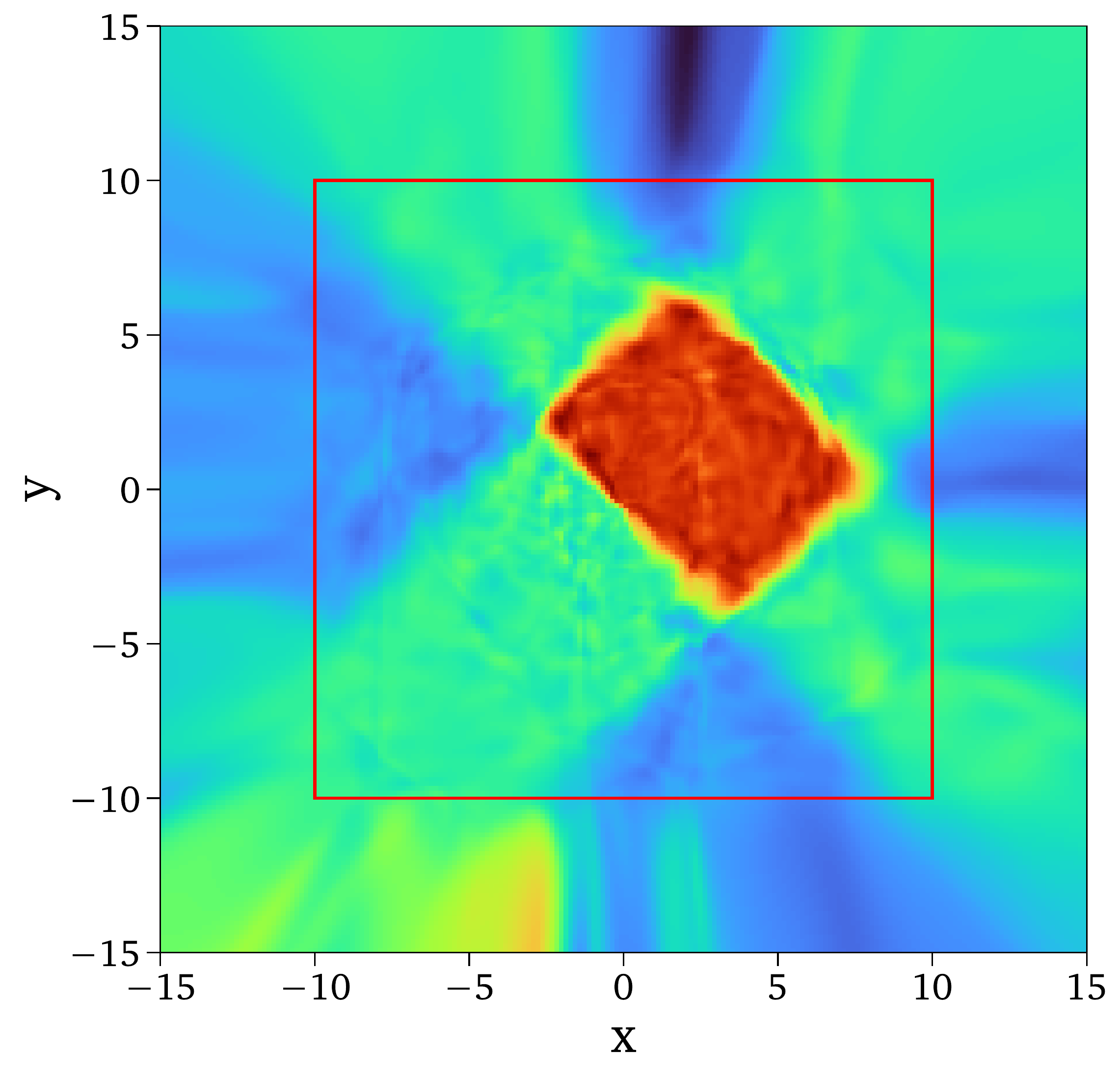}
    \end{subfigure}
    \begin{subfigure}{0.24\textwidth}
        \centering
        \caption{}
        \label{fig:Rect_DeepONet_Supp_10}
        \includegraphics[width=1.5in]{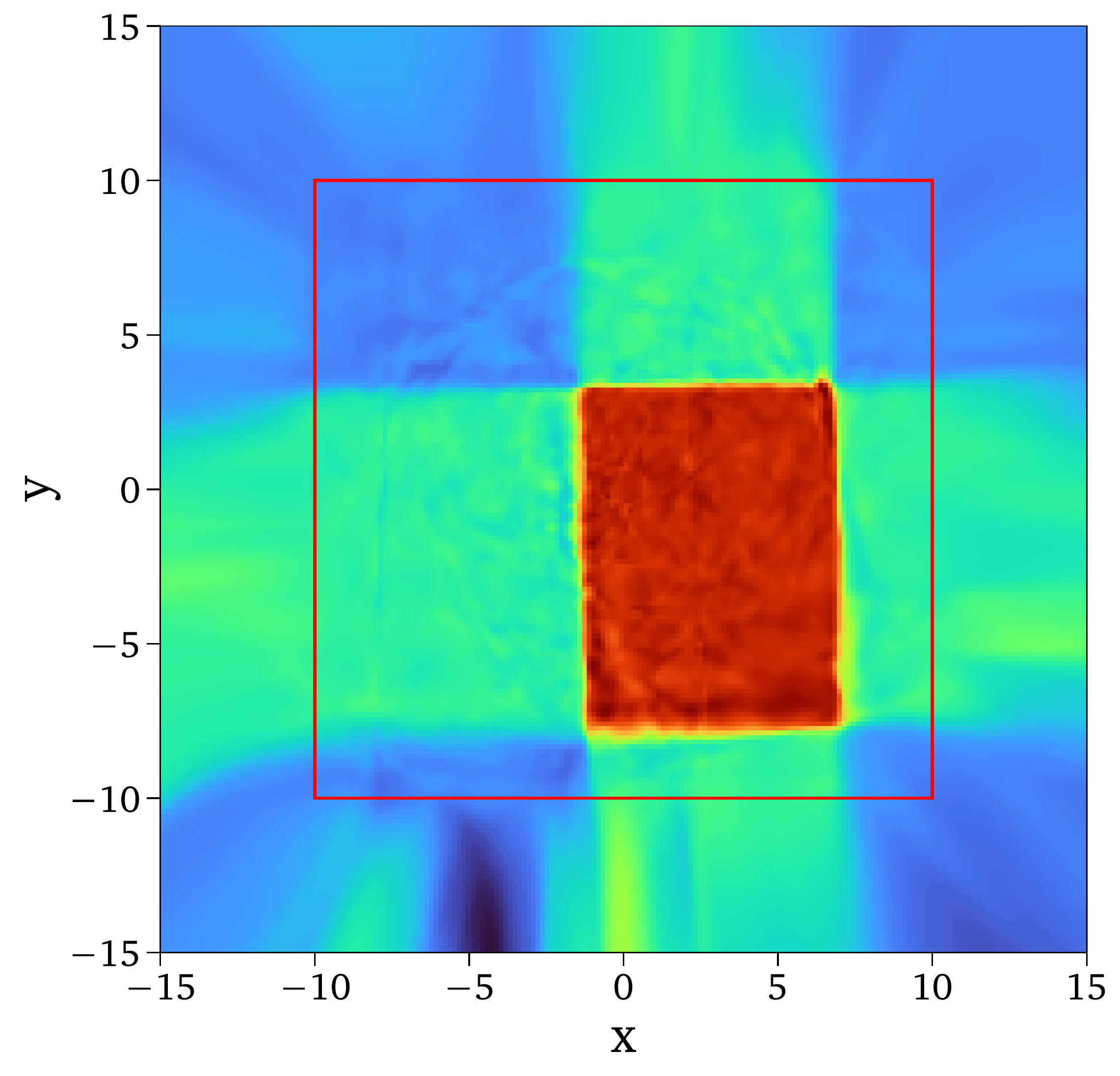}
    \end{subfigure}
    \begin{subfigure}{0.24\textwidth}
        \centering
        \caption{}
        \label{fig:Rect_DeepONet_Supp_11}
        \includegraphics[width=1.5in]{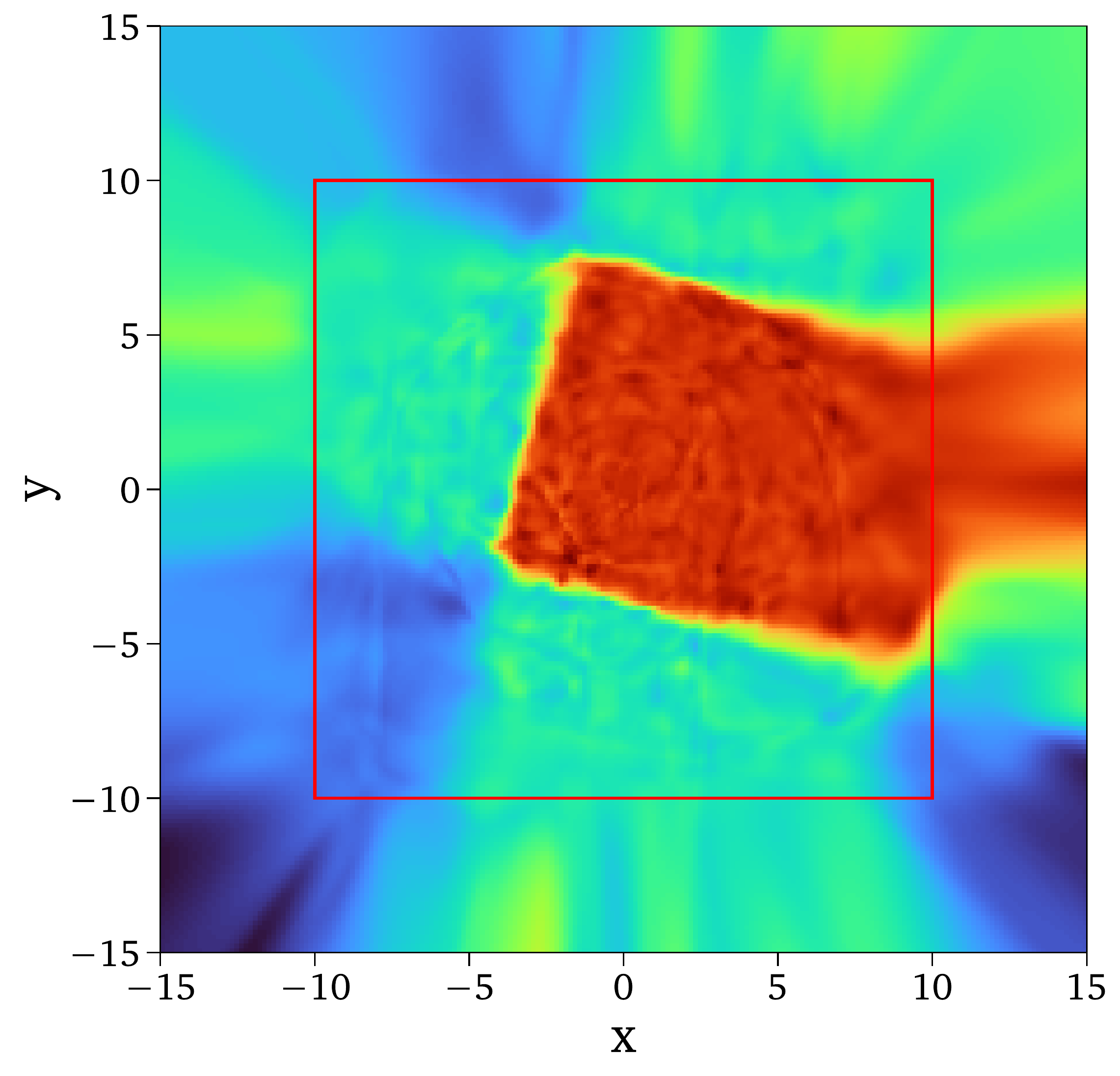}
    \end{subfigure}
    \begin{subfigure}{0.24\textwidth}
        \centering
        \caption{}
        \label{fig:Rect_DeepONet_Supp_12}
        \includegraphics[width=1.5in]{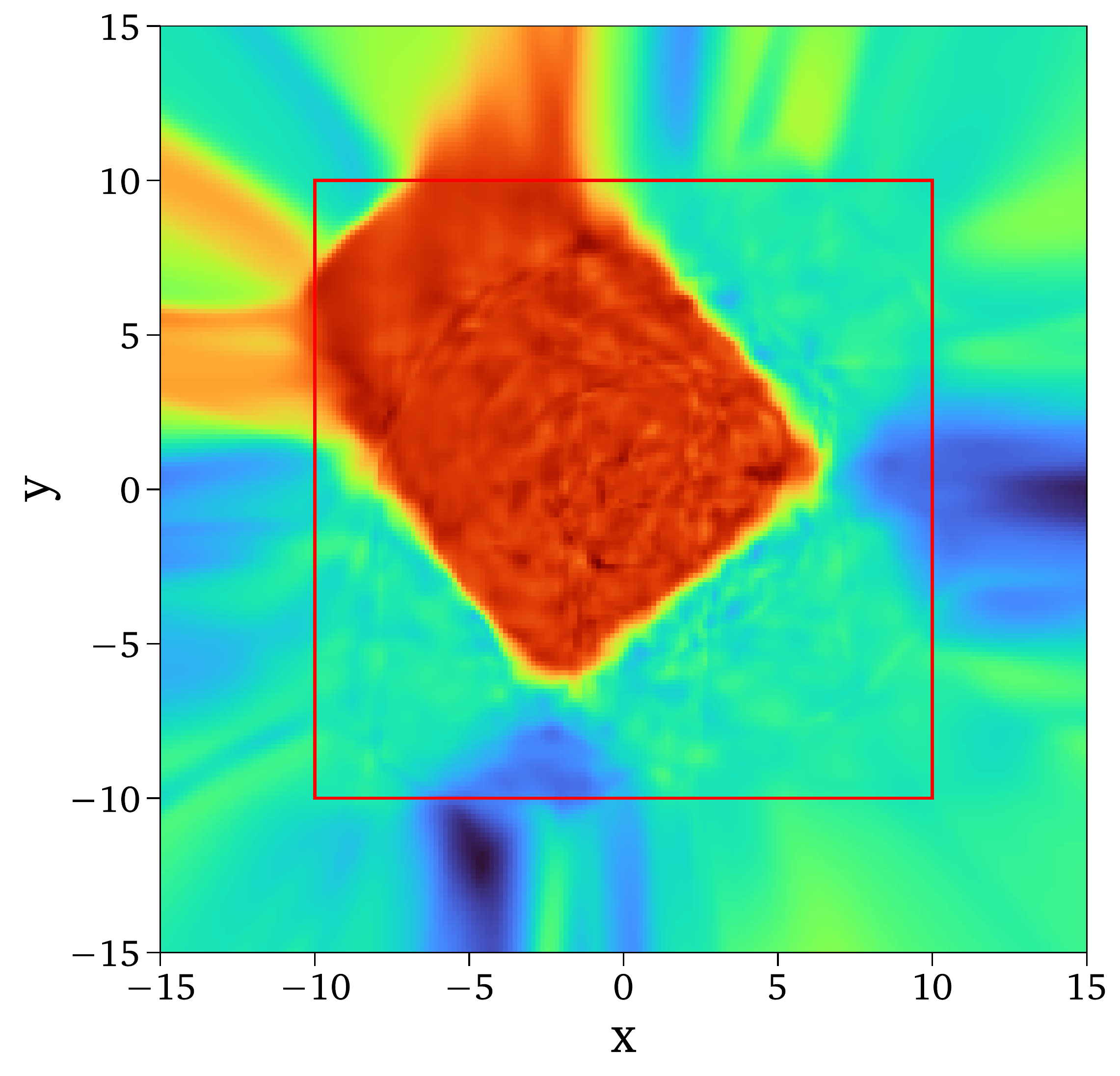}
    \end{subfigure}
    
    \begin{subfigure}{0.24\textwidth}
        \centering
        \caption{}
        \label{fig:Rect_DeepONet_Supp_13}
        \includegraphics[width=1.5in]{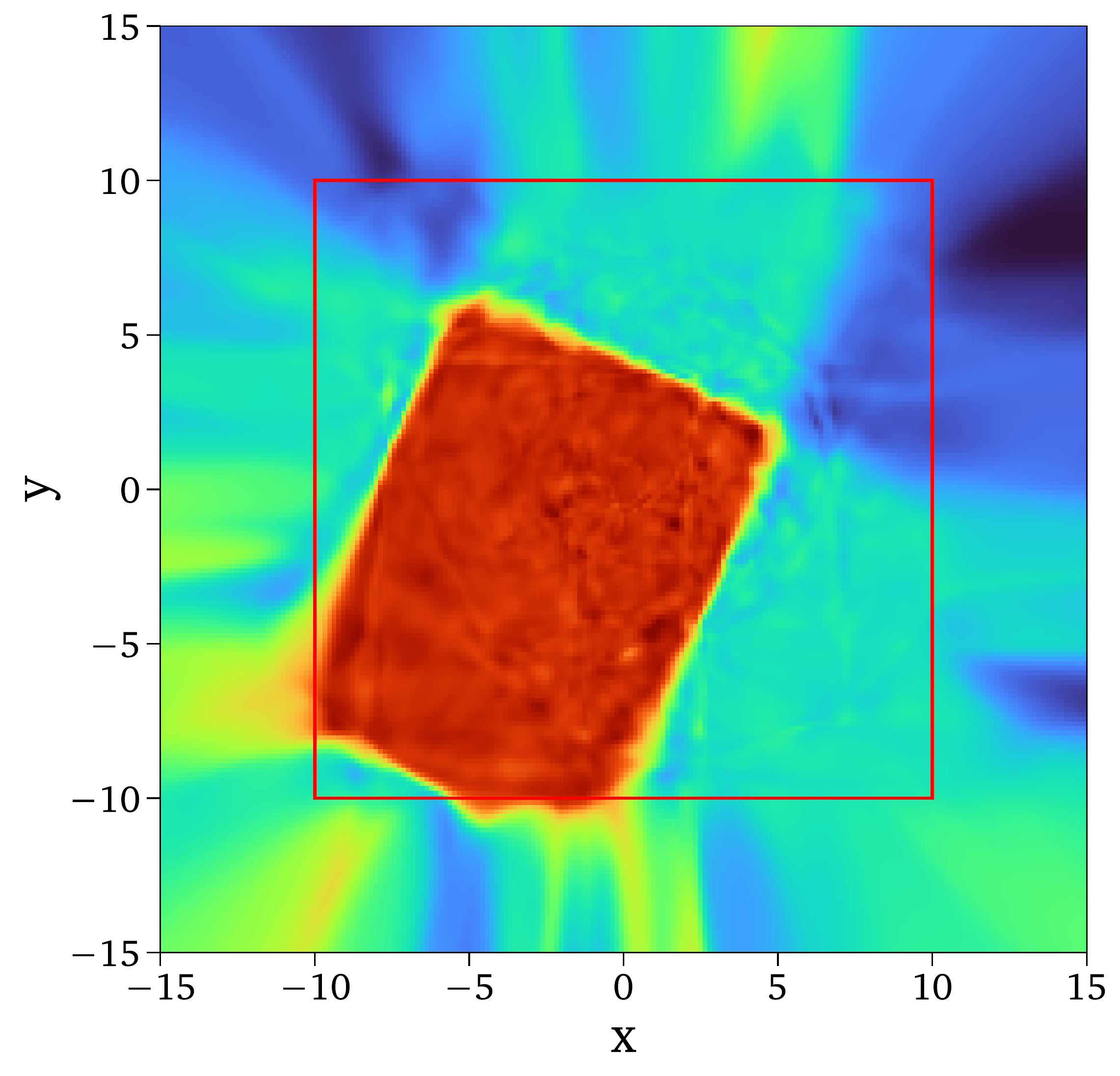}
    \end{subfigure}
    \begin{subfigure}{0.24\textwidth}
        \centering
        \caption{}
        \label{fig:Rect_DeepONet_Supp_14}
        \includegraphics[width=1.5in]{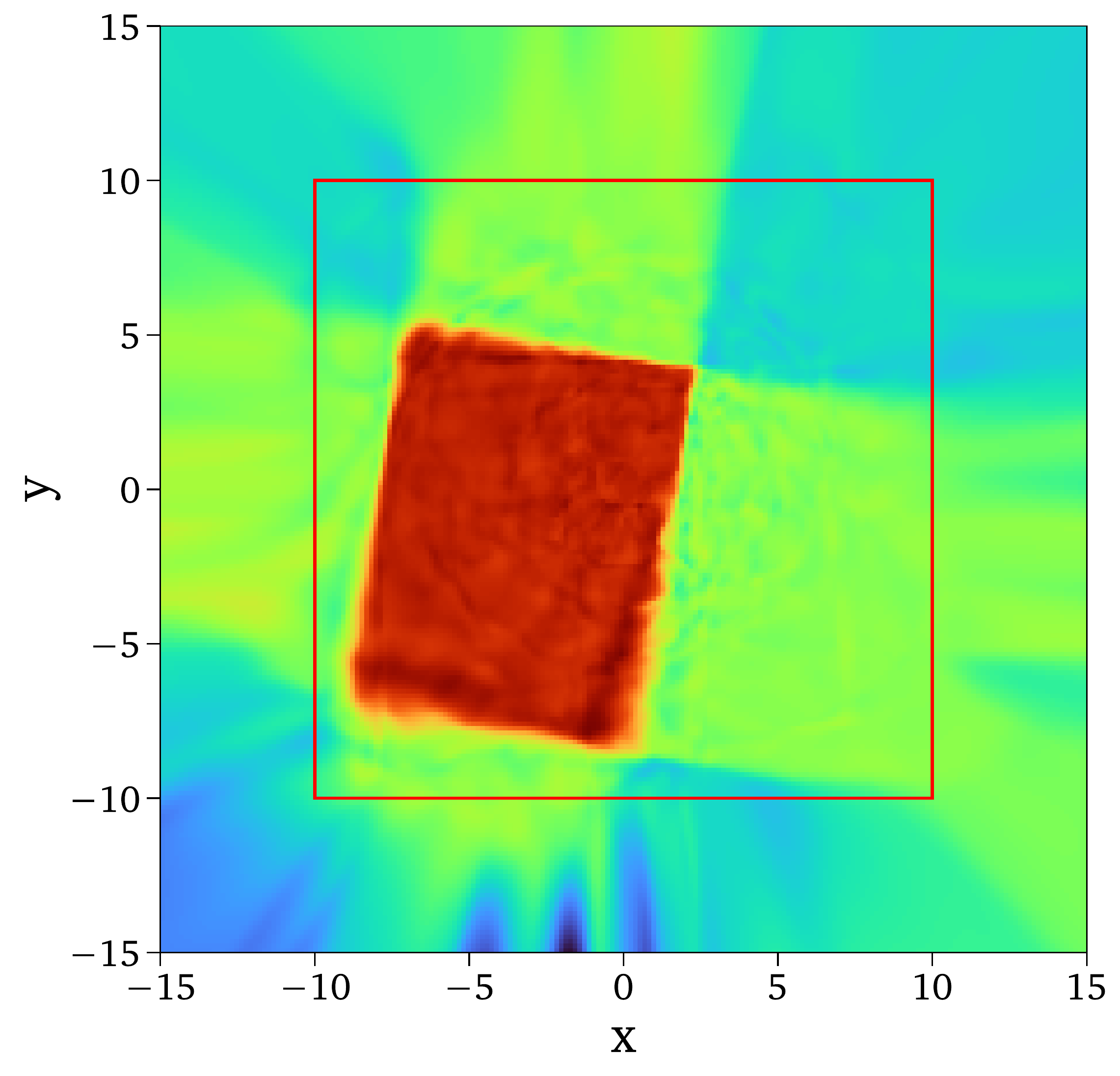}
    \end{subfigure}
    \begin{subfigure}{0.24\textwidth}
        \centering
        \caption{}
        \label{fig:Rect_DeepONet_Supp_15}
        \includegraphics[width=1.5in]{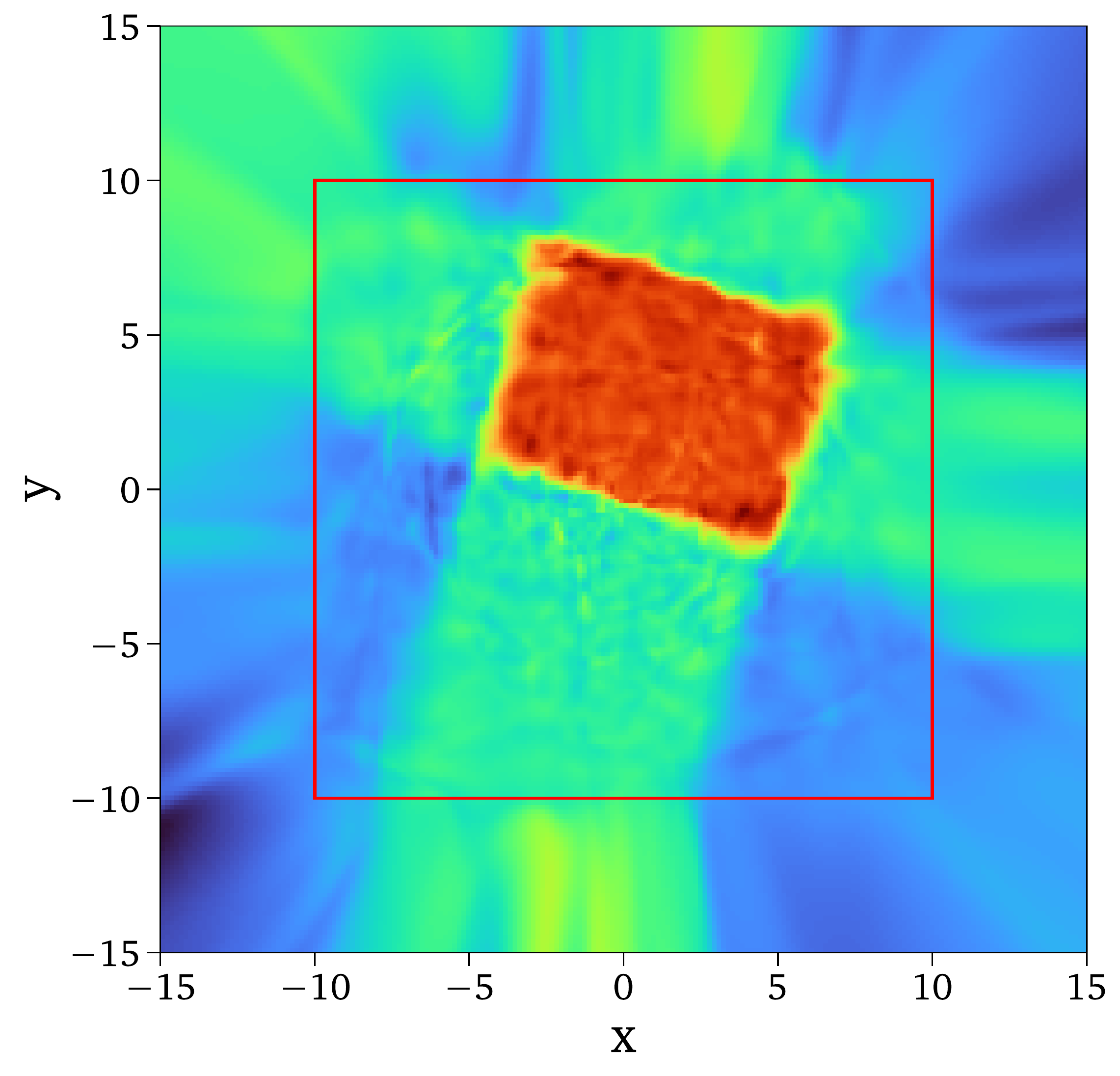}
    \end{subfigure}
    \begin{subfigure}{0.24\textwidth}
        \centering
        \caption{}
        \label{fig:Rect_DeepONet_Supp_16}
        \includegraphics[width=1.5in]{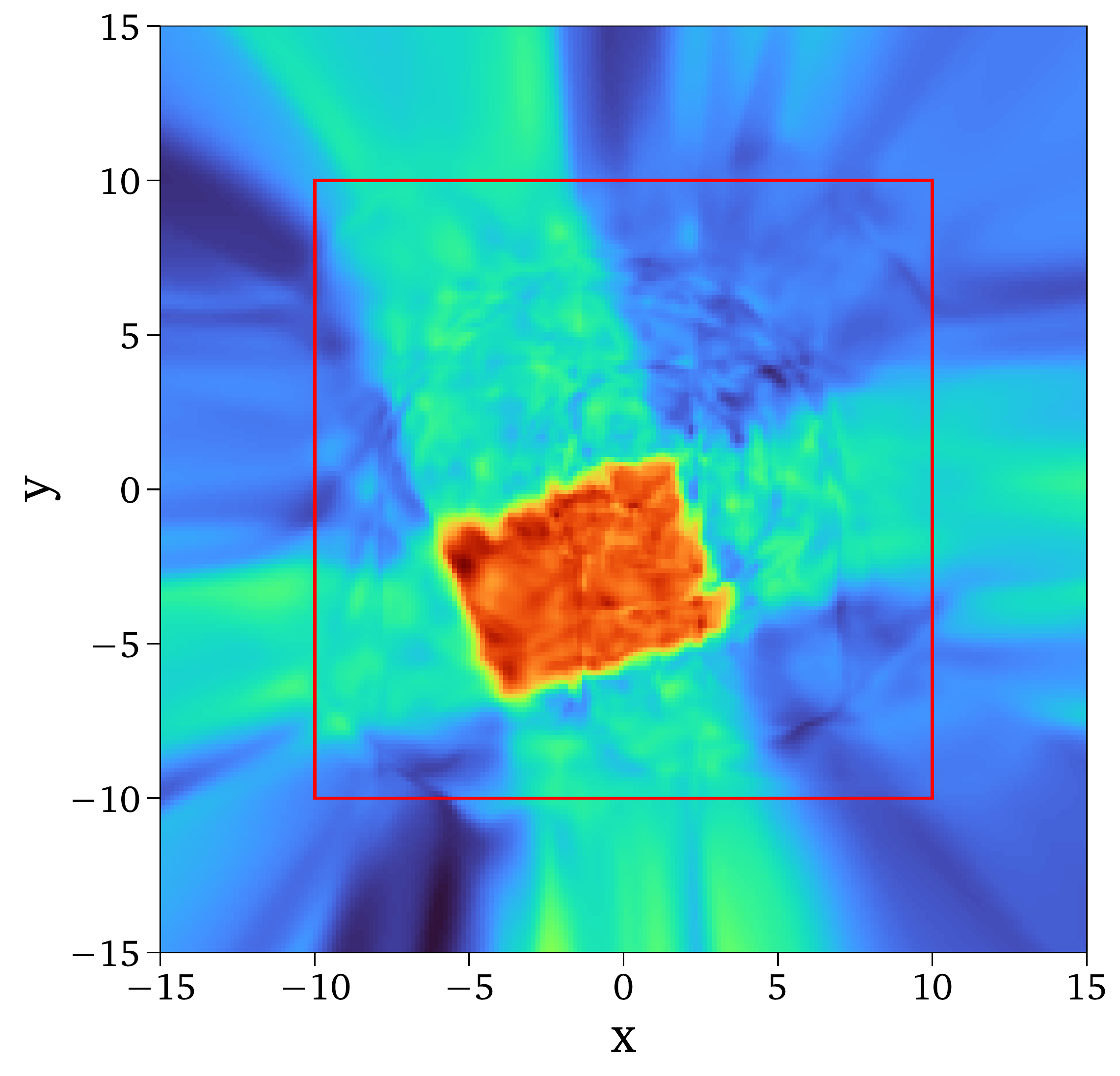}
    \end{subfigure}
    \caption{\textbf{Vanilla DeepONet predictions of the rigid body's dynamics}. The snapshots are produced at sixteen time instants uniformly spaced between $t=0$ [s] and $t=10$ [s]. Note: the training of the vanilla DeepONet relied only on data randomly selected inside the $x-y$ domain represented by the red squares. Predictions outside the red squares are extrapolations.}
    \label{fig:Rect_DeepONet_Supp}
\end{figure}


\clearpage
\subsection{flexDeepONet}

\begin{figure}[!htb]
    \begin{subfigure}{0.24\textwidth}
        \centering
        \caption{}
        \label{fig:Rect_flexDeepONet_Supp_1}
        \includegraphics[width=1.5in]{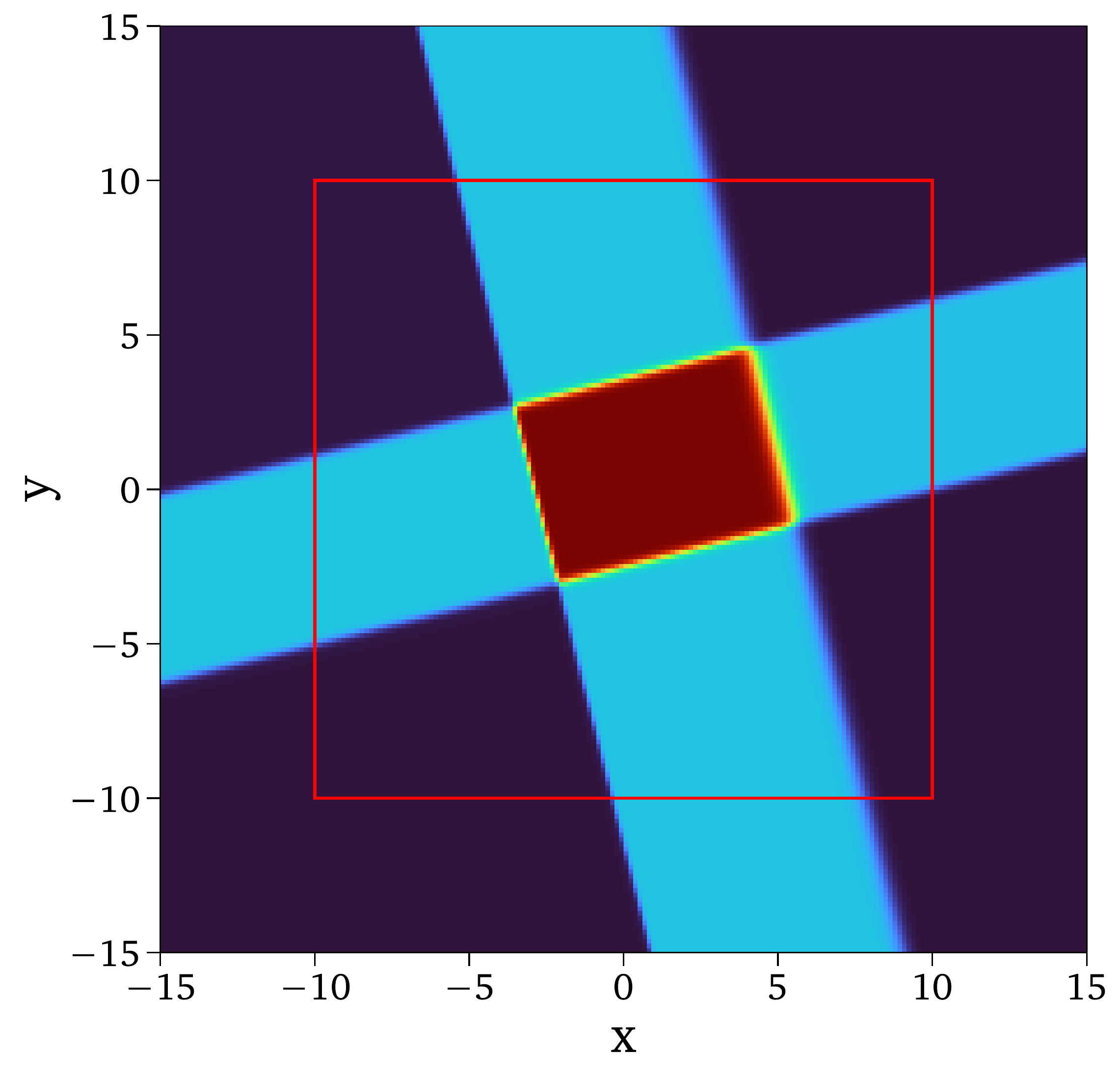}
    \end{subfigure}
    \begin{subfigure}{0.24\textwidth}
        \centering
        \caption{}
        \label{fig:Rect_flexDeepONet_Supp_2}
        \includegraphics[width=1.5in]{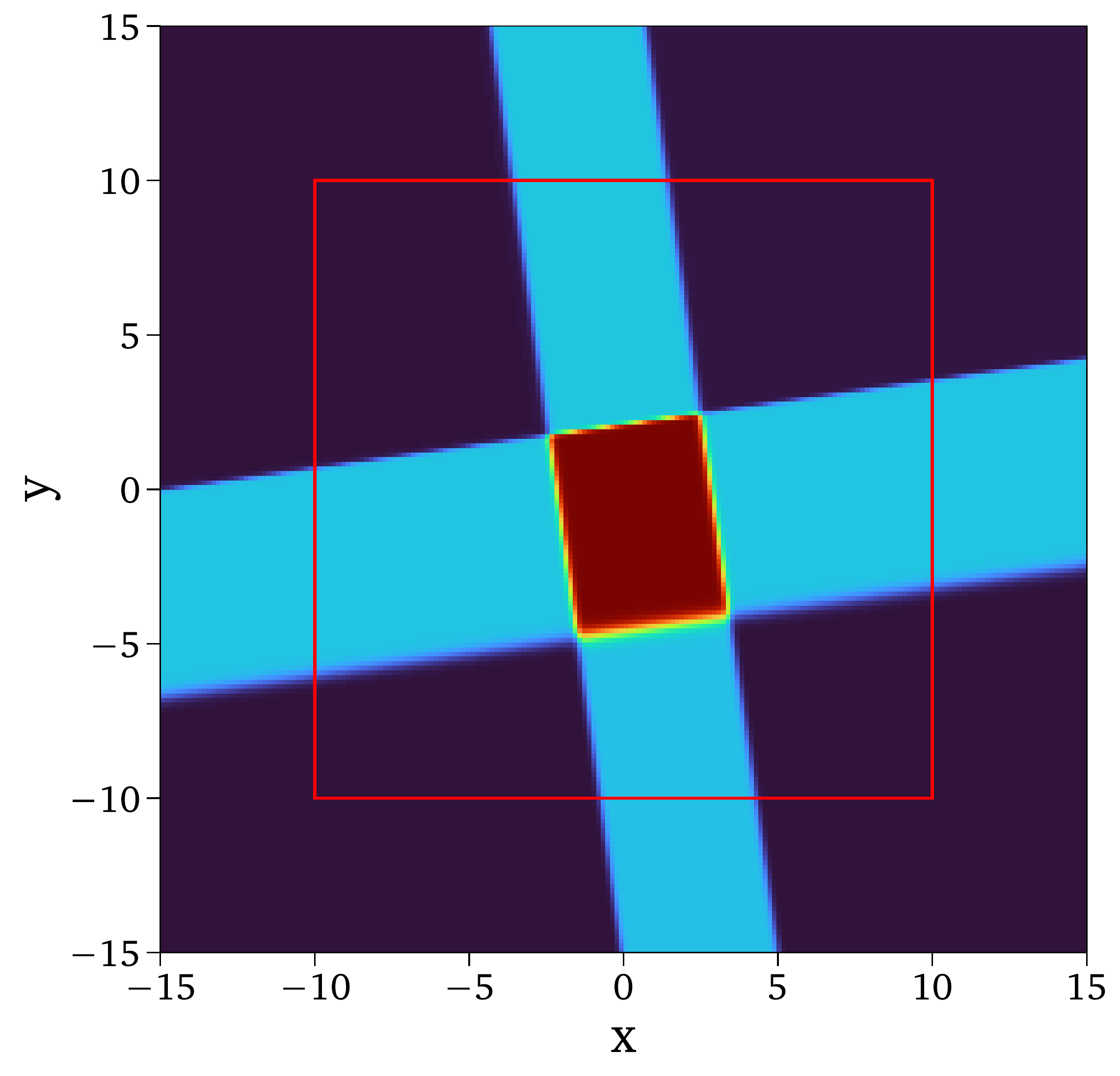}
    \end{subfigure}
    \begin{subfigure}{0.24\textwidth}
        \centering
        \caption{}
        \label{fig:Rect_flexDeepONet_Supp_3}
        \includegraphics[width=1.5in]{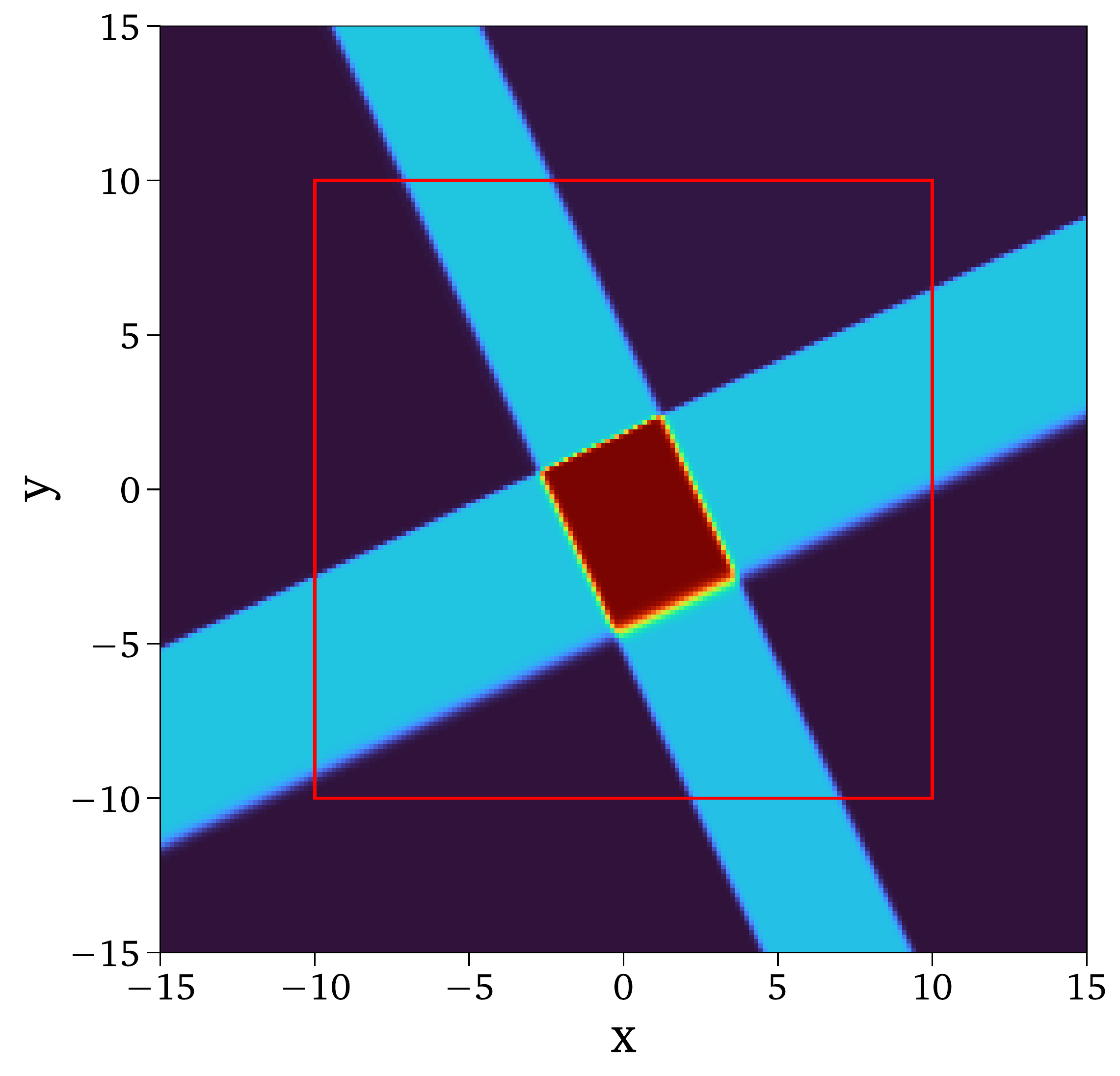}
    \end{subfigure}
    \begin{subfigure}{0.24\textwidth}
        \centering
        \caption{}
        \label{fig:Rect_flexDeepONet_Supp_4}
        \includegraphics[width=1.5in]{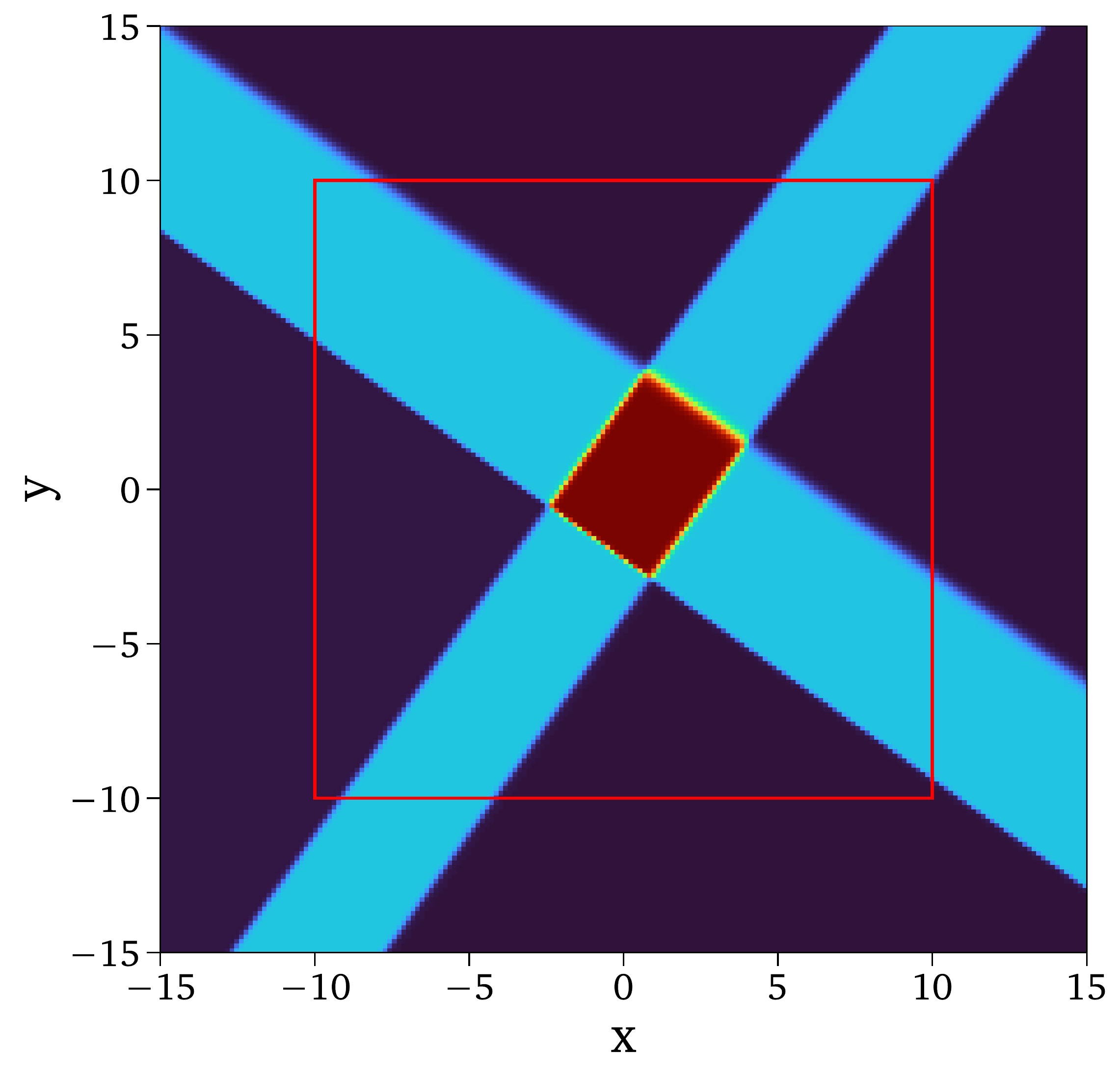}
    \end{subfigure}
    
    \begin{subfigure}{0.24\textwidth}
        \centering
        \caption{}
        \label{fig:Rect_flexDeepONet_Supp_5}
        \includegraphics[width=1.5in]{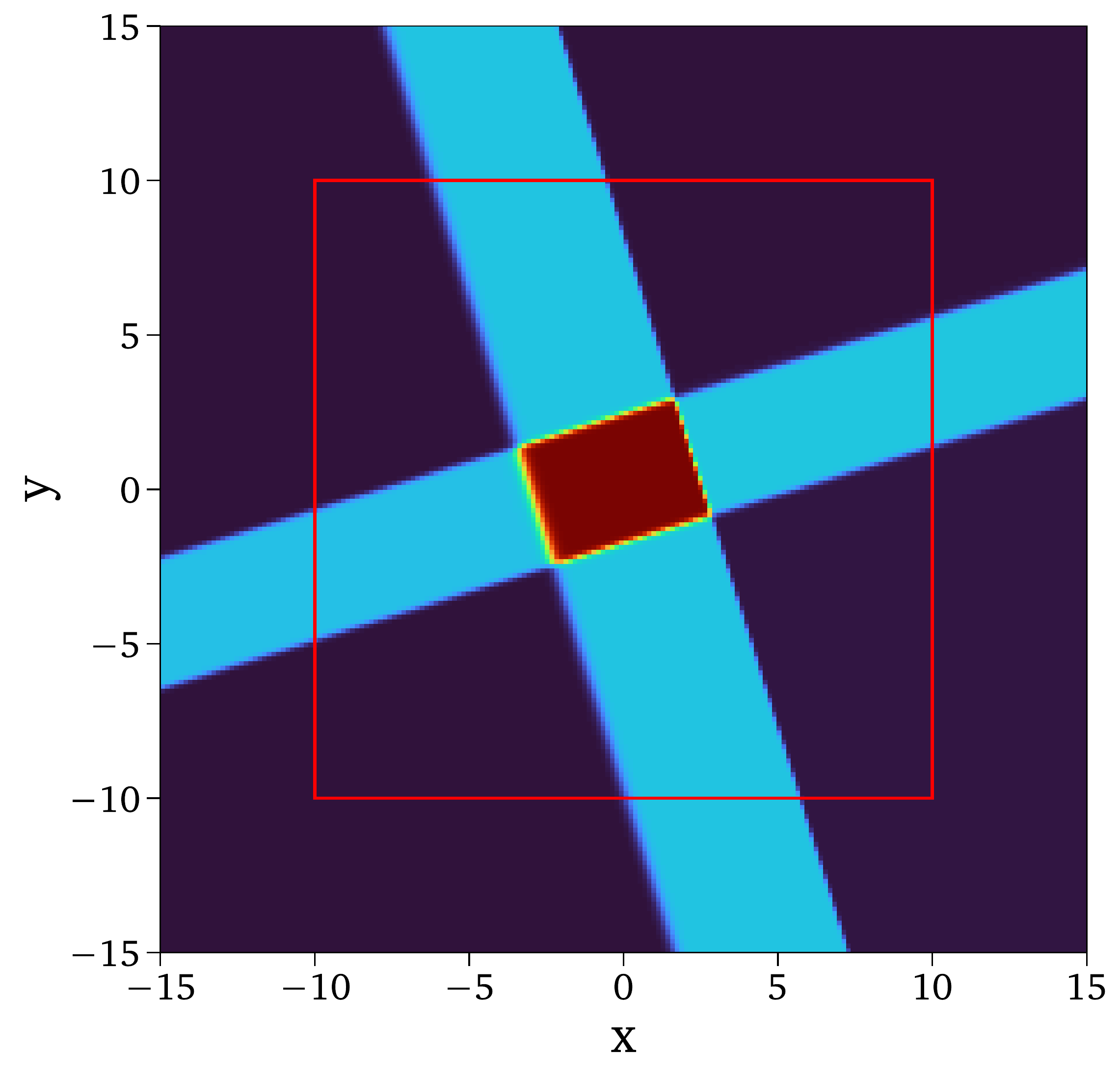}
    \end{subfigure}
    \begin{subfigure}{0.24\textwidth}
        \centering
        \caption{}
        \label{fig:Rect_flexDeepONet_Supp_6}
        \includegraphics[width=1.5in]{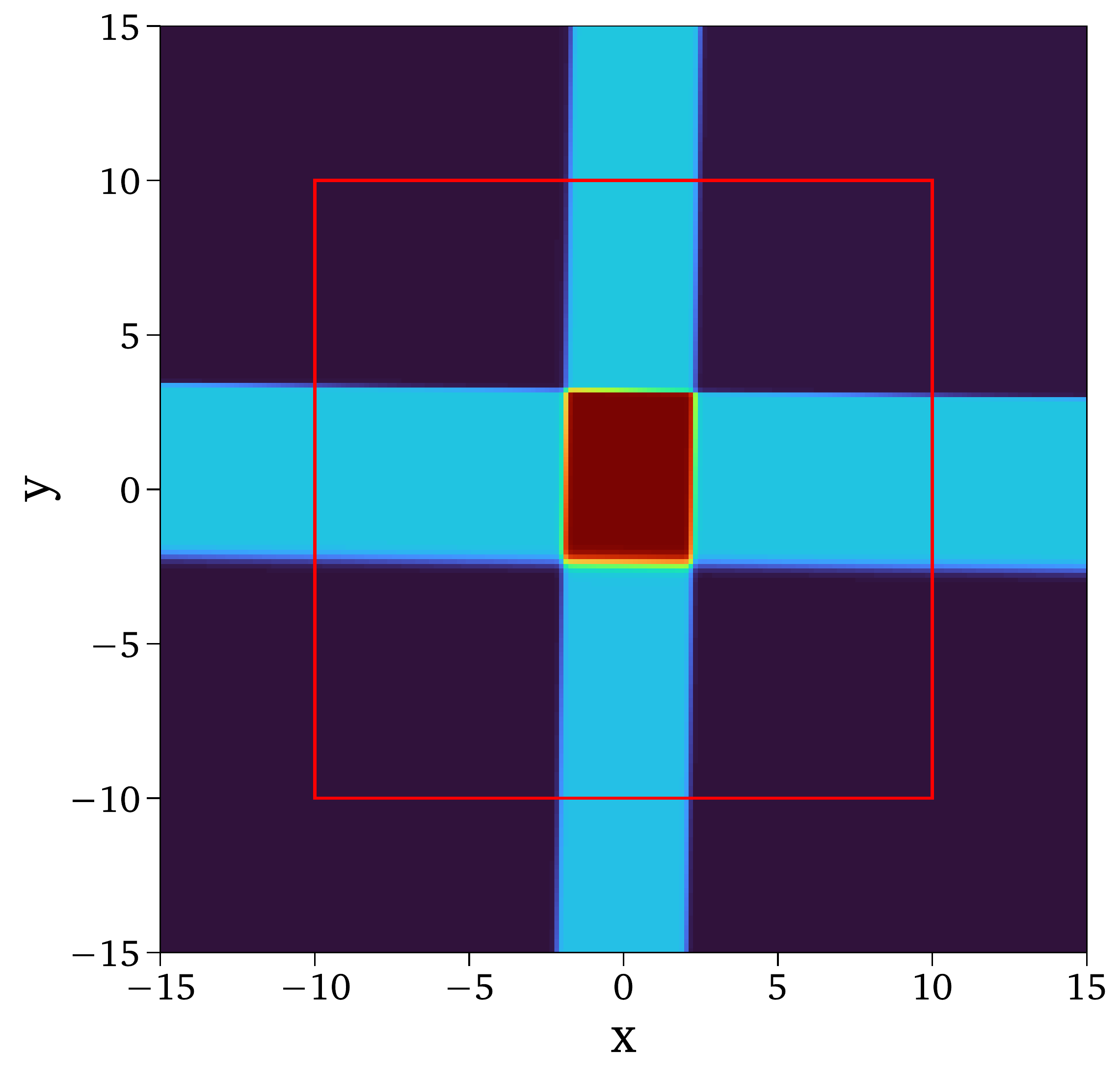}
    \end{subfigure}
    \begin{subfigure}{0.24\textwidth}
        \centering
        \caption{}
        \label{fig:Rect_flexDeepONet_Supp_7}
        \includegraphics[width=1.5in]{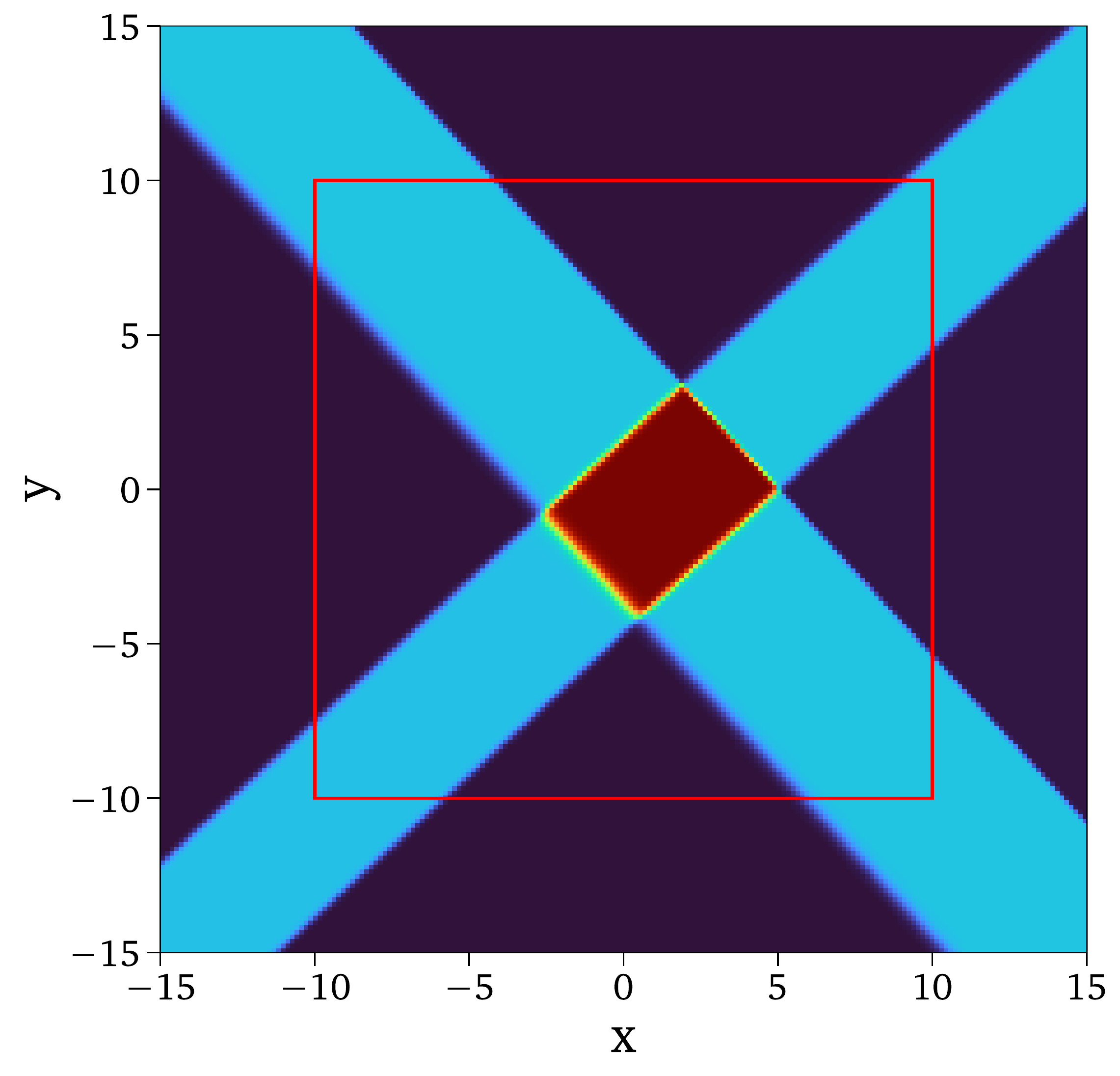}
    \end{subfigure}
    \begin{subfigure}{0.24\textwidth}
        \centering
        \caption{}
        \label{fig:Rect_flexDeepONet_Supp_8}
        \includegraphics[width=1.5in]{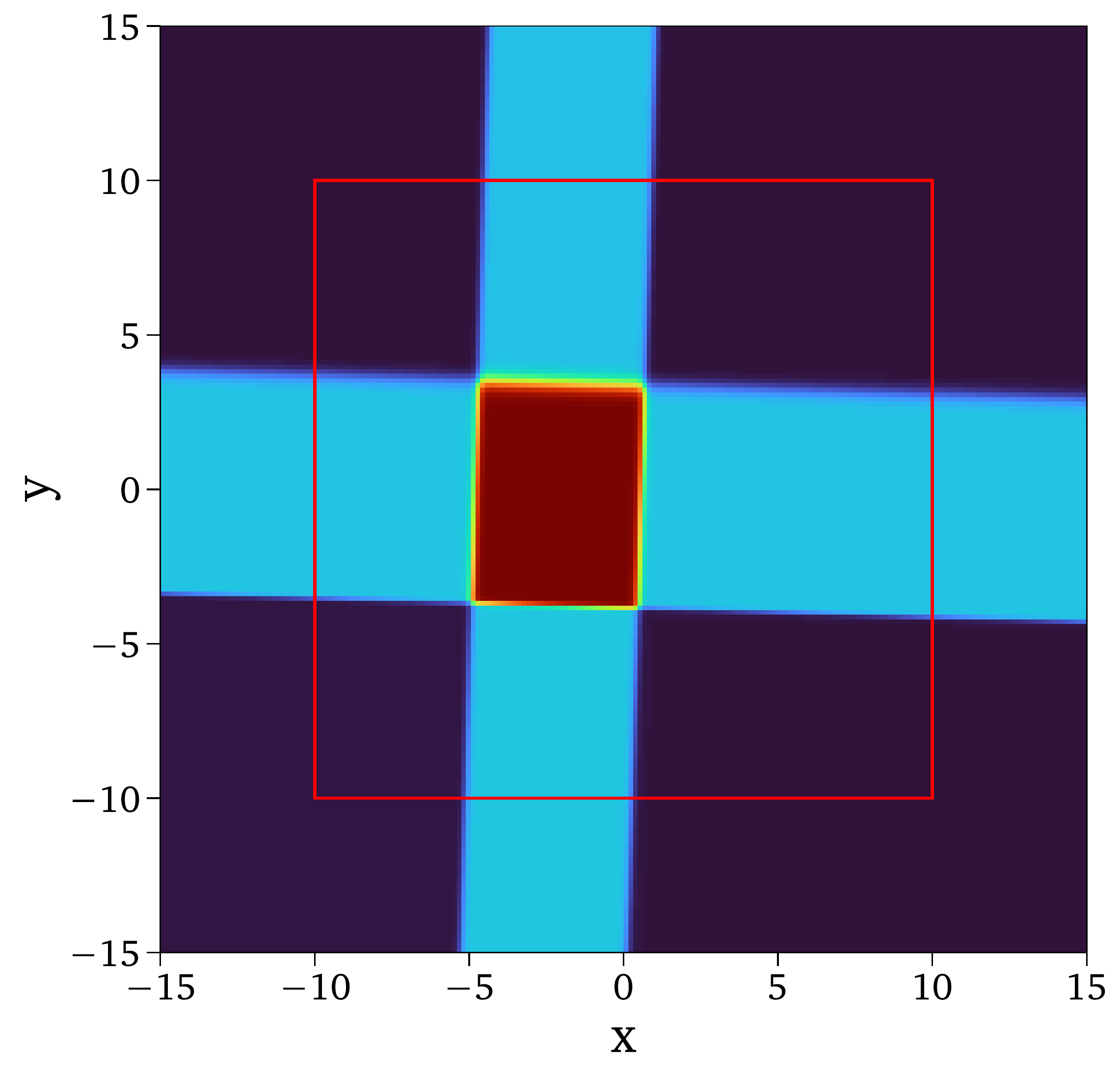}
    \end{subfigure}
    
    \begin{subfigure}{0.24\textwidth}
        \centering
        \caption{}
        \label{fig:Rect_flexDeepONet_Supp_9}
        \includegraphics[width=1.5in]{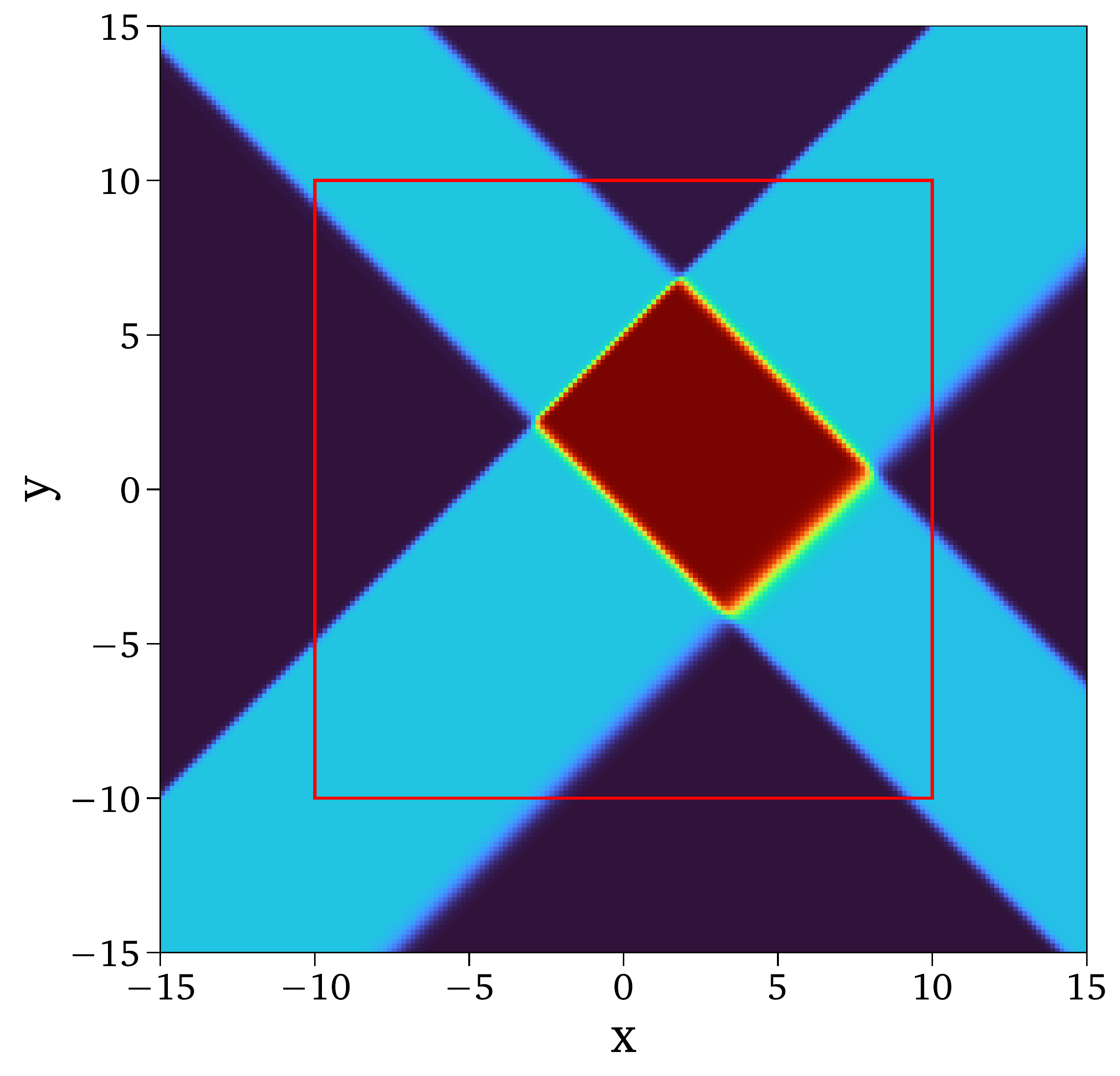}
    \end{subfigure}
    \begin{subfigure}{0.24\textwidth}
        \centering
        \caption{}
        \label{fig:Rect_flexDeepONet_Supp_10}
        \includegraphics[width=1.5in]{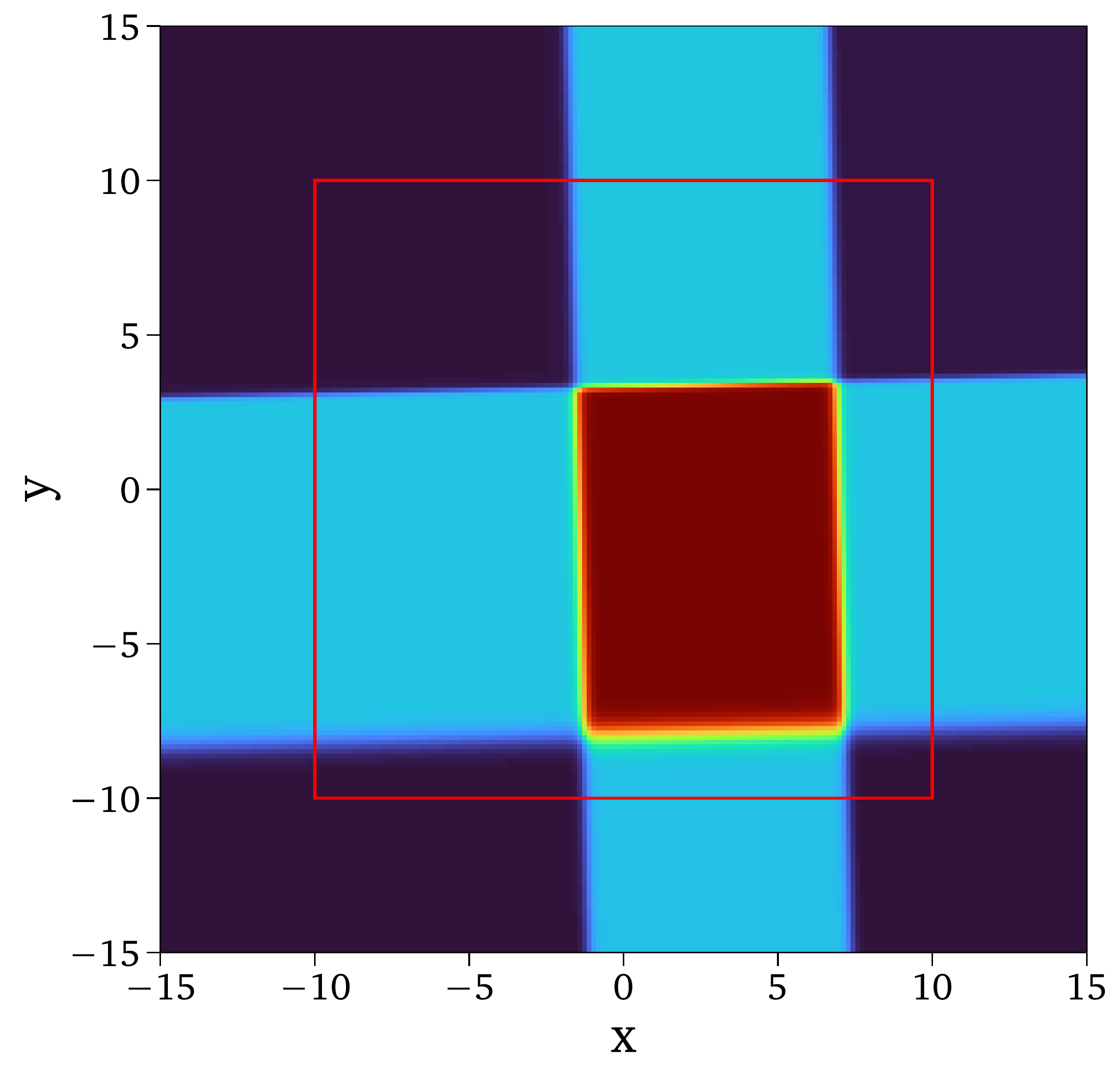}
    \end{subfigure}
    \begin{subfigure}{0.24\textwidth}
        \centering
        \caption{}
        \label{fig:Rect_flexDeepONet_Supp_11}
        \includegraphics[width=1.5in]{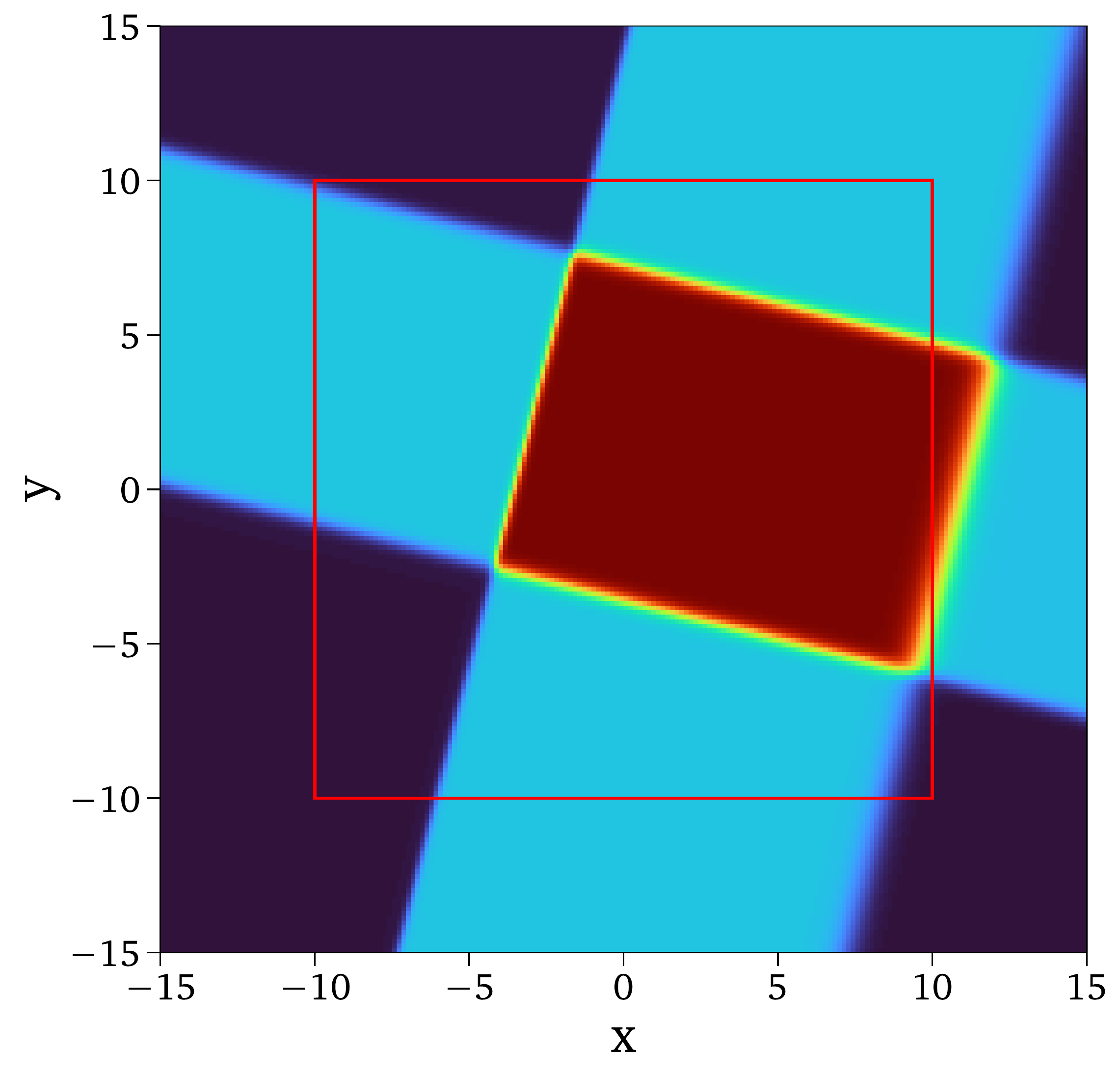}
    \end{subfigure}
    \begin{subfigure}{0.24\textwidth}
        \centering
        \caption{}
        \label{fig:Rect_flexDeepONet_Supp_12}
        \includegraphics[width=1.5in]{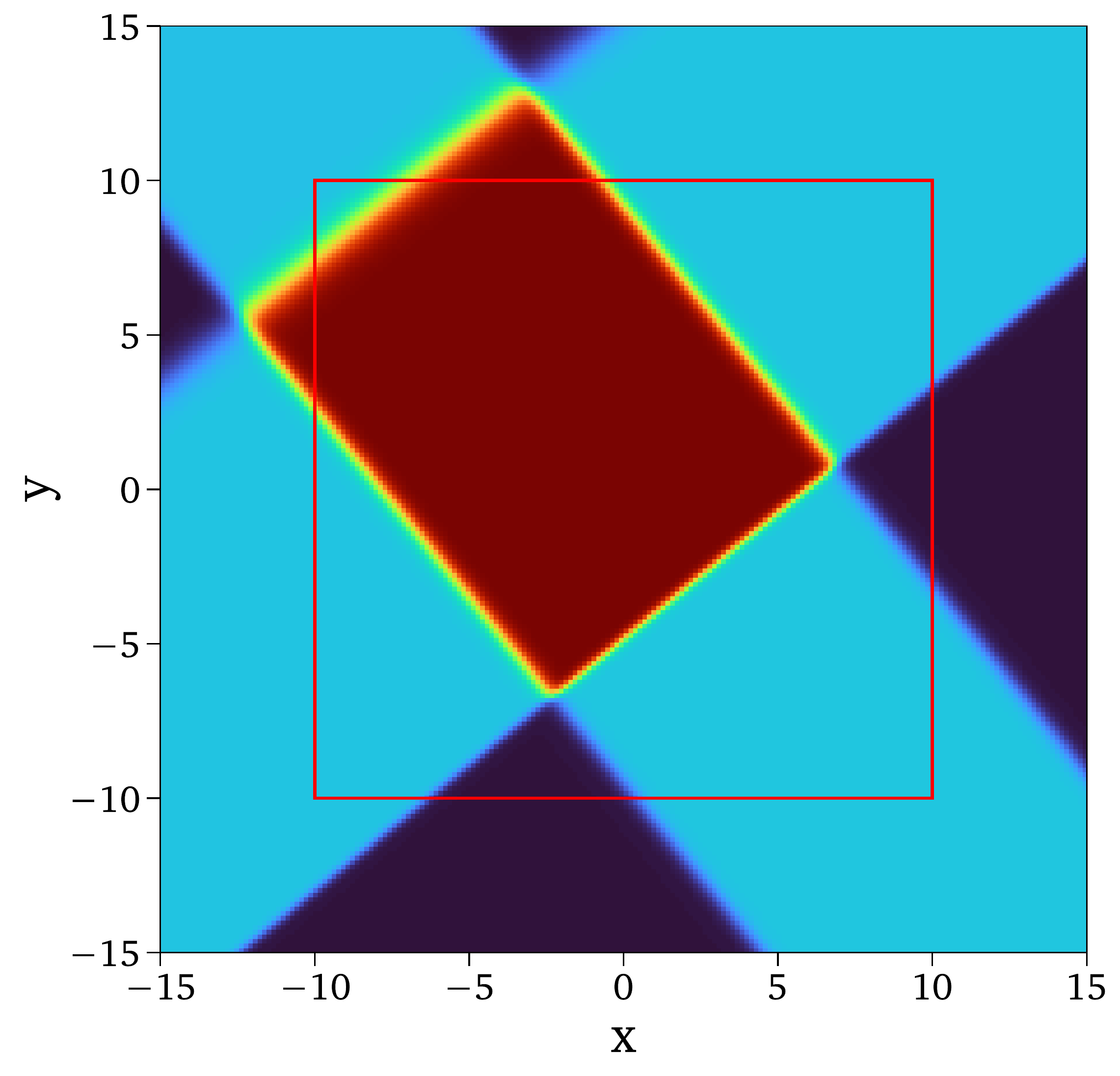}
    \end{subfigure}
    
    \begin{subfigure}{0.24\textwidth}
        \centering
        \caption{}
        \label{fig:Rect_flexDeepONet_Supp_13}
        \includegraphics[width=1.5in]{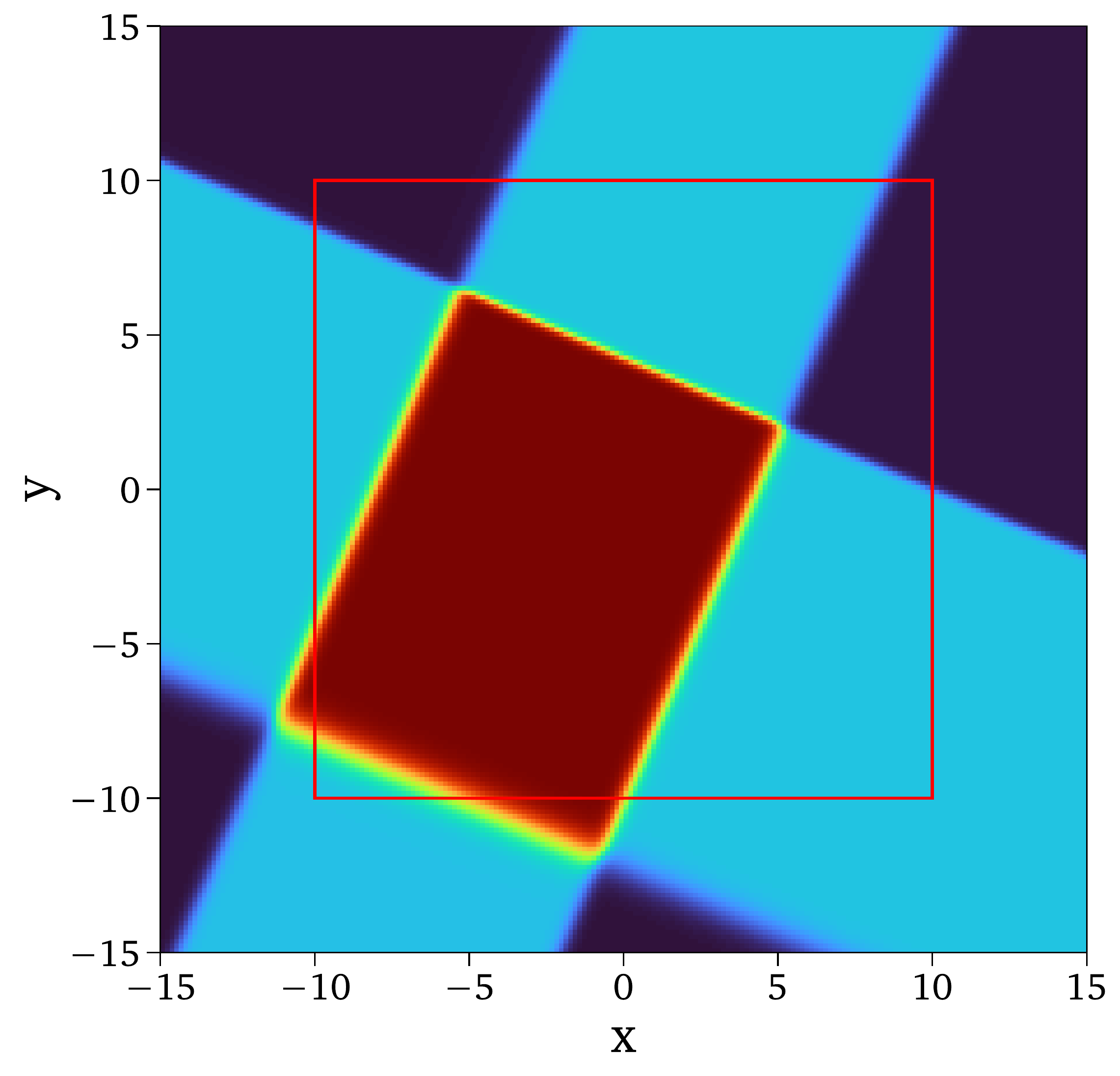}
    \end{subfigure}
    \begin{subfigure}{0.24\textwidth}
        \centering
        \caption{}
        \label{fig:Rect_flexDeepONet_Supp_14}
        \includegraphics[width=1.5in]{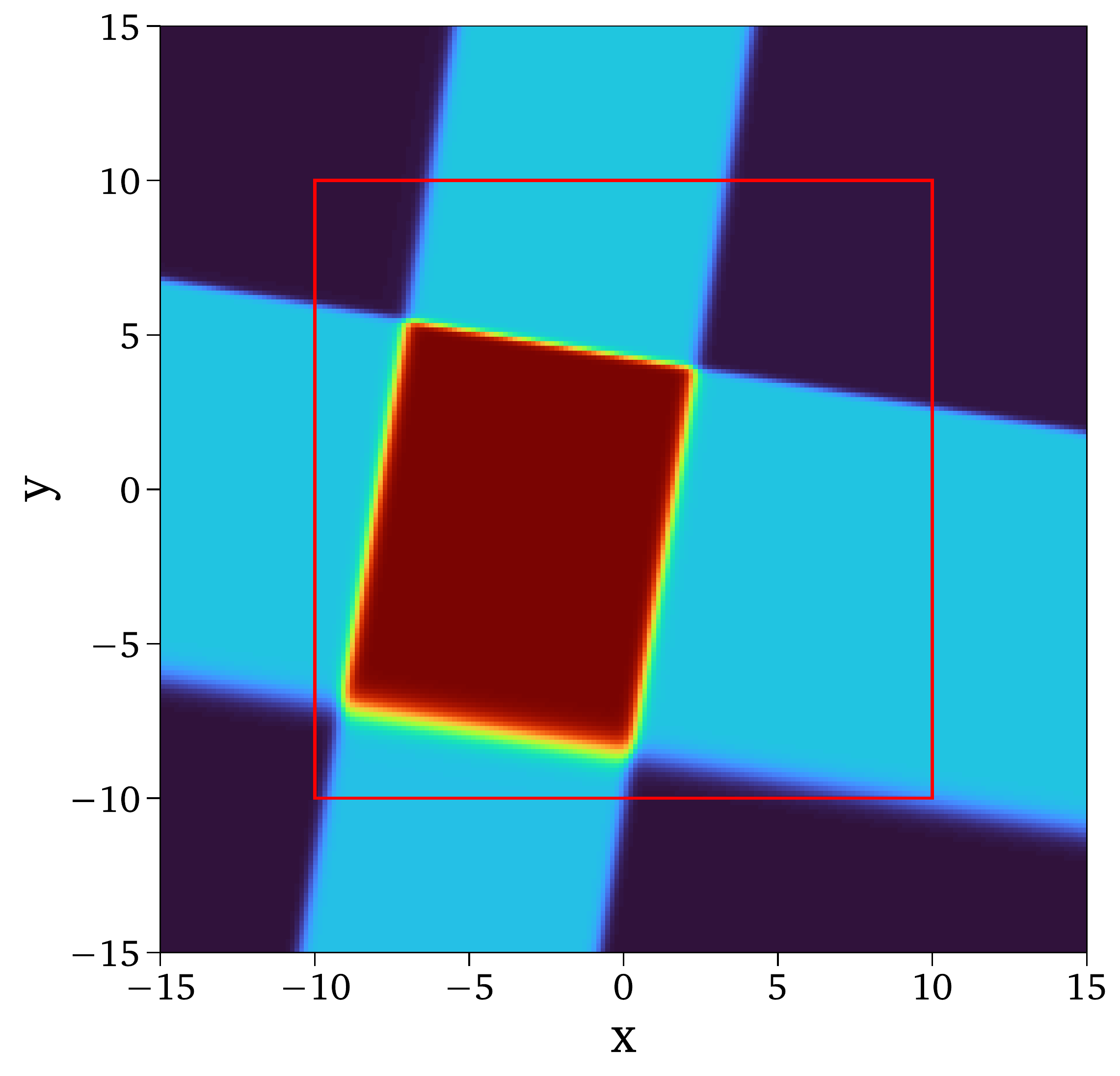}
    \end{subfigure}
    \begin{subfigure}{0.24\textwidth}
        \centering
        \caption{}
        \label{fig:Rect_flexDeepONet_Supp_15}
        \includegraphics[width=1.5in]{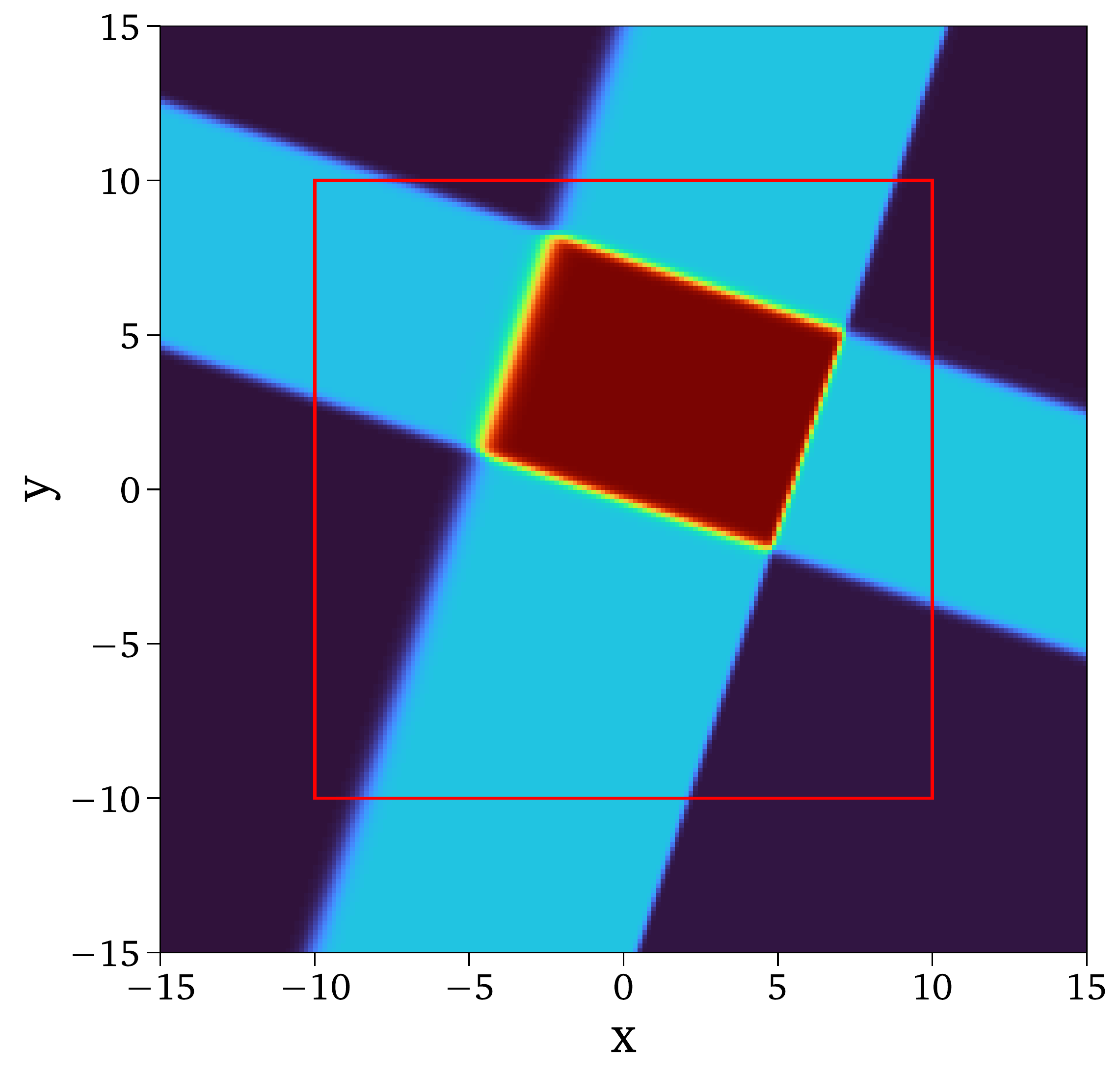}
    \end{subfigure}
    \begin{subfigure}{0.24\textwidth}
        \centering
        \caption{}
        \label{fig:Rect_flexDeepONet_Supp_16}
        \includegraphics[width=1.5in]{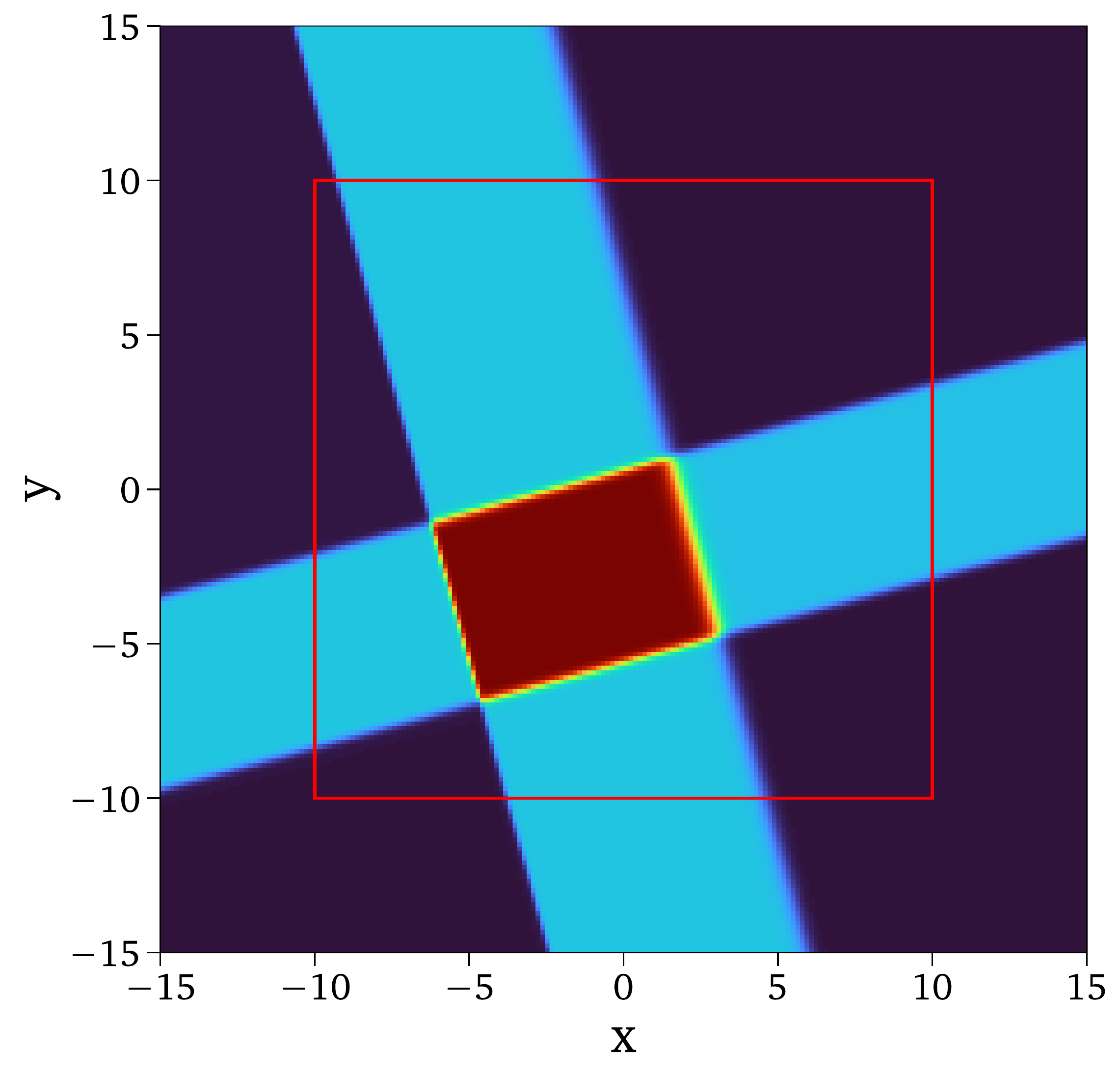}
    \end{subfigure}
    \caption{\textbf{FlexDeepONet predictions of the rigid body's dynamics}. The snapshots are produced at sixteen time instants uniformly spaced between $t=0$ [s] and $t=10$ [s]. Note: FlexDeepONet's training relied only on data randomly selected inside the $x-y$ domain represented by the red squares. Predictions outside the red squares are extrapolations.}
    \label{fig:Rect_flexDeepONet_Supp}
\end{figure}

\clearpage
\subsection{Architectures Comparison}

\begin{table}[!htp] 
    \centering
    \begin{tabular}{||l|c|c|c||} 
        \hline
         & Vanilla DeepONet & flexDeepONet \\[1.2ex] 
         & (Fig.~\ref{fig:Rect_DeepONet_1}) & (Fig.~\ref{fig:Rect_DeepONet_2}) \\[1.2ex] 
        \hline
        $p$ & 128 & 1 \\[1.8ex]
        \hline
        No. of Parameters & 165,761 & 1,921 \\[1.8ex]
        \hline
        \hline
        RMSE & 0.939761 & 0.318290 \\[1.2ex] 
        \hline
        \hline
        Training Time & 81,103 [s] & 14,927 [s] \\[1.2ex] 
        \hline
        Prediction Time & 148 [ms/snapshot] & 21 [ms/snapshot] \\[1.2ex] 
        \hline
    \end{tabular}
    \vspace{5mm}
    \caption{\textbf{Comparison between architectures}. Comparisons between the number of trunks' outputs ($p$), the number of trainable parameters, the root mean squared errors (RMSEs) at the one hundred test time-snapshots, the training times, and the prediction times. Training has been performed via Tensorflow 2.8. The prediction times refer to the overall times necessary for the architectures to predict the $z$ output at a time-snapshot made of 10,000 spatial locations, and they have been computed by relying on architecture's implementations based on the NumPy 1.22.3 library. It should be noted that POD-DeepONet has not been applied to this test case, given that the data is not aligned on the spatial grid and the modes cannot be computed via classical POD/SVD}\label{table:Rect_TestErrors}
\end{table}


\end{document}